\documentclass[preprint,11pt]{elsarticle}

\usepackage{amssymb}
\usepackage{lineno}

\usepackage{multirow}
\usepackage{tabularx}
\usepackage{caption}

\usepackage{caption} 
\usepackage{subcaption}

\usepackage{rotating}
\usepackage{booktabs}

\usepackage[hyphens]{url}

\usepackage{xcolor}

\usepackage{enumitem}

\journal{Journal of Artificial Intelligence}

\begin{document}

% \begin{linenumbers}
\begin{frontmatter}

%% Title, authors and addresses

%% use the tnoteref command within \title for footnotes;
%% use the tnotetext command for theassociated footnote;
%% use the fnref command within \author or \address for footnotes;
%% use the fntext command for theassociated footnote;
%% use the corref command within \author for corresponding author footnotes;
%% use the cortext command for theassociated footnote;
%% use the ead command for the email address,
%% and the form \ead[url] for the home page:
%% \title{Title\tnoteref{label1}}
%% \tnotetext[label1]{}
%% \author{Name\corref{cor1}\fnref{label2}}
%% \ead{email address}
%% \ead[url]{home page}
%% \fntext[label2]{}
%% \cortext[cor1]{}
%% \affiliation{organization={},
%%             addressline={},
%%             city={},
%%             postcode={},
%%             state={},
%%             country={}}
%% \fntext[label3]{}

\title{NovPhy: A Testbed for Physical Reasoning in Open-world Environments}

%% use optional labels to link authors explicitly to addresses:
%% \author[label1,label2]{}
%% \affiliation[label1]{organization={},
%%             addressline={},
%%             city={},
%%             postcode={},
%%             state={},
%%             country={}}
%%
%% \affiliation[label2]{organization={},
%%             addressline={},
%%             city={},
%%             postcode={},
%%             state={},
%%             country={}}
\author[inst1,label2]{Chathura Gamage}
\author[inst1,label2]{Vimukthini Pinto}%\fnref{myfootnote}
\author[inst1,label2]{Cheng Xue} 

\author[inst1]{\\Peng Zhang}
\author[inst1]{Ekaterina Nikonova}
\author[inst2]{Matthew Stephenson}
\author[inst1]{Jochen Renz}
\affiliation[inst1]{organization={School of Computing},%Department and Organization
            addressline={The Australian National University}, 
            city={Canberra},
            country={Australia}}
\affiliation[inst2]{organization={College of Science and Engineering},%Department and Organization
            addressline={Flinders University}, 
            city={Adelaide},
            country={Australia}}

\affiliation[label2]{These authors contributed equally}

\begin{abstract}
Due to the emergence of AI systems that interact with the physical environment, there is an increased interest in incorporating physical reasoning capabilities into those AI systems. 
But is it enough to only have physical reasoning capabilities to operate in a real physical environment?
In the real world, we constantly face novel situations we have not encountered before. As humans, we are competent at successfully adapting to those situations.
Similarly, an agent needs to have the ability to function under the impact of novelties in order to properly operate in an open-world physical environment.
To facilitate the development of such AI systems, we propose a new testbed, NovPhy, that requires an agent to reason about physical scenarios in the presence of novelties and take actions accordingly. 
% We develop eight novelties representing a diverse novelty space. Also, we create a variety of task templates for commonly encountered five physical scenarios in a physical environment related to applying forces and motions such as rolling, falling, and sliding of objects.
% Then we applied the developed novelties to those task templates, which results in 40 novel task templates.
The testbed consists of tasks that require agents to detect and adapt to novelties in physical scenarios.
To create tasks in the testbed, we develop eight novelties representing a diverse novelty space and apply them to five commonly encountered scenarios in a physical environment, related to applying forces and motions such as rolling, falling, and sliding of objects.
%We evaluate a range of agents on their novelty detection and adaptation performance using these tasks.
%The novelty detection and adapatation abilities of different physcial reasoning agents can then be measured using tasks based on these common novelties.
According to our testbed design, we evaluate two capabilities of an agent: the performance on a novelty when it is applied to different physical scenarios and the performance on a physical scenario when different novelties are applied to it. 
We conduct a thorough evaluation with human players, learning agents, and heuristic agents. Our evaluation shows that humans' performance is far beyond the agents' performance. Some agents, even with good normal task performance, perform significantly worse when there is a novelty, and the agents that can adapt to novelties typically adapt slower than humans.
We promote the development of intelligent agents capable of performing at the human level or above when operating in open-world physical environments. Testbed website: \url{https://github.com/phy-q/novphy}
\end{abstract}

%%Graphical abstract
% \begin{graphicalabstract}
% \includegraphics{grabs}
% \end{graphicalabstract}

%%Research highlights
% \begin{highlights}
% \item Research highlight 1
% \item Research highlight 2
% \end{highlights}

\begin{keyword}
%% keywords here, in the form: keyword \sep keyword
Physical Reasoning \sep Open-world Learning \sep Novelty Testbed 
%% PACS codes here, in the form: \PACS code \sep code
%\PACS 0000 \sep 1111
%% MSC codes here, in the form: \MSC code \sep code
%% or \MSC[2008] code \sep code (2000 is the default)
%\MSC 0000 \sep 1111
\end{keyword}

\end{frontmatter}

%% \linenumbers

%% main text
\section{Introduction}
\label{sec:introduction}
A key aspect of human intelligence is the ability to reason about the physical behaviour of objects and make decisions in the physical environment \cite{Davis2006}. Research suggests that humans develop physical reasoning capabilities within just the first year of birth \cite{Renee1995,Renee2009}. Even though it has been challenging to develop AI systems that can do general physical reasoning as good as humans do \cite{Chollet19,PhyQ}, there is work that shows AI systems that could achieve human performance in some physical reasoning tasks \cite{VirtualTools,Physion}. But the question is: is it adequate to only have physical reasoning capabilities to successfully work in the real physical world?

\begin{figure}[h!]
\centering
\includegraphics[width=1\textwidth]{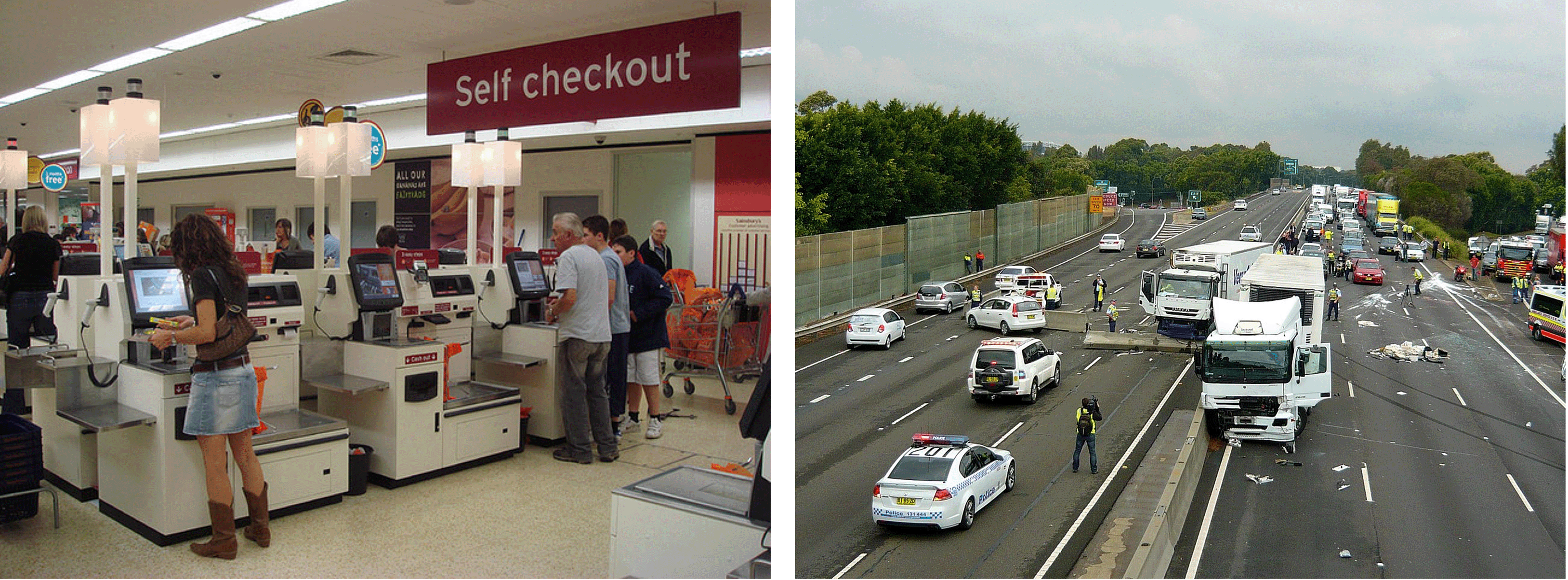}
\caption{Example novelties that could be encountered in the real world. Left: self-checkout machines started appearing in supermarkets after the early 2000s, until then customers were used to traditional checkout methods, hence using self-checkout machines was a novelty to the customers \cite{WikiSelfCheckout}. Right: traffic accidents are generally novel situations for self-driving cars as such incidents are rare in the training data and usually visually unique in each incident \cite{WikiRoadAccident}.}
\label{fig_real_world_novelties}
\end{figure}

Encountering novel situations is an inherent characteristic of the real world (Figure \ref{fig_real_world_novelties}). As humans, we are adept at working in such novel situations that we constantly face in our day-to-day life. For example, consider a person who can ride a bicycle on a normal road. On a rainy day when the roads are slick, the person can still ride the bicycle safely by adjusting the speed and applying the brakes smoothly without slipping. Even though sometimes a human may initially take a suboptimal decision when facing a novel situation, they usually quickly recover successfully after continuing to work for a short period of time under the influence of novelty.
%Out of many definitions to describe a novelty (some references), one definition is, a novelty is states or situations that violate (implicit or explicit) assumptions about agents, the environment, and agent–agent and agent–environment interactions (ref BAA or any other relevant definition).
Similarly, for an agent that operates in an open-world physical environment, along with physical reasoning capabilities, it is crucial to possess capabilities that are required to handle novel situations \cite{Davis2022}, i.e., novelty detection and adaptation capabilities.

In contrast to the intelligence of humans, current AI systems tend to struggle when they are presented with situations that were not available during their training stage or if the situation was not anticipated by the developers \cite{Ben2007}. This could be due to the fact that the research field, Open World Learning (OWL), attempting to address this issue is relatively new \cite{Senator2019,Langley2020,MonopolyNoveltyGenerator}.
Apart from that, not having adequate testbeds to experiment and evaluate such AI systems also hinders their advancement. There are frameworks such as Monopoly \cite{MonopolyNoveltyGenerator}, Polycraft \cite{NovelCraft}, Cartpole \cite{CartPole}, etc, that treat novelties as first-class citizens and facilitate agent experimentation. Even though some of them are physics based environments \cite{CartPole}, none of them specifically focus on introducing novelties to the physical scenarios that an agent would encounter in the real world. Also, it is out of their context to evaluate agents in real-world physical tasks in the presence of novelties.

To fill the above gaps, we propose a new novelty-centric testbed, NovPhy, where agents need to perform in real-world physical scenarios in the presence of novelties. NovPhy includes a wide variety of novelties applied to different physical scenarios. We implement our testbed on the physics-based video game Angry Birds as it has realistic physics and it is a versatile domain to introduce physics-based novelties. Moreover, Angry Birds is popular in both physical reasoning research \cite{PhyQ,Renzet2015,RenzNature} and OWL research \cite{SBFramework, SBNoveltyGenerator, SBDifficulty}. The main contributions of this work can be summarized as follows:

\begin{itemize}
    \item \textbf{NovPhy - A testbed for novelty detection and adaptation in physical environments: } We consider five commonly encountered physical scenarios in NovPhy: applying a single force and multiple forces, and rolling, falling, and sliding objects. We developed eight novelties representing a diverse novelty space. We designed task templates by applying the eight novelties to the five physical scenarios separately, resulting in 40 novel task templates. A task template is used to generate related tasks by varying task template parameters such as the locations of the objects. We also created 40 corresponding normal task templates without novelties to facilitate our evaluation protocol. Further, we developed a task variation generator that can generate an unlimited amount of tasks from these task templates.

    \item \textbf{Agent evaluation setups for open-world physical environments: } We propose a comprehensive evaluation setup to evaluate the novelty detection and adaptation of AI systems in open-world physical environments. In this setup the novelties are orthogonal to the physical scenarios, hence facilitating us to evaluate agents in two settings: first, the same novelty is applied separately to the tasks of multiple physical scenarios, and second, multiple novelties are applied separately to the tasks of the same physical scenario. The former is used to evaluate how well an agent can deal with the same novelty in different physical reasoning tasks and the latter is used to evaluate how well an agent can perform the same physical reasoning task under different novel situations.

    \item \textbf{Establishing results for baseline agents: } We evaluate 11 baseline agents. Three heuristic-based agents, two standard online learning agents, two standard offline agents, three adaptive learning agents, and a random agent. We report their novelty detection and novelty adaptation performance.

    \item \textbf{Establishing baseline human performance: } In order to show our novelties are adaptable for humans, we conducted an experiment using human players in our testbed. These results also show that humans can detect and adapt to novelties better and faster compared to AI agents, thus acting as a milestone performance for AI to achieve.
\end{itemize}

\section{Background and Related Work}
\label{sec:background_and_related_work}

In this section, we discuss the background and related work regarding physical reasoning research, novelty theories, and the existing novelty-centric benchmarks/testbeds. 

\subsection{Physical Reasoning Research}
\label{sub-sec:physical_reasoning_research}

Physical reasoning has become an important aspect of AI research due to the increased reliance on autonomous AI systems in day-to-day operations. There are multiple physical reasoning benchmarks and testbeds available to assist AI systems enhance physical reasoning capabilities in order to perform securely in the real world.

The physical reasoning benchmarks such as Physion \cite{Physion}, IntPhys \cite{IntPhys2019}, CLEVERER \cite{CLEVRER} are based on videos while COPHY \cite{COPHY} is an image-based physical reasoning benchmark. Physion comprises eight physical reasoning scenarios, including rolling, sliding, and projectile motion, which are important capabilities to work in the physical world \cite{Physion}. IntPhys, on the other hand, is concerned with physical reasoning abilities acquired during infancy, such as object permanence, spatio-temporal continuity, and shape consistency \cite{IntPhys2019}. The CLEVERER benchmark presents videos and asks questions inspired by the theory of human causal judgement \cite{CLEVRER}. COPHY benchmark presents a sequence of images based on physical scenarios and asks to predict the outcome if you make a modification to the initial image. COPHY is developed to test counterfactual reasoning applied to the physical world \cite{COPHY}. 

To bring physical reasoning abilities of AI systems closer to reality, researchers have created action based benchmarks that require agents to take an action in order to accomplish the goal. Examples of such benchmarks include PHYRE \cite{Phyre} and Virtual Tools \cite{VirtualTools}. PHYRE is a benchmark consisting of simple 2D physics based tasks, aimed to foster the development of efficient models capable of generalisation across tasks \cite{Phyre}. The Virtual Tools game, focuses on evaluating agents on selecting appropriate tools and taking the correct action using the tool to solve the tasks \cite{VirtualTools}.

Phy-Q \cite{PhyQ} is also an action based physical reasoning testbed that consists of a broad variety of 15 physical reasoning scenarios. Some physical scenarios in Phy-Q (applying single or multiple forces, rolling, falling, sliding) are inspired by physical reasoning abilities developed during childhood, whereas some scenarios (adequate timing, clearing paths, and manoeuvring) are required to overcome challenges for robots to work safely in physical environments. As an improvement to the previously mentioned benchmarks, Phy-Q testbed supports different evaluation settings based on different generalization levels (local generalization and broad generalization) and also they have established human performance on the scenarios in the testbed. Furthermore, this testbed allows us to compute the physical reasoning quotient, which reflects an agent's physical reasoning aptitude.

Even though all the preceding benchmarks support the development of AI systems with advanced physical reasoning capabilities, none of them focuses on physical reasoning under novel circumstances, which is the setting an agent in the real world would frequently encounter. In our testbed NovPhy, we combine physical reasoning scenarios with novel situations, to create a real-world like setting to evaluate the agent's performance. To introduce novel situations, we consider the first five physical reasoning scenarios from the Phy-Q tesbed: single force, multiple forces, rolling, falling, and sliding.

\subsection{Theories of Novelty}
AI systems have already shown superhuman performance in a wide range of closed-world domains \cite{silver2017, Silver2018, Vinyals2019}. However, compared to a closed-world, in an open world, the agents may struggle to perform due to encountering novel situations. Even though some novel situations can be predicted by the developers, some cannot be anticipated, making it impossible to integrate all the possible novel situations into an agent model.
DARPA has launched the Science of Artificial Intelligence and Learning for Open-world Novelty (SAIL-ON) program to investigate and develop the underlying scientific principles, general engineering techniques, and algorithms required to create AI systems that adapt appropriately when a novel situation arises \cite{Senator2019, SAILON}. 

Researchers have looked at different directions to formalize what it means to be a novelty. The novel situations are sometimes referred to as anomalies or out-of-distribution data by some researchers \cite{Boult2021, NovelCraft}. SAIL-ON program defines novelty as situations that violate implicit or explicit assumptions in an agent’s model of the external world, including other agents, the environment, and their interactions \cite{Doctor2022, MonopolyNoveltyGenerator}. Alternatively, Langley describes novelty as transformations of the elements in the environment \cite{Langley2020}. Examples of such transformations are spatio-temporal transformations, structural transformations, process transformations, and constraint transformations. In addition, Molineaux and Dannenhauer \cite{Molineaux2022} formally define different environmental transformations. On the other hand, Boult et al \cite{Boult2021}, have introduced a unifying framework of novelty in the context of AI. Boult et al, focus on the world space, observation space, and the agent state to formally define diverse types of novelties. 

A working group in the SAIL-ON program, the novelty working group (see \cite{Doctor2022} for participant list), has also developed a novelty hierarchy that enables the categorization of the novelties considering their properties. This categorization of novelties facilitates a solid novelty evaluation as it helps to design novelties covering a large novelty space and helps to identify different categories of novelties an agent model would fail to perform. In this paper, we use the novelty hierarchy developed at the SAIL-ON novelty working group. The levels in this novelty hierarchy are objects, agents, actions, relations, interactions, environments, goals, and events \cite{Doctor2022}. Table \ref{Tab:Novelty Hierarchy} provides a description of the novelty hierarchy levels and representative novelties we have designed in NovPhy. In Section \ref{sec:novelty_theories}, we have a detailed discussion of our desiderata on how we designed novelties and novel tasks in a physical environment such that it allows us a comprehensive agent evaluation.  

\subsection{Novelty-centric Domains}

The availability of novelty-centric domains that facilitate agents to evaluate and compare agents' performance is a critical factor that contributes to the advancement of open-world learning. Several testbeds/ frameworks/ benchmarks have been introduced to facilitate OWL by considering novelties as first-class citizens. This section describes related research on domains for OWL and situates NovPhy within them.  

GNOME is a novelty-centric simulator tool that facilitates developing and evaluating AI systems in multi-agent environments such as strategic board games \cite{MonopolyNoveltyGenerator}.
GNOME is applied to the popular board game monopoly to inject novelties from the first three levels of the novelty hierarchy described above. Example novelties from GNOME are adding more dice to the game, shuffling the order of slots on the board, etc.  

NovGrid \cite{NovGrid} and NovelGridworlds \cite{NovelGridworlds} are two frameworks developed for grid environments. NovGrid is a toolkit developed for novelty generation in MiniGrid environment \cite{MiniGrid}. NovGrid extends the MiniGrid environment by enhancing the functionality of existing objects and enabling agents to detect and adapt to novelty. 
NovelGridworlds implements a Minecraft \cite{Minecraft} inspired grid world to study novelties. 
In NovGrid, authors have introduced novelties based on the first two novelty hierarchy levels while in NovelGridworlds authors introduce novelties based on the first three levels of the novelty hierarchy.

NovelCraft \cite{NovelCraft} is a benchmark dataset for novelty detection and adaptation based on a modified version of Minecraft. NovelCraft is a 3D environment where an agent needs to select a sequence of actions to turn available resources into a pogo stick \cite{Polycraft}. In this work, authors have established agent performance in novelties based on the first level of the novelty hierarchy: objects \cite{NovelCraft}.  

None of the above-mentioned novelty domains is based on physics, which is an important characteristic to consider when developing agents that work in the real world.
Cartpole with novelty \cite{CartPole} and Science Birds Novelty \cite{SBFramework} are two physics based domains developed for novelty detection and adaptation. In Cartpole, an agent must keep the pole balanced by pushing the cart forward, backward, left, or right. In Science Birds Novelty, which is based on the physics game Angry Birds \cite{Rovio}, an agent needs to destroy the pigs by shooting birds from a slingshot. NovPhy uses the Science Birds framework \cite{SBFramework} and introduces various realistic physical reasoning tasks, as our focus is on evaluating agents' physical reasoning ability under the influence of novelty. As mentioned in Section \ref{sub-sec:physical_reasoning_research}, we use five physical reasoning scenarios and combine them with novelties from all eight levels of the novelty hierarchy. % We discuss NovPhy in detail in Section \ref{sec:NovPhy_Testbed}. 

%Authors propose example novelties in the first three levels of the novelty hierarchy. NovPhy also proposes a testbed in Angry Birds but in contrast to Science Birds Novelty, NovPhy combines physical scenarios of Phy-Q and explicitly focuses on physical reasoning under the influence of novelty. Moreover, NovPhy includes novelties from all eight levels of the novelty hierarchy along with specifically designed normal-novel task pairs. We discuss NovPhy in detail in Section \ref{sec:NovPhy_Testbed}. 

\section{Designing Novel Tasks and Agent Evaluations in Open-world Physical Environments}
\label{sec:novelty_theories}

In this section, we discuss the desiderata we satisfied when designing tasks in our testbed and when setting up agent evaluations for those tasks. We term the tasks that have novelties in them as novel tasks and the tasks without novelties (i.e., tasks in the normal environment) as normal tasks. In our testbed, we consider a constrained environment setting where an agent's action only applies to the state of the environment when the action is taken, and that action determines the subsequent states of the environment (i.e., the agent cannot control the subsequent states of the environment after the action is taken).  

\subsection{Designing Novel Tasks}
\label{sec:novelty_design_theory}

An agent working in a physical environment has to encounter different physical scenarios such as applying a force to an object, moving an object from one place to another, avoiding an obstacle in its path, etc. We introduce novelties to these scenarios that an agent could encounter in the physical environment. 

%Due to the multiple ways a novelty can occur in the environment, the novelty can either make a difference to the actions that the agent has to take to perform the task or sometimes the agent could still perform the task without changing the original actions (refer to the literature about impactful vs nuisance novelties etc boults, Weibull). In this work, we consider only the novelties that make an impact on the task performance such that the agent has to adjust its actions to successfully perform the task. 

When designing novel tasks, \textbf{we ensure that the agent has to work under the effects of the novelty to solve the task}. In other words, in the novel tasks, there are no solutions to the task that skip the effects of the novelty.
%there are no any paths that the agent can solve the task by skipping the effects of the novelty. 
%For example, consider a scenario where the agent has to push an object on the floor from one place to another, and when the novelty occurs the floor becomes slippery. 
%In this scenario, the agent is required to interact with the novel element (the floor) in order to complete the task.
%Here, we eliminate paths where the agent can skip the slippery floor and push the object normally as it could do in the normal task. 
%Therefore, the only way to successfully perform this task is to adapt to push the object on the slippery floor. 
For example, consider a bowling game where the player has to roll a ball on a surface to knock over the pins. Assume that when novelty occurs, the surface on which the ball rolls becomes slippery. In this scenario, the player is required to interact with the novel element (the surface) in order to complete the task. 
Therefore, the only way to successfully perform this task is to adapt to roll the ball on the slippery surface.

To ensure that the agent has to go through the novelty when completing a task, when designing novelties we consider the physical interactions in the solution of a physical scenario. We categorize these physical interactions into three phases: the initial phase, the middle phase, and the final phase. We only design novelties that at least affect one of these three interaction phases, to guarantee that the agent has to work along with the effects of the novelty.
%and guarantee that the effect of the novelty affects at least one of the three phases when designing novel tasks. 
The initial phase includes the immediate impact on the objects by the agent's action, the middle phase includes the consequences of the immediate impact of the action, and the final phase includes the interactions that complete the task. In all the phases, the objects that are involved in those physical interactions are also considered.
% When designing a novel task for a scenario, to ensure the agent has to work along with the effects of the novelty, we only design novelties that at least affect one of these three interaction phases. 

%For example, consider a bowling game where the player has to roll a ball on a surface to knock over the pins. In this scenario, the possible physical interactions are, the player throws the ball giving a starting velocity to the ball, the ball rolls on the surface, and the ball hits the pins knocking them down. 
For example, consider the previously mentioned bowling game. In this scenario, the possible physical interactions are, the player throws the ball giving a starting velocity to the ball, the ball rolls on the surface, and the ball hits the pins knocking them down. In this instance, the initial interaction phase includes the ball and the velocity the player applies to the ball. The middle phase includes the ball and the surface, and the rolling movement of the ball. The final phase includes the ball and the pins, and the collision between the ball and the pins. Therefore, here, the novelty can be applied to the ball, to the surface, to the pins, or to something that affects the rolling of the ball and collision of the ball and pins, in order to make sure the agent has to bowl under the effect of the novelty.

\subsection{Designing Agent Performance Evaluations}
\label{sec:evaluation_theory}

In OWL, agent evaluations are conducted to measure two capabilities of agents: the novelty detection capability and the novelty adaptation capability \cite{Pinto2022, Jafarzadeh2020, Peng2021}. The novelty detection measures evaluate whether an agent could successfully detect a novelty in the environment and the novelty adaptation measures evaluate whether an agent could successfully perform the task in the presence of the novelty. Novelty adaptation is generally more emphasized than novelty detection, as performing the task under the influence of a novelty is more important than merely detecting something is novel for an agent that actually works in an open-world environment. In this work, we also prioritize evaluating agents' novelty adaptation performance.

The standard agent evaluation setup in OWL consists of a set of trials, where each trial consists of a sequence of normal tasks followed by a sequence of novel tasks \cite{Pinto2022, Peng2021,Muhammad2021}. We follow the same setup in NovPhy.
We believe that, in order to measure whether an agent genuinely adapts to a novelty, \textbf{there should be a change in the solution path of the tasks when moving from the normal tasks to the novel tasks}. In an abstract form, we define a solution path as a sequence of physical interactions including the associated objects, initiated by an agent's action, that leads to solving the task. 
When designing a novel task for a physical scenario we also design a corresponding normal task that has a different solution path compared to the novel task. Then, when defining the trials for the evaluation, we select these normal and novel task pairs to guarantee that there is a solution path change from normal to novel tasks.

In this setting, since there is an obvious change in the solution path from the normal tasks to the novel tasks, novelty detection becomes trivial. To detect whether there is a novelty, the agent has to simply monitor whether the solution in the normal tasks is no longer working. To avoid this consequence, one could define separate trials that are only used to evaluate the novelty detection performance by including the tasks that have the same solution path in both normal and novel tasks. In this work, we do not include such trials in the evaluation as our main focus is evaluating the novelty adaptation performance of the agents.

We consider another desideratum when evaluating the performance of an agent in an open-world physical environment. From the perspective of OWL, we believe that, \textbf{if an agent can truly perform under a novelty, the agent should be able to perform with that novelty when the novelty is applied to different physical scenarios}. Also, from the perspective of physical reasoning, we believe that, \textbf{if an agent is robust at performing in a physical scenario, that agent should be able to perform in that scenario under the effect of different novelties}. To achieve these two evaluation setups, we designed the  novelties orthogonally to the physical scenarios, such that the same novelty can be applied to multiple scenarios and the same scenario can get affected by multiple novelties.

\section{NovPhy Testbed}
\label{sec:NovPhy_Testbed}
In this section, we introduce our testbed, the physical scenarios we consider, the novelties we designed, the tasks in the testbed, and explain the evaluation settings we have used in the testbed.

\subsection{Introduction to NovPhy}
Based on different physical scenarios and novelties, we develop the novelty-centric testbed NovPhy using Angry Birds. We use an open-source research clone of the game developed in Unity called Science Birds \cite{Ferreira2014}. Our testbed is adapted from a framework that can be used to inject novelties and conduct agent evaluations, developed from Science Birds \cite{SBFramework}.

In Angry Birds, the goal of the player is to kill all the pigs in the game level by shooting a given number of birds from a slingshot. In the normal game environment, along with the slingshot, the player will encounter four types of game objects: birds, pigs, blocks, and platforms. Additionally, we have also introduced an external agent to the normal environment called Air Turbulence that applies an upward force to any object that travels through it. An external agent is an agent with goal-oriented behaviour and having external agents enables situations that hinder or support the action that a player takes.
%Introducing the Air Turbulence external agent is also useful when defining some novelties according to the novelty hierarchy we use.
In the game, birds, pigs, and blocks are dynamic objects, which behave according to Newtonian physics, while platforms are static and are not affected by external forces. The dynamic objects have health points that get reduced in the collisions and they get destroyed when the health points become zero.
The blocks have 12 variations in shape and they are made of one of 3 types of materials: wood, stone, and ice. There are three types of pigs with three different sizes; the larger the size the higher the health points are. All objects in NovPhy are shown in Appendix Figure \ref{appendix_fig:all_normal_objects}.

% In NovPhy, we allow the same game playing procedure in the Phy-Q testbed for agents. 
An agent playing the game can request the current game state anytime as a screenshot and/or as a symbolic representation. The screenshot is a 480x640 coloured image of the game. The symbolic representation is in JSON format and contains all the objects in the screenshot. Here, an object is represented as a polygon of ordered vertices along with the percentages of its 8-bit quantized colours. The full world state is not provided to the agent such as the exact positions of the objects and their physical parameters such as mass, coefficient of friction, etc. as they are not directly observable in the real world.
The action of an agent is the release point of the bird relative to the slingshot. Sometimes when there is more than one bird in the game level the agent takes a sequence of actions. We provide a trajectory planner that can be used to calculate the release point of the bird to reach a target, under the normal settings in the environment, when the target point is given. The agent passes the game level if it destroys all the pigs with the provided number of birds or fails if not.

The physical scenarios and the novelties we consider in this testbed are discussed in the next subsections \ref{sec:scenarios_in_NovPhy} and \ref{sec:novelties_in_NovPhy} respectively.
\subsection{Physical Scenarios in NovPhy}
\label{sec:scenarios_in_NovPhy}

%The physical scenarios considered in NovPhy are obtained from the scenarios in the physical reasoning benchmark Phy-Q (ref). 
As discussed in Section \ref{sec:background_and_related_work}, we use the first five physical scenarios introduced in Phy-Q as they are the most basic and frequently encountered scenarios in a physical environment. The scenarios include applying forces directly on target objects - the effect of a single force and the effect of multiple forces \cite{Force}. The motion-related scenarios: rolling, falling, and sliding, inspired by the physical reasoning capabilities developed in human infancy \cite{RollingFalling}. The five scenarios and the corresponding physical rules that can be used to achieve the goal of the associated tasks are:

\begin{enumerate}
    \item Single force: Target objects have to be destroyed with a single force.
    \item Multiple forces: Target objects have to be destroyed with multiple forces.
    \item Rolling: Circular objects have to be rolled along a surface to a target.
    \item Falling: Objects have to fall onto a target.
    \item Sliding: Non-circular objects have to be slid along a surface to a target.
\end{enumerate}

\begin{center}
\begin{table}
    \begin{tabular}{ |p{2.8cm}|p{4cm}|p{5cm}| } 
        
        \hline
        Novelty Level & Description & Representative Novelty \\
        \hline
        1. Objects & New classes, attributes, or representations of non-volitional entities. & A new pig/block that has a different colour to the normal pigs/blocks.\\
        \hline
        
        2. Agents & New classes, attributes, or representations of volitional entities. & A novel external agent, Fan, that blows air (horizontally from left to right) affecting the moving path of objects.\\
        \hline
        
        3. Actions & New classes, attributes, or representations of external agent behavior. & The non-novel external agent, Air Turbulence, increases the magnitude of its upward force.\\
        \hline
        
        4. Interactions & New classes, attributes, or representations of dynamic properties of behaviors impacting multiple entities. &  Existing circular wood object now has magnetic properties: repels objects of its type and attracts other object types.\\
        \hline
        
        5. Relations & New classes, attributes, or representations of static properties of the relationships between multiple entities. & The slingshot which is at the left side of the tasks is now at the right side of the tasks (i.e., the spatial relationship between the slingshot and other objects is changed).\\
        \hline
        
        6. Environments & New classes, attributes, or representations of elements independent of specific entities. & The gravity in the environment is now inverted, which affects the behaviour of the dynamic objects.\\
        \hline
        
        7. Goals & New classes, attributes, or representations of external agent objectives. & The non-novel external agent, Air Turbulence, changes its goal from pushing objects up to pushing objects down.\\
        \hline
        
        8. Events & New classes, attributes, or representations of series of state changes. & When the first bird is dead, a storm occurs that affects the motion of the objects (by applying a force to the right direction).\\
        
        %8. Events & New classes, attributes, or representations of series of state changes that are not the direct result of volitional action by an external agent or the SAIL-ON agent. & When the first bird is dead, a storm comes affecting the motion of the objects.\\
        \hline
        
        \end{tabular}
    \captionof{table}{SAIL-ON Open-world novelty hierarchy \cite{Doctor2022} and the representative novelties in NovPhy for each hierarchy level.} 
    \label{Tab:Novelty Hierarchy}

\end{table}
\end{center}

\subsection{Novelties used in NovPhy}
\label{sec:novelties_in_NovPhy}

We design a representative novelty for each hierarchy level in the open-world novelty hierarchy proposed by the SAIL-ON program novelty working group. The novelty hierarchy consists of eight novelty levels that cover a wide range of novelty types that could occur in an open-world environment. Table \ref{Tab:Novelty Hierarchy} shows the open-world novelty hierarchy and descriptions of representative novelties in NovPhy. Appendix Figure \ref{appendix_fig:all_novel_objects} shows the new game objects that are introduced to the game for the novelties associated with a game object.

\subsection{Task Templates}

\begin{figure}[h!]
  \centering
  \begin{subfigure}[b]{0.26\columnwidth}
    \includegraphics[width=\linewidth]{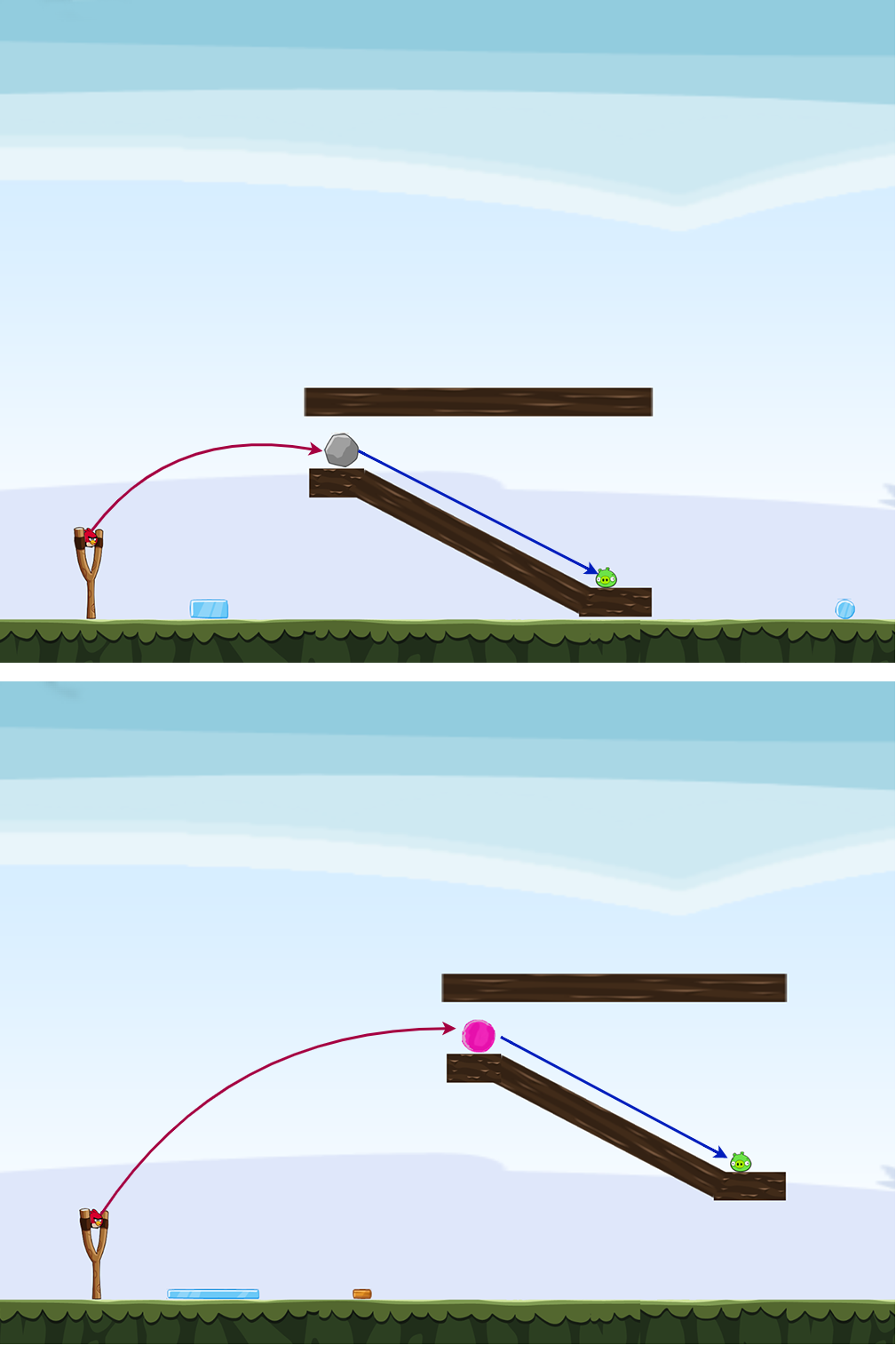}
    \caption{Objects}
  \end{subfigure}
  \begin{subfigure}[b]{0.26\columnwidth}
    \includegraphics[width=\linewidth]{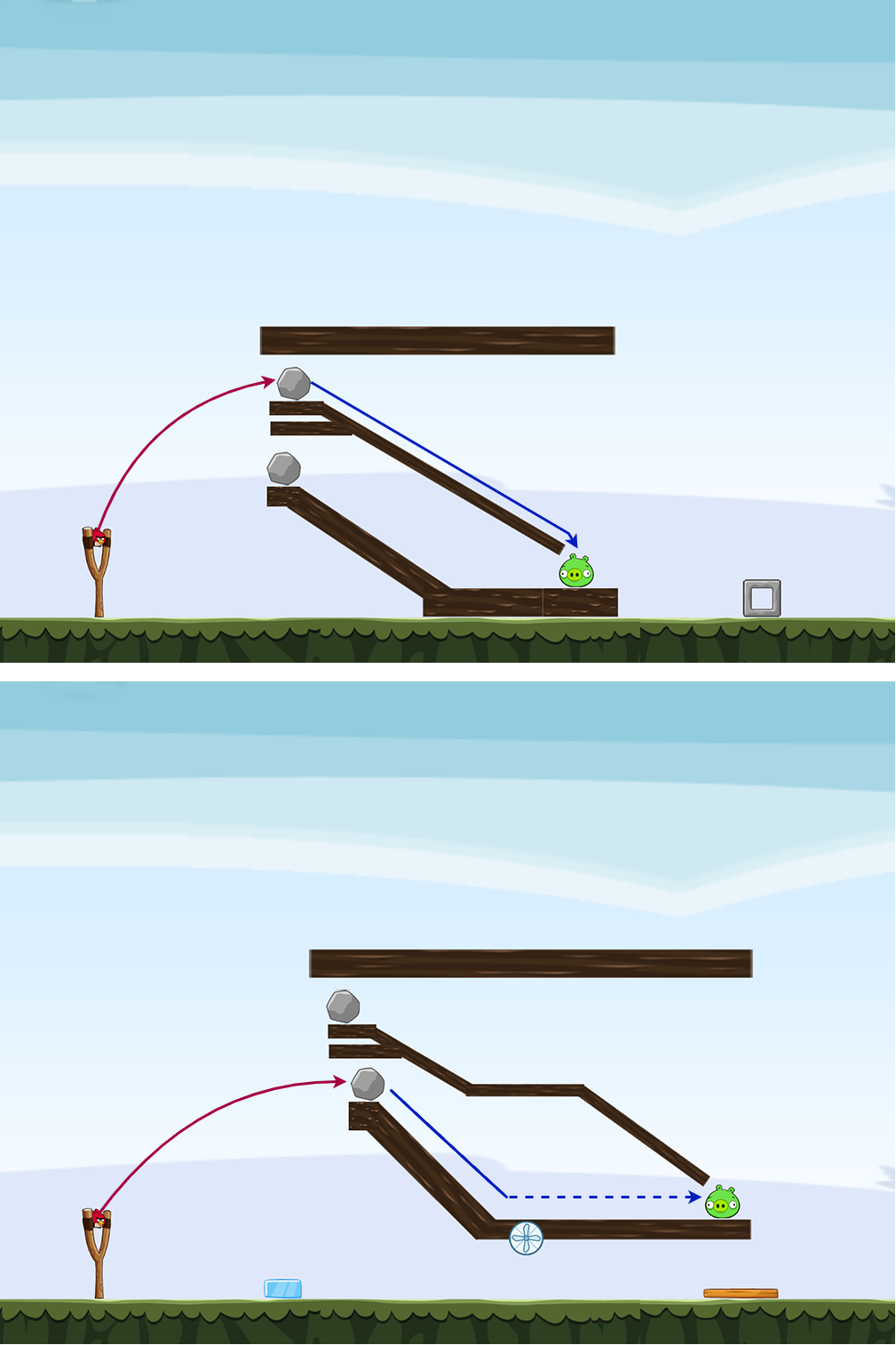}
    \caption{Agents}
  \end{subfigure}
    \begin{subfigure}[b]{0.26\columnwidth}
    \includegraphics[width=\linewidth]{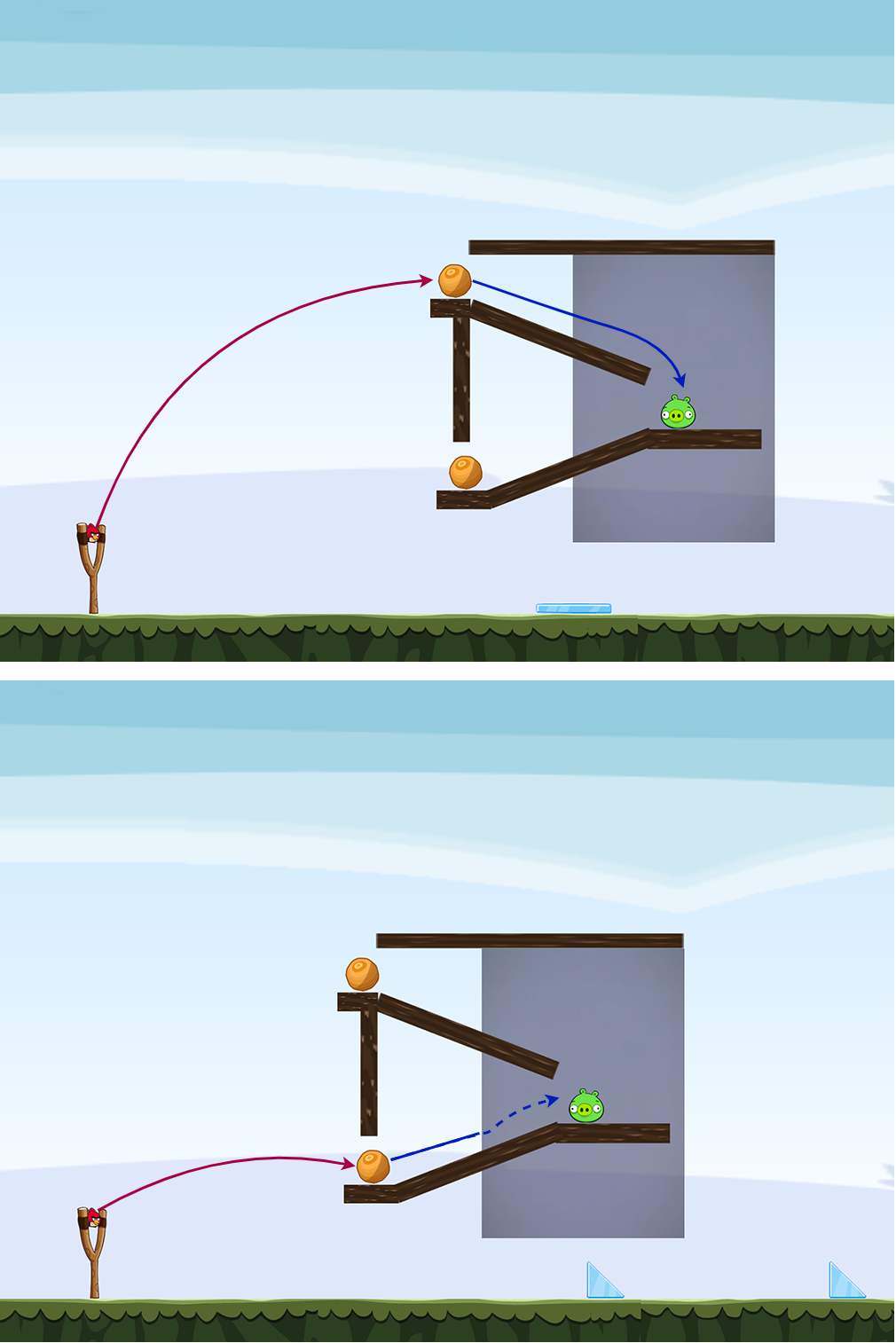}
    \caption{Actions}
  \end{subfigure}
  \begin{subfigure}[b]{0.26\columnwidth}
    \includegraphics[width=\linewidth]{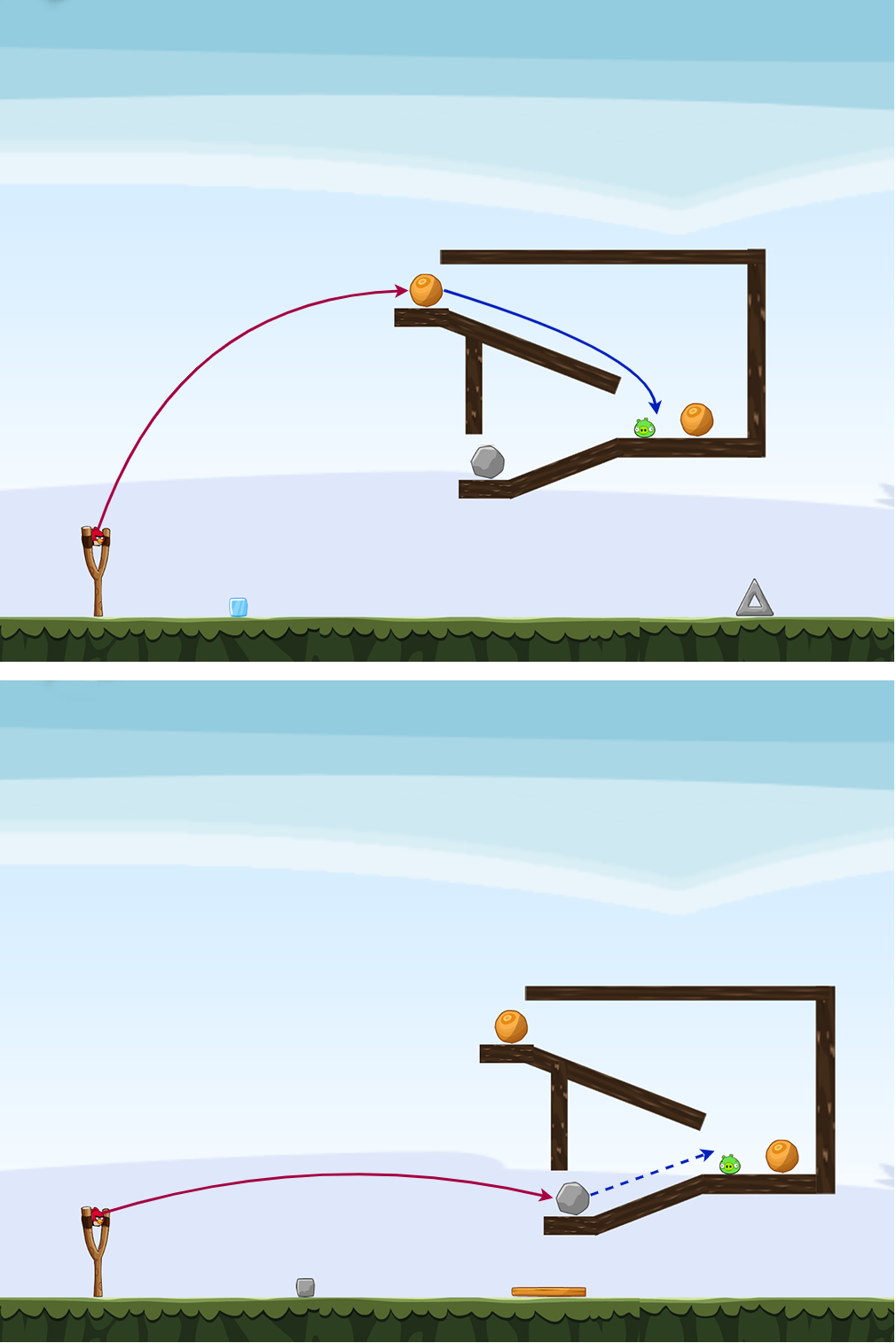}
    \caption{Interactions}
  \end{subfigure}
    \begin{subfigure}[b]{0.26\columnwidth}
    \includegraphics[width=\linewidth]{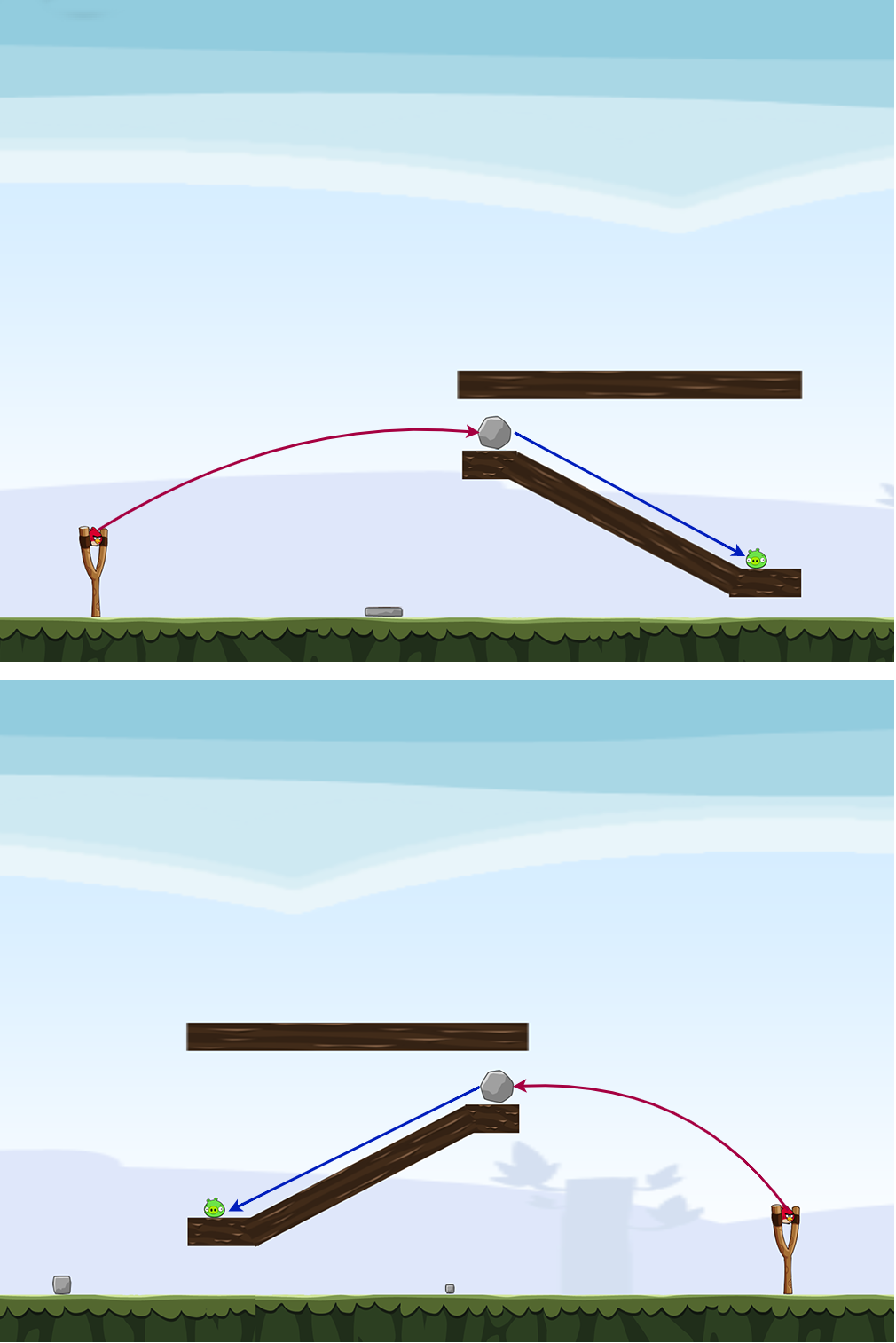}
    \caption{Relations}
  \end{subfigure}
  \begin{subfigure}[b]{0.26\columnwidth}
    \includegraphics[width=\linewidth]{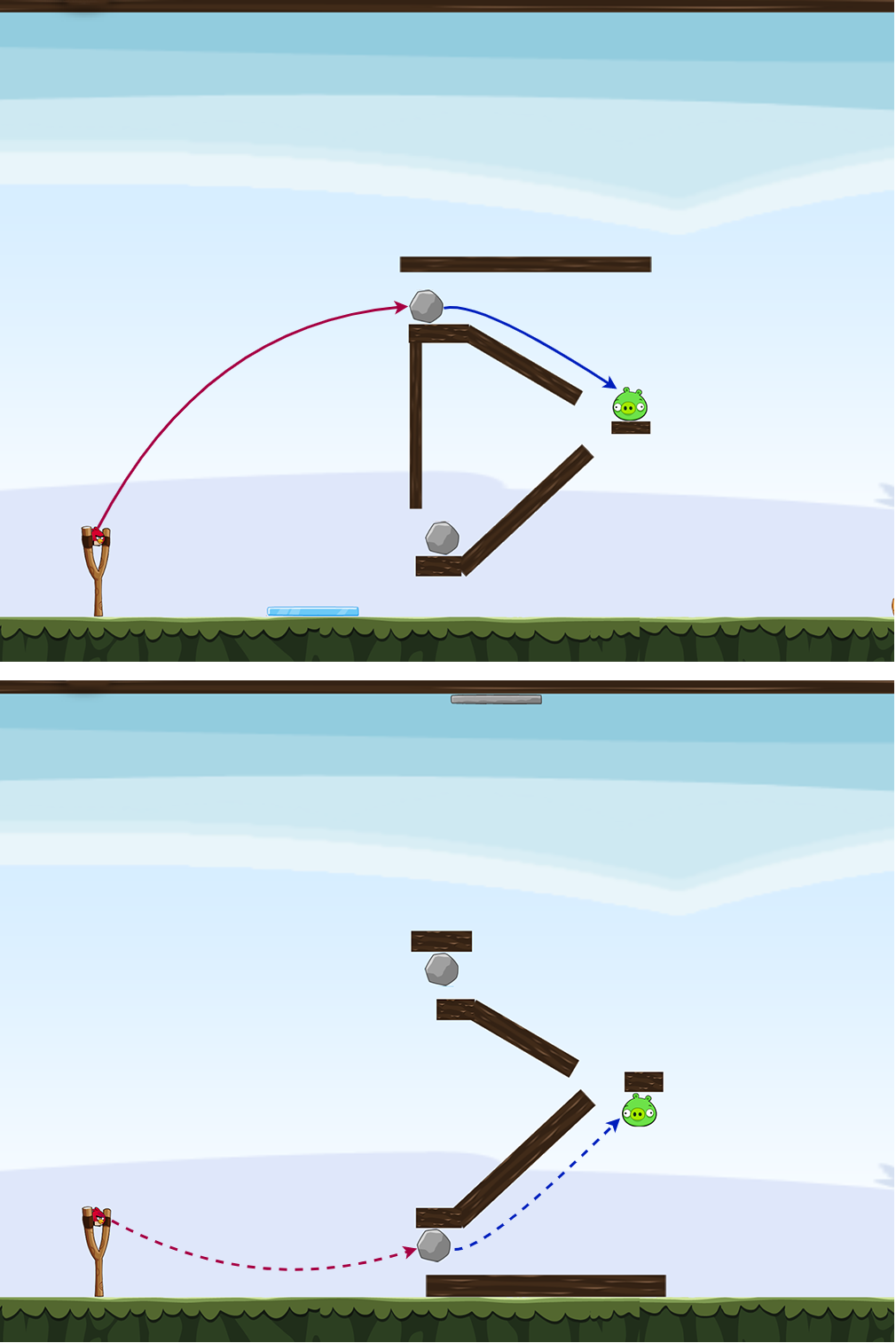}
    \caption{Environments}
  \end{subfigure}
  \begin{subfigure}[b]{0.26\columnwidth}
    \includegraphics[width=\linewidth]{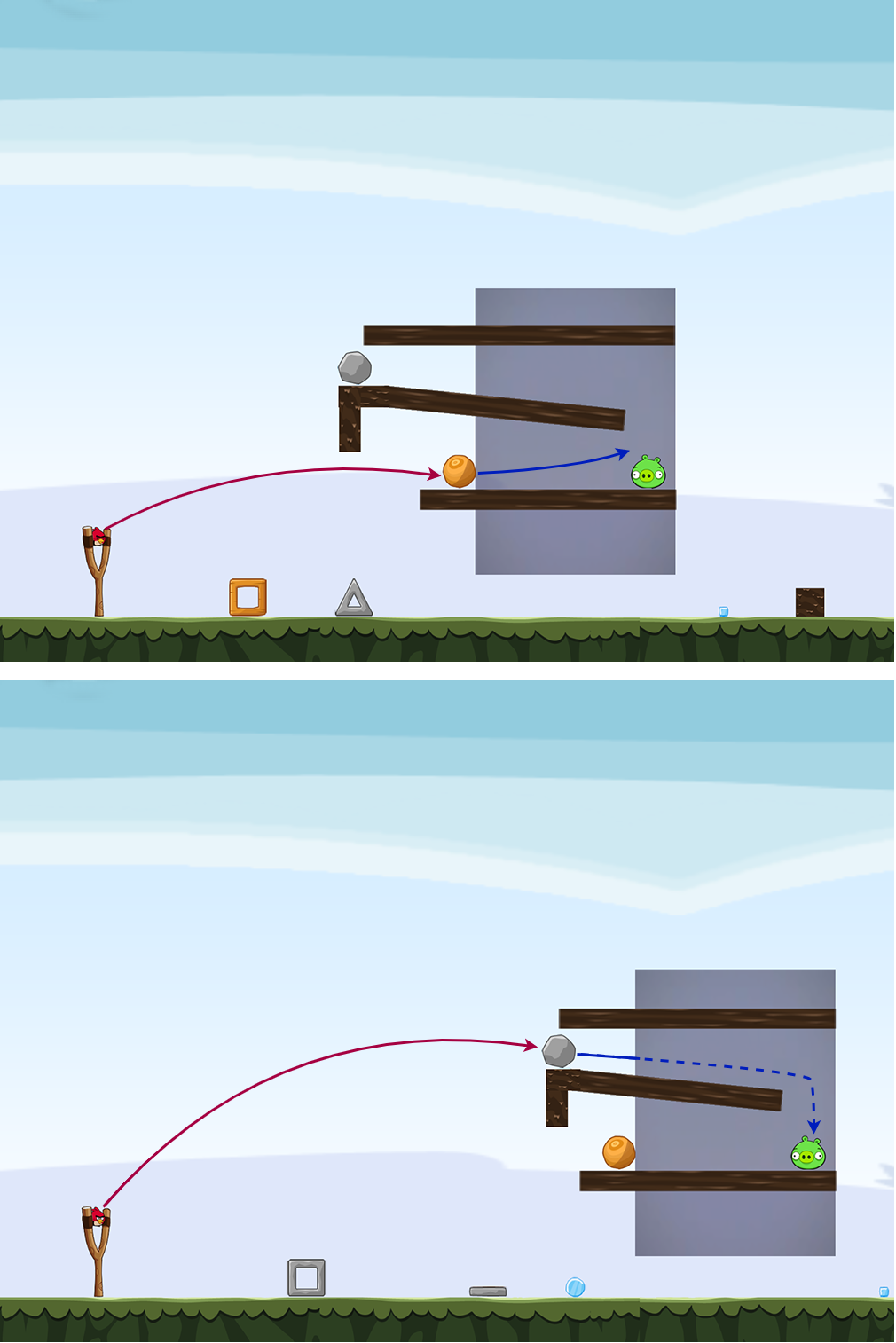}
    \caption{Goals}
  \end{subfigure}  
  \begin{subfigure}[b]{0.52\columnwidth}
    \includegraphics[width=\linewidth]{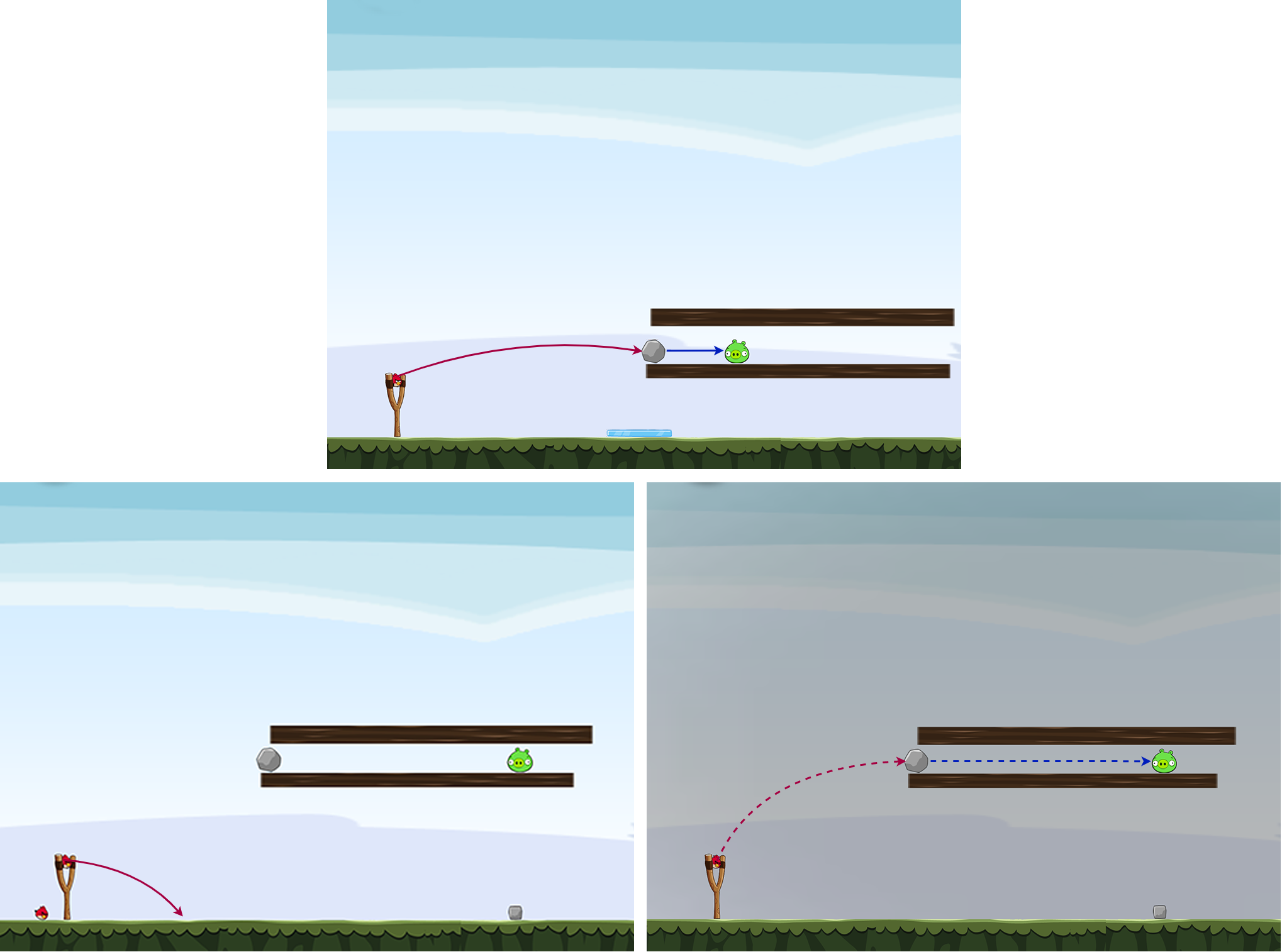}
    \caption{Events}
  \end{subfigure}  
\caption{Example tasks of the rolling scenario for the eight novelties. In each subfigure, the top figure is the normal task and the bottom figure is the corresponding task with the novelty. The arrows show the trajectories of the objects when the solution is executed.}
\label{eight_novelties_on_rolling}
\end{figure}
%  (in red: bird trajectory, in blue: other objects' trajectories when the bird is hit). Dotted arrows represent the trajectories affected by the novelty. Example tasks of the rolling scenario with the eight novelties applied to them. In each subfigure, the left figure is the normal task and the right figure is the corresponding novel task with the novelty applied. (a) objects: the new pink coloured object has to be rolled, (b) agents: the force of the new Fan agent has to be used to roll the object further, (c) actions: the magnitude of the upward force of the Air Turbulence agent is increased, which helps to roll the ball upwards in the ramp, (d) interactions: the circular wood (brown) objects have magnetic properties, thus repels the object of the same type and attracts the stone (grey) object helping to roll the stone upwards in the ramp, (e) relations: the slingshot is now placed in the right side of the task, instead of shooting left to right now the shooting has to be done from right to left to roll the object to the left direction, (f) environments: the inverted gravity makes the objects attract towards the sky, hence objects can be rolled upwards in ramps (g) goals: the goal changed Air Turbulence agent (from the goal of pushing objects up to pushing objects down) hinders rolling on flat surfaces while helps rolling on inclined surfaces  (h) events: when the first bird is dead, the storm occurs which applies a force to the right direction of the moving objects, hence shooting the second bird to the circular object makes the object to roll further to reach the pig.

A task template defines a set of related tasks that can be created by varying task template parameters such as the locations of the objects. We design task templates by applying each of the eight novelties to each of the five physical scenarios discussed in the above sections. For example, for the rolling scenario we design templates that require rolling an object when: it is a new object, there is an effect from the Fan, the Air Turbulence agent changes the magnitude of the upward force, the slingshot is in the right side of the object, there is an effect from a magnetic field, the gravity is inverted, Air Turbulence pushes the object down, and there is a storm after shooting the first bird. We term a physical scenario with a novelty applied as a novelty-scenario. Since we consider eight novelties and five scenarios, we have 40 novelty-scenarios. 

When designing novel task templates for NovPhy, we follow the desiderata discussed in Section \ref{sec:novelty_theories}. For each novelty-scenario, we designed and handcrafted a novel task template. A novel task template is a task template where a novelty is present.
In the design, we ensure the novel tasks meet our desideratum, that it is necessary to work under the effects of the novelty to complete the task. Appendix Table \ref{Tab:Interaction Affected by Novelty} contains more details on how this desideratum is satisfied when designing the tasks, considering the physical interaction phases of the solution that are affected when the novelty is introduced. Then, for each novel task template, we design a corresponding normal task template as well. This is done by removing the novelty from the novel task template and adjusting the template accordingly to make it solvable without the novelty. Considering our next desideratum, when designing the normal task templates, we also ensure that there is a solution path change from the normal task to the novel task.

In the task template design, we also ensure that all the templates of a given scenario can be solved by the associated physical rule of that scenario discussed in Section \ref{sec:scenarios_in_NovPhy}. Figure \ref{eight_novelties_on_rolling} shows how the eight novelties are applied to the tasks of the rolling scenario. In each subfigure,
the top figure is the normal task and the bottom figure is the corresponding task with the novelty. The arrows show the trajectories of the objects when the solution is executed (in red: bird's trajectory, in blue: other objects' trajectories when the bird is hit). Dotted arrows represent the trajectories affected by the novelty. To solve the novel task shown in each subfigure: 
\begin{enumerate}[label=(\alph*)]
    \item objects: the new pink coloured object has to be rolled.
    \item agents: the force of the new Fan agent has to be used to roll the object further.
    \item actions: the magnitude of the upward force of the Air Turbulence agent is increased, which helps to roll the ball upwards in the ramp.
    \item interactions: the circular wood (brown) objects have magnetic properties, thus repels the object of the same type and attracts the stone (grey) object helping to roll the stone upwards in the ramp.
    \item relations: the slingshot is now placed in the right side of the task, instead of shooting left to right now the shooting has to be done from right to left to roll the object to the left direction.
    \item environments: the inverted gravity makes the objects attract towards the sky, hence objects can be rolled upwards in ramps.
    \item goals: the goal changed Air Turbulence agent (from the goal of pushing objects up to pushing objects down) hinders rolling on flat surfaces while helps rolling on inclined surfaces.
    \item events (the novel template has two birds as shown in the first bottom figure and when the first bird dies it activates the storm as shown in the second bottom figure): when the first bird is wasted, the storm occurs which applies a force to the right direction of the moving objects, hence shooting the second bird to the circular object makes the object to roll further to reach the pig.    
\end{enumerate}
% (b) agents: the force of the new Fan agent has to be used to roll the object further,(c) actions: the magnitude of the upward force of the Air Turbulence agent is increased, which helps to roll the ball upwards in the ramp, (d) interactions: the circular wood (brown) objects have magnetic properties, thus repels the object of the same type and attracts the stone (grey) object helping to roll the stone upwards in the ramp, (e) relations: the slingshot is now placed in the right side of the task, instead of shooting left to right now the shooting has to be done from right to left to roll the object to the left direction, (f) environments: the inverted gravity makes the objects attract towards the sky, hence objects can be rolled upwards in ramps (g) goals: the goal changed Air Turbulence agent (from the goal of pushing objects up to pushing objects down) hinders rolling on flat surfaces while helps rolling on inclined surfaces  (h) events (the novel template has two birds as shown in the first bottom figure and when the first bird dies it activates the storm as shown in the second bottom figure): when the first bird is wasted, the storm occurs which applies a force to the right direction of the moving objects, hence shooting the second bird to the circular object makes the object to roll further to reach the pig.

\begin{figure}[t!]
  \centering
  \begin{subfigure}[b]{0.26\columnwidth}
    \includegraphics[width=\linewidth]{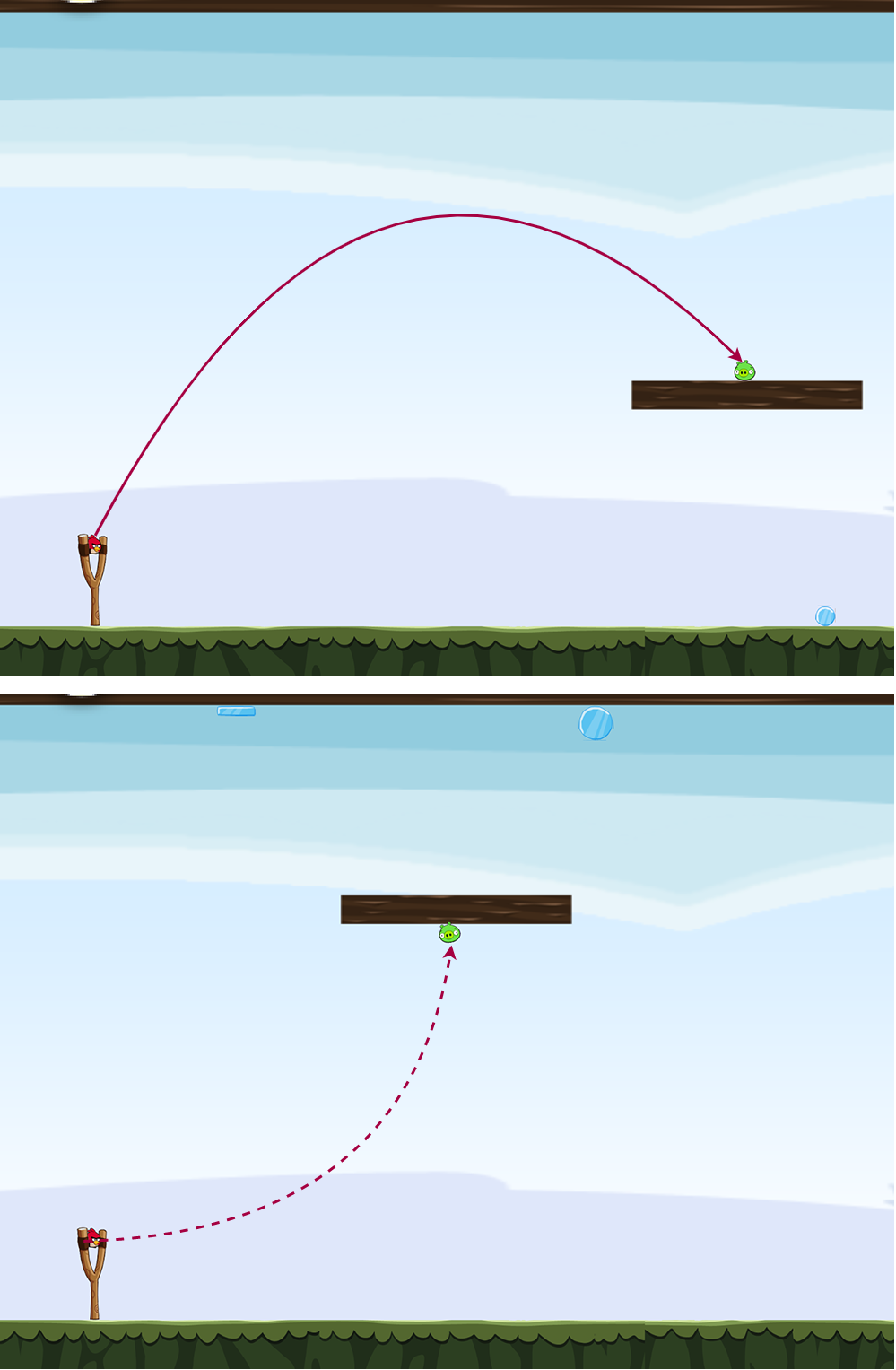}
    \caption{Single force}
  \end{subfigure}
  \begin{subfigure}[b]{0.26\columnwidth}
    \includegraphics[width=\linewidth]{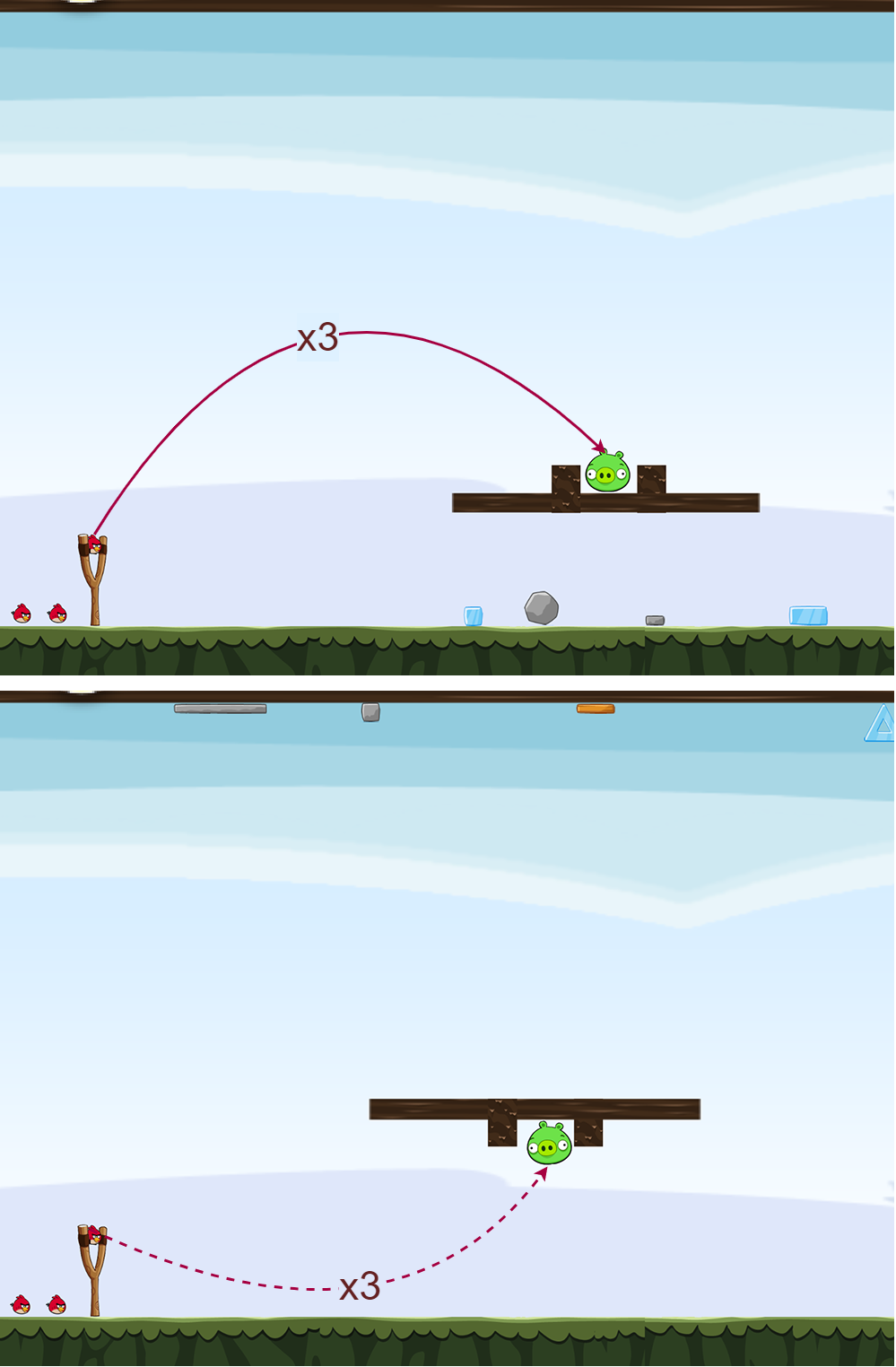}
    \caption{Multiple forces}
  \end{subfigure}
    \begin{subfigure}[b]{0.26\columnwidth}
    \includegraphics[width=\linewidth]{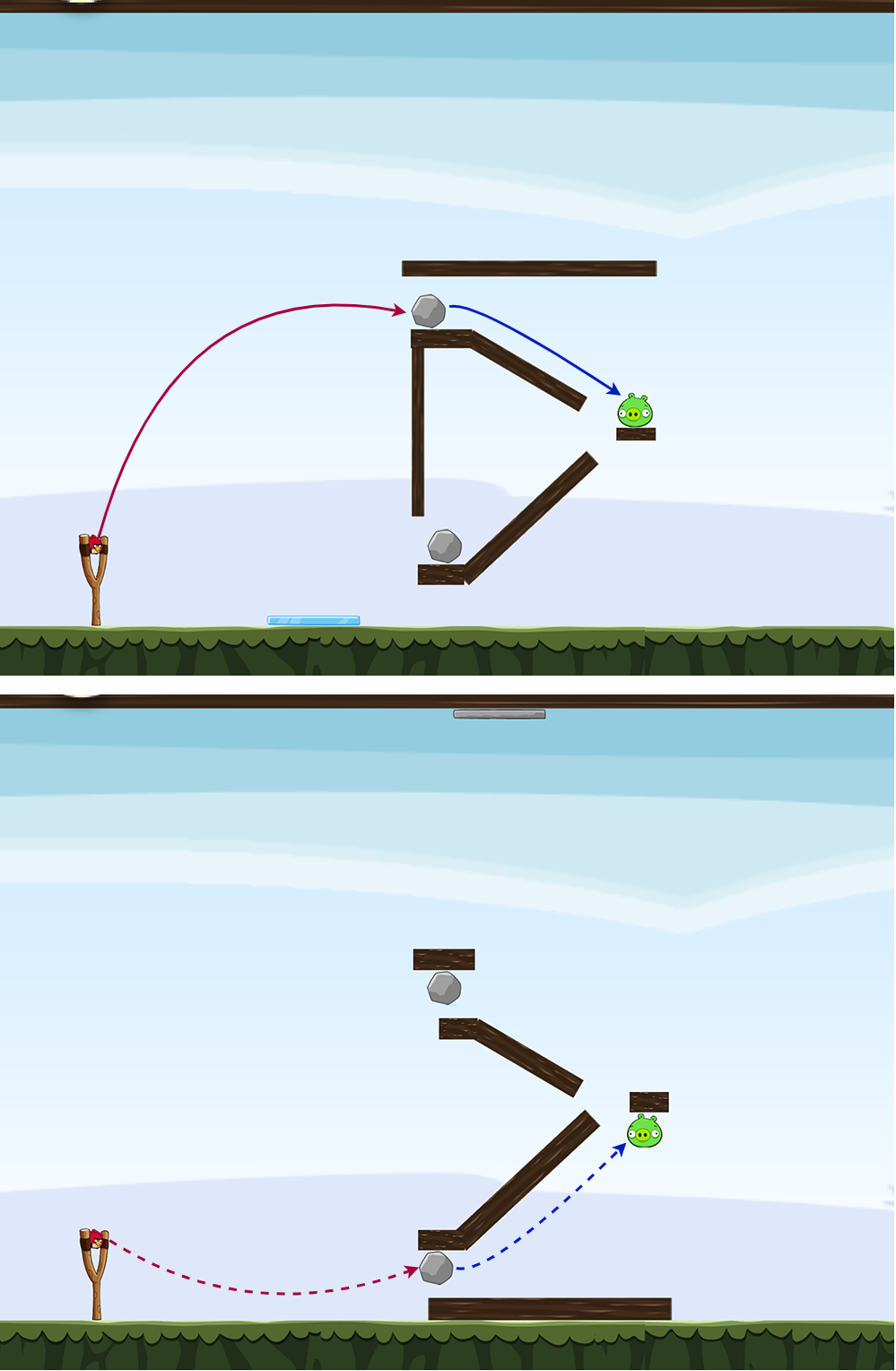}
    \caption{Rolling}
  \end{subfigure}
  \begin{subfigure}[b]{0.26\columnwidth}
    \includegraphics[width=\linewidth]{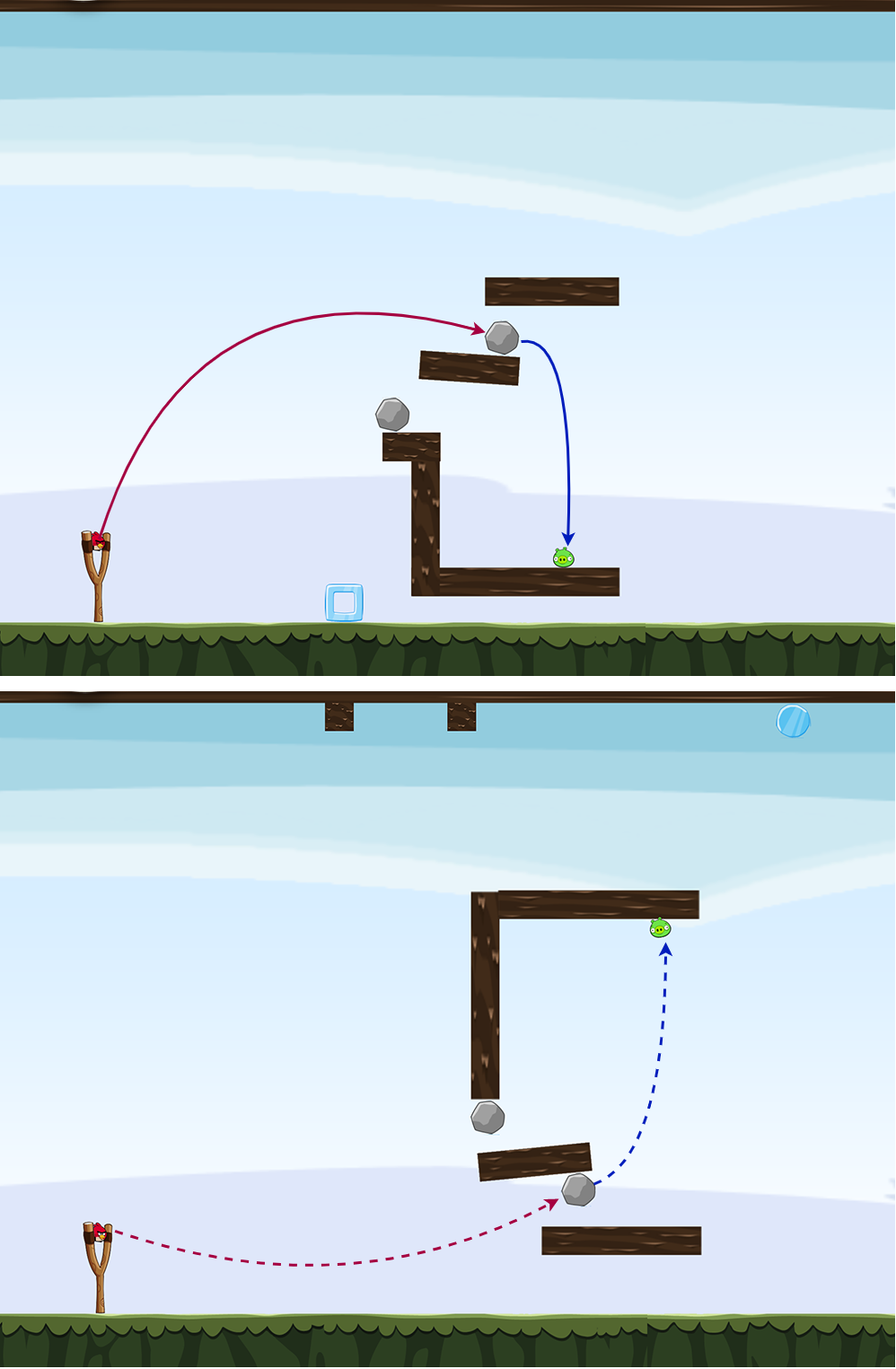}
    \caption{Falling}
  \end{subfigure}
    \begin{subfigure}[b]{0.26\columnwidth}
    \includegraphics[width=\linewidth]{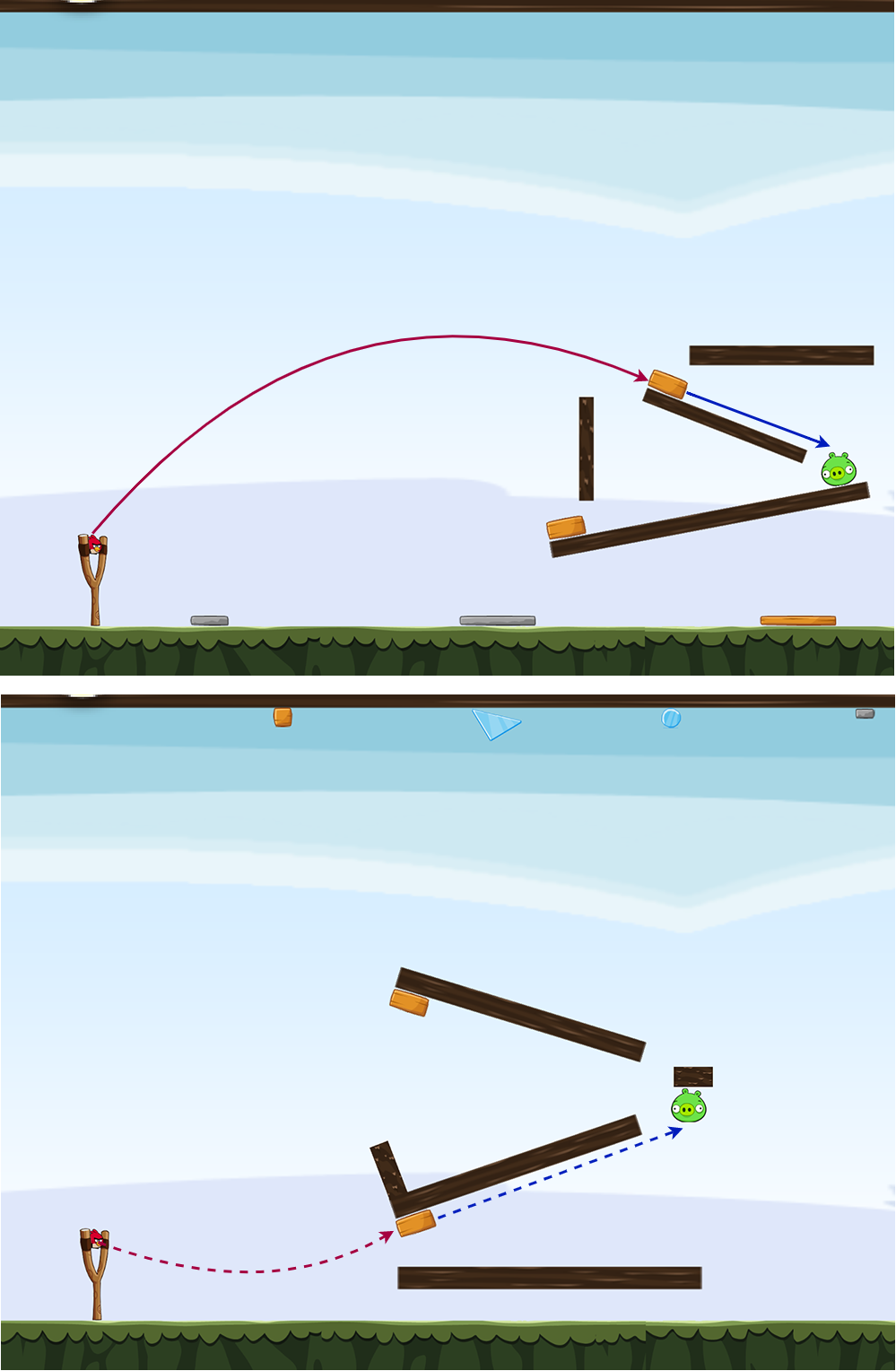}
    \caption{Sliding}
  \end{subfigure}
\caption{Example tasks of the inverted gravity novelty applied to the five physical scenarios. In each scenario, the top figure is the normal task and the bottom figure is the novel task. The arrows show the trajectories of the objects when the solution is executed. Solid arrows are the trajectories of the objects that are not affected by the novelty and the dotted arrows are the trajectories of the objects that are affected by the novelty. The inverted gravity has made all the dynamic objects attract towards the sky, hence they have been stopped from platforms to avoid flying away. The motion of the dynamic objects is also affected by the inverted gravity.}
\label{five_scenarios_on_inverse_gravity}
\end{figure}

Figure \ref{five_scenarios_on_inverse_gravity} shows how the inverted gravity novelty is applied across the tasks of the five physical scenarios. All 40 task templates in NovPhy can be found in \ref{Appendix: Section Tasks in NovPhy}.

\subsection{Task Generation}

We developed a task generator that can generate an unlimited number of tasks from a given template. The game levels generated from a task template are termed as the tasks of that template. When generating the tasks we vary the locations of the game objects within a suitable range in the level space. Additionally, some random game objects are added as distraction objects at random positions of the game level to trick the agents. In the generation, we ensure that the task can still be solved by the solution path in the original template. To achieve this, we define template specific constraints such as, which game objects are reachable/unreachable to the bird, which objects should be excluded from the distraction objects, what are the feasible regions to place specific objects, etc. These constraints are determined by the template designers and are input to the task generator.

We provide 350 generated tasks for each task template, but we also provide the task generator in case it is necessary to generate more tasks. Appendix Figure \ref{appendix_fig:task_variations_in_NovPhy} shows the variations of the tasks generated from the rolling scenario template with the inverse gravity novelty applied.

\subsection{Evaluation Protocol}

\begin{figure}[t]
\centering
\includegraphics[width=0.9\textwidth]{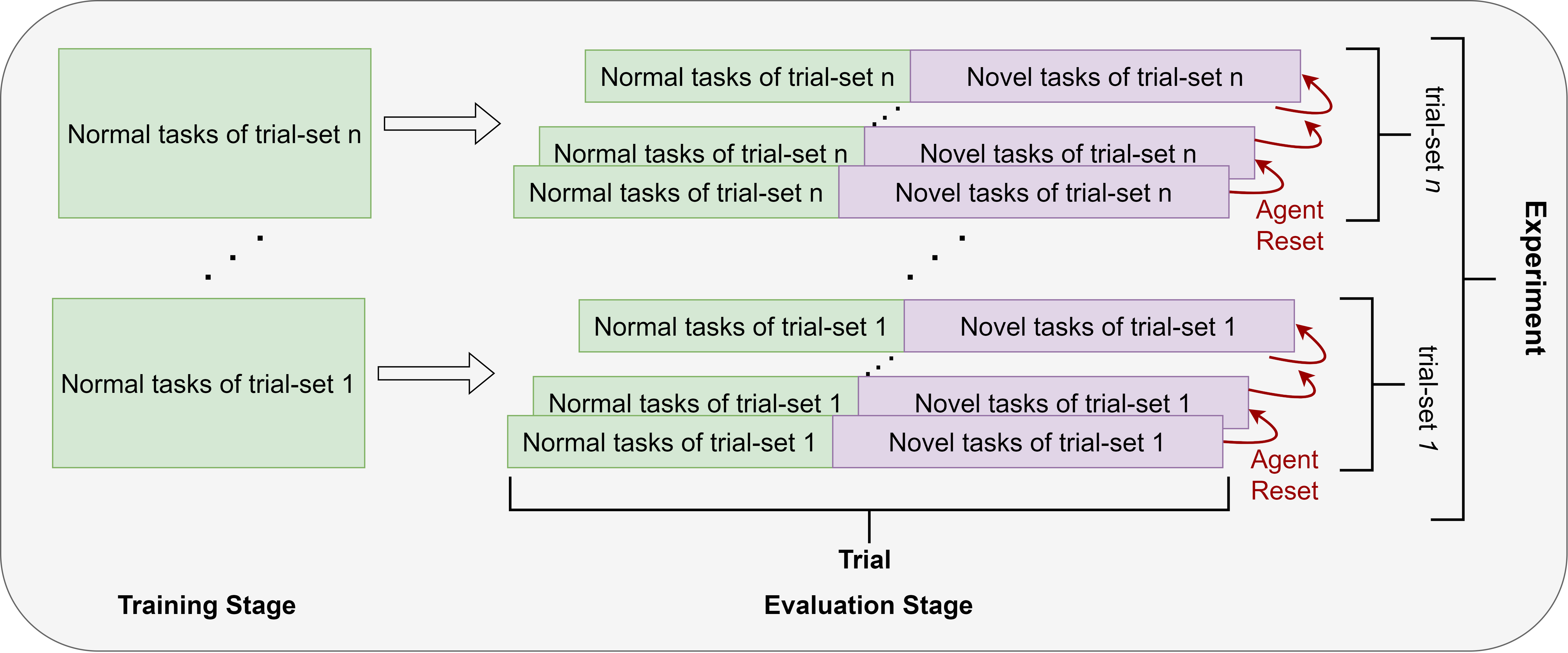}
\caption{The trial-based evaluation protocol used in NovPhy. The evaluation stage follows the training stage. An experiment contains {\em trial-sets} where a trial-set contains multiple trials. A trial contains a variable number of tasks drawn first from the normal task distribution and then from the novel task distribution. In a given trial-set the agent is only evaluated on a single novelty-scenario.}
\label{fig_evaluation_protocol}
\end{figure}

In NovPhy testbed, as in a standard OWL evaluation, we evaluate the novelty detection and novelty adaptation capabilities of the agents. In the novelty detection evaluation, we measure if an agent can detect if a novelty is present in the given task. In novelty adaptation evaluation, we measure the task performance of the agent in the presence of a novelty. Both novelty detection and novelty adaptation evaluation are done by using a trial setting \cite{Pinto2022}. A trial is a sequence of tasks, which starts from normal tasks and after a random number of normal tasks switches to novel tasks. After switching to novel tasks, all the subsequent tasks until the end of the trial are novel tasks. Figure \ref{fig_evaluation_protocol} shows how evaluations are done through the trial setup. A trial-set is a set of trials. A given trial set consists of trials of the same novelty-scenario. i.e., all the trials of a given trial-set only have normal and novel tasks from the same novelty-scenario. The agent is not allowed to share knowledge in between trials, i.e., at the start of each trial of a novelty-scenario, the agent is in the same initial state, as the agent was at the beginning of the evaluation.

To evaluate an agent on a novelty-scenario, the agent is trained on the normal task distribution of that novelty-scenario and tested using a trial-set of that novelty-scenario. This evaluation resembles the local generalization evaluation of the agents, i.e., the agent trains on the tasks of a normal template and is tested on the tasks of the same template and the corresponding novel template. Even though the testbed facilitates broad generalization evaluations (i.e., the agent trains on the tasks of a normal template of a physical scenario. Then the agent is tested on the tasks of a different template of the same scenario and its corresponding novel template), we use this local generalization based evaluation setup as it is proven that learning agents still struggle to generalize broadly in physical reasoning \cite{PhyQ}. Moreover, it is a precondition for agents to have a good normal task performance before adapting to novel tasks. 
%, and having good normal task performance is assumed when evaluating agents in novel tasks.

When the agent is playing a trial, the task completion status (whether the task is passed/failed) and the score the agent achieved at the end of the task are recorded. This data is used to calculate the novelty adaptation performance of the agent. For the novelty detection performance calculation, the agent has to inform in which task of the trial it believes the novelty occurred. 
% For the novelty detection, at the end of each task the agent has to provide a probability that it believes that there was a novelty in the task.

% NovPhy enables three main types of evaluations. The agent developers can choose which evaluation to perform on their agent. The three evaluations are summarized as follows.
In this work, we focus on evaluating agents in the below two evaluation settings.

\begin{enumerate}
    % \item Novelty Detection Evaluation: In this evaluation, the agent will only be evaluated on the novelty detection ability. 
    \item Novelty Informed Evaluation: In this evaluation, an agent will be informed when the novelty appears in each trial. The agent will only be evaluated on the novelty adaptation ability.
    \item Novelty Uninformed Evaluation: In this evaluation, an agent will be evaluated on both novelty detection and novelty adaptation. The agent will not be informed when the novelty appears in a given trial. 
\end{enumerate}

\subsection{Evaluation Measures}
%In NovPhy, to measure the novelty detection performance we use the standard OWL measures, and for the novelty adaptation, we use slightly changed standard OWL measures to fit into our domain.

%In NovPhy, we use the standard OWL evaluation measures that are used to measure the novelty detection and novelty adaptation performance. We use the asymptotic pass rate of the agent (Equation \ref{eq:ap}) and the area under the success curve. . Derived from them, we also introduce two new measures for open-world physical environments: 1) to measure an agent's performance on a novelty when the same novelty is applied to different physical reasoning scenarios (NPS - Novelty Performance under Scenarios) and 2) to measure the physical reasoning performance of an agent on a physical scenario when different novelties are applied to the scenario (SPN - Scenario Performance under Novelties).

For novelty detection, we use standard OWL measures used in the SAIL-ON program: the percentage of correctly detected trials (CDT) and the detection delay (DD) calculated using the average number of tasks taken to detect the novelty \cite{Pinto2022}. 
Consider a trial $t \in T$, where $T$ represents a set of trials for a novelty-scenario, $FP_t$ represents the number of normal tasks in trial $t$ where the agent incorrectly detected as a novel task. $TP_t$ represents the number of novel tasks in trial $t$ where the agent correctly detected as a novel task. A correctly detected trial is a trial where the agent detected novelty only after entering the novel task sequence. {\textit{CDT}} is defined in Equation \ref{eq:CDT}. CDT varies between 0 and 1 and 1 is the best result. 

\begin{equation} \label{eq:CDT}
\resizebox{.650\linewidth}{!}{$
CDT = \frac{1}{|T|} \sum_{t=1}^{|T|} \bigg\{ 
\begin{array}{ll}
      1, & \mbox{if \textit{$FP_t$ = 0} and \textit{$TP_t$ $\neq$ 0}}\\
      0, & \mbox{otherwise}
    \end{array}
$}
\end{equation}

{\textit{DD}} quantifies the delay in detection using the number of tasks required to correctly detect the novelty. {\textit{DD}} is defined in Equation \ref{eq: DD}. The lower the DD, the better the detection performance (in terms of timeliness), and the best possible DD is 1. 

\begin{equation} \label{eq: DD}
\resizebox{.650\linewidth}{!}{$
DD=\frac{1}{N_{cdt}} \sum_{t=1}^{|T|} \bigg\{ 
\begin{array}{ll}
      d_t, &\mbox{if \textit{$FP_t$=0} and \textit{$TP_t$ $\neq$0}}\\
      0, &\mbox{otherwise}
    \end{array}
$}
\end{equation}
where,
\begin{equation}
\resizebox{.550\linewidth}{!}{$
N_{cdt} = \sum_{t=1}^{|T|} \bigg\{ 
\begin{array}{ll}
      1, & \mbox{if \textit{$FP_t$=0} and \textit{$TP_t$ $\neq$0}}\\
      0, & \mbox{otherwise}
    \end{array}
$}
\end{equation}
and $d_{t}$ is the number of novel tasks taken until the agent informs a novelty detection in trial {\em t} (including the task that the agent detected novelty). 

To measure novelty adaptation performance, we use the area under the pass rate curve (success curve). First, in a given novelty-scenario, to measure the performance of the agent after adapting to novelty, we use the pass rate of the asymptotic tasks. We refer to this measure as the asymptotic performance ({\textit{AP}}). In equation \ref{eq:ap} for {\textit{AP}}, $n$ represents the length of the novel task sequence and $m$ represents the asymptotic length we consider. The asymptotic length can be adjusted based on the percentage of novel tasks in the trial. 

\begin{equation}\label{eq:ap}
\resizebox{.850\linewidth}{!}{$
AP = \frac{1}{m} \sum_{i=n-m}^{n} \bigg( \frac{1}{|T|} 
\sum_{t=1}^{|T|}
\bigg\{ 
\begin{array}{ll}
      1, & \mbox{if $i^{th}$ novel task in $t^{th}$ trial is passed}\\
      0, & \mbox{otherwise}
    \end{array}
\bigg)
$}    
\end{equation}

Second, as \textit{AP} does not capture the timeliness of adaptation, we compute the total area under the success curve (\textit{AUS}). \textit{AUS} can be defined as follows.  

\begin{equation}
\resizebox{.850\linewidth}{!}{$
AUS = \frac{1}{n} \sum_{i=1}^{n} \bigg( \frac{1}{|T|} 
\sum_{t=1}^{|T|}
\bigg\{ 
\begin{array}{ll}
      1, & \mbox{if $i^{th}$ novel task in $t^{th}$ trial is passed}\\
      0, & \mbox{otherwise}
    \end{array}
\bigg)
$}    
\end{equation}

\begin{figure}[t]
    \centering
    \includegraphics[width=0.7\textwidth,height=\textheight,keepaspectratio]{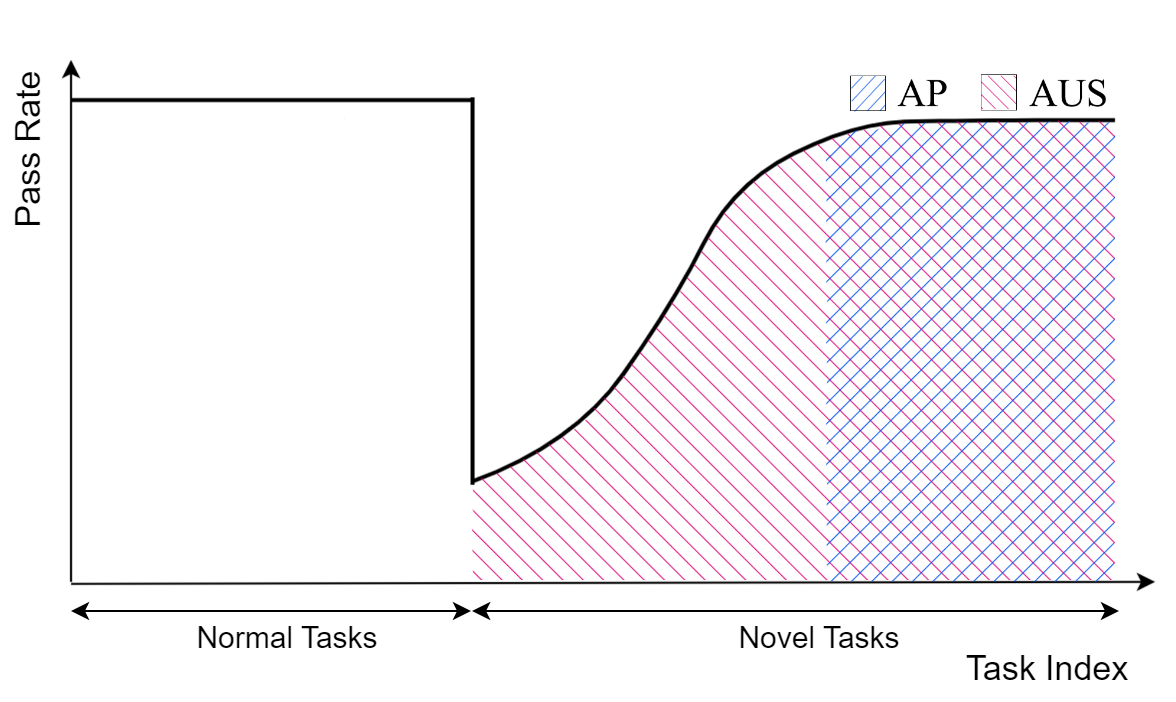}
    \caption{An example pass rate curve of an agent that played a set of trials. The area shaded in blue is considered for the asymptotic performance (AP) and the area shaded in red is considered for the area under success curve performance (AUS).}
    \label{fig:aps_au_measures}
\end{figure}

Both AP and AUS vary between 0 and 1 and 1 is the best achievable adaptation performance. Figure \ref{fig:aps_au_measures} graphically illustrates the areas considered in the pass rate curve of an agent when calculating AP and AUS.

Deriving from the above measures for novelty detection and novelty adaptation, we define \textit{NPS} (Novelty Performance in Scenarios) and \textit{SPN} (Scenario Performance under Novelties). \textit{NPS} measures how agents perform in a single novelty, when the novelty is applied to different physical scenarios. \textit{SPN} measures how agents perform in a single physical scenario, when different novelties are applied to the scenario. 
To calculate the \textit{NPS} measures, the average performance is taken across all scenarios for a single novelty. 
Formally, consider $i$ to represent the $i^{th}$ novelty, and $j$ to represent the $j^{th}$ scenario. The NPS measures, $NPS_{CDT,i}$, $NPS_{DD,i}$, $NPS_{AP,i}$, and $NPS_{AUS,i}$ are defined for a given novelty $i$ as the average of $CDT_{ij}$, $DD_{ij}$, $AP_{ij}$, and $AUS_{ij}$ respectively for all the scenarios $j$ with the novelty $i$.
Similarly, for \textit{SPN} measures, for a given scenario $j$, the average performance is taken across all the novelties $i$ for the scenario $j$. i.e., \textit{SPN} measures $SPN_{CDT,j}$, $SPN_{DD,j}$, $SPN_{AP,j}$, and $SPN_{AUS,j}$ are calculated from the average performance taken for $CDT_{ij}$, $DD_{ij}$, $AP_{ij}$, and $AUS_{ij}$ respectively for all the novelties $i$ in the scenario $j$.

\section{Experiments}
\label{sec:experiments}

We conduct experiments on baseline agents on the 40 novelty-scenarios to measure how agents detect and adapt to novelty in each of those novelty-scenarios. In addition, we establish human performance on all the novelty-scenarios. This section describes the baseline agents we provide and the experimental setups used for each experiment. 

\subsection{Baseline Agents} \label{sec:baseline_agents}

We include experimental results of 11 baseline agents which consist of three heuristic agents, seven learning agents, and a random agent. 

\paragraph{\textbf{Heuristic Agents}}

The heuristic agents are based on hard-coded physical rules developed by agent developers. All the agents were participating agents from the AIBIRDS competition \cite{AIBirds}, which is an annual competition held to find the best Angry Birds game-playing AI agent. % We use the heuristic agents to show that hard-coded heuristics may not work when novelties are present. Moreover, we use the heuristic agent's results to compute standard SAIL-ON performance evaluation measures. 
Following is the list of heuristic agents evaluated in NovPhy. 

\begin{itemize}
    \item Datalab: Datalab is a planning agent that has six strategies. The strategies include destroying pigs, destroying physical structures, and shooting at round blocks. The agent selects which action to take based on the game objects available, possible trajectories, the bird on the sling, and the birds remaining \cite{Datalab}. 
    \item Eagle's Wing: Eagle's Wing agent selects from a suit of five strategies based on structural analysis. The five strategies include: shooting at unprotected pigs, destroying many blocks as possible, and shooting at objects close to round objects \cite{EagleWings}.
    \item Pig Shooter: Pig Shooter has only one strategy: shooting at pigs. The agent randomly selects which pig to shoot and which trajectory to use \cite{Matthew2018}.
\end{itemize}

All these heuristic agents work under the uninformed evaluation setting, in which the agent is not informed when the first novel task appears in the trial.
 
\paragraph{\textbf{Learning Agents}}
In this work, we evaluate seven learning agents/versions of agents. All seven learning agents we present here work under the informed evaluation setting in which the agent is informed when the novelty appears in the trial. Therefore, we do not evaluate the novelty detection performance of these agents. The seven agents are \textit{DQN Offline/ Online/ Adapt}, \textit{Relational Offline/ Online/ Adapt}, and \textit{Naive Adapt}. Same as the deep reinforcement learning agents used in \cite{PhyQ}, we train a DQN \cite{wang2016dueling, van2016deep} agent and the Relational agent that contains a relational module \cite{zambaldi2018relational}. Both agents are trained on the tasks generated from a normal task template and are evaluated on the trials that contain tasks from the corresponding novel task template. We evaluate the DQN and Relational agents in three versions: offline, online, and adapt. With the offline version, the two deep reinforcement learning agents DQN and Relational, always select the action with the highest q-value throughout the trial. On the other hand, online learning agents update the q-network after novelty is introduced and try to relearn the policy to solve novel tasks. We also evaluate the recently developed open-world learning component \textit{NAPPING} \cite{napping} together with DQN and Relational agents. We call these agents DQN Adapt and Relational Adapt. 

The \textit{Naive Adapt} is built on top of the \textit{Pig Shooter} agent in \cite{PhyQ}, which shoots only at the pigs. \textit{Naive Adapt} uses the strategy of the \textit{Pig Shooter} in the pre-novelty game tasks. After the agent is informed that the novelty has occurred, it searches for a combination of (objects, trajectories, and delays) that solve a game level and keeps a record of each triplet tried. Once a solution triplet (e.g., a solution triplet could be (pig$_2$, high trajectory, delay 5 seconds)) is found for a trial, the \textit{Naive Adapt} will keep using the triplet until it does not solve the tasks anymore, where the agent starts to search for another triplet.

% \begin{itemize}
%     \item DQN Offline: Same as the DRL agents used in \cite{PhyQ},
%     \item DQN Online
%     \item DQN Adapt
%     \item Relational Offline
%     \item Relational Online
%     \item Relational Adapt
%     \item Naive Adapt:
% \end{itemize}

\paragraph{\textbf{Random Agent}}

The Random Agent selects a random release point (\textit{x},\textit{y}) relative to the slingshot. The \textit{x} is sampled from [-200, 200] and \textit{y} is sampled from [-200, 200]. This agent works under the uninformed evaluation setting. 

\subsection{Experimental Setups}

\subsubsection{Human Experiment Setup}
The experiments with human participants were approved by the Australian National University committee on human ethics under the protocol 2021/293. Participation was entirely voluntary, and no monetary compensation was provided. There were 47 participants with ages ranging from 20 to 35 years and there were both males and females. They were not experienced Angry Birds players. Some of the participants have never played the game and some of them knew the general game mechanics and had played the game on an occasional basis in the past, but did not have an extensive understanding of the game's strategies. Participants provided their consent to use their play-data. 

For a single participant, we provided 10 trials from 10 novelty-scenarios. We had four such trial sets to cover all 40 novelty-scenarios. In a single trial, there were 1-4 normal tasks and 4 novel tasks. On average participants spent 25-30 minutes to complete the experiment. Participants attempted to solve the tasks and at the end of each task they indicated if they detected a novelty or not. 

\subsubsection{Agent Experiment Setup}
We use the standard SAIL-ON evaluation setup for all agents. As mentioned previously, we have eight novelties and five physical reasoning scenarios which results in 40 novelty-scenarios. For a single novelty-scenario, we test the agent on 40 trials. A trial consists of 1-40 normal tasks and 40 novel tasks. All heuristic agents and the Random agent does not require any training. However, learning agents are trained on the normal tasks of the corresponding novelty-scenario and then the agent is evaluated on the trial set.    

\section{Results and Analysis}
\label{sec:results_and_analysis}
In this section, we present and discuss the results of the experiments we conducted: the human player experiment and the baseline agent experiment. For both experiments, we report the novelty detection and novelty adaptation performance.

\subsection{Human Performance}

\begin{figure}
    \centering
    \includegraphics[width=\textwidth,height=\textheight,keepaspectratio]{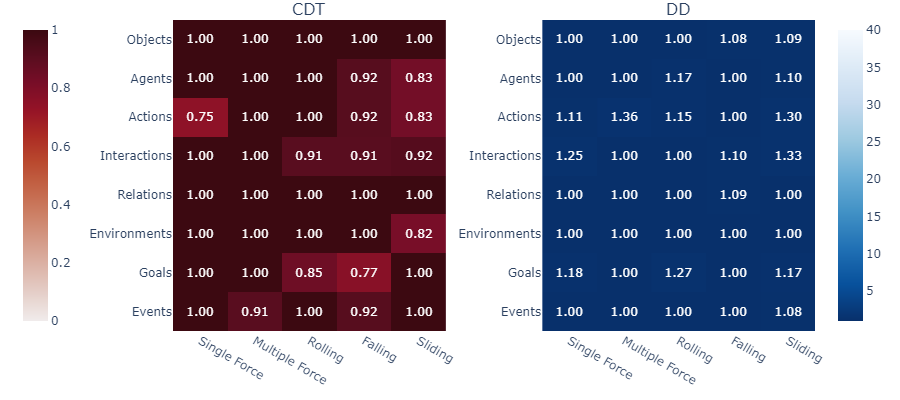}
    \caption{CDT (left) and DD (right) results of the human players. In the heat maps, the x-axis is the physical scenario and the y-axis is the novelty applied.}
    \label{fig:human_detection_results}
\end{figure}

Figure \ref{fig:human_detection_results} shows CDT and DD results of the human players. Overall, the participants were able to correctly detect the novelties in almost all the trials (CDT is close to 1). The lowest CDT is for the novelty-scenario actions-single force, in which the upward force of the Air Turbulence agent is increased. The reason for this is likely because this novelty is not visually detectable until interacted with. Also, when this novelty is applied to the single force scenario, the player has only to slightly adjust the shooting angle of the bird compared to the shooting angle in the normal tasks. As the impact of this novelty is subtle, it might not be perceivable to humans. This is also likely to be the case with the novelty-scenario that has the second lowest CDT: goals-falling.
When we look at the DD results, in most cases it can be seen that the participants could detect the novelty in the first game level where the novelty was encountered (DD is close to 1). Generally, it can also be seen that for the novelties that are not visually detectable without interaction (actions, interactions, and goals), humans have a higher detection delay compared to the other novelties that can be visually detected before interaction.

\begin{figure}
    \centering
    \includegraphics[width=\textwidth,height=\textheight,keepaspectratio]{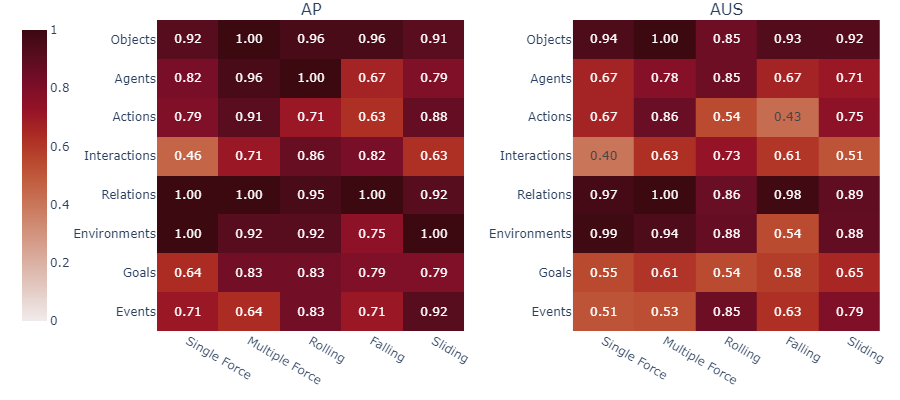}
    \caption{AP (left) and AUS (right) results of the human players. In the heat maps, the x-axis is the physical scenario and the y-axis is the novelty applied. The asymptotic length considered for the AP calculation is 2. }
    \label{fig:human_adaptation_results}
\end{figure}

The AP and AUS performances of the human players are shown in Figure \ref{fig:human_adaptation_results}. For the AP calculation, we used the asymptotic length as 2 (i.e., the performance of the last two tasks in the trial). As the AP results depict, the participants obtained above 80\% performance for most of the novelty-scenarios. Generally, for both the AP and AUS, participants showed lower results for the novelty-scenarios which required highly accurate actions. For example, interactions-single force, goals-single force, events-multiple forces, interactions-sliding, and actions-falling have slightly lower results for both the measures as it is required to shoot the bird with high accuracy to solve the tasks in those novelty-scenarios. The rate of adaptation is captured from the AUS results. When the accuracy requirements are higher, it takes more tasks for humans to adapt to the novelty. This is reflected in the relatively lower AUS results for the novelties that demand accurate actions: interactions, goals, and events. On the other hand, novelties that do not require accurate actions such as objects and relations have nearly perfect AUS results, as humans can adapt to such novelties straight away.

Moreover, we analyzed the relationship between the detection performance and adaptation performance of human players using the non-parametric Mann-Whitney test and Spearman's rank correlation (see \ref{Appendix: Section Human Detection vs Adaptation} for more details). Considering the relationship between CDT and adaptation (both the AP and AUS), only the results of the actions-single force novelty-scenario shows that, the adaptation performance is independent of whether participants detected it or not. As discussed in the above paragraphs, this is because, even though humans have successfully adapted to the actions-single force, the novelty may not be perceivable to humans.
The correlation between detection delay and adaptation performance (both the AP and AUS) shows that there are some novelty-scenarios such as goals-single force, goals-rolling, and interactions-sliding have moderately negative correlations implying that the longer a player takes to detect the novelty, the lower the adaptation performance. The correlation plots are shown in Appendix Figure \ref{fig:DD_vs_adaptation_correlation}.

\subsection{Baseline Agent Performance}

In this section, we discuss the performance of the 11 baseline agents and humans in terms of novelty detection and novelty adaptation. 

\subsubsection{Baseline Agent Novelty Detection Performance}
As discussed in Section \ref{sec:baseline_agents} some of the agents in NovPhy work under the uninformed evaluation. We compute the detection performance measures for those agents. The detection modules in those baseline agents are based on the pass rate deviation (discussed in detail in \ref{Appendix: Novelty Detection Performance}).
The results of the detection measures per novelty are presented in Table \ref{tab:NPS} and detection measures per scenario are presented in Table \ref{tab:SPN} for all the heuristic agents, the random agent, and humans.

\begin{sidewaystable}
%\sidewaystablefn%
\begin{center}
\resizebox{650pt}{!}{%  
    %\begin{minipage}{\textheight}
    \begin{tabular}{|c|c|c|c|c|c|c|c|c|c|c|c|c|c|}
        \hline
           Novelty & Measure & Human & DQN Offline & DQN Online & DQN Adapt & Rel Offline & Rel Online & Rel Adapt & Naive Adapt & Datalab & Eagle'sWing & PigShooter & Random  \\
          \hline
         1. Objects  & $NPS_{CDT,1}$ & 1.00 $\pm$ 0.00 & -  & - & - & - & - & - & - & \textcolor{blue}{0.65 $\pm$ 0.13} & 0.57 $\pm$ 0.08 & 0.25 $\pm$ 0.08 & 0.23 $\pm$ 0.09 \\
         & $NPS_{DD,1}$ & 1.03 $\pm$ 0.02 & -  & - & - & - & - & - & - & 7.85 $\pm$ 4.04 & \textcolor{blue}{5.46 $\pm$ 1.17} & 6.07 $\pm$ 2.53 & 12.79 $\pm$ 4.31 \\
         & $NPS_{AP,1}$ & 0.95 $\pm$ 0.02 & 0.65 $\pm$  0.17	&	0.66 $\pm$  0.14	&	0.79 $\pm$  0.10	&	0.54 $\pm$  0.10	&	0.50 $\pm$  0.10	&	\textcolor{blue}{0.80 $\pm$  0.06}	&	0.61 $\pm$  0.19	&	0.37 $\pm$  0.19	&	0.40 $\pm$  0.21	&	0.02 $\pm$  0.02	&	0.01 $\pm$  0.01 \\
         & $NPS_{AUS,1}$ & 0.93 $\pm$ 0.02 & 0.66 $\pm$ 0.17	&	0.67 $\pm$ 0.16	&	\textcolor{blue}{0.81 $\pm$ 0.10}	&	0.56 $\pm$ 0.11	&	0.55 $\pm$ 0.10	&	0.80 $\pm$ 0.06	&	0.61 $\pm$ 0.19	&	0.38 $\pm$ 0.19	&	0.41 $\pm$ 0.21	&	0.02 $\pm$ 0.02	&	0.01 $\pm$ 0.01  \\
        \hline
        2. Agents  & $NPS_{CDT,2}$ & 0.95 $\pm$ 0.03 & -  & - & - & - & - & - & - & \textcolor{blue}{0.49 $\pm$ 0.17} & 0.42 $\pm$ 0.12 & 0.17 $\pm$ 0.07 & 0.24 $\pm$ 0.07 \\
        & $NPS_{DD,2}$ & 1.05 $\pm$ 0.03 & -  & - & - & - & - & - & - & 10.96 $\pm$ 4.29 & 10.01 $\pm$ 2.25 & \textcolor{blue}{4.57 $\pm$ 0.39} & 19.59 $\pm$ 1.42 \\
        & $NPS_{AP,2}$ & 0.85 $\pm$ 0.02 & 0.06 $\pm$  0.03	&	0.21 $\pm$  0.07	&	\textcolor{blue}{0.73 $\pm$  0.13}	&	0.07 $\pm$  0.04	&	0.11 $\pm$  0.04	&	0.67 $\pm$  0.09	&	0.11 $\pm$  0.06	&	0.05 $\pm$  0.03	&	0.05 $\pm$  0.03	&	0.01 $\pm$  0.01	&	0.01 $\pm$  0.00 \\
        & $NPS_{AUS,2}$ & 0.73 $\pm$ 0.03 & 0.06 $\pm$ 0.03	&	0.18 $\pm$ 0.06	&	\textcolor{blue}{0.62 $\pm$ 0.10}	&	0.07 $\pm$ 0.04	&	0.10 $\pm$ 0.04	&	0.56 $\pm$ 0.09	&	0.09 $\pm$ 0.05	&	0.05 $\pm$ 0.03	&	0.05 $\pm$ 0.03	&	0.00 $\pm$ 0.00	&	0.01 $\pm$ 0.00  \\
        \hline
        3. Actions  & $NPS_{CDT,3}$ & 0.90 $\pm$ 0.05 & -  & - & - & - & - & - & - & 0.44 $\pm$ 0.15 & \textcolor{blue}{0.49 $\pm$ 0.19} & 0.24 $\pm$ 0.02 & 0.17 $\pm$ 0.06 \\
        & $NPS_{DD,3}$ & 1.19 $\pm$ 0.07 & -  & - & - & - & - & - & - & 10.11 $\pm$ 1.69 & 8.12 $\pm$ 2.20 & \textcolor{blue}{5.67 $\pm$ 0.57} & 19.27 $\pm$ 1.73 \\
        & $NPS_{AP,3}$ & 0.78 $\pm$ 0.05 & 0.07 $\pm$  0.06	&	0.13 $\pm$  0.08	&	0.56 $\pm$  0.17	&	0.05 $\pm$  0.03	&	0.12 $\pm$  0.07	&	\textcolor{blue}{0.63 $\pm$  0.16}	&	0.10 $\pm$  0.07	&	0.03 $\pm$  0.02	&	0.08 $\pm$  0.04	&	0.00 $\pm$  0.00	&	0.01 $\pm$  0.00  \\
        & $NPS_{AUS,3}$ & 0.65 $\pm$ 0.08 & 0.08 $\pm$ 0.06	&	0.13 $\pm$ 0.08	&	0.53 $\pm$ 0.15	&	0.05 $\pm$ 0.03	&	0.10 $\pm$ 0.06	&	\textcolor{blue}{0.55 $\pm$ 0.14}	&	0.08 $\pm$ 0.05	&	0.03 $\pm$ 0.02	&	0.08 $\pm$ 0.04	&	0.00 $\pm$ 0.00	&	0.01 $\pm$ 0.00  \\
        \hline
        4. Interactions  & $NPS_{CDT,4}$ & 0.95 $\pm$ 0.02 & -  & - & - & - & - & - & - & 0.42 $\pm$ 0.15  & 0.46 $\pm$ 0.18 & 0.18 $\pm$ 0.05 & \textcolor{blue}{0.55 $\pm$ 0.12} \\
        & $NPS_{DD,4}$ & 1.14 $\pm$ 0.07 & -  & - & - & - & - & - & - & 5.87 $\pm$ 1.32 & 16.31 $\pm$ 7.38 & \textcolor{blue}{5.25 $\pm$ 0.35} & 18.40 $\pm$ 5.13 \\
        & $NPS_{AP,4}$ & 0.69 $\pm$ 0.07 & 0.10 $\pm$  0.04	&	0.24 $\pm$  0.07	&	\textcolor{blue}{0.75 $\pm$  0.16}	&	0.08 $\pm$  0.03	&	0.19 $\pm$  0.05	&	0.72 $\pm$  0.15	&	0.33 $\pm$  0.14	&	0.10 $\pm$  0.06	&	0.3 $\pm$  0.17	&	0.03 $\pm$  0.02	&	0.06 $\pm$  0.02 \\
        & $NPS_{AUS,4}$ & 0.58 $\pm$ 0.05 & 0.10 $\pm$ 0.04	&	0.25 $\pm$ 0.08	&	\textcolor{blue}{0.64 $\pm$ 0.14}	&	0.08 $\pm$ 0.03	&	0.19 $\pm$ 0.04	&	0.63 $\pm$ 0.13	&	0.27 $\pm$ 0.11	&	0.18 $\pm$ 0.07	&	0.35 $\pm$ 0.18	&	0.02 $\pm$ 0.02	&	0.06 $\pm$ 0.02  \\
        \hline
        5. Relations  & $NPS_{CDT,5}$ & 1.00 $\pm$ 0.00 & -  & - & - & - & - & - & - & 0.27 $\pm$ 0.18 & \textcolor{blue}{0.28 $\pm$ 0.17} & 0.08 $\pm$ 0.07 & 0.01 $\pm$ 0.01 \\
        & $NPS_{DD,5}$ & 1.02 $\pm$ 0.02 & -  & - & - & - & - & - & - & \textcolor{blue}{3.47 $\pm$ 0.40} & 7.49 $\pm$ 0.02 & 3.86 $\pm$ 0.54 & 4.00 $\pm$ 0.00 \\
        & $NPS_{AP,5}$ & 0.97 $\pm$ 0.02 & 0.00 $\pm$  0.00	&	0.00 $\pm$  0.00	&	0.00 $\pm$  0.00	&	0.00 $\pm$  0.00	&	0.00 $\pm$  0.00	&	0.00 $\pm$  0.00	&	0.00 $\pm$  0.00	&	0.00 $\pm$  0.00	&	0.00 $\pm$  0.00	&	0.00 $\pm$  0.00	&	0.00 $\pm$  0.00 \\
        & $NPS_{AUS,5}$ & 0.94 $\pm$ 0.03 & 0.00 $\pm$ 0.00	&	0.00 $\pm$ 0.00	&	0.00 $\pm$ 0.00	&	0.00 $\pm$ 0.00	&	0.00 $\pm$ 0.00	&	0.00 $\pm$ 0.00	&	0.00 $\pm$ 0.00	&	0.00 $\pm$ 0.00	&	0.00 $\pm$ 0.00	&	0.00 $\pm$ 0.00	&	0.00 $\pm$ 0.00 \\
        \hline
        6. Environments  & $NPS_{CDT,6}$ & 0.96 $\pm$ 0.04 & -  & - & - & - & - & - & - & \textcolor{blue}{0.44 $\pm$ 0.15} & 0.35 $\pm$ 0.13 & 0.01 $\pm$ 0.01 & 0.40 $\pm$ 0.10 \\
        & $NPS_{DD,6}$ & 1.00 $\pm$ 0.00 & -  & - & - & - & - & - & - & 4.75 $\pm$ 0.19 & 5.50 $\pm$ 1.03 & \textcolor{blue}{2.50 $\pm$ 0.95} & 14.64 $\pm$ 0.83 \\
        & $NPS_{AP,6}$ & 0.92 $\pm$ 0.05 & 0.00 $\pm$  0.00	&	0.00 $\pm$  0.00	&	0.01 $\pm$  0.00	&	0.00 $\pm$  0.00	&	0.00 $\pm$  0.00	&	0.00 $\pm$  0.00	&	0.00 $\pm$  0.00	&	0.00 $\pm$  0.00	&	0.00 $\pm$  0.00	&	0.00 $\pm$  0.00	&	\textcolor{blue}{0.03 $\pm$  0.01}  \\
        & $NPS_{AUS,6}$ & 0.85 $\pm$ 0.08 & 0.00 $\pm$ 0.00	&	0.00 $\pm$ 0.00	&	0.01 $\pm$ 0.00	&	0.00 $\pm$ 0.00	&	0.00 $\pm$ 0.00	&	0.00 $\pm$ 0.00	&	0.00 $\pm$ 0.00	&	0.00 $\pm$ 0.00	&	0.00 $\pm$ 0.00	&	0.00 $\pm$ 0.00	&	\textcolor{blue}{0.03 $\pm$ 0.01}  \\
        \hline
        7. Goals  & $NPS_{CDT,7}$ & 0.92 $\pm$ 0.05 & -  & - & - & - & - & - & - & 0.43 $\pm$ 0.13 & \textcolor{blue}{0.55 $\pm$ 0.10} & 0.19 $\pm$ 0.05 & 0.39 $\pm$ 0.09 \\
        & $NPS_{DD,7}$ & 1.12 $\pm$ 0.05 & -  & - & - & - & - & - & - & 11.62 $\pm$ 1.48 & 12.28 $\pm$ 2.99 & \textcolor{blue}{5.73 $\pm$ 0.22} & 15.06 $\pm$ 2.40 \\
        & $NPS_{AP,7}$ & 0.78 $\pm$ 0.04 & 0.07 $\pm$  0.02	&	0.09 $\pm$  0.03	&	0.45 $\pm$  0.15	&	0.04 $\pm$  0.02	&	0.14 $\pm$  0.08	&	\textcolor{blue}{0.54 $\pm$  0.18}	&	0.29 $\pm$  0.12	&	0.08 $\pm$  0.04	&	0.09 $\pm$  0.03	&	0.01 $\pm$  0.01	&	0.02 $\pm$  0.01  \\
        & $NPS_{AUS,7}$ & 0.59 $\pm$ 0.02 & 0.08 $\pm$ 0.03	&	0.10 $\pm$ 0.03	&	0.42 $\pm$ 0.13	&	0.04 $\pm$ 0.02	&	0.12 $\pm$ 0.05	&	\textcolor{blue}{0.49 $\pm$ 0.16}	&	0.29 $\pm$ 0.12	&	0.08 $\pm$ 0.04	&	0.09 $\pm$ 0.04	&	0.01 $\pm$ 0.01	&	0.02 $\pm$ 0.01  \\
        \hline
        8. Events  & $NPS_{CDT,8}$ & 0.97 $\pm$ 0.02 & -  & - & - & - & - & - & - & \textcolor{blue}{0.46 $\pm$ 0.13} & \textcolor{blue}{0.46 $\pm$ 0.12} & 0.07 $\pm$ 0.03 & 0.26 $\pm$ 0.80 \\
        & $NPS_{DD,8}$ & 1.02 $\pm$ 0.02 & -  & - & - & - & - & - & - & 7.36 $\pm$ 1.73 & 11.00 $\pm$ 2.36 & \textcolor{blue}{4.80 $\pm$ 1.35} & 18.14 $\pm$ 2.18 \\
        & $NPS_{AP,8}$ & 0.76 $\pm$ 0.05 & 0.27 $\pm$  0.12	&	0.21 $\pm$  0.10	&	0.21 $\pm$  0.1	&	\textcolor{blue}{0.29 $\pm$  0.11}	&	0.2 $\pm$  0.09	&	0.2 $\pm$  0.09	&	0.00 $\pm$  0.00	&	0.13 $\pm$  0.05	&	0.12 $\pm$  0.04	&	0.01 $\pm$  0.00	&	0.02 $\pm$  0.01  \\
        & $NPS_{AUS,8}$ & 0.66 $\pm$ 0.07 & 0.27 $\pm$ 0.12	&	0.24 $\pm$ 0.11	&	0.23 $\pm$ 0.1	&	\textcolor{blue}{0.29 $\pm$ 0.11}	&	0.22 $\pm$ 0.10	&	0.21 $\pm$ 0.09	&	0.00 $\pm$ 0.00	&	0.13 $\pm$ 0.05	&	0.12 $\pm$ 0.04	&	0.01 $\pm$ 0.00	&	0.02 $\pm$ 0.01  \\
        \hline
    \end{tabular}}
    \caption{Results of the NPS measures of the agents and humans for the eight novelties.}
    \label{tab:NPS}
    %\end{minipage}
    \end{center}
\end{sidewaystable}

\begin{sidewaystable}
\begin{center}
    \resizebox{650pt}{!}{%
    \begin{tabular}{|c|c|c|c|c|c|c|c|c|c|c|c|c|c|}
        \hline
          Scenario & Measure & Human & DQN Offline & DQN Online & DQN Adapt & Rel Offline & Rel Online & Rel Adapt & Naive Adapt & Datalab & Eagle'sWing & PigShooter & Random  \\
          \hline
        \centering
       1. Single Force  & $SPN_{CDT,1}$ & 0.97 $\pm$ 0.03 & -  & - & - & - & - & - & - & 0.59 $\pm$ 0.07 & \textcolor{blue}{0.67 $\pm$ 0.05} & 0.08 $\pm$ 0.04 & 0.31 $\pm$ 0.10 \\
         & $SPN_{DD,1}$ & 1.07 $\pm$ 0.04 & -  & - & - & - & - & - & - & 6.61 $\pm$ 0.88 & 7.42 $\pm$ 1.46 & \textcolor{blue}{4.82 $\pm$ 1.02} & 14.21 $\pm$ 2.62 \\
         & $SPN_{AP,1}$ & 0.79 $\pm$ 0.07 & 0.11 $\pm$ 0.05	&	0.13 $\pm$ 0.05	&	\textcolor{blue}{0.37 $\pm$ 0.14}	&	0.12 $\pm$ 0.06	&	0.13 $\pm$ 0.05	&	0.36 $\pm$ 0.13	&	0.13 $\pm$ 0.11	&	0.08 $\pm$ 0.04	&	0.16 $\pm$ 0.11	&	0.01 $\pm$ 0.00	&	0.03 $\pm$ 0.02 \\
         & $SPN_{AUS,1}$ & 0.71 $\pm$ 0.08 & 0.11 $\pm$ 0.05	&	0.13 $\pm$ 0.05	&	\textcolor{blue}{0.33 $\pm$ 0.12}	&	0.12 $\pm$ 0.06	&	0.13 $\pm$ 0.06	&	\textcolor{blue}{0.33 $\pm$ 0.11} &	0.13 $\pm$ 0.12	&	0.08 $\pm$ 0.05	&	0.17 $\pm$ 0.12	&	0.01 $\pm$ 0.00	&	0.03 $\pm$ 0.02  \\
        \hline
        2. Multiple Force & $SPN_{CDT,2}$ & 0.99 $\pm$ 0.01 & -  & - & - & - & - & - & - & \textcolor{blue}{0.55 $\pm$ 0.14} & \textcolor{blue}{0.55 $\pm$ 0.10} & 0.14 $\pm$ 0.03 & 0.14 $\pm$ 0.09 \\
        & $SPN_{DD,2}$ & 1.05 $\pm$ 0.05 & -  & - & - & - & - & - & - & \textcolor{blue}{4.29 $\pm$ 1.28} & 4.88 $\pm$ 0.47 & 5.16 $\pm$ 0.38 & 17.48 $\pm$ 1.26 \\
         & $SPN_{AP,2}$ & 0.87 $\pm$ 0.05 & 0.07 $\pm$ 0.03	&	0.15 $\pm$ 0.06	&	0.11 $\pm$ 0.05	&	0.05 $\pm$ 0.03	&	0.06 $\pm$ 0.03	&	0.16 $\pm$ 0.08	&	\textcolor{blue}{0.17 $\pm$ 0.12}	&	0.01 $\pm$ 0	&	0.08 $\pm$ 0.05	&	0.00 $\pm$ 0.00	&	0.01 $\pm$ 0.00  \\
         & $SPN_{AUS,2}$ & 0.79 $\pm$ 0.07 & 0.08 $\pm$ 0.03	&	0.14 $\pm$ 0.06	&	0.12 $\pm$ 0.06	&	0.04 $\pm$ 0.02	&	0.07 $\pm$ 0.03	&	0.15 $\pm$ 0.07	&	\textcolor{blue}{0.17 $\pm$ 0.12}	&	0.05 $\pm$ 0.05	&	0.10 $\pm$ 0.07	&	0.00 $\pm$ 0.00	&	0.01 $\pm$ 0.00  \\
        \hline
        3. Rolling & $SPN_{CDT,3}$ & 0.97 $\pm$ 0.02 & -  & - & - & - & - & - & - & 0.26 $\pm$ 0.07 & 0.24 $\pm$ 0.08 & 0.20 $\pm$ 0.04 & \textcolor{blue}{0.34 $\pm$ 0.07} \\
        & $SPN_{DD,3}$ & 1.07 $\pm$ 0.04 & -  & - & - & - & - & - & - & 10.49 $\pm$ 3.11 & 12.02 $\pm$ 2.28 & \textcolor{blue}{4.45 $\pm$ 0.42} & 16.13 $\pm$ 1.71 \\
         & $SPN_{AP,3}$ & 0.88 $\pm$ 0.03 & 0.24 $\pm$ 0.13	&	0.27 $\pm$ 0.12	&	0.59 $\pm$ 0.14	&	0.24 $\pm$ 0.11	&	0.22 $\pm$ 0.10	&	\textcolor{blue}{0.60 $\pm$ 0.14}	&	0.18 $\pm$ 0.10	&	0.07 $\pm$ 0.03	&	0.07 $\pm$ 0.03	&	0.00 $\pm$ 0.00	&	0.02 $\pm$ 0.00  \\
         & $SPN_{AUS,3}$ & 0.76 $\pm$ 0.05 & 0.25 $\pm$ 0.13	&	0.28 $\pm$ 0.12	&	\textcolor{blue}{0.57 $\pm$ 0.13}	&	0.25 $\pm$ 0.11	&	0.23 $\pm$ 0.11	&	\textcolor{blue}{0.57 $\pm$ 0.13}	&	0.15 $\pm$ 0.08	&	0.07 $\pm$ 0.02	&	0.07 $\pm$ 0.03	&	0.00 $\pm$ 0.00	&	0.02 $\pm$ 0.00  \\
        \hline
        4. Falling & $SPN_{CDT,4}$ & 0.93 $\pm$ 0.03 & -  & - & - & - & - & - & - & \textcolor{blue}{0.48 $\pm$ 0.11} & 0.44 $\pm$ 0.11 & 0.19 $\pm$ 0.05 & 0.36 $\pm$ 0.08 \\
        & $SPN_{DD,4}$ & 1.03 $\pm$ 0.02 & -  & - & - & - & - & - & - & 10.04 $\pm$ 1.75 & 12.51 $\pm$ 2.37 & \textcolor{blue}{4.89 $\pm$ 0.48} & 13.35 $\pm$ 1.98 \\
         & $SPN_{AP,4}$ & 0.79 $\pm$ 0.05 & 0.17 $\pm$ 0.11	&	0.20 $\pm$ 0.1	&	\textcolor{blue}{0.57 $\pm$ 0.14}	&	0.14 $\pm$ 0.09	&	0.2 $\pm$ 0.08	&	0.55 $\pm$ 0.14	&	0.21 $\pm$ 0.09	&	0.19 $\pm$ 0.10	&	0.19 $\pm$ 0.10	&	0.02 $\pm$ 0.01	&	0.02 $\pm$ 0.01  \\
         & $SPN_{AUS,4}$ & 0.67 $\pm$ 0.07 & 0.18 $\pm$ 0.11	&	0.22 $\pm$ 0.10	&	\textcolor{blue}{0.51 $\pm$ 0.13}	&	0.14 $\pm$ 0.09	&	0.19 $\pm$ 0.08	&	0.50 $\pm$ 0.13	&	0.19 $\pm$ 0.09	&	0.20 $\pm$ 0.10	&	0.19 $\pm$ 0.10	&	0.02 $\pm$ 0.01	&	0.02 $\pm$ 0.01  \\
        \hline
        5. Sliding & $SPN_{CDT,5}$ & 0.93 $\pm$ 0.03 & -  & - & - & - & - & - & - & \textcolor{blue}{0.37 $\pm$ 0.13} & 0.33 $\pm$ 0.13 & 0.12 $\pm$ 0.06 & 0.25 $\pm$ 0.05 \\
        & $SPN_{DD,5}$ & 1.13 $\pm$ 0.04 & -  & - & - & - & - & - & - & 9.10 $\pm$ 1.72 & 12.97 $\pm$ 4.41 & \textcolor{blue}{6.99 $\pm$ 2.25} & 22.12 $\pm$ 2.83 \\
         & $SPN_{AP,5}$ & 0.85 $\pm$ 0.04 & 0.17 $\pm$ 0.11	&	0.21 $\pm$ 0.12	&	\textcolor{blue}{0.55 $\pm$ 0.15}	&	0.13 $\pm$ 0.08	&	0.18 $\pm$ 0.07	&	\textcolor{blue}{0.55 $\pm$ 0.14}	&	0.21 $\pm$ 0.10	&	0.13 $\pm$ 0.10	&	0.16 $\pm$ 0.11	&	0.02 $\pm$ 0.01	&	0.01 $\pm$ 0.00 \\
         & $SPN_{AUS,5}$ & 0.76 $\pm$ 0.05 & 0.18 $\pm$ 0.12	&	0.21 $\pm$ 0.12	&	\textcolor{blue}{0.49 $\pm$ 0.13}	&	0.13 $\pm$ 0.08	&	0.18 $\pm$ 0.07	&	0.48 $\pm$ 0.12	&	0.19 $\pm$ 0.10	&	0.14 $\pm$ 0.10	&	0.16 $\pm$ 0.12	&	0.01 $\pm$ 0.01	&	0.01 $\pm$ 0.00  \\
        \hline
        \end{tabular}}
        \caption{Results of the SPN measures of the agents and humans for the five scenarios.}
        \label{tab:SPN}
\end{center}
\end{sidewaystable}

As the results indicate, Datalab has the highest overall CDT while Pig Shooter has the lowest. It is expected for Pig Shooter to have the lowest CDT as the agent only directly shoots at the pigs, generally resulting in the same outcome in the tasks of a given trial, hence does not show a significant deviation in pass rates. Considering the overall DD, Random Agent has the highest DD. This is because the agent's randomness in the actions results in novelty detection at random positions of a trial causing the overall DD to fall around the mean number of the tasks in the trial. The Pig Shooter has the best (lowest) DD, this is because, for the few trials it correctly detects, it is done rapidly due to the deterministic nature of action selection. All the agents are far below the humans' novelty detection performance which is near perfect (CDT = 0.96 and DD = 1.07).

\subsubsection{Baseline Agent Novelty Adaptation Performance}

\begin{figure}[t]
    \centering
    \includegraphics[width=\textwidth,height=\textheight,keepaspectratio]{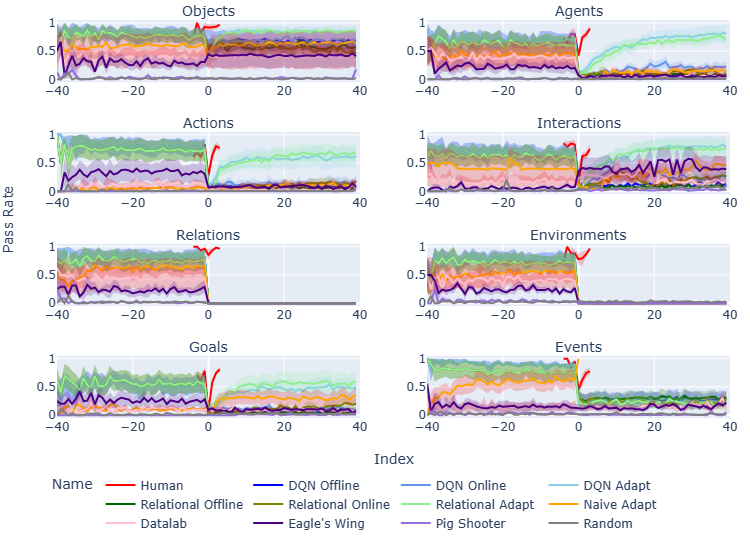}
    \caption{The pass rate of the agents per novelty. The x-axis represents the index of the task in trials. Indexes -40 to -1 represent normal tasks and 0 to 40 represent novel tasks. The y-axis shows the pass rate averaged across all trials relevant to the respective novelty.}
    \label{fig:adaptation_plot_per_novelty}
\end{figure}

\begin{figure}[t]
    \centering
    \includegraphics[width=\textwidth,height=\textheight,keepaspectratio]{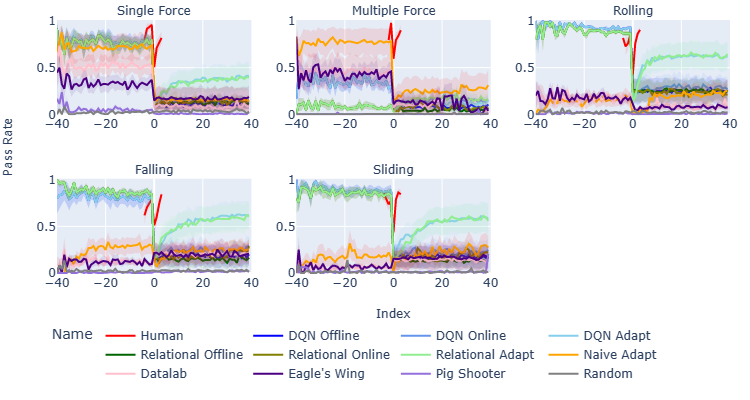}
    \caption{The pass rate of the agents per scenario. x-axis represents the index of the task in trials. Indexes -40 to -1 represent normal tasks and 0 to 40 represent novel tasks. The y-axis shows the pass rate averaged across all trials relevant to the respective scenario.}
    \label{fig:adaptation_plot_per_scenario}
\end{figure}

The adaptation plots per novelty are shown in Figure \ref{fig:adaptation_plot_per_novelty} and the adaptation plots per scenario are shown in Figure \ref{fig:adaptation_plot_per_scenario}. The novelty adaptation measures derived from the adaptation curves, AP and AUS results per scenario are presented in Table \ref{tab:NPS} and per novelty results are presented in Table \ref{tab:SPN}. The AP results are based on the last 50\% of the novel tasks (m=20).  
%The novelty adaptation results per scenario are presented in Table \ref{tab:NPS} and the adaptation results per novelty are presented in Table \ref{tab:SPN}.
%The x-axis of the two figures represents the index of the task. Indexes -40 to -1 represent normal tasks and 0 to 40 represent novel tasks. The y-axis shows the pass rate averaged across all trials relevant to the respective scenario or the respective novelty. For humans, the indexes are available only from -4 to 3 as the human experiment contained 1-4 normal tasks and 1-4 novel tasks. 
Ideally, an agent would have a higher pass rate in normal tasks and when novel tasks begin (at index 0 in Figures \ref{fig:adaptation_plot_per_novelty} and \ref{fig:adaptation_plot_per_scenario}), the performance would drop and recover its performance to reach the normal task performance within a few tasks. Therefore, an ideal agent would have AP and AUS results close to 1.

In Figure \ref{fig:adaptation_plot_per_scenario}, the agents DQN Adapt and Relational Adapt show a better adaptation behaviour compared to other agents in four scenarios out of five, the only exception is multiple forces. In those four scenarios, out of the two agents, DQN Adapt shows a slightly better performance (at 10\% level of significance) than Relational Adapt when we look at the $SPN_{AP,j}$, and $SPN_{AUS,j}$.
Similarly as shown in Figure \ref{fig:adaptation_plot_per_novelty}, the two agents adapt in five novelties except the novelties Relations, Environments, and  Events. This is because that to adapt to the Relations and Environments novelties, the agent needs to adjust the pre-defined trajectory planner. For example, in the Environments novelty, the trajectory is upside down. However, the -Adapt agents currently use the provided trajectory planner. Similarly, Naive Adapt agent also shows the adaptation behaviour in all scenarios except for single force. This agent shows an adaptation behaviour in all the novelties except Relations, Environments, and Events. However, as the Naive Adapt agent searches for just one solution tuple through a trial, it can not adapt to novelty trials that require different solution tuples under different situations. As a result, the Naive Adapt agent does not reach the performance level of the two agents, DQN Adapt and Relational Adapt, who learn efficiently how to handle novelties in different scenarios through each trial. Overall, the agents show the best adaptation performance in the Objects novelty. This is because the Objects novelty tested here is a change of the colour of an existing object, which has not impacted the agents' actions drastically. It is interesting to note that none of the agents has reached the humans' pass rate in any scenario or in any novelty, except for interactions novelty where Relational Adapt and DQN Adapt exceed the humans' $NPS_{AP,j}$, and $NPS_{AUS,j}$. However, within the same number of tasks that were given to humans, those two agents have not reached the performance the humans could achieve, showing that there is room for improvement in terms of adaptation efficiency.

\section{Conclusion and Future Work}
\label{sec:conculsion}

The objective of NovPhy is to facilitate the development of AI systems that can perform physical reasoning tasks in the presence of novelties, which is the condition that a system in an open-world physical environment would encounter. Towards this objective, NovPhy was designed to evaluate the abilities of an agent to detect novelties and adapt to perform under those novelties in a physical environment. We designed task templates for five commonly encountered real-world physical scenarios. Then, we designed novel tasks by introducing a diverse set of novelties to those task templates. This design enables to measure novelty detection and adaptation of agents in two directions: 1) how an agent performs in a novelty when the novelty is applied to different physical scenarios, and 2) how an agent performs in a physical scenario when different novelties are applied to it. To measure the true novelty adaptation performance of the agents, when designing the tasks we ensure that the agent has to work under the influence of the novelty rather than bypassing the novelties to solve the tasks. We evaluated the agents using a trial setting, in which the agent has to play a sequence of tasks of a scenario without novelties followed by a sequence of tasks of that scenario with novelties. 

We have established the baseline results of the testbed using human players, learning agents, and heuristic agents. The results show that novelties affect the agents' performance severely and some agents can recover as they play more and more tasks. However, agents' solving rate and efficiency in adaptation are subpar compared to humans' performance.
Although our results show that DQN Adapt and Relational Adapt agents are able to adapt to most novelties, there are still some novelties that the agents fail to adapt to. The main reason is that the agents still use the provided trajectory planner to plan for the shot (the release point). Future work on agents may focus on easing the need of using a trajectory planner to allow the agent to adapt to a wider range of novelties. 

We foresee different directions of improvement for NovPhy. NovPhy can be advanced by introducing more novelties representing the levels of the novelty hierarchy. Further, more complex physical reasoning scenarios such as relative height, relative weight, and clearing paths can be introduced to the testbed after agents show efficient novelty detection and novelty adaptation in the existing scenarios. Moreover, the testbed can be extended to assess the novelty characterization ability of agents (i.e., to evaluate whether an agent correctly detects `what is novel' in a task). Additionally, the concept in NovPhy, evaluating the physical reasoning capabilities under the influence of novelties, can be extended to other physical reasoning domains. For example, novelties can be introduced to physics-based robotic benchmarks such as CausalWorld \cite{ahmed2020causalworld} and RLBench \cite{james2020rlbench}, which will facilitate evaluating agents on physical scenarios such as pulling, picking and placing, and stacking, which are not seen in the Angry Birds domain.
We believe that NovPhy builds the foundation for future research on developing agents that can efficiently detect and adapt to novelty in the physical world as humans do. 

%% If you have bibdatabase file and want bibtex to generate the
%% bibitems, please use
%%
\section*{Acknowledgments}

This research was sponsored by the Defense Advanced Research Projects Agency (DARPA) and the Army Research Office (ARO) and was accomplished under Cooperative Agreement Number W911NF-20-2-0002. The views and conclusions contained in this document are those of the authors and should not be interpreted as representing the official policies, either expressed or implied, of the DARPA or ARO, or the U.S. Government. The U.S. Government is authorized to reproduce and distribute reprints for Government purposes notwithstanding any copyright notation herein.

\bibliographystyle{bibliography} 
\bibliography{cas-refs}

% \iffalse
\newpage
\appendix

\section{The Objects in NovPhy Tasks} \label{appendix:The Objects in NovPhy Tasks}

\begin{figure}[h!]
  \centering
  \includegraphics[width=0.6\linewidth]{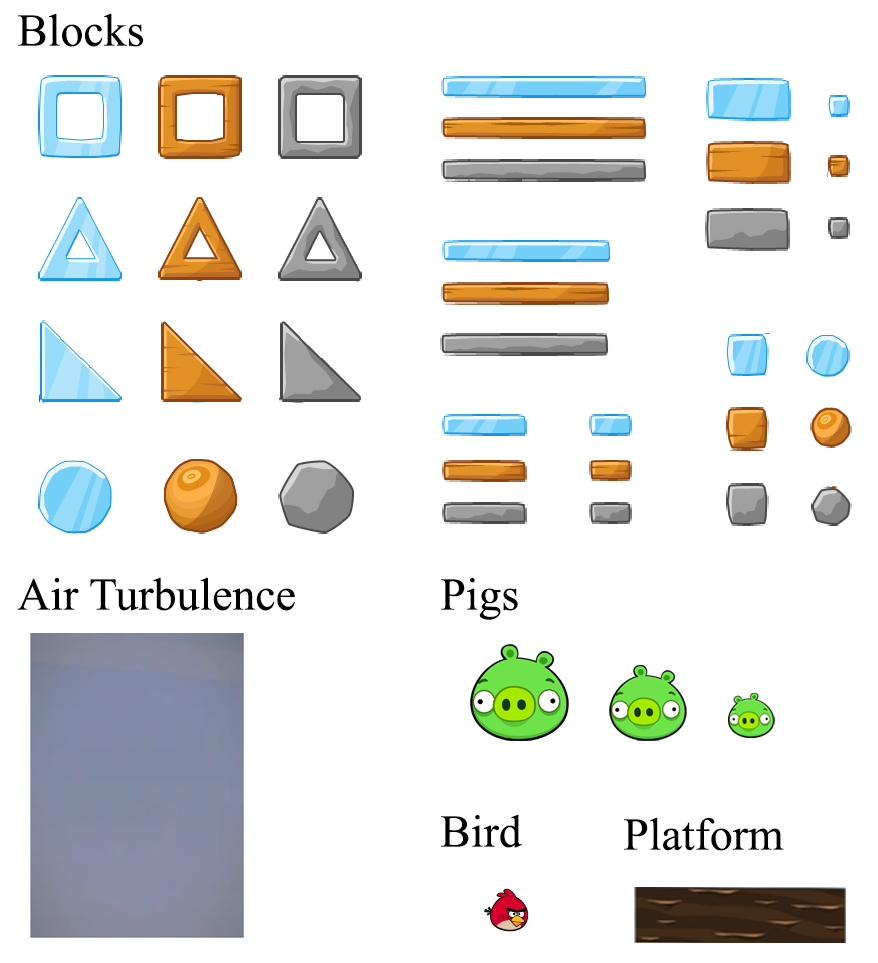}
  \caption{Objects that appear in normal tasks in  NovPhy (not to the scale). The shape and the size of the objects are fixed, except the platform object that can appear in different shapes and sizes.}
  \label{appendix_fig:all_normal_objects}
\end{figure}

\newpage

\begin{figure}[h!]
  \centering
  \includegraphics[width=0.6\linewidth]{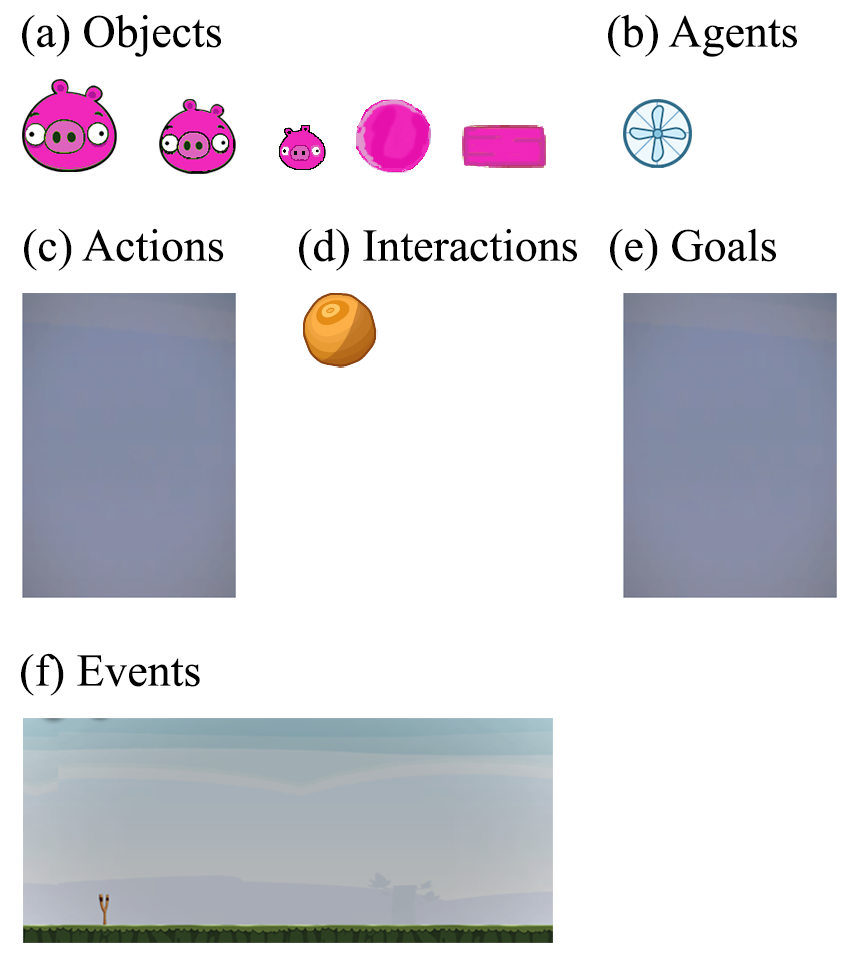}
  \caption{Objects associated with the novelties in NovPhy (not to the scale). Relations (the slingshot which is normally at the left side of the tasks is now at the right side of the tasks) and Environments (the gravity in the environment is now inverted) novelties are not shown here as they are not associated with a specific game object. Shown in the figures are, (a) Objects: a new pig/block that has a different colour (pink) to the normal pigs/blocks, (b) Agents: a novel external agent, Fan, that blows air (horizontally from left to right) affecting the moving path of objects, (c) Actions: the Air Turbulence external agent increases the magnitude of its upward force (has the same appearance of the Air Turbulence agent), (d) Interactions:  existing circular wood object now has magnetic properties: repels objects of its type and attracts other object types (has the same appearance of the circular wood object), (e) Goals: the Air Turbulence external agent changes its goal from pushing objects up to pushing objects down (has the same appearance of the Air Turbulence agent), (f) Events: when the first bird is dead, a storm occurs that applies a force to the right direction affecting the motion of the objects (figure shows the appearance of the environment when the storm has occurred).}

  \label{appendix_fig:all_novel_objects}
\end{figure}

\newpage
\section{Tasks in NovPhy} \label{Appendix: Section Tasks in NovPhy}

\begin{figure}[h!]
  \centering
  \begin{subfigure}[b]{0.49\columnwidth}
      \begin{subfigure}[b]{0.49\columnwidth}
        \includegraphics[width=\linewidth]{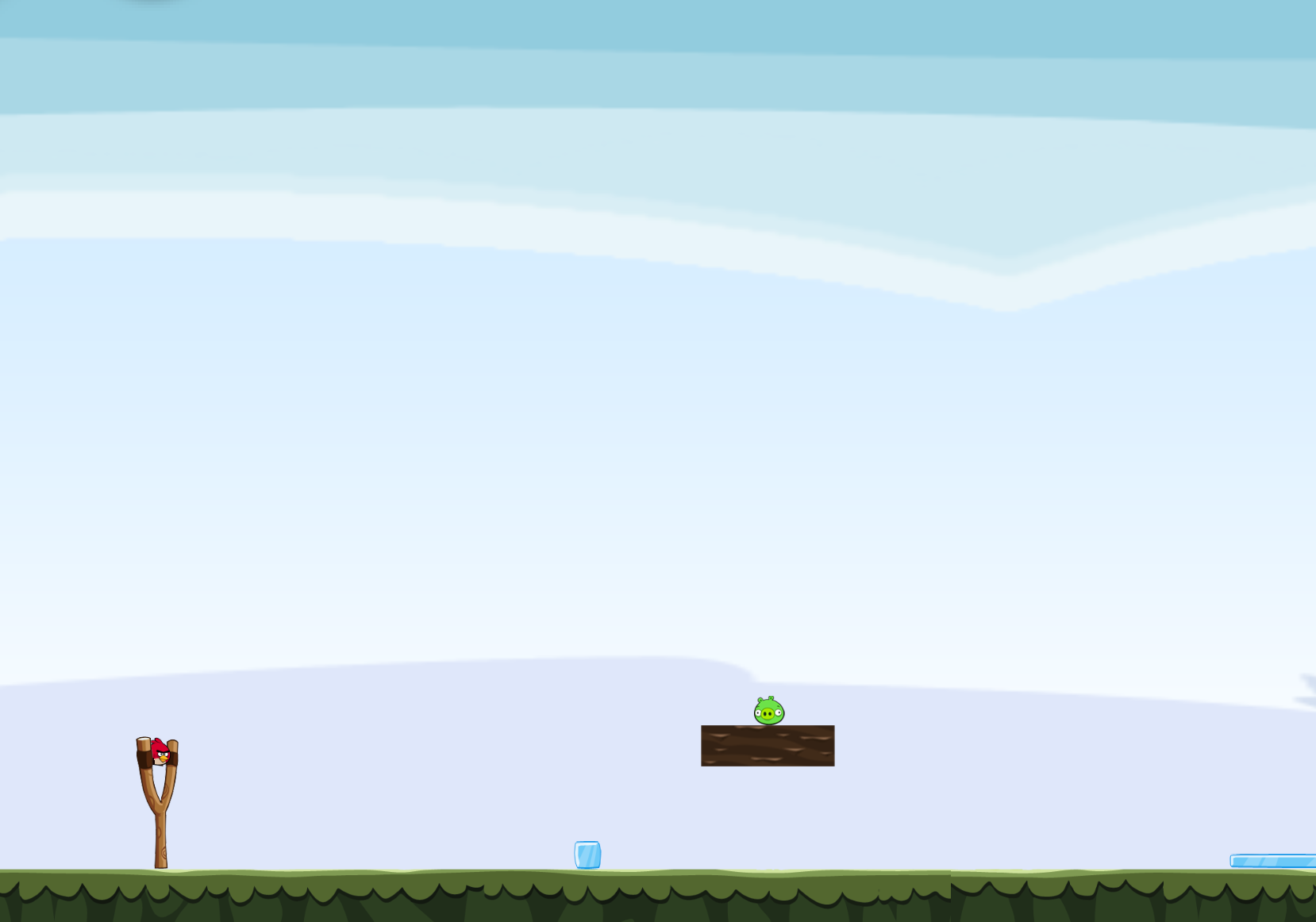}
      \end{subfigure}
      \begin{subfigure}[b]{0.49\columnwidth}
        \includegraphics[width=\linewidth]{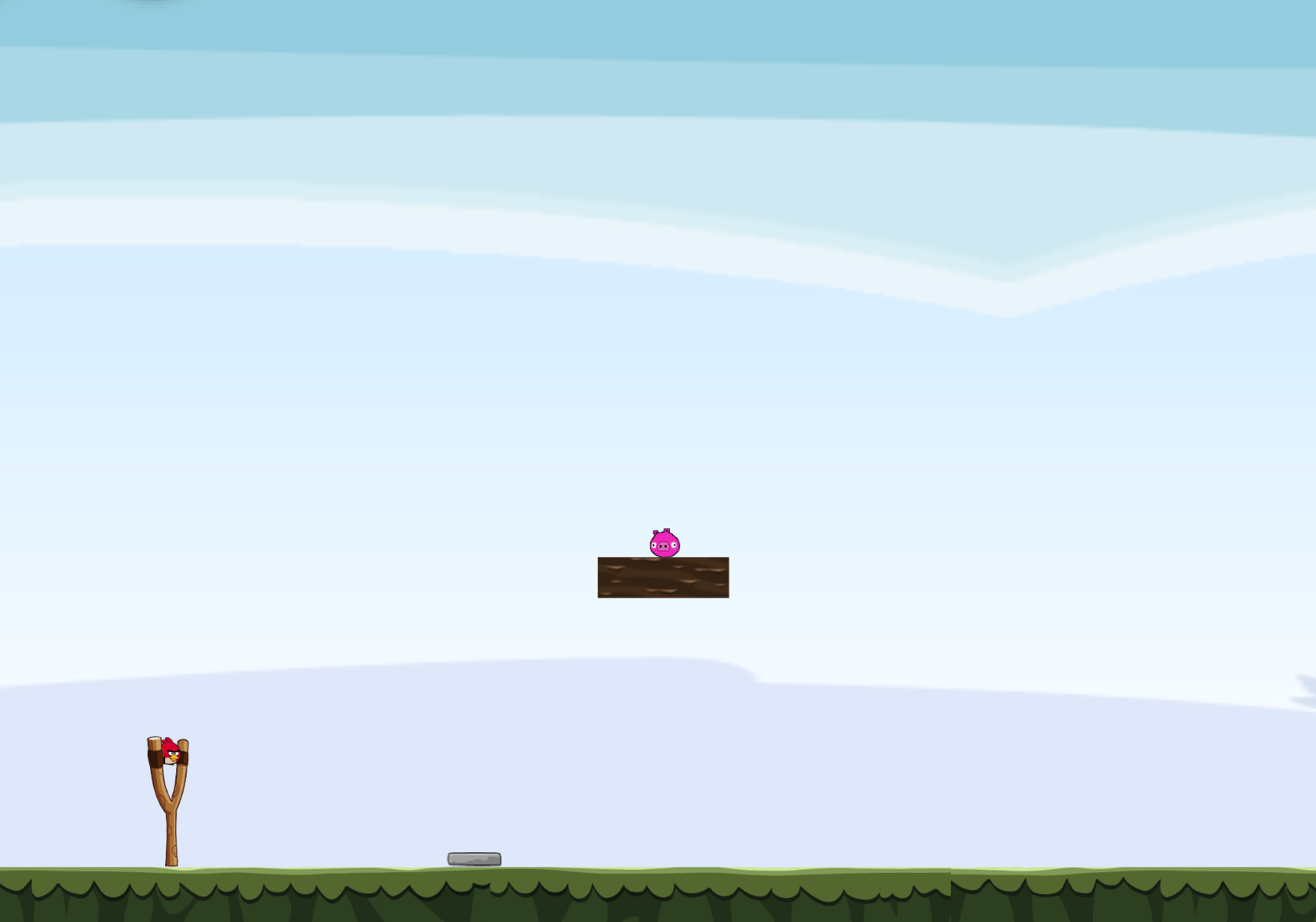}
      \end{subfigure}
  \caption{Objects}
  \end{subfigure}
  \begin{subfigure}[b]{0.49\columnwidth}
      \begin{subfigure}[b]{0.49\columnwidth}
        \includegraphics[width=\linewidth]{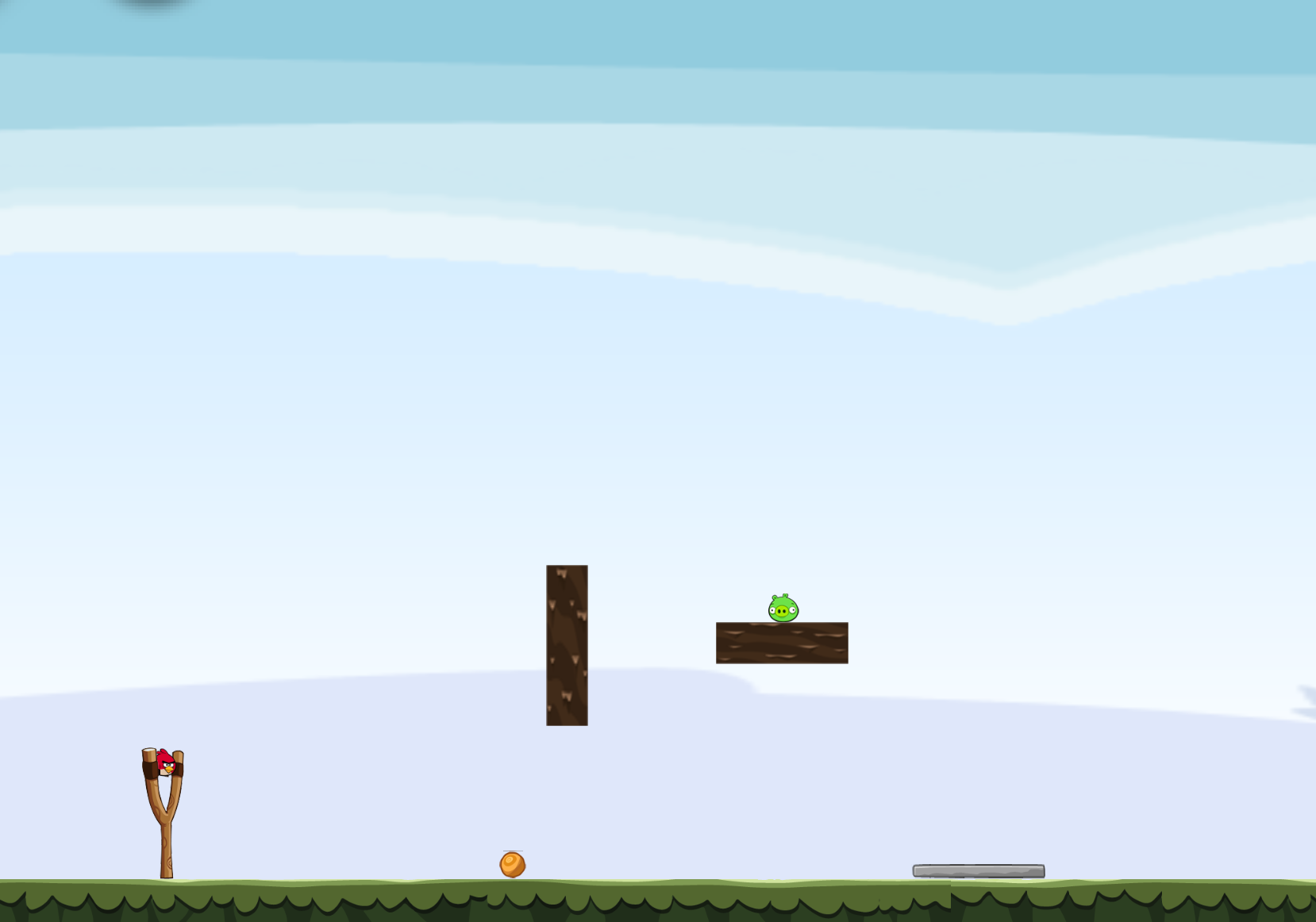}
      \end{subfigure}
      \begin{subfigure}[b]{0.49\columnwidth}
        \includegraphics[width=\linewidth]{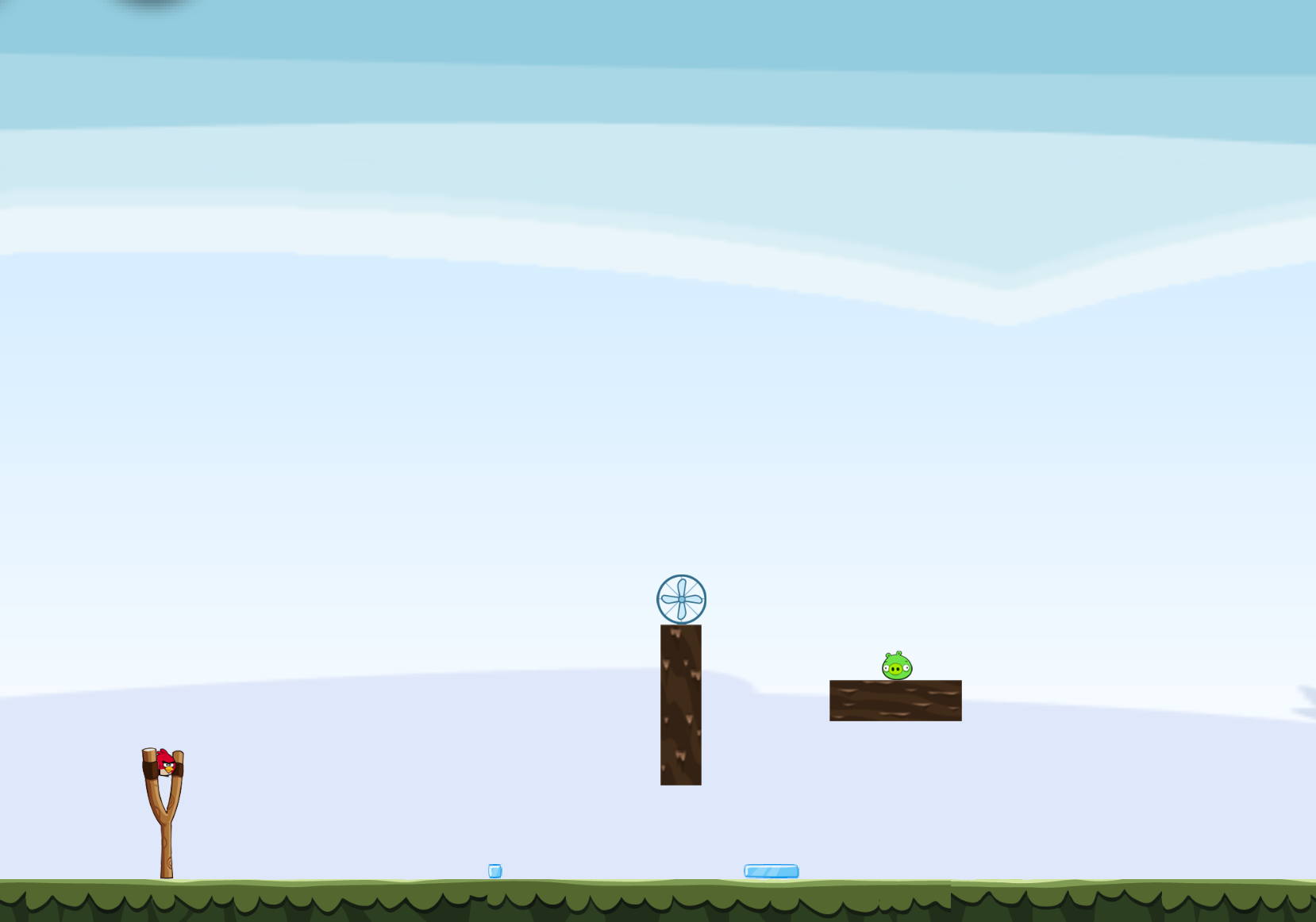}
      \end{subfigure}
  \caption{Agents}
  \end{subfigure}
  \begin{subfigure}[b]{0.49\columnwidth}
      \begin{subfigure}[b]{0.49\columnwidth}
        \includegraphics[width=\linewidth]{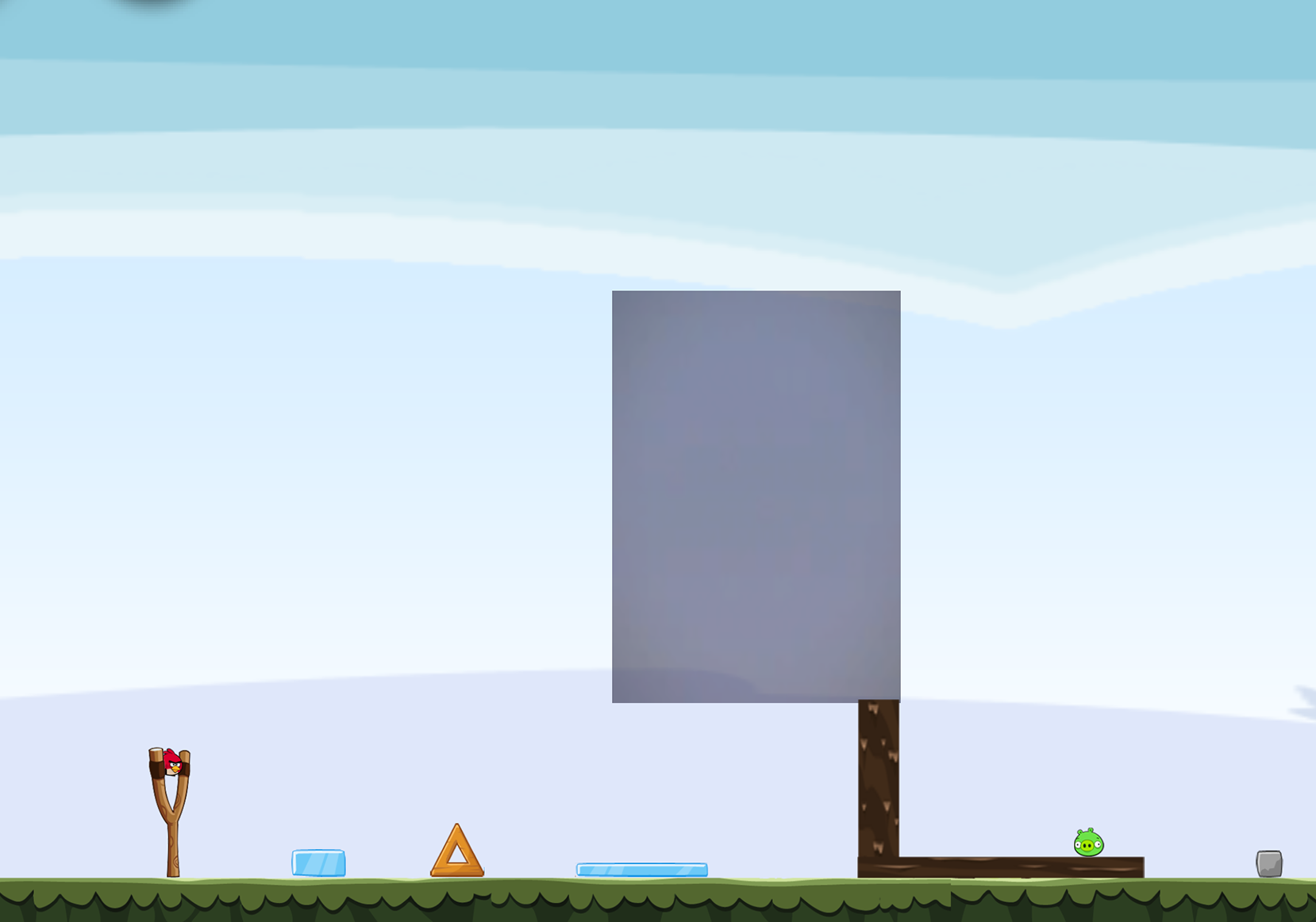}
      \end{subfigure}
      \begin{subfigure}[b]{0.49\columnwidth}
        \includegraphics[width=\linewidth]{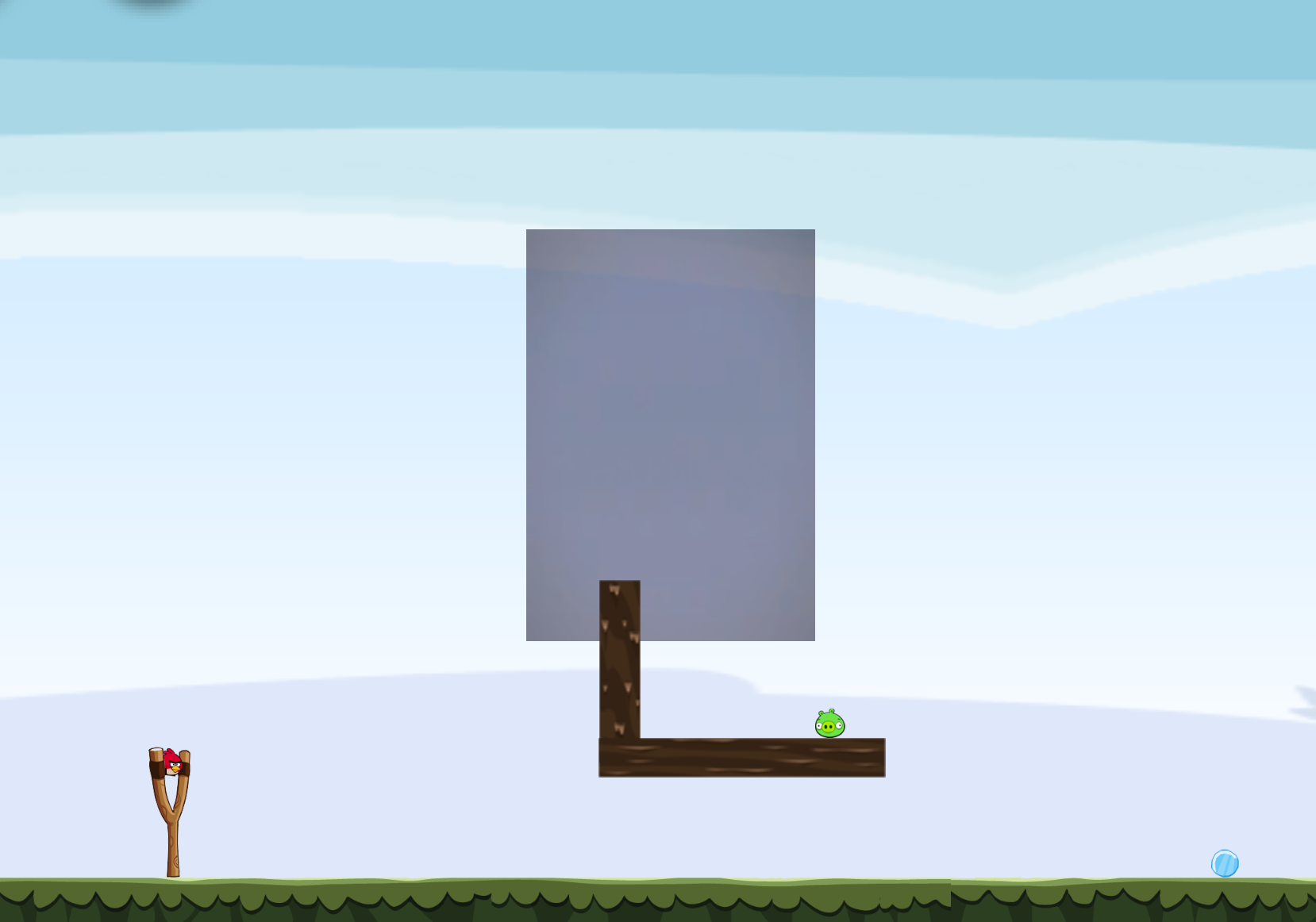}
      \end{subfigure}
  \caption{Actions}
  \end{subfigure}
  \begin{subfigure}[b]{0.49\columnwidth}
      \begin{subfigure}[b]{0.49\columnwidth}
        \includegraphics[width=\linewidth]{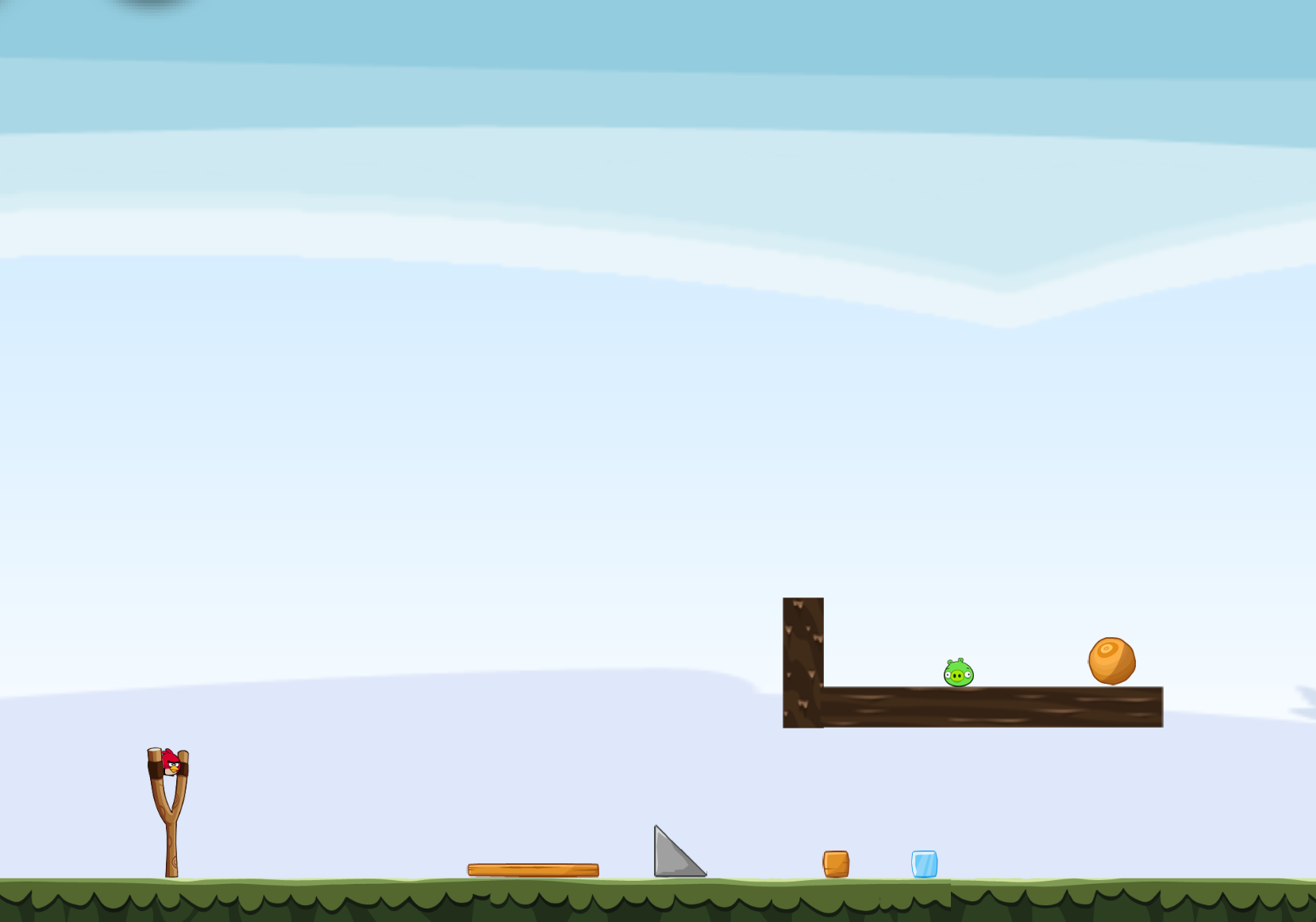}
      \end{subfigure}
      \begin{subfigure}[b]{0.49\columnwidth}
        \includegraphics[width=\linewidth]{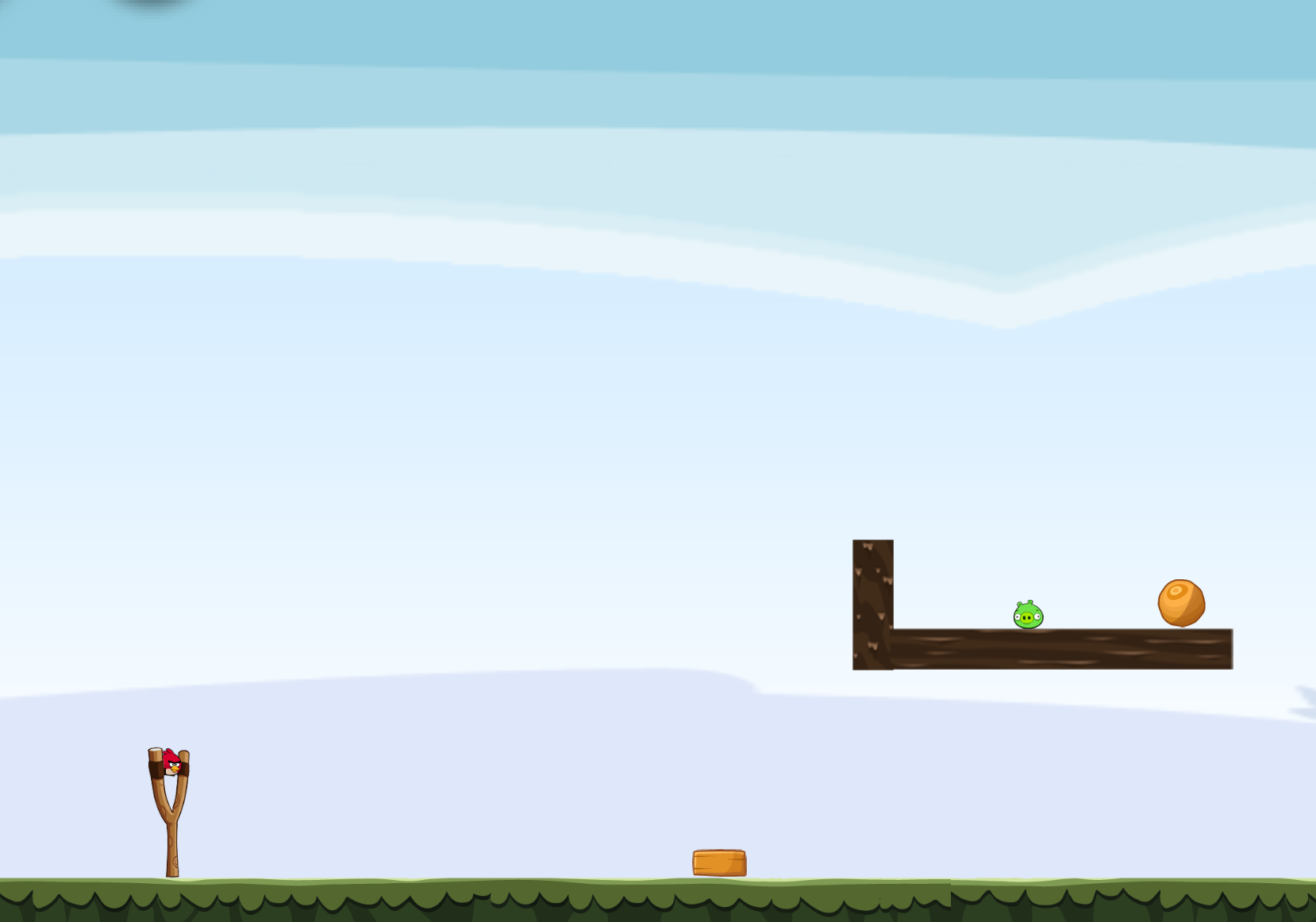}
      \end{subfigure}
  \caption{Interactions}
  \end{subfigure}
    \begin{subfigure}[b]{0.49\columnwidth}
      \begin{subfigure}[b]{0.49\columnwidth}
        \includegraphics[width=\linewidth]{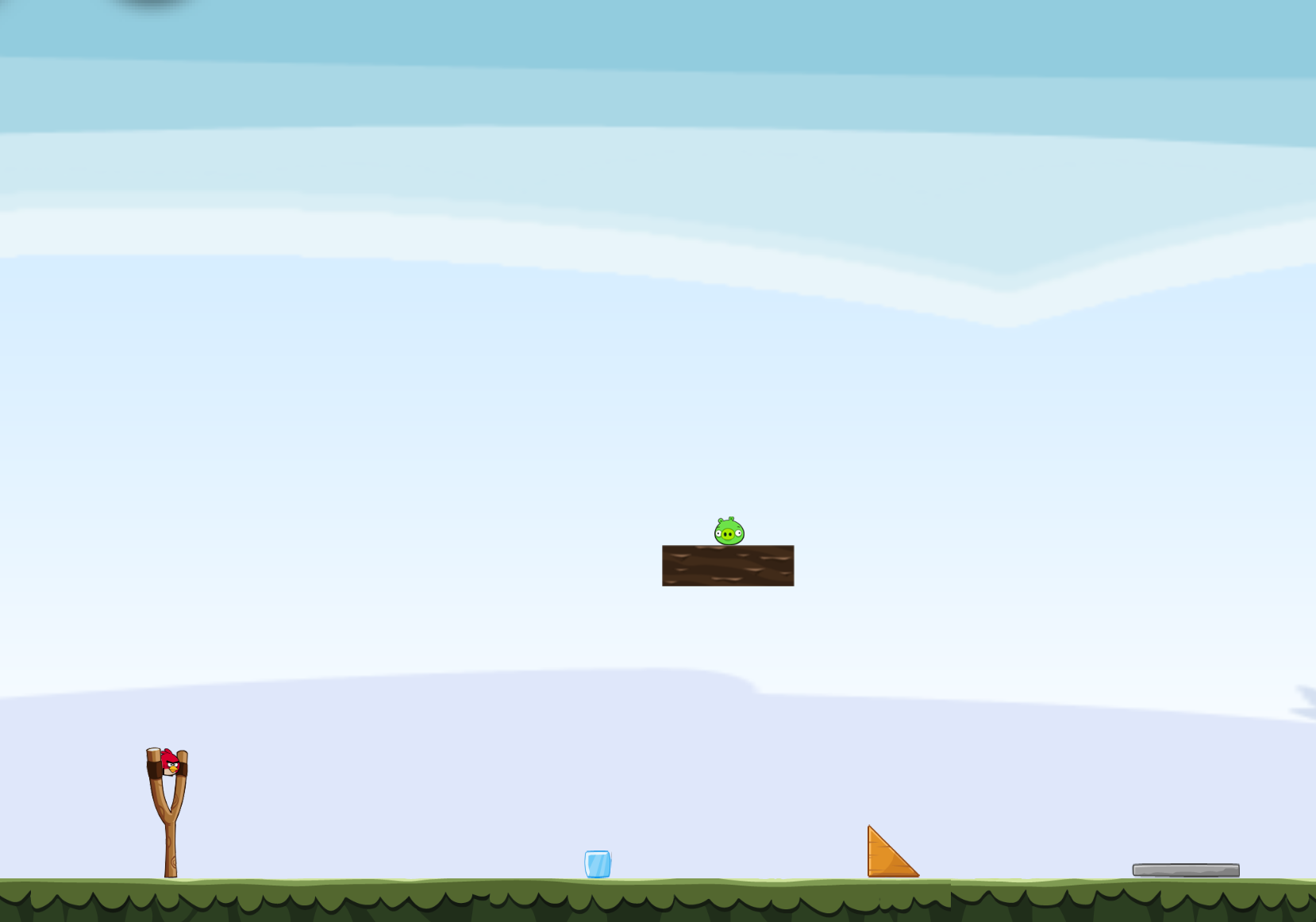}
      \end{subfigure}
      \begin{subfigure}[b]{0.49\columnwidth}
        \includegraphics[width=\linewidth]{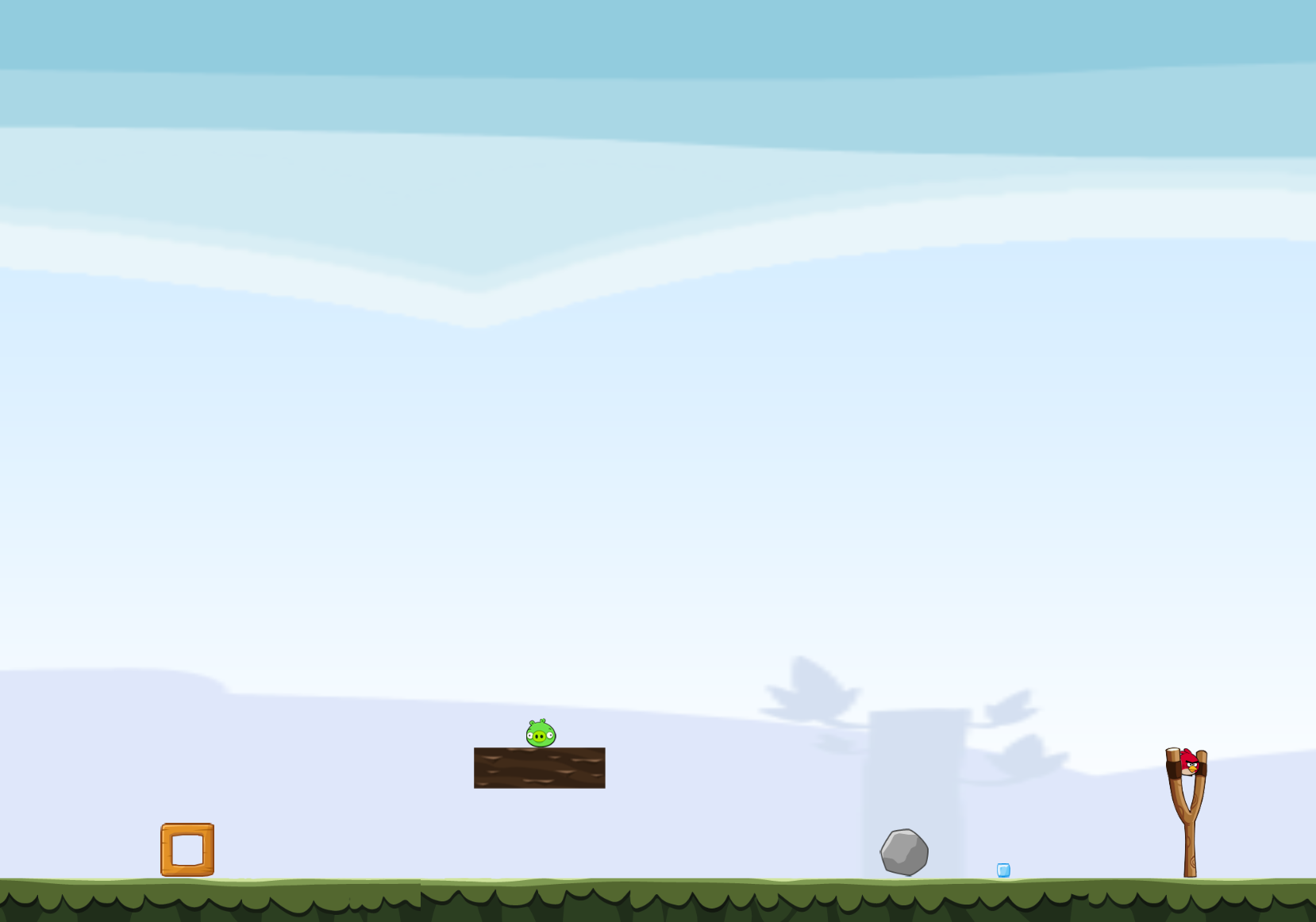}
      \end{subfigure}
  \caption{Relations}
  \end{subfigure}
  \begin{subfigure}[b]{0.49\columnwidth}
      \begin{subfigure}[b]{0.49\columnwidth}
        \includegraphics[width=\linewidth]{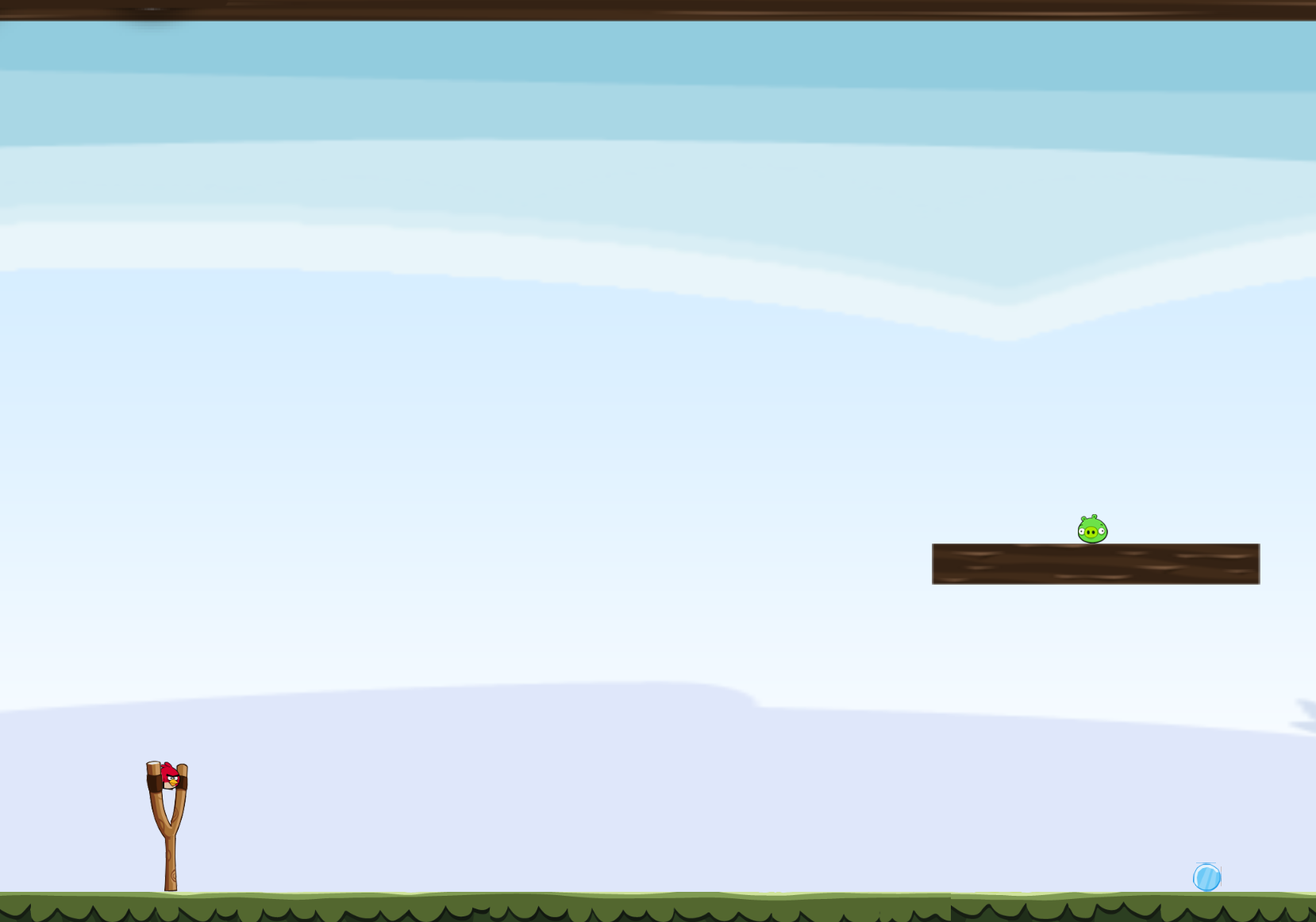}
      \end{subfigure}
      \begin{subfigure}[b]{0.49\columnwidth}
        \includegraphics[width=\linewidth]{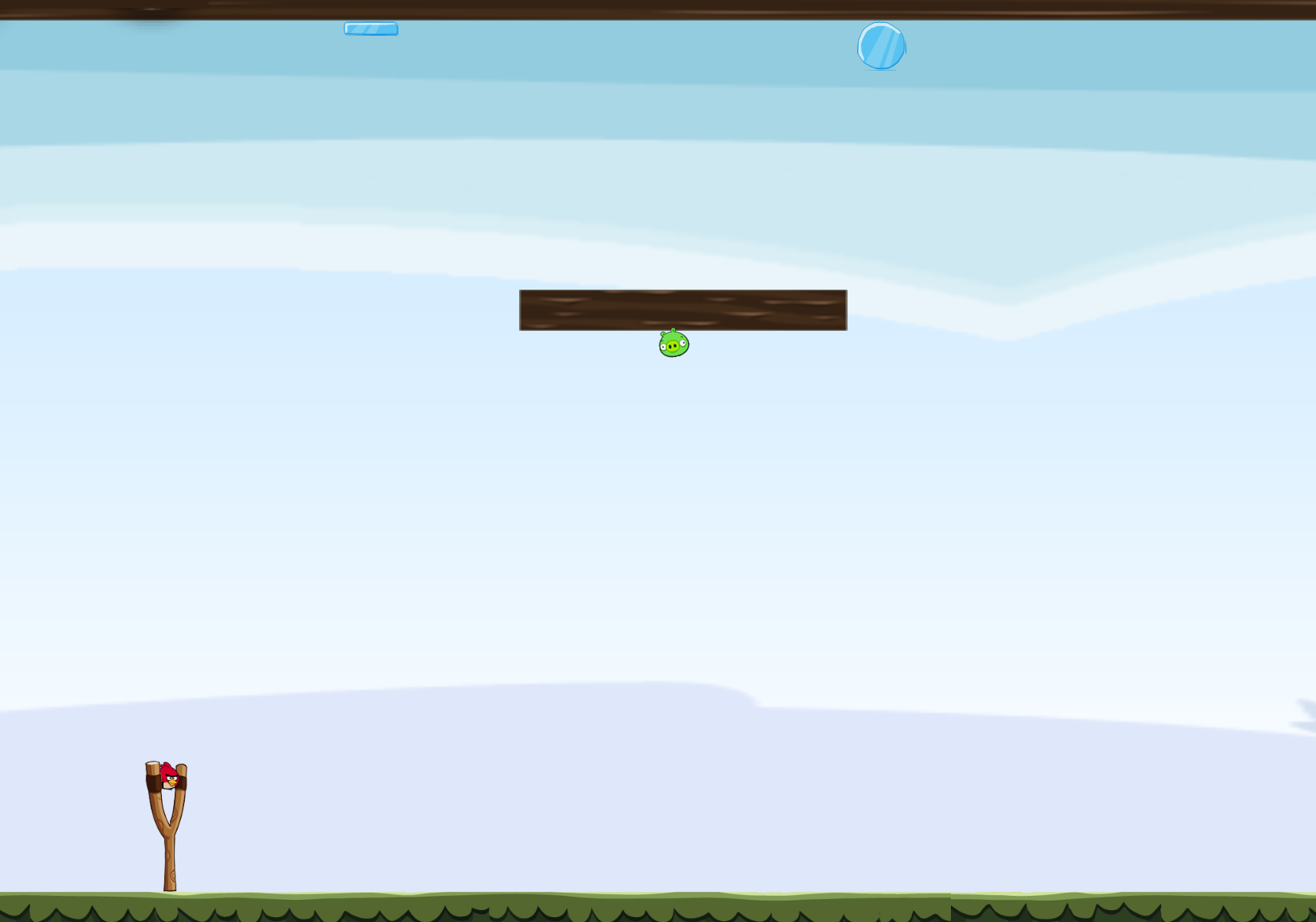}
      \end{subfigure}
  \caption{Environments}
  \end{subfigure}
  \begin{subfigure}[b]{0.49\columnwidth}
      \begin{subfigure}[b]{0.49\columnwidth}
        \includegraphics[width=\linewidth]{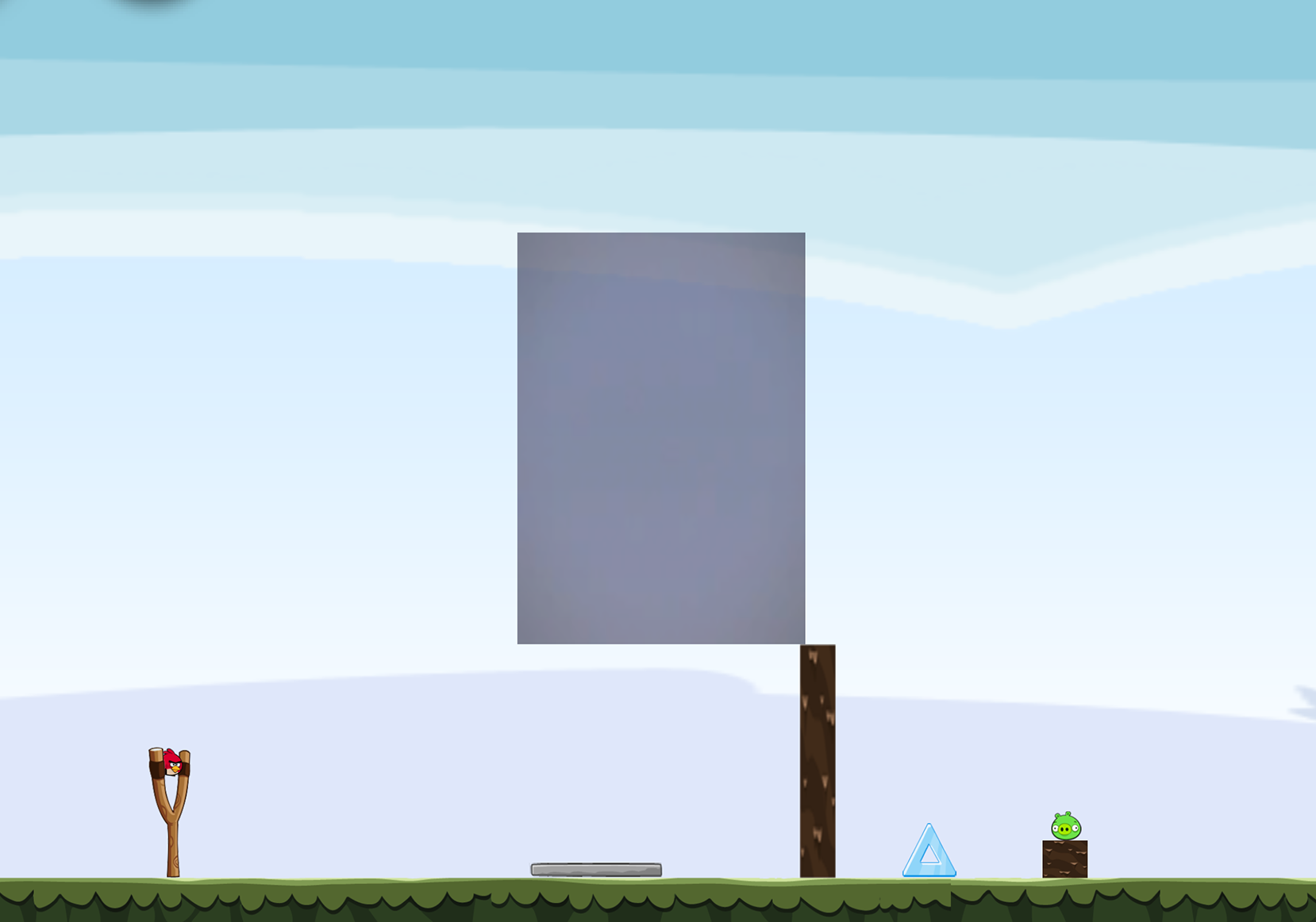}
      \end{subfigure}
      \begin{subfigure}[b]{0.49\columnwidth}
        \includegraphics[width=\linewidth]{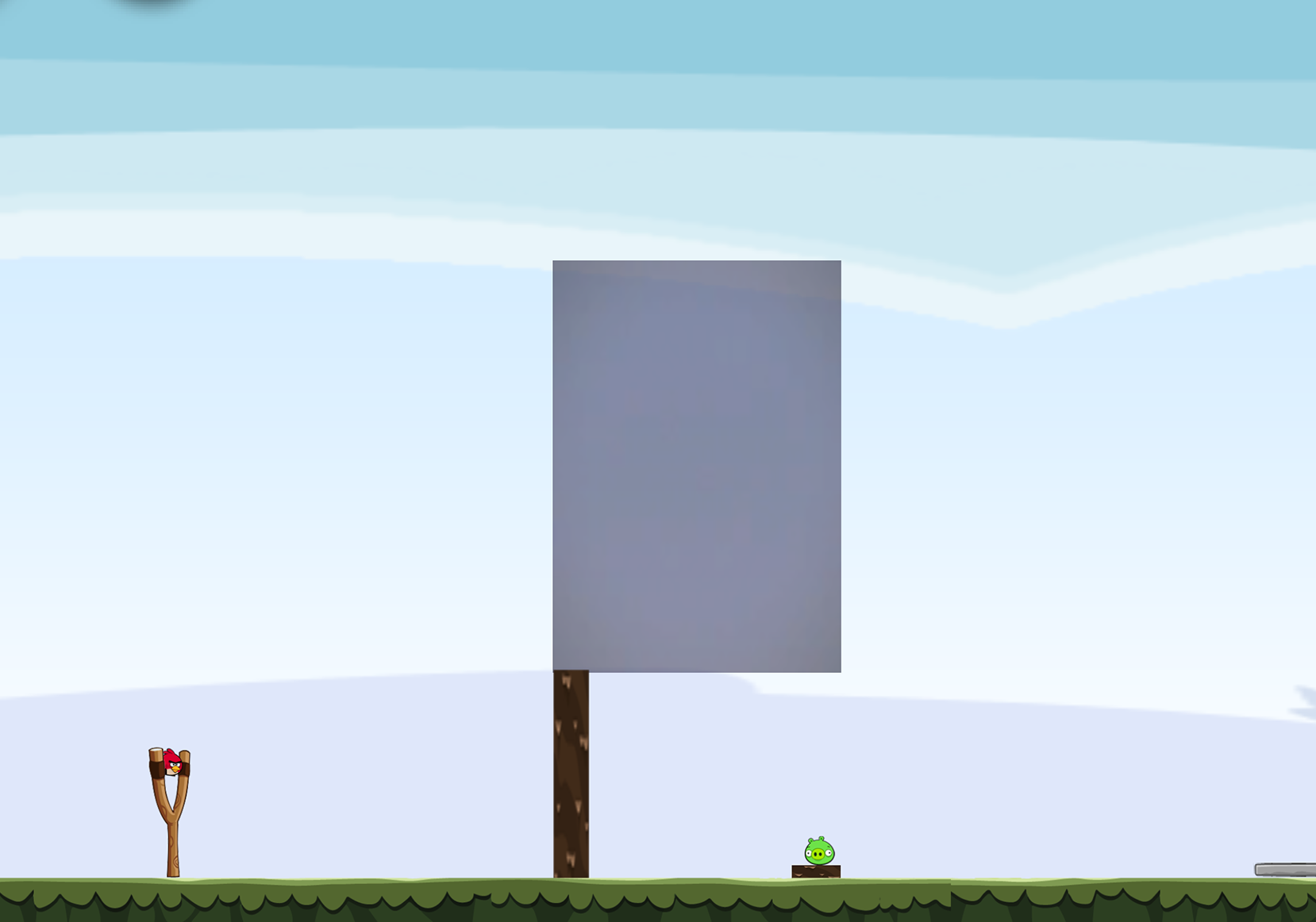}
      \end{subfigure}
  \caption{Goals}
  \end{subfigure}
  \begin{subfigure}[b]{0.49\columnwidth}
      \begin{subfigure}[b]{0.49\columnwidth}
        \includegraphics[width=\linewidth]{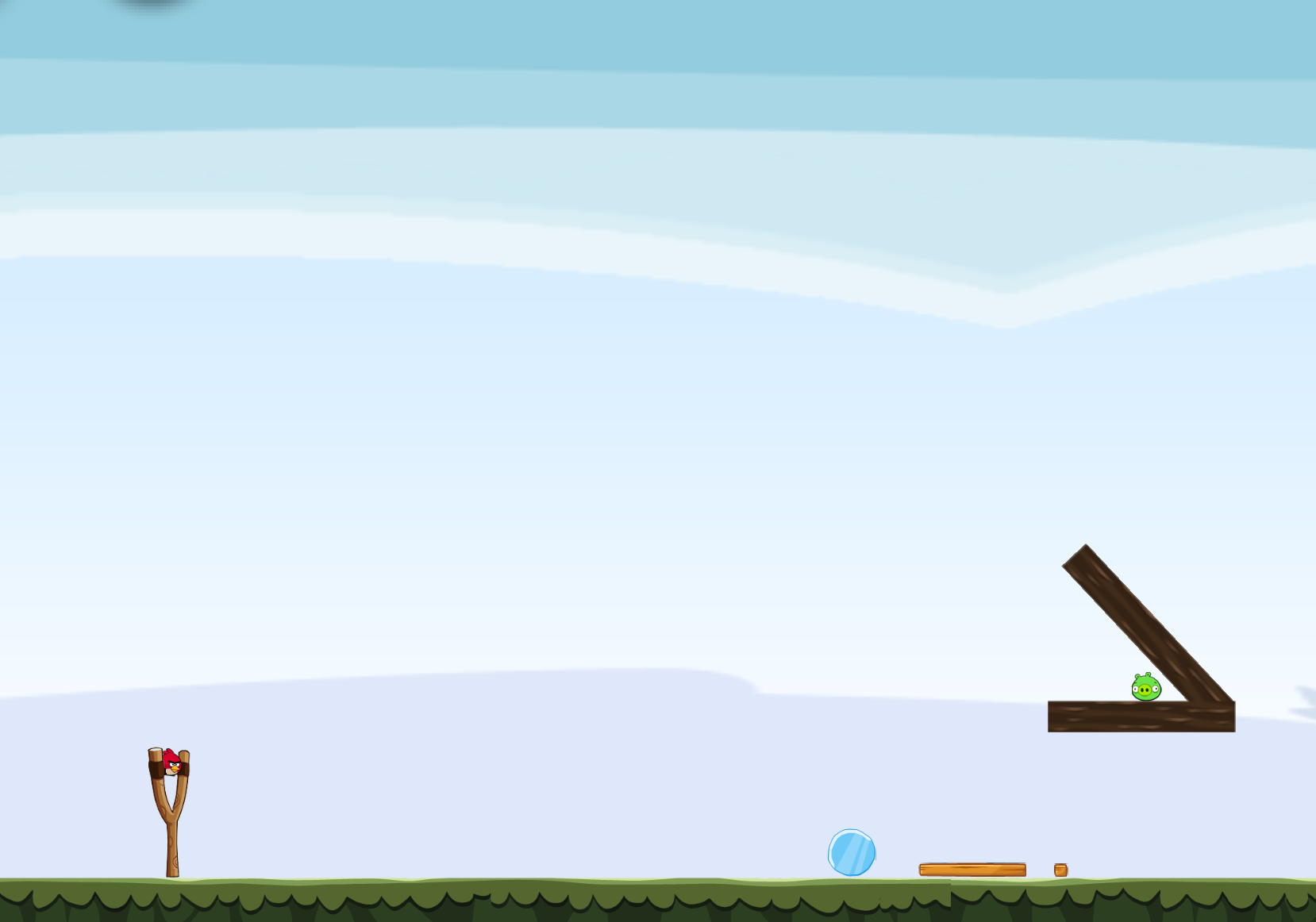}
      \end{subfigure}
      \begin{subfigure}[b]{0.49\columnwidth}
        \includegraphics[width=\linewidth]{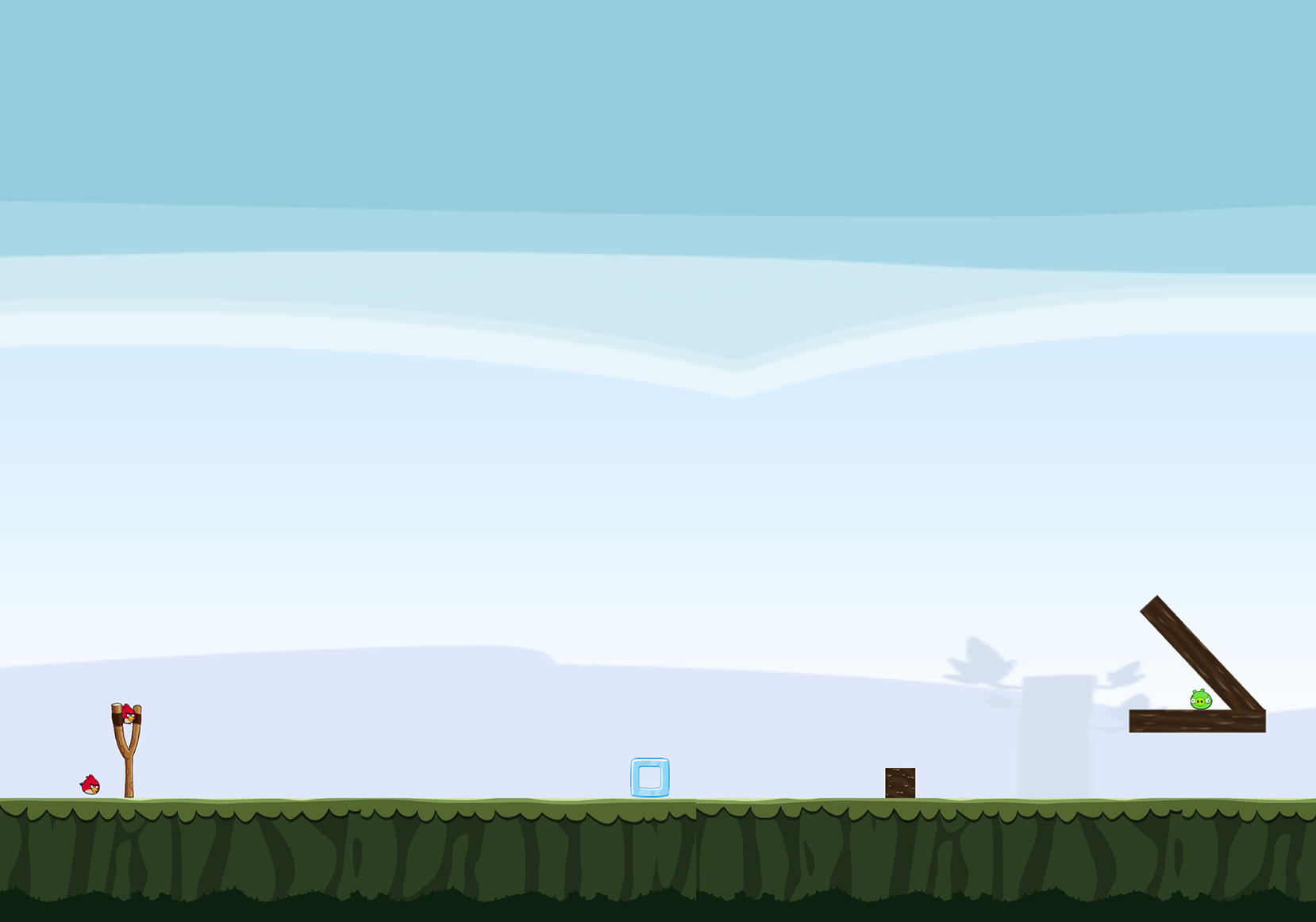}
      \end{subfigure}
  \caption{Events}
  \end{subfigure}
\caption{Task templates of the single force scenario with eight novelties applied to them. In each subfigure, the left figure is the normal task and the right figure is the corresponding novel task with the novelty applied.}
\label{appendix_fig:single_force}
\end{figure}

\begin{figure}[h!]
  \centering
  \begin{subfigure}[b]{0.49\columnwidth}
      \begin{subfigure}[b]{0.49\columnwidth}
        \includegraphics[width=\linewidth]{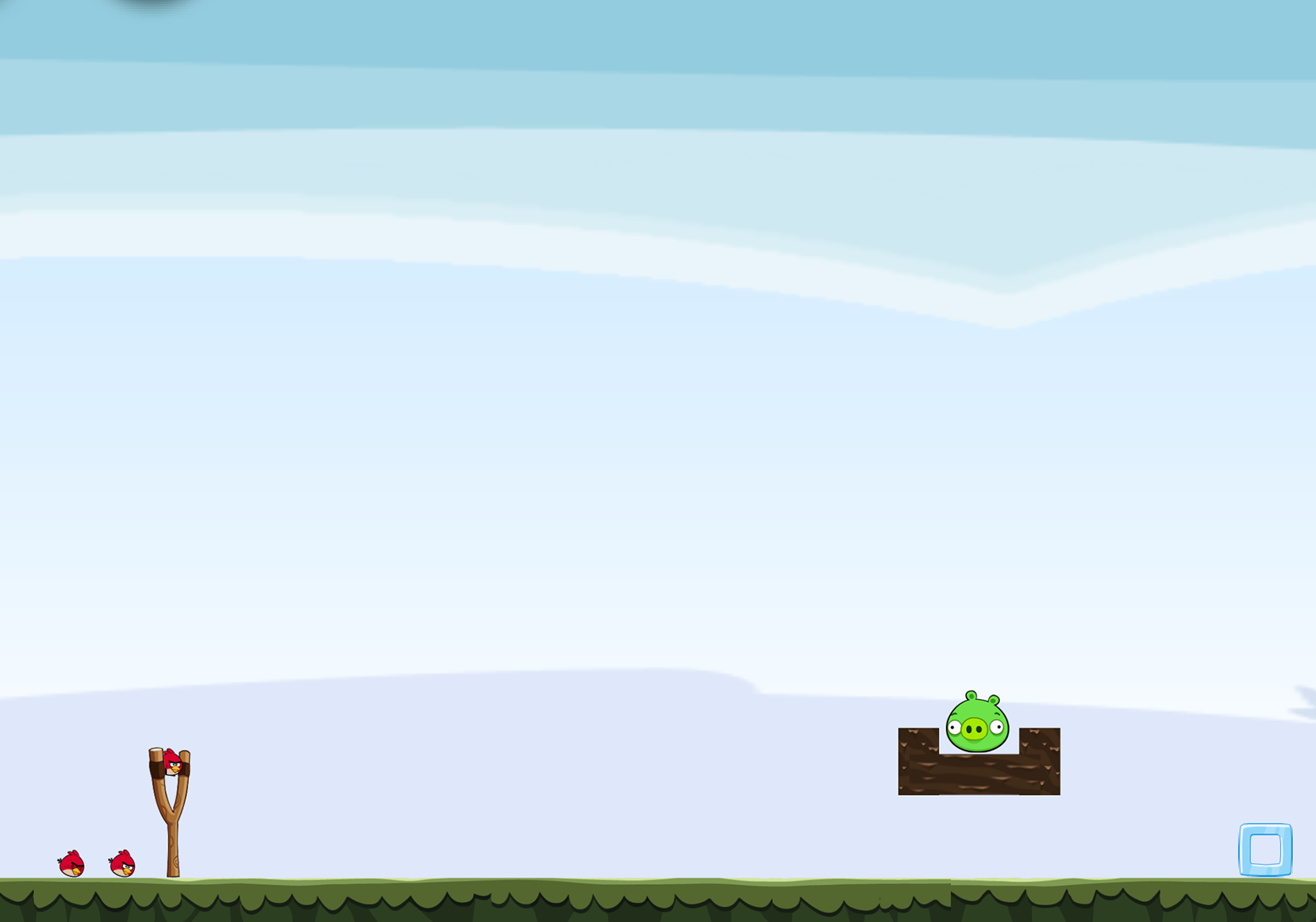}
      \end{subfigure}
      \begin{subfigure}[b]{0.49\columnwidth}
        \includegraphics[width=\linewidth]{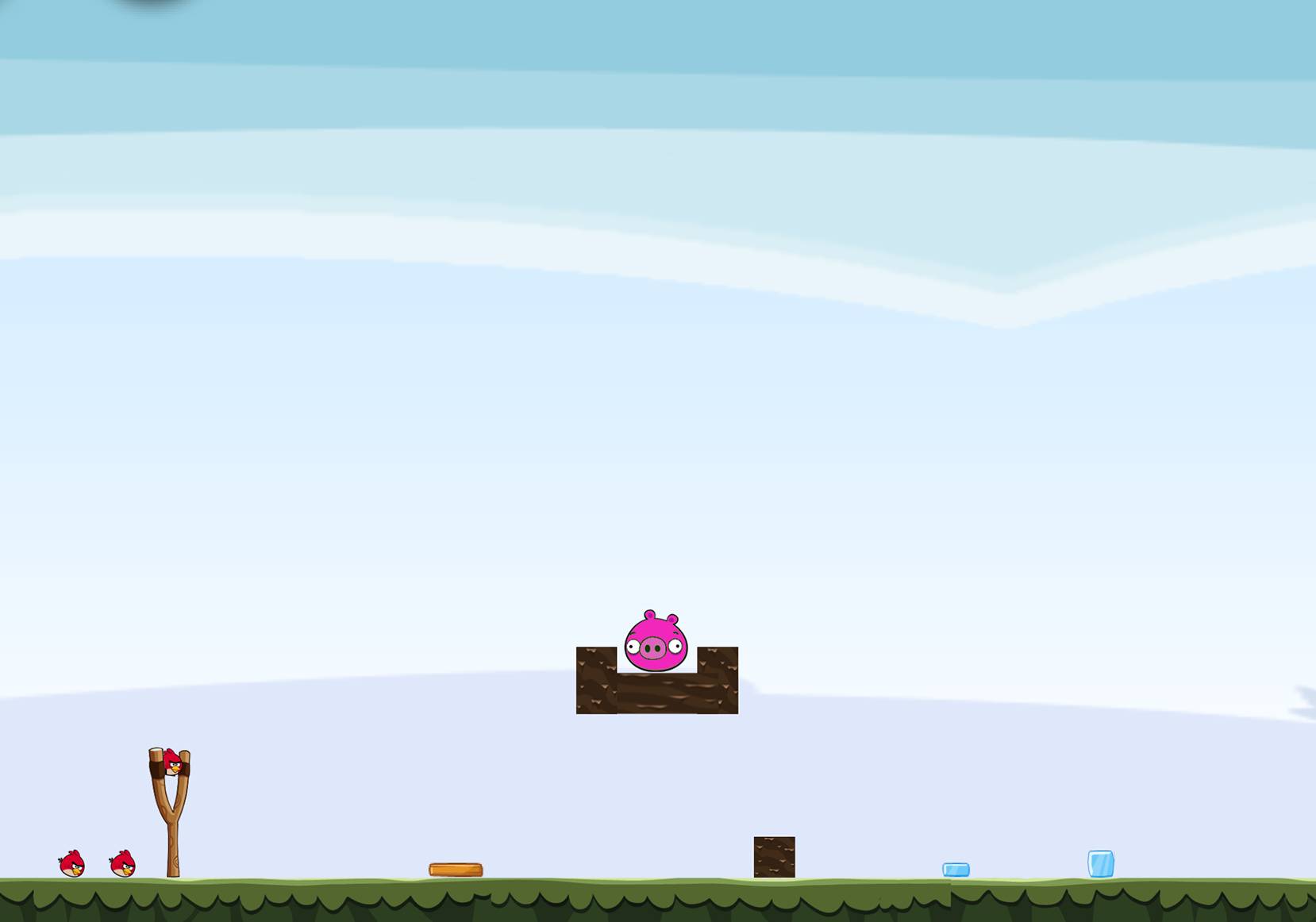}
      \end{subfigure}
  \caption{Objects}
  \end{subfigure}
  \begin{subfigure}[b]{0.49\columnwidth}
      \begin{subfigure}[b]{0.49\columnwidth}
        \includegraphics[width=\linewidth]{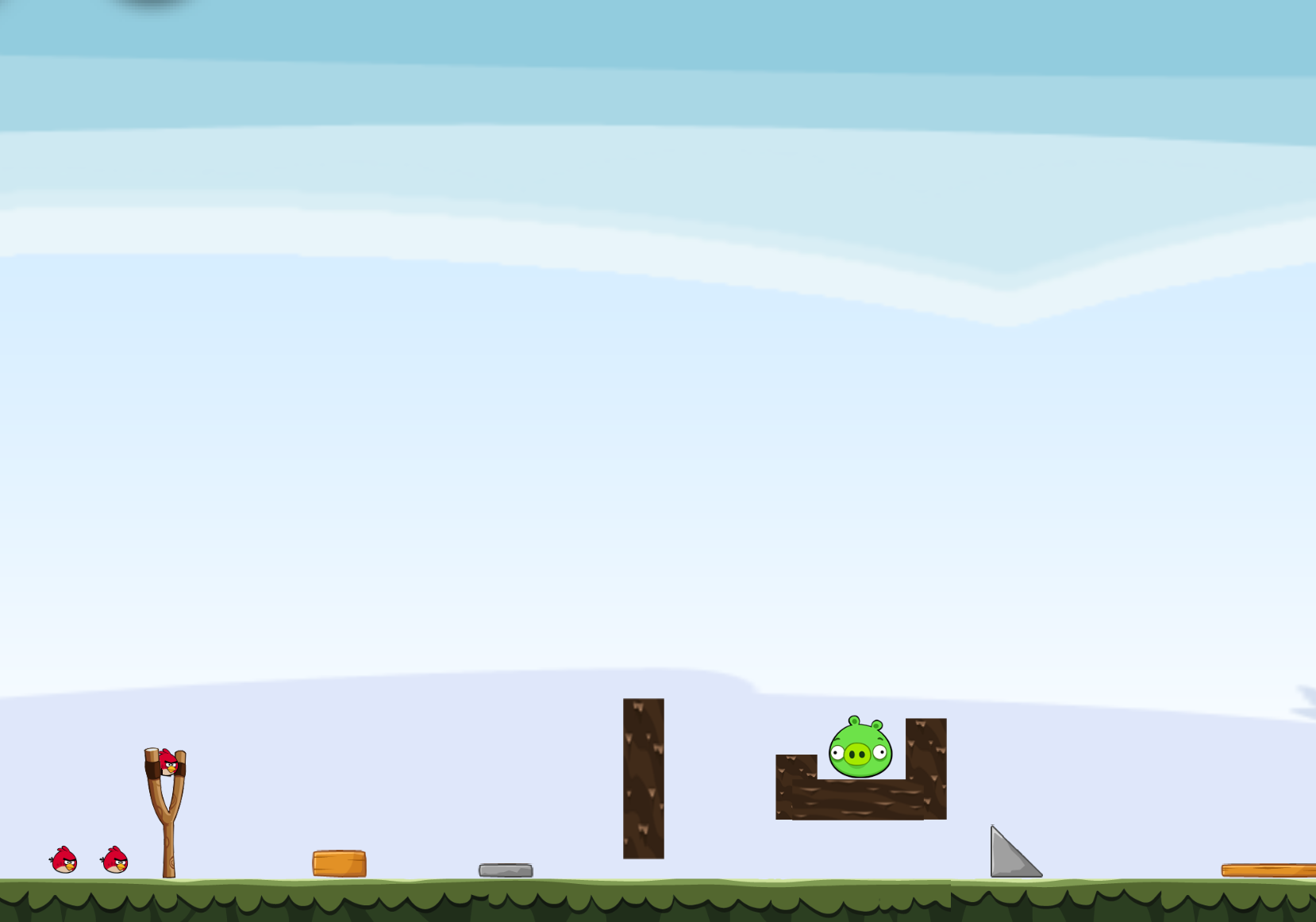}
      \end{subfigure}
      \begin{subfigure}[b]{0.49\columnwidth}
        \includegraphics[width=\linewidth]{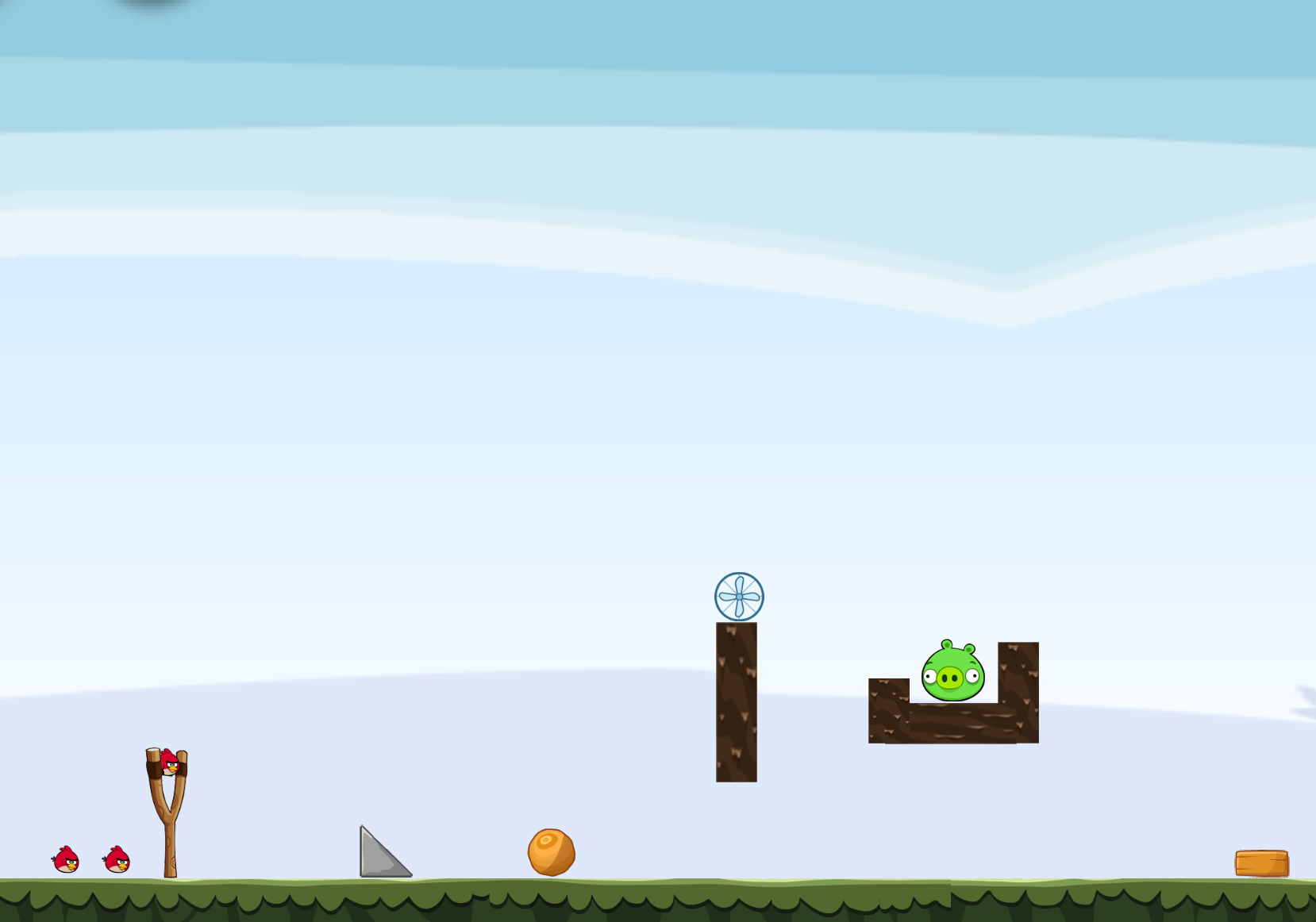}
      \end{subfigure}
  \caption{Agents}
  \end{subfigure}
  \begin{subfigure}[b]{0.49\columnwidth}
      \begin{subfigure}[b]{0.49\columnwidth}
        \includegraphics[width=\linewidth]{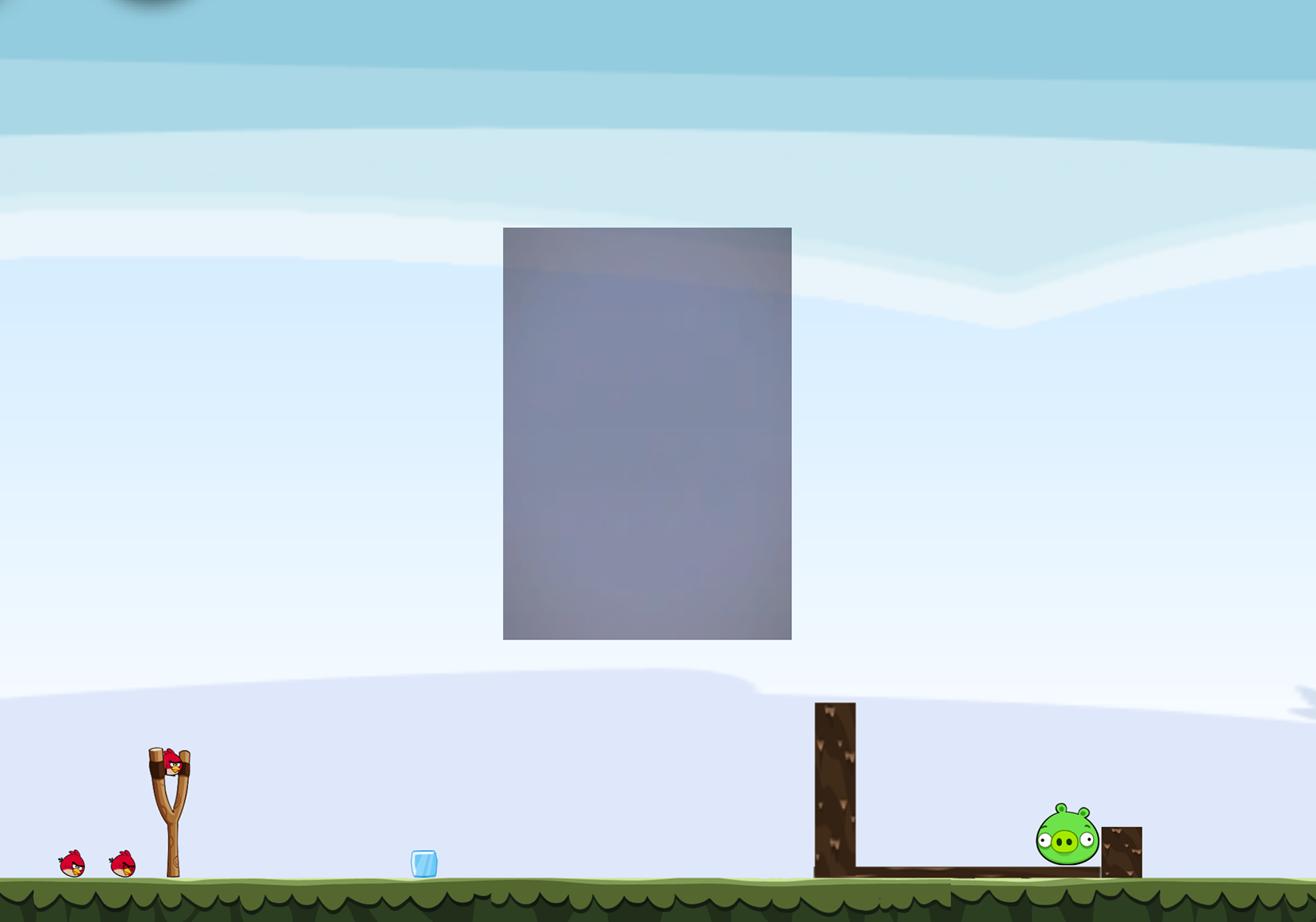}
      \end{subfigure}
      \begin{subfigure}[b]{0.49\columnwidth}
        \includegraphics[width=\linewidth]{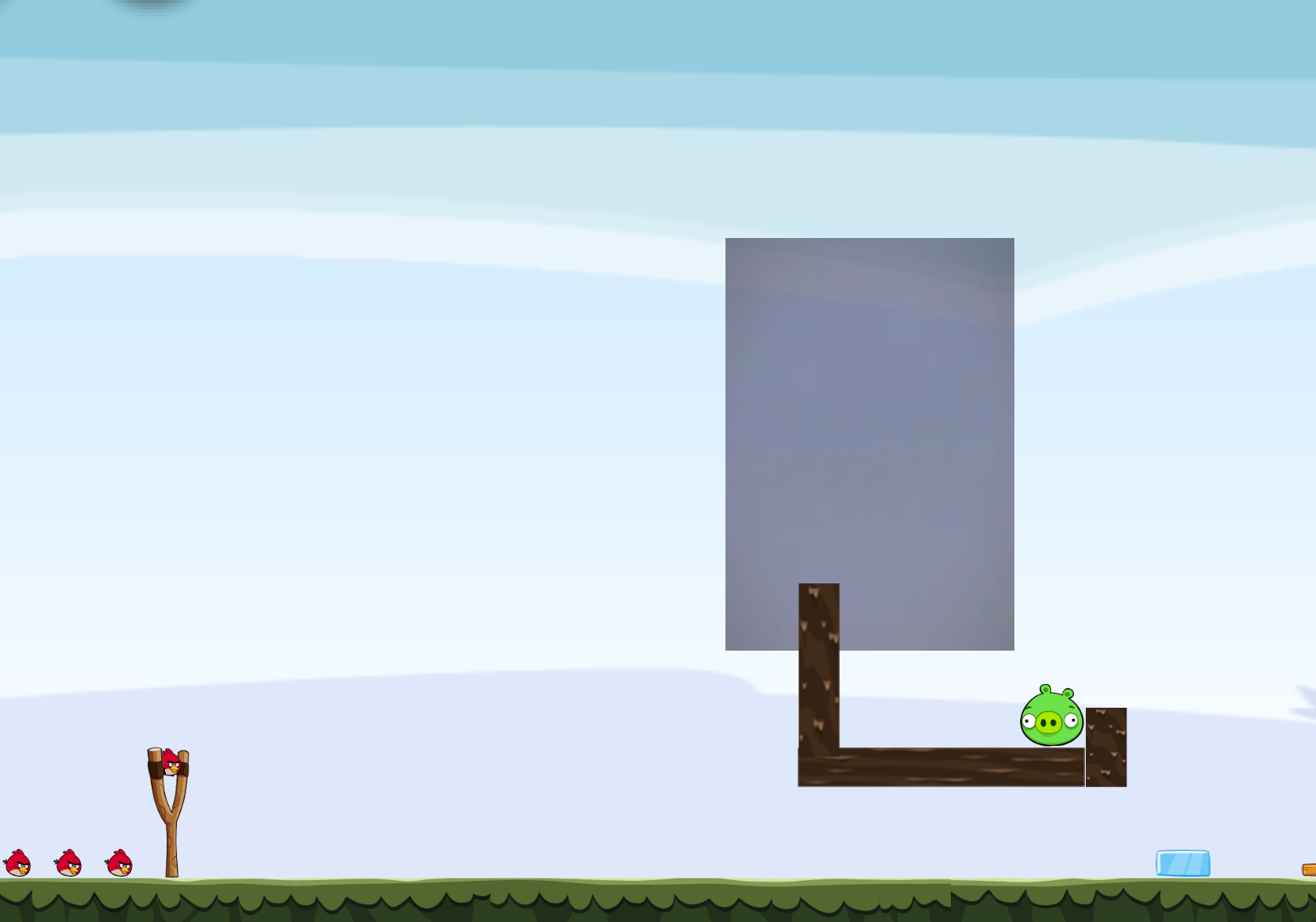}
      \end{subfigure}
  \caption{Actions}
  \end{subfigure}
  \begin{subfigure}[b]{0.49\columnwidth}
      \begin{subfigure}[b]{0.49\columnwidth}
        \includegraphics[width=\linewidth]{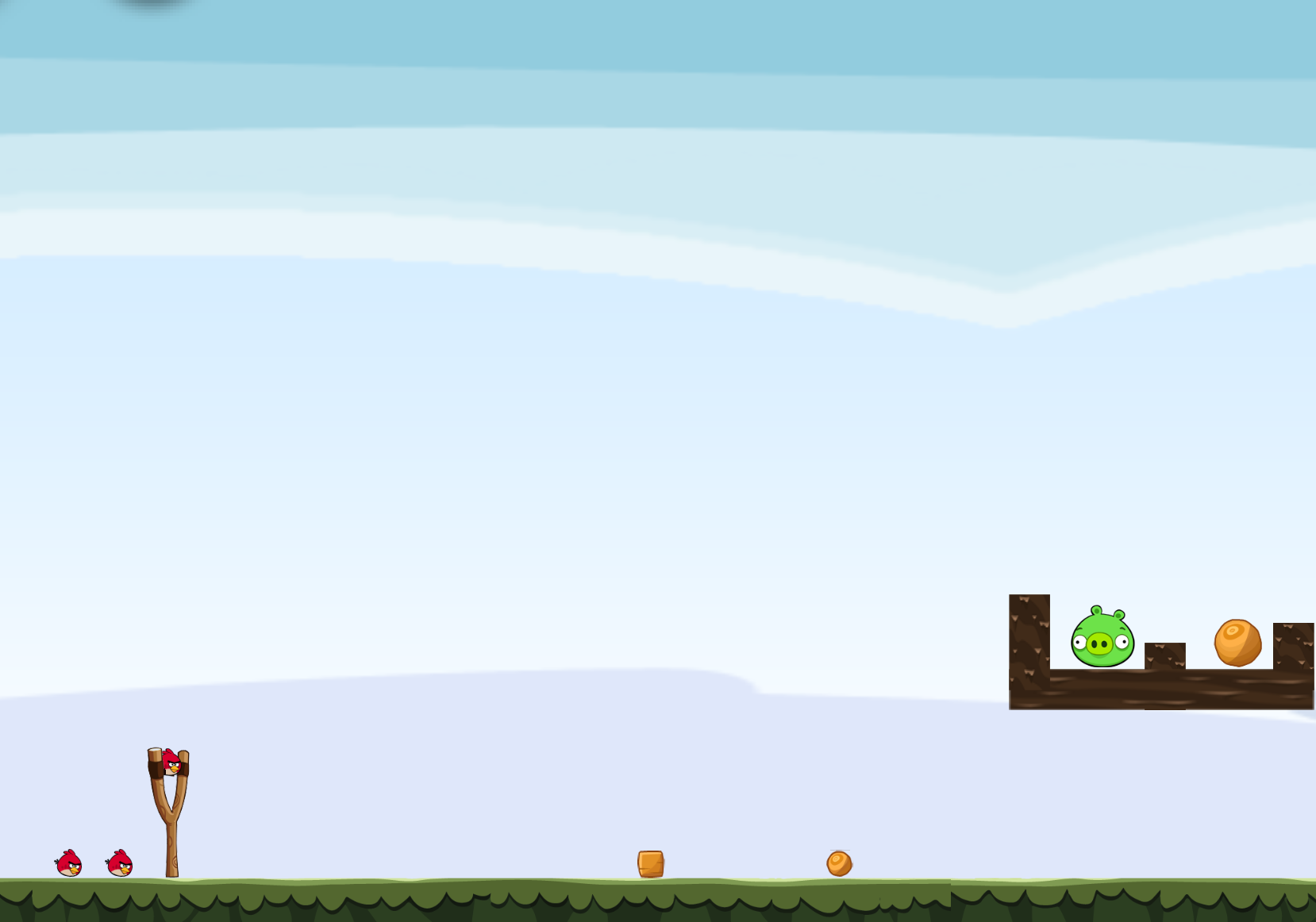}
      \end{subfigure}
      \begin{subfigure}[b]{0.49\columnwidth}
        \includegraphics[width=\linewidth]{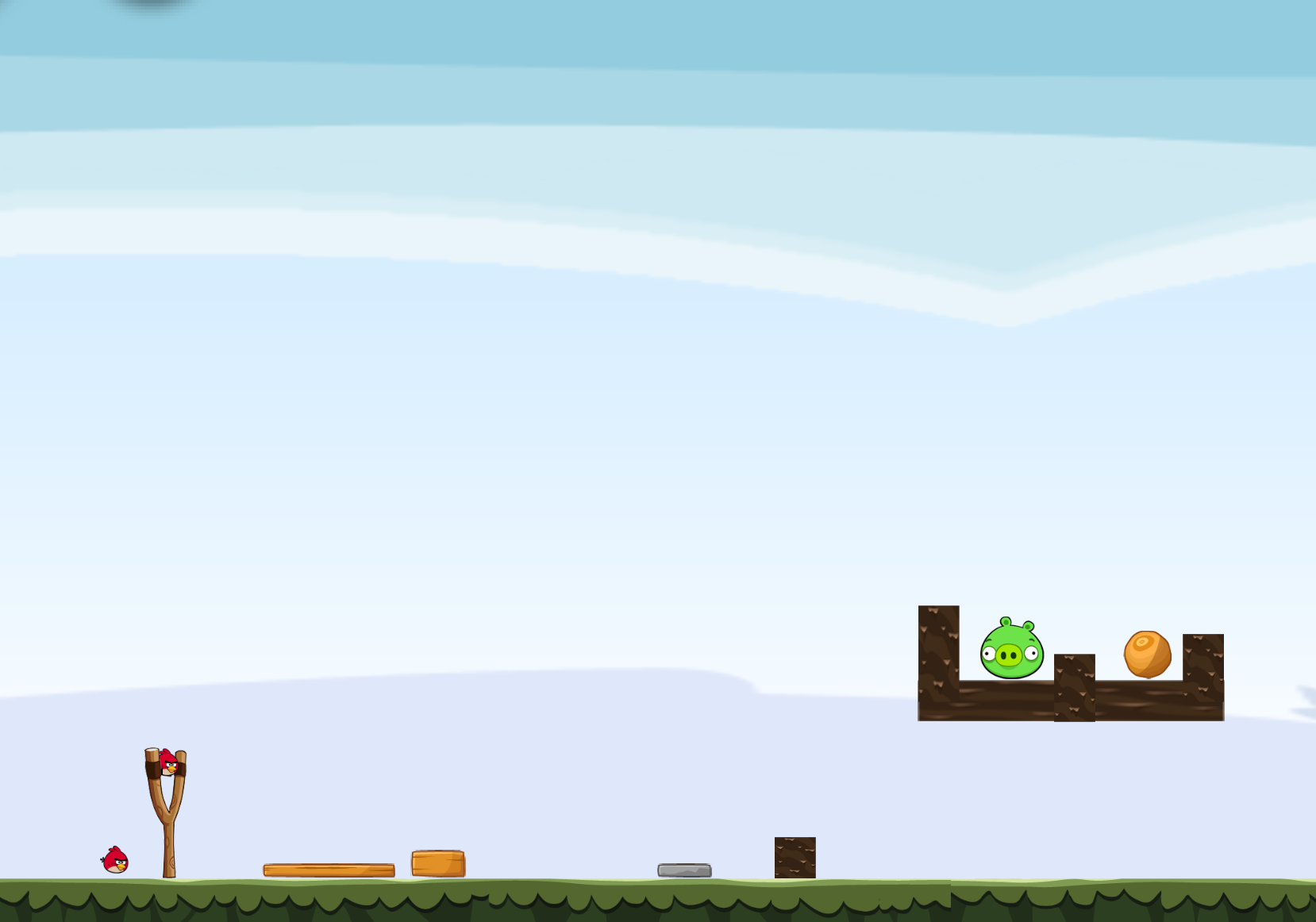}
      \end{subfigure}
  \caption{Interactions}
  \end{subfigure}
    \begin{subfigure}[b]{0.49\columnwidth}
      \begin{subfigure}[b]{0.49\columnwidth}
        \includegraphics[width=\linewidth]{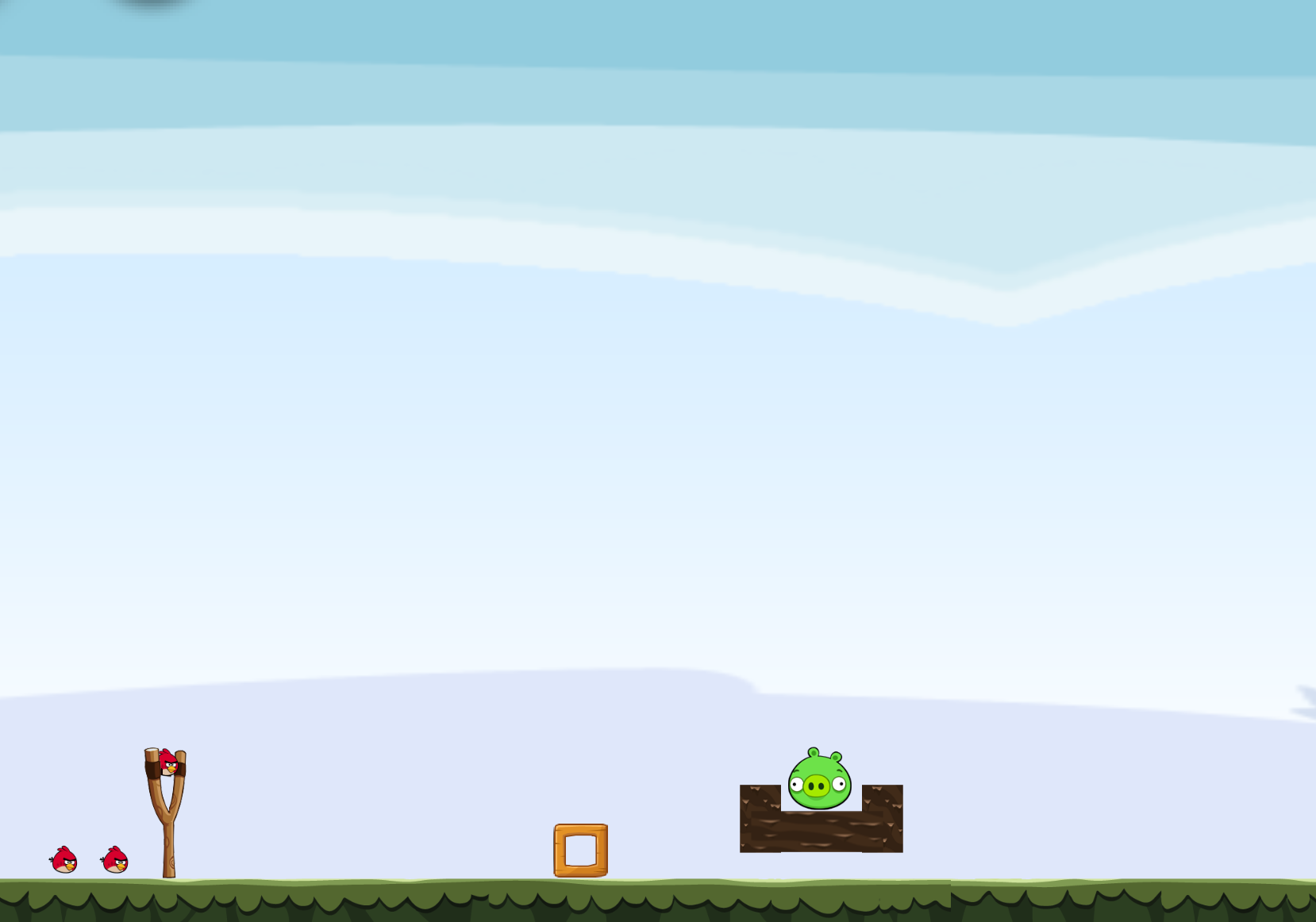}
      \end{subfigure}
      \begin{subfigure}[b]{0.49\columnwidth}
        \includegraphics[width=\linewidth]{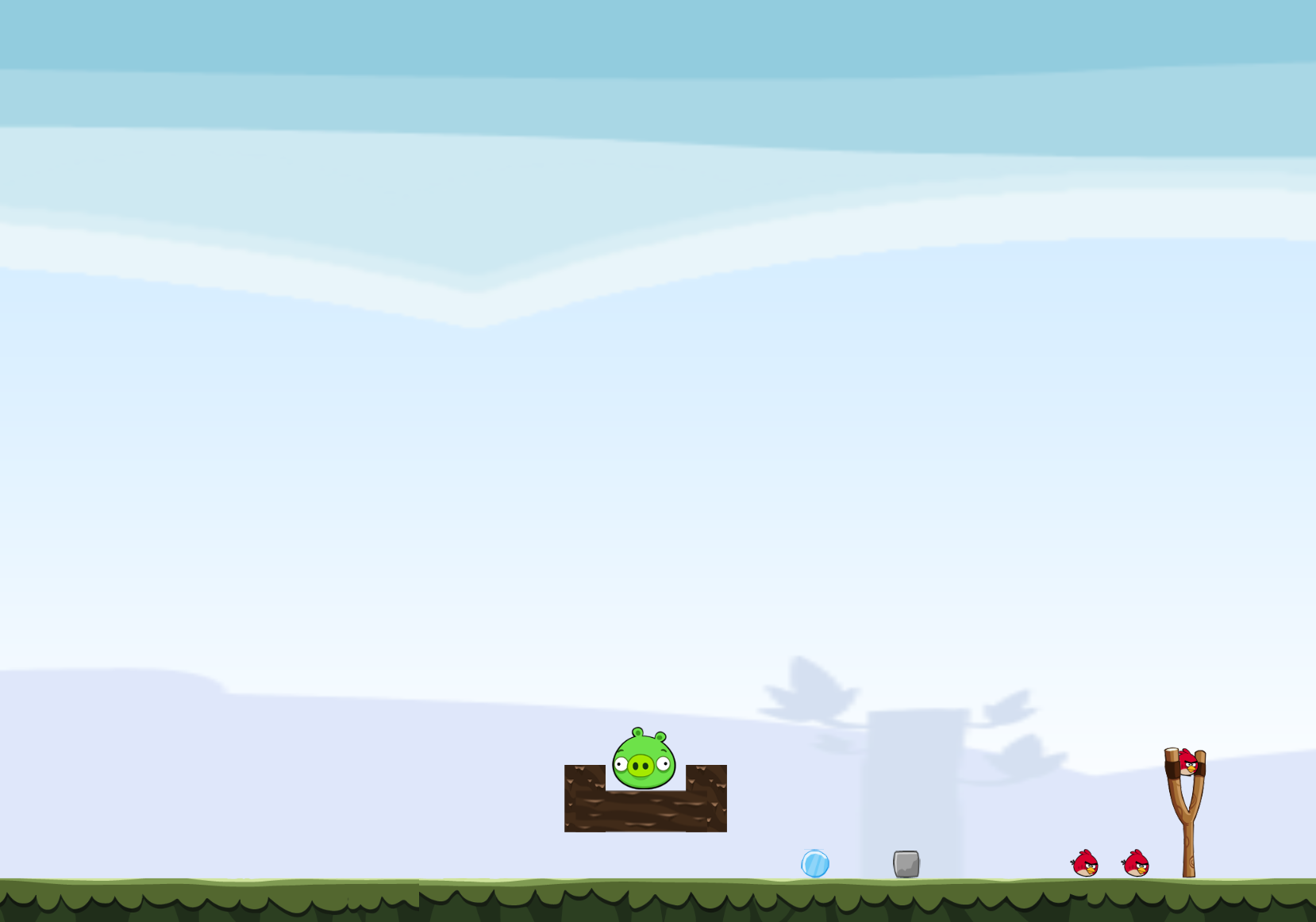}
      \end{subfigure}
  \caption{Relations}
  \end{subfigure}
  \begin{subfigure}[b]{0.49\columnwidth}
      \begin{subfigure}[b]{0.49\columnwidth}
        \includegraphics[width=\linewidth]{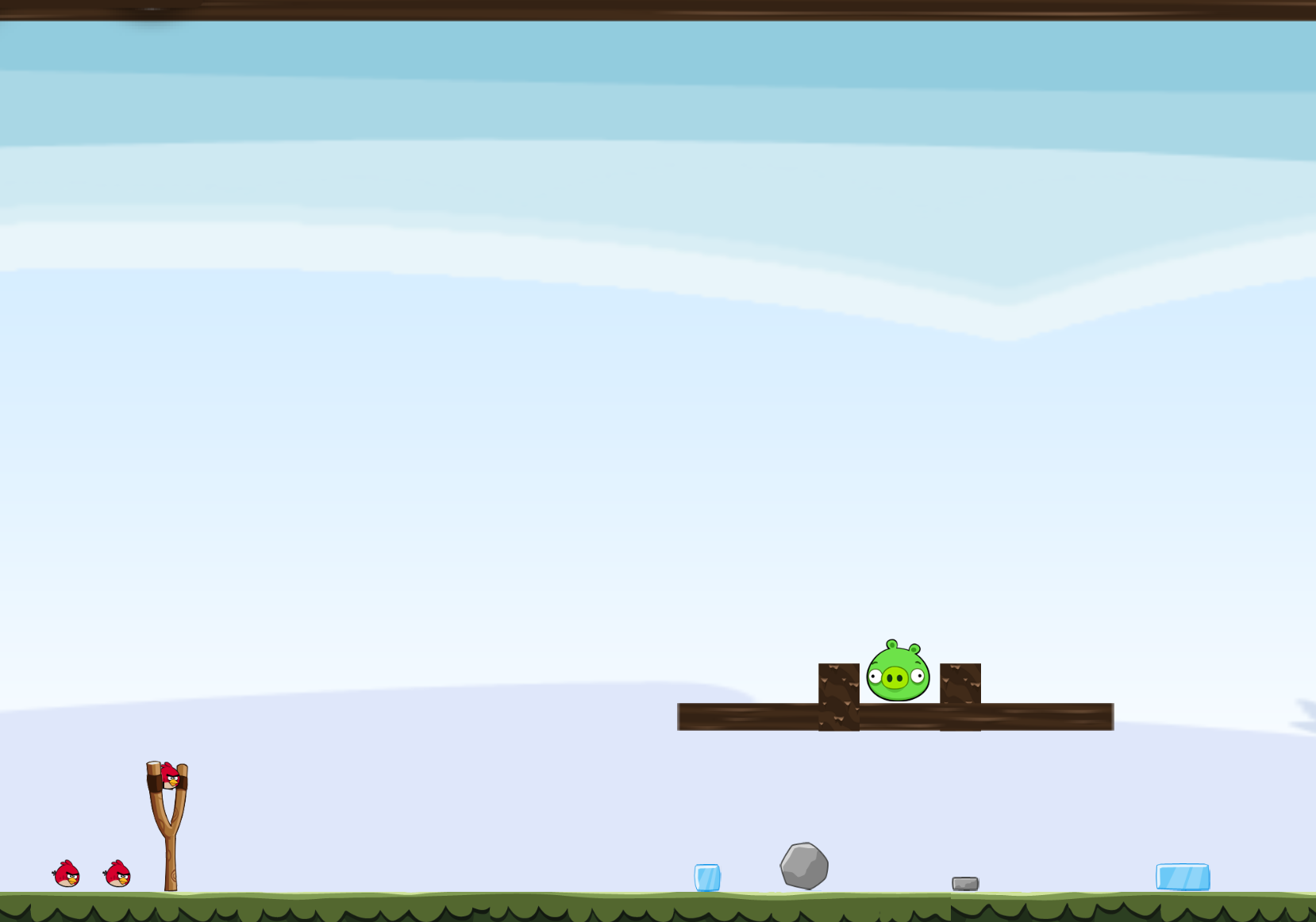}
      \end{subfigure}
      \begin{subfigure}[b]{0.49\columnwidth}
        \includegraphics[width=\linewidth]{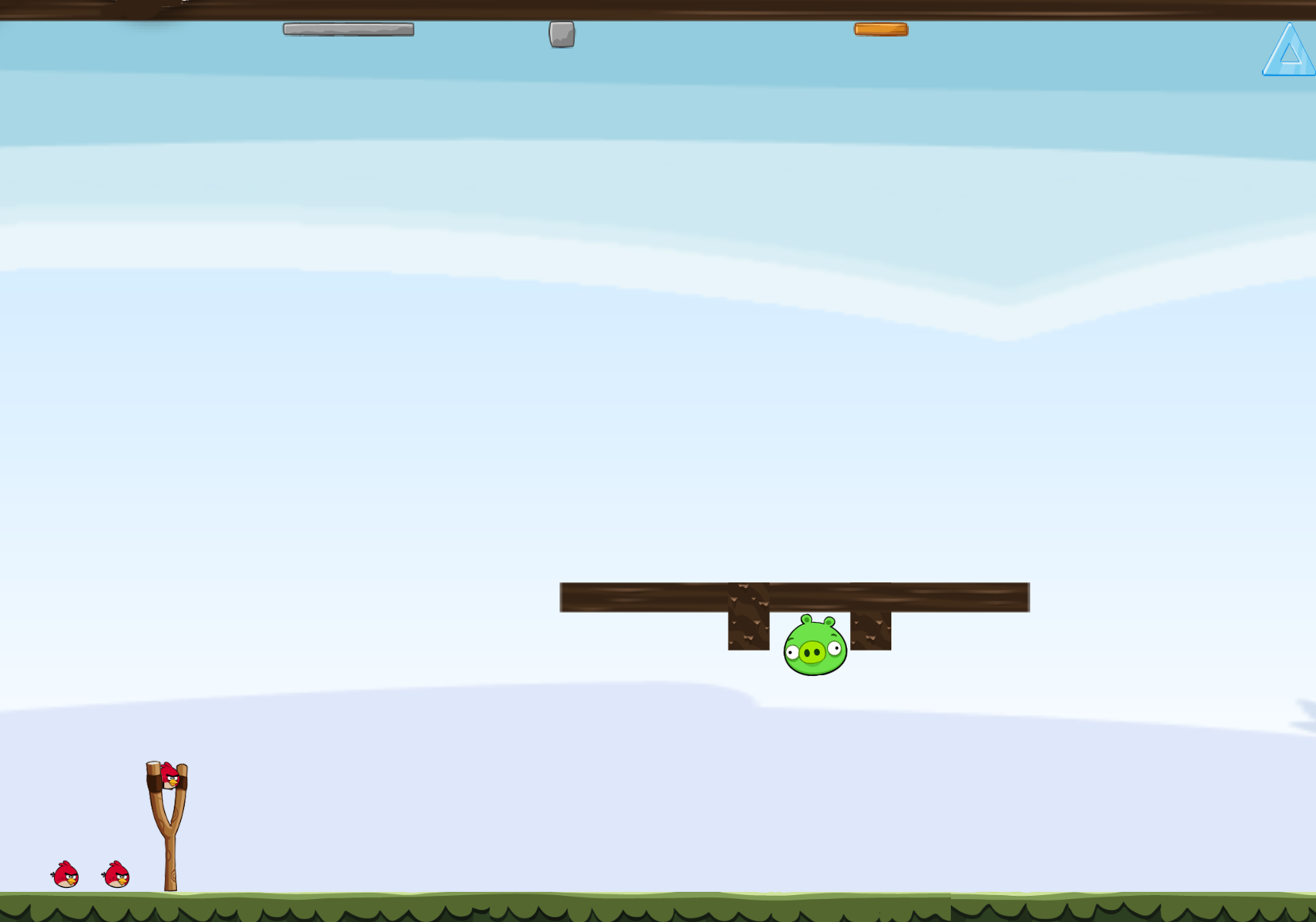}
      \end{subfigure}
  \caption{Environments}
  \end{subfigure}
  \begin{subfigure}[b]{0.49\columnwidth}
      \begin{subfigure}[b]{0.49\columnwidth}
        \includegraphics[width=\linewidth]{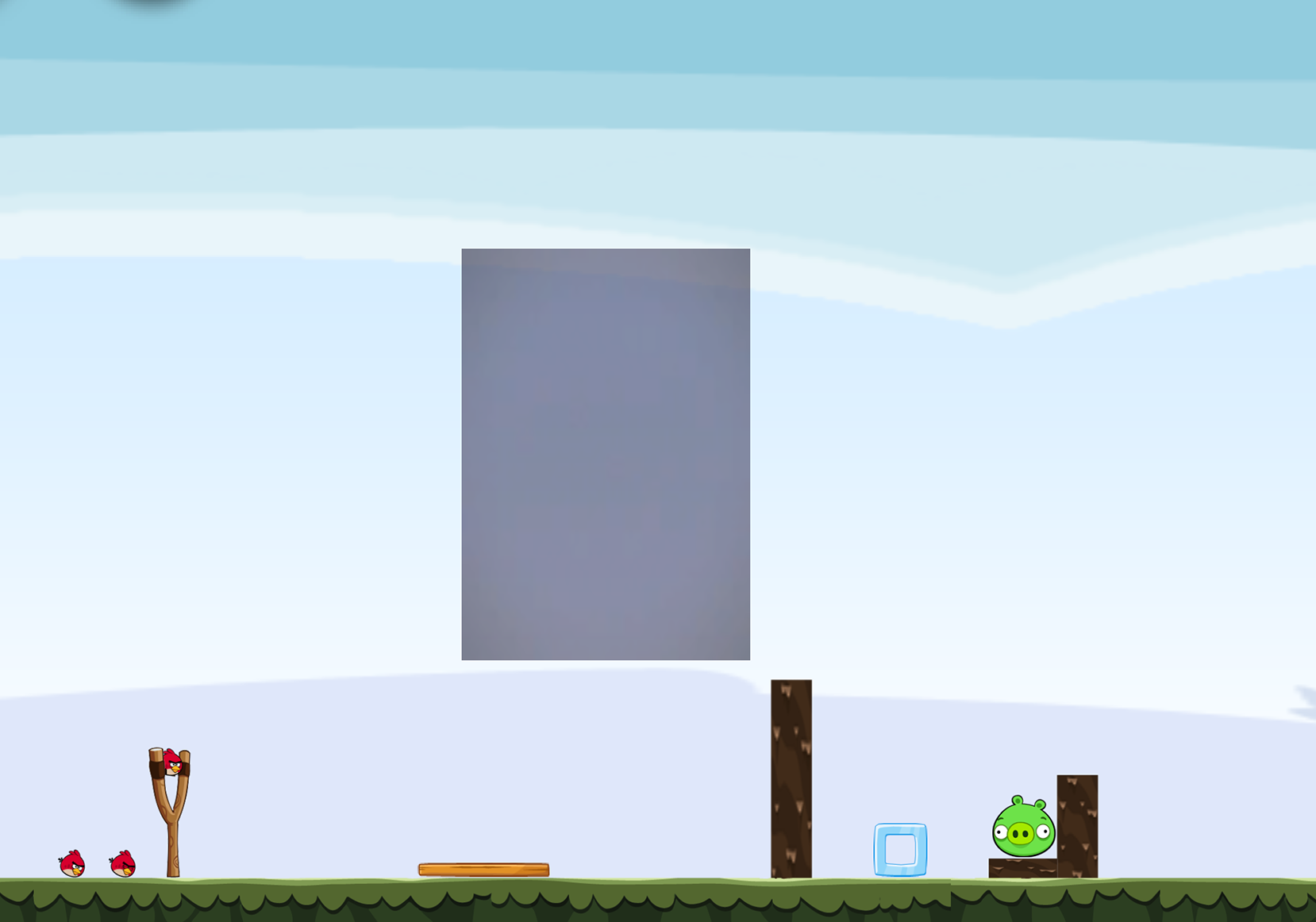}
      \end{subfigure}
      \begin{subfigure}[b]{0.49\columnwidth}
        \includegraphics[width=\linewidth]{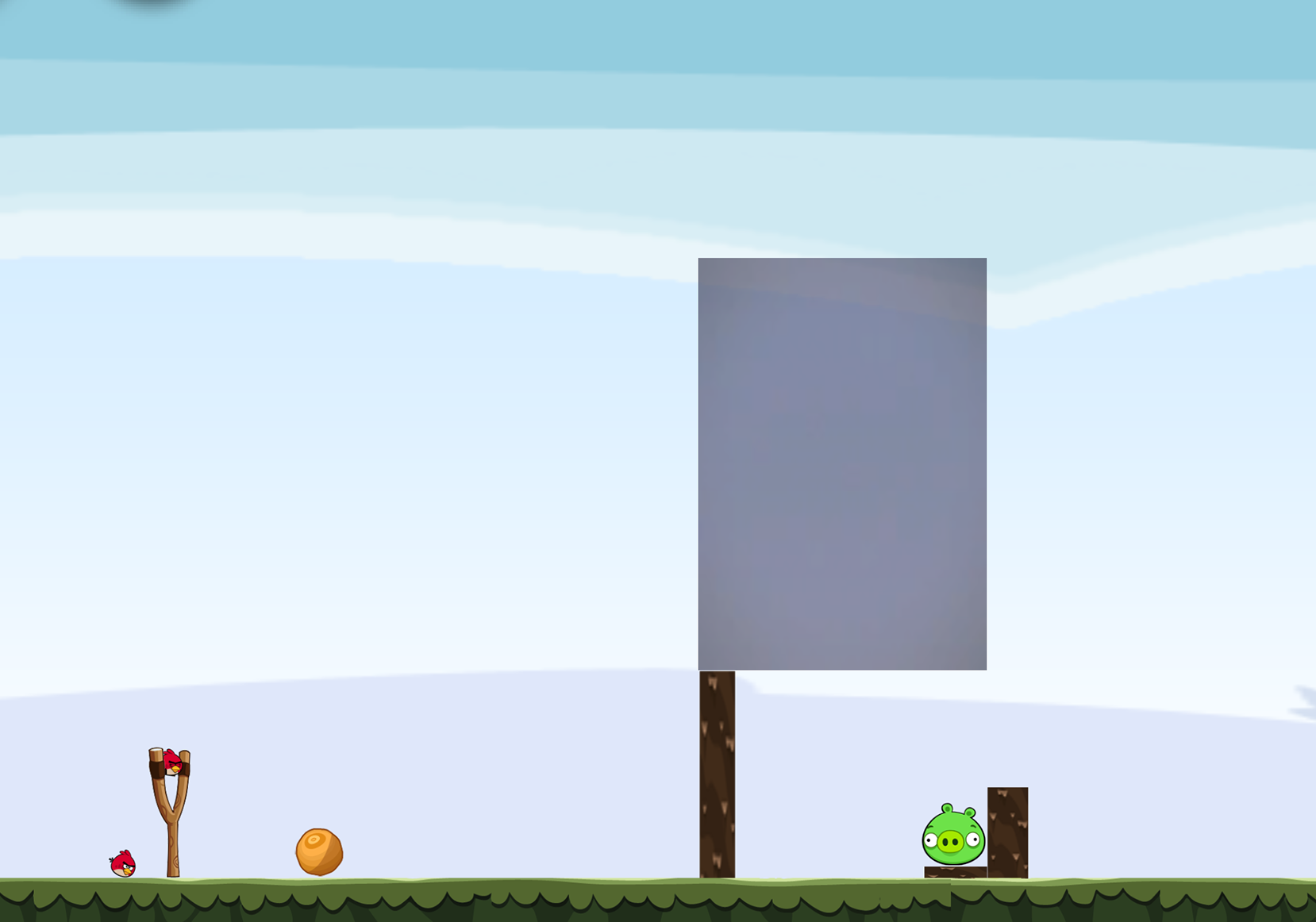}
      \end{subfigure}
  \caption{Goals}
  \end{subfigure}
  \begin{subfigure}[b]{0.49\columnwidth}
      \begin{subfigure}[b]{0.49\columnwidth}
        \includegraphics[width=\linewidth]{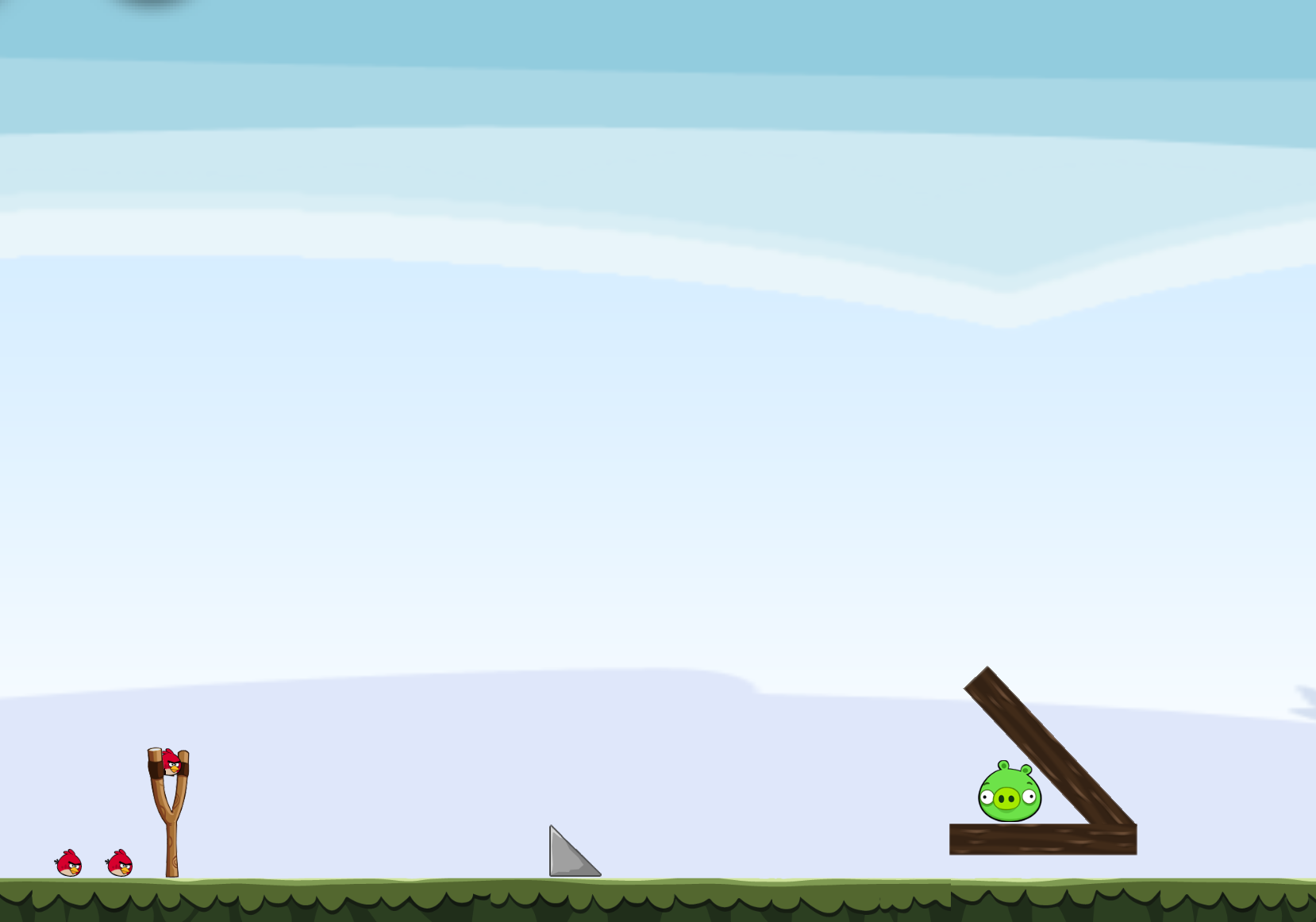}
      \end{subfigure}
      \begin{subfigure}[b]{0.49\columnwidth}
        \includegraphics[width=\linewidth]{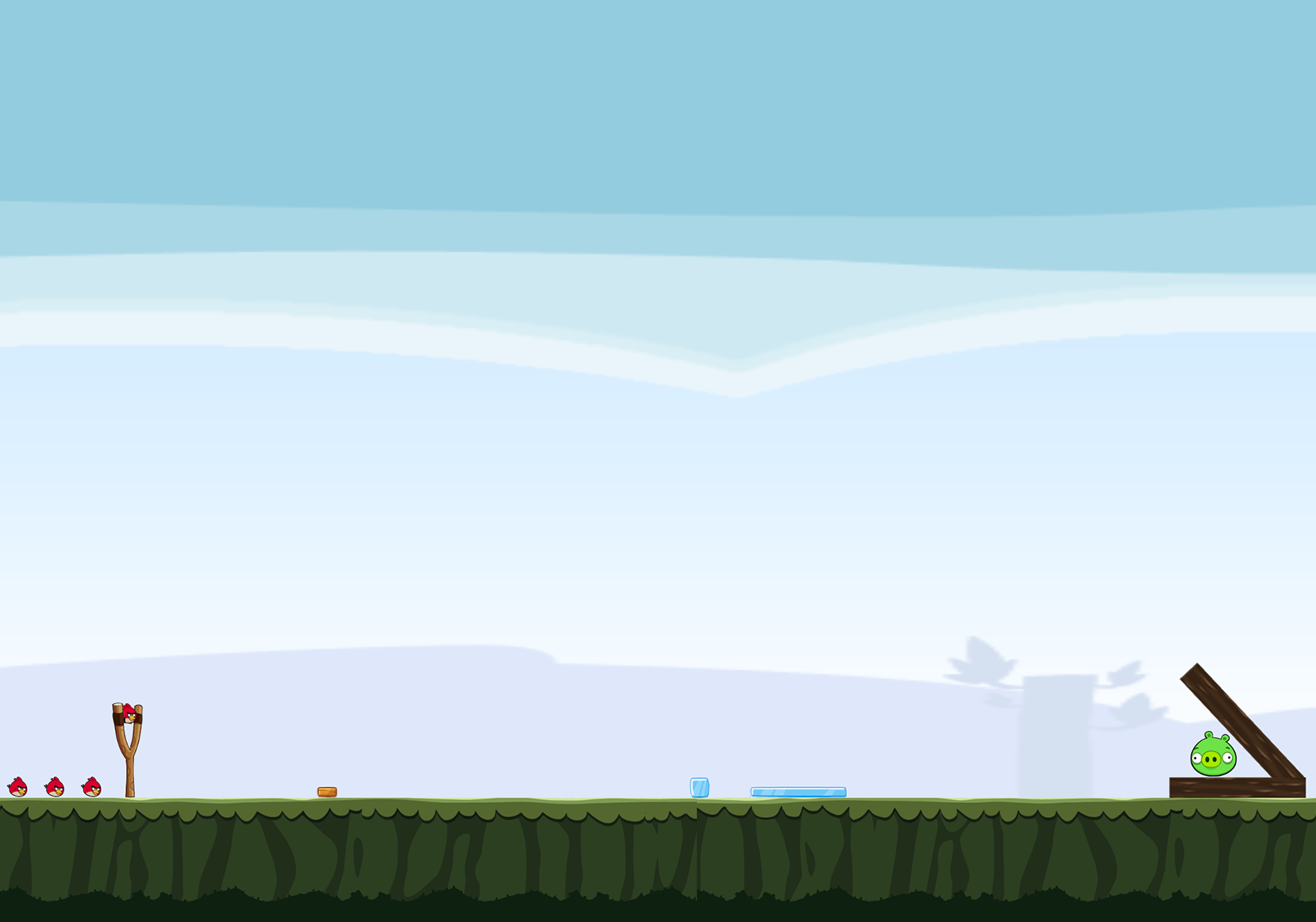}
      \end{subfigure}
  \caption{Events}
  \end{subfigure}
\caption{Task templates of the multiple forces scenario with eight novelties applied to them. In each subfigure, the left figure is the normal task and the right figure is the corresponding novel task with the novelty applied.}
\label{appendix_fig:multiple_forces}
\end{figure}

\begin{figure}[h!]
  \centering
  \begin{subfigure}[b]{0.49\columnwidth}
      \begin{subfigure}[b]{0.49\columnwidth}
        \includegraphics[width=\linewidth]{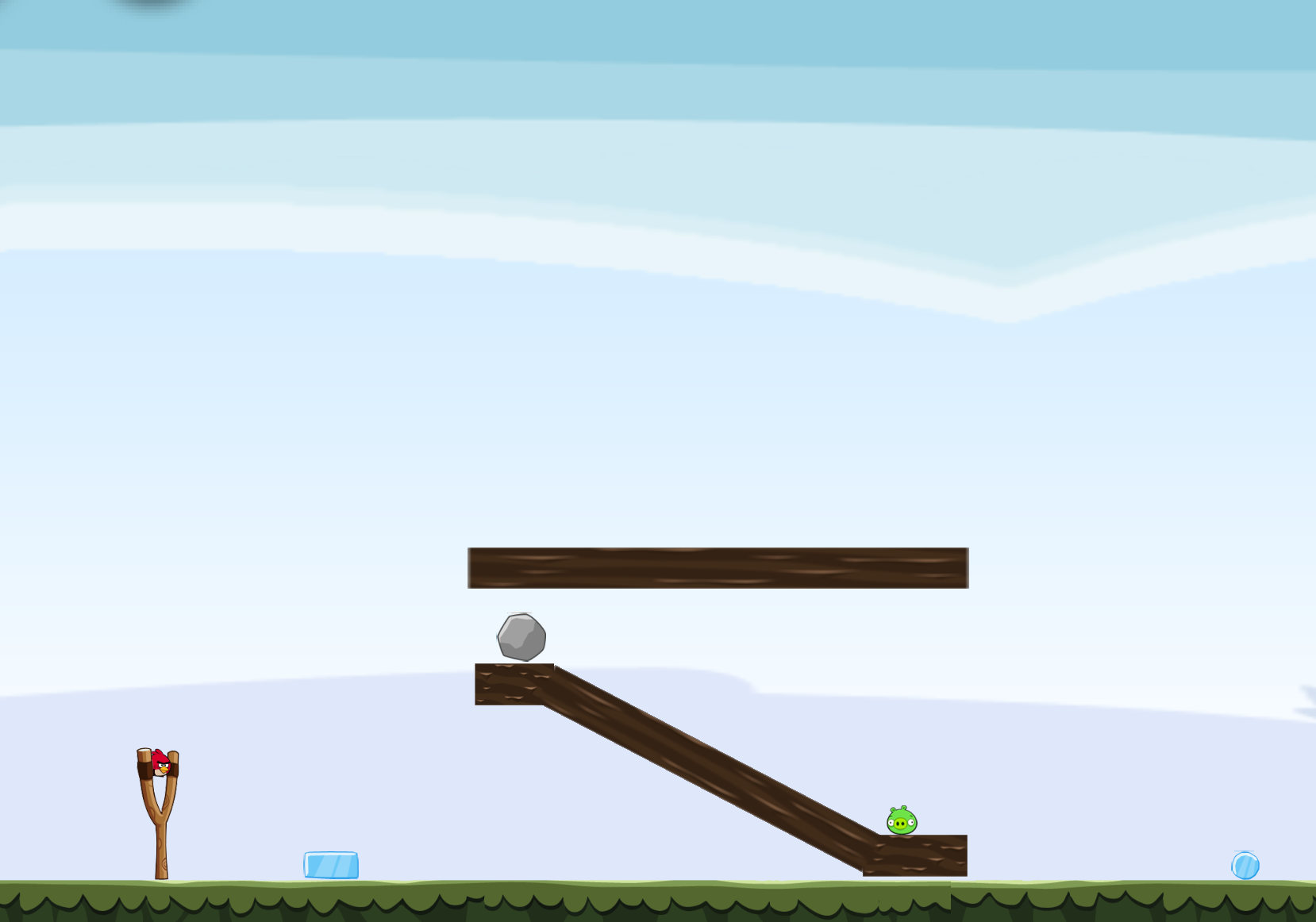}
      \end{subfigure}
      \begin{subfigure}[b]{0.49\columnwidth}
        \includegraphics[width=\linewidth]{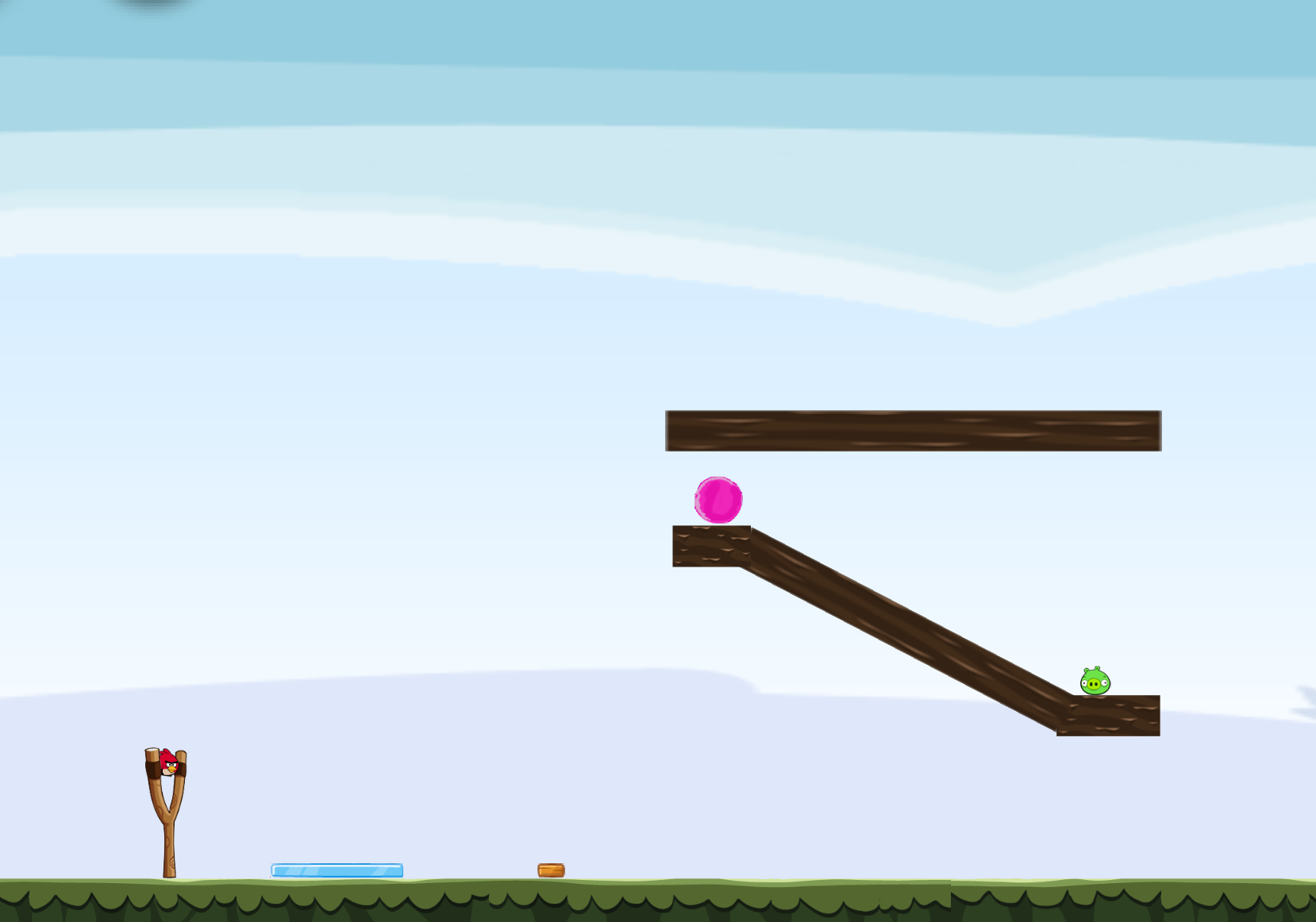}
      \end{subfigure}
  \caption{Objects}
  \end{subfigure}
  \begin{subfigure}[b]{0.49\columnwidth}
      \begin{subfigure}[b]{0.49\columnwidth}
        \includegraphics[width=\linewidth]{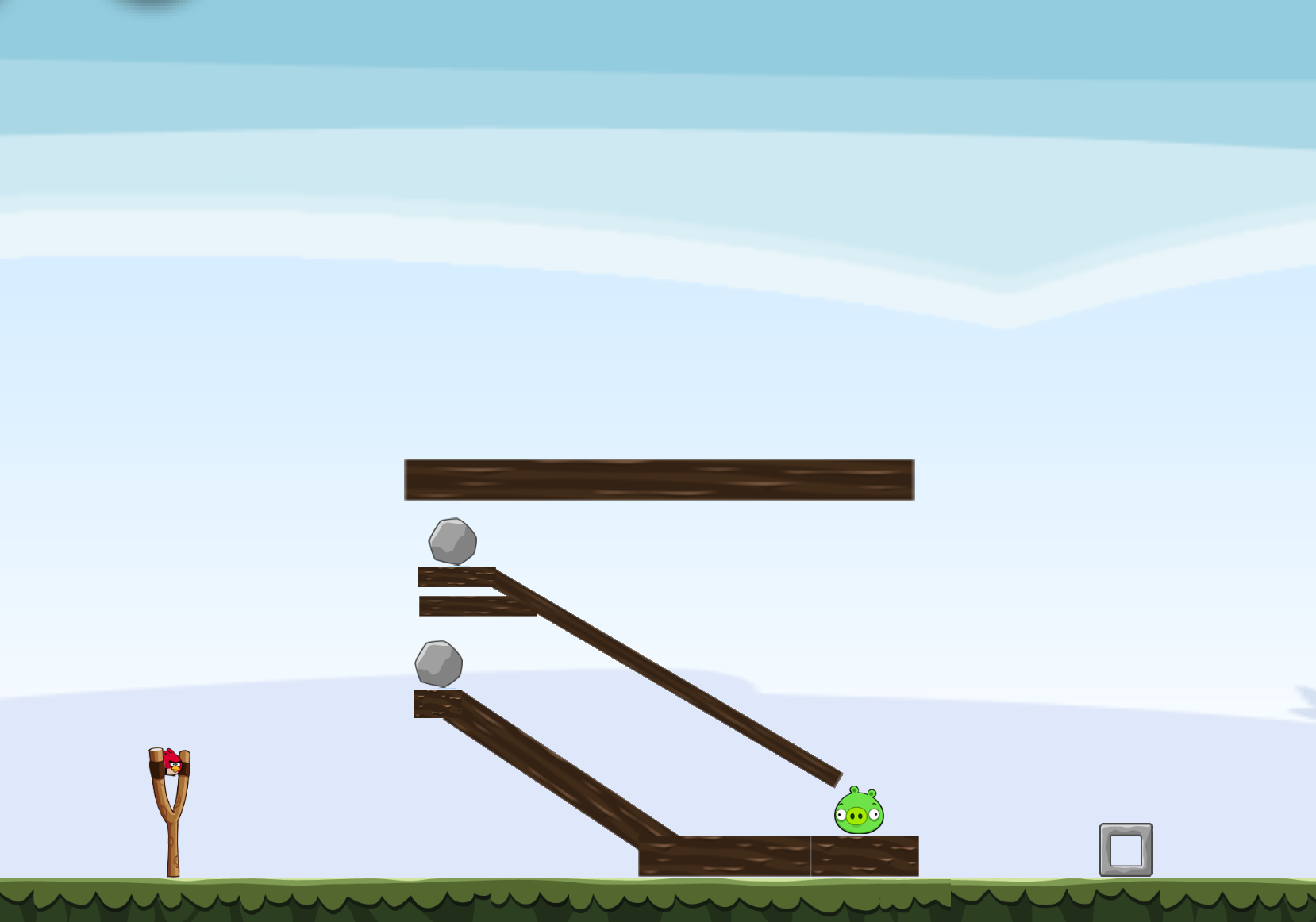}
      \end{subfigure}
      \begin{subfigure}[b]{0.49\columnwidth}
        \includegraphics[width=\linewidth]{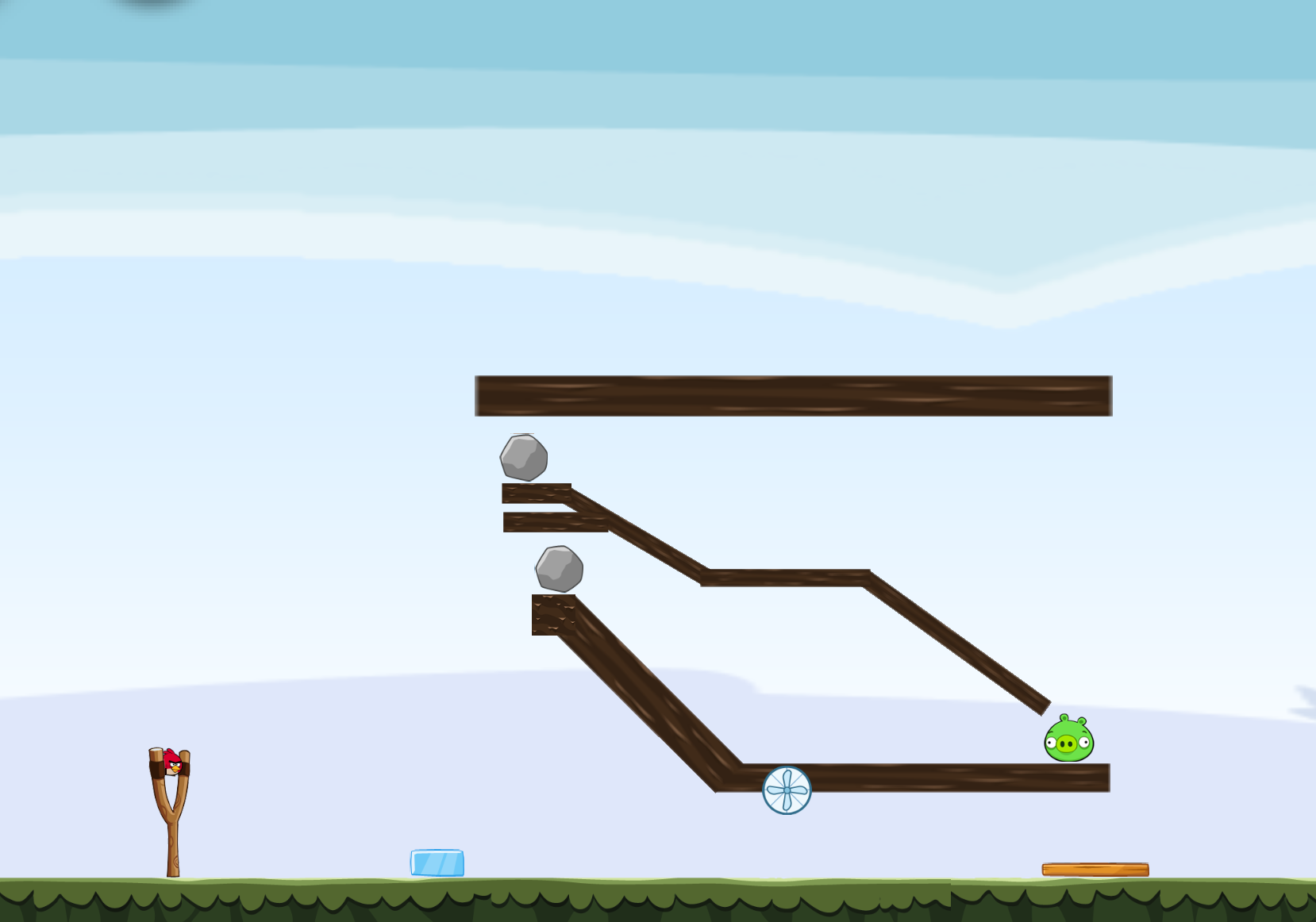}
      \end{subfigure}
  \caption{Agents}
  \end{subfigure}
  \begin{subfigure}[b]{0.49\columnwidth}
      \begin{subfigure}[b]{0.49\columnwidth}
        \includegraphics[width=\linewidth]{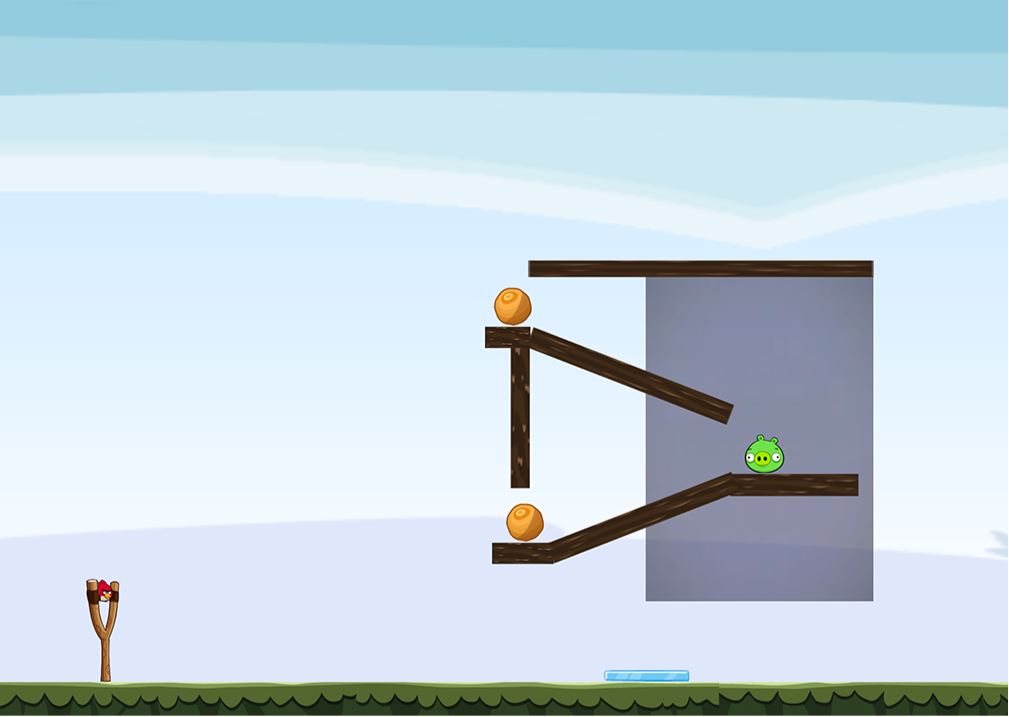}
      \end{subfigure}
      \begin{subfigure}[b]{0.49\columnwidth}
        \includegraphics[width=\linewidth]{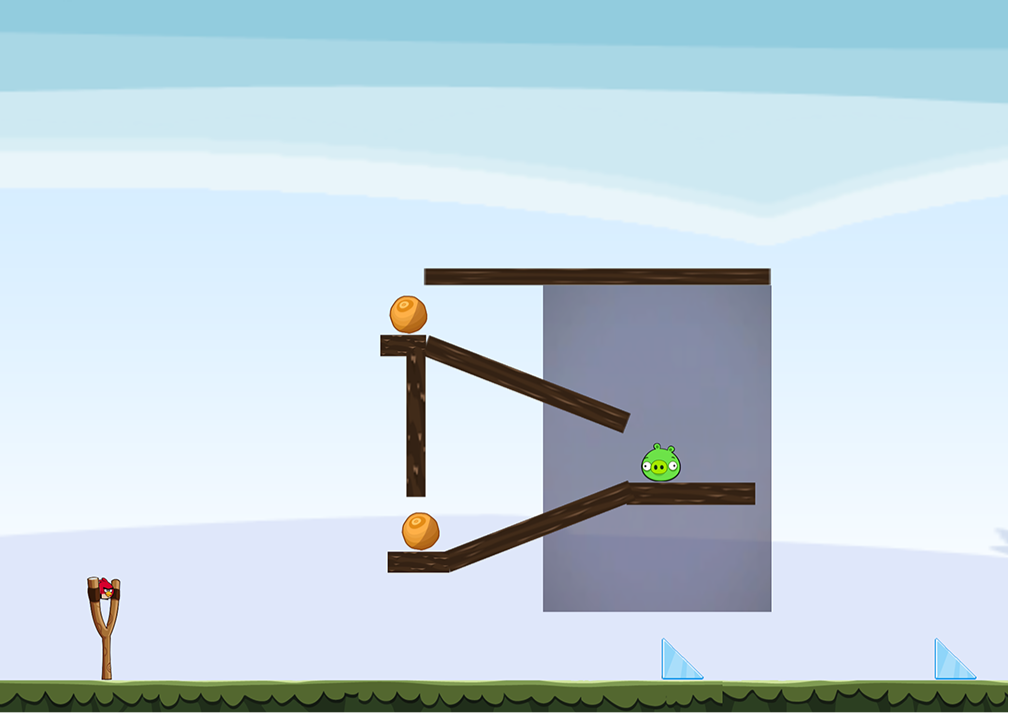}
      \end{subfigure}
  \caption{Actions}
  \end{subfigure}
  \begin{subfigure}[b]{0.49\columnwidth}
      \begin{subfigure}[b]{0.49\columnwidth}
        \includegraphics[width=\linewidth]{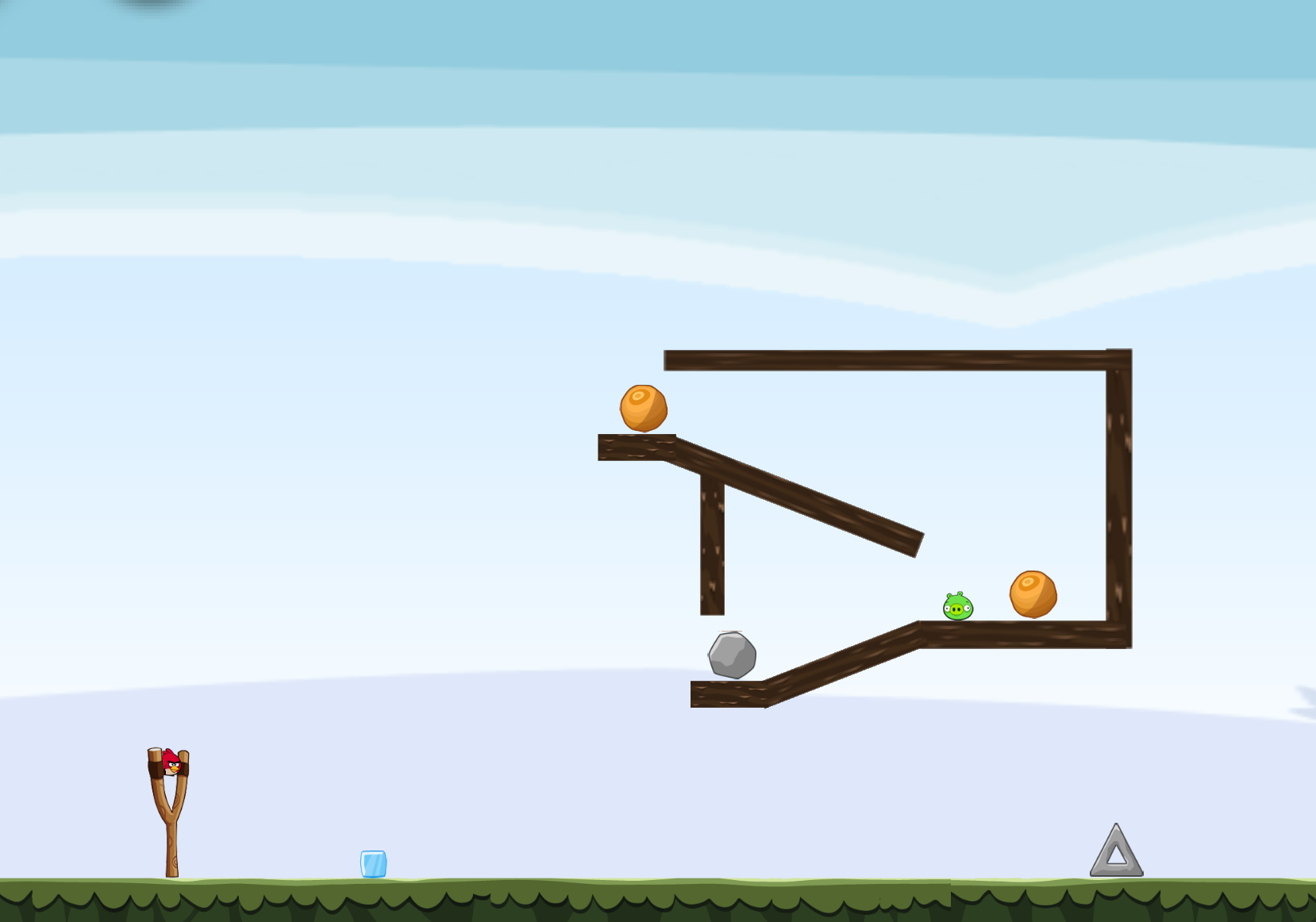}
      \end{subfigure}
      \begin{subfigure}[b]{0.49\columnwidth}
        \includegraphics[width=\linewidth]{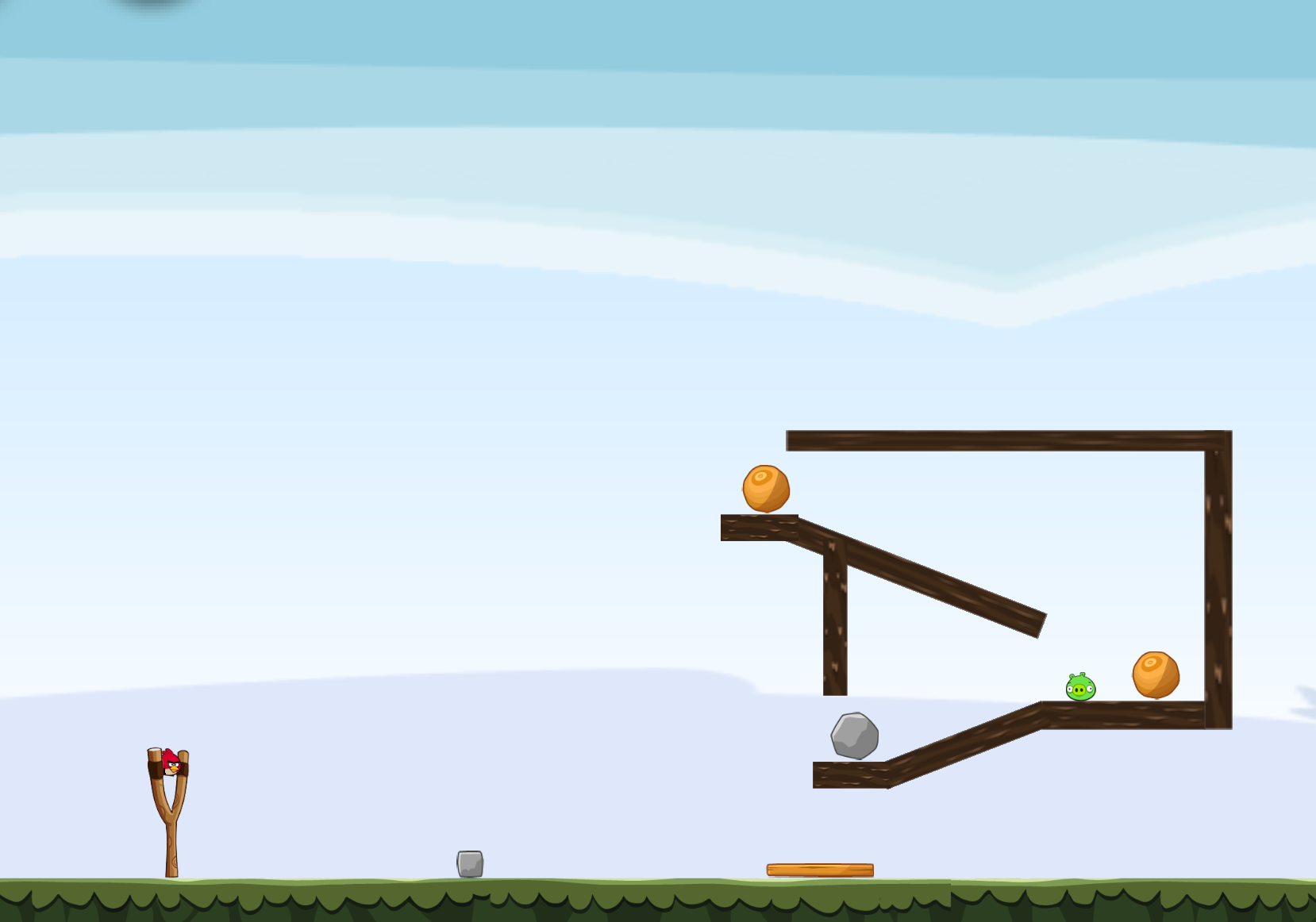}
      \end{subfigure}
  \caption{Interactions}
  \end{subfigure}
    \begin{subfigure}[b]{0.49\columnwidth}
      \begin{subfigure}[b]{0.49\columnwidth}
        \includegraphics[width=\linewidth]{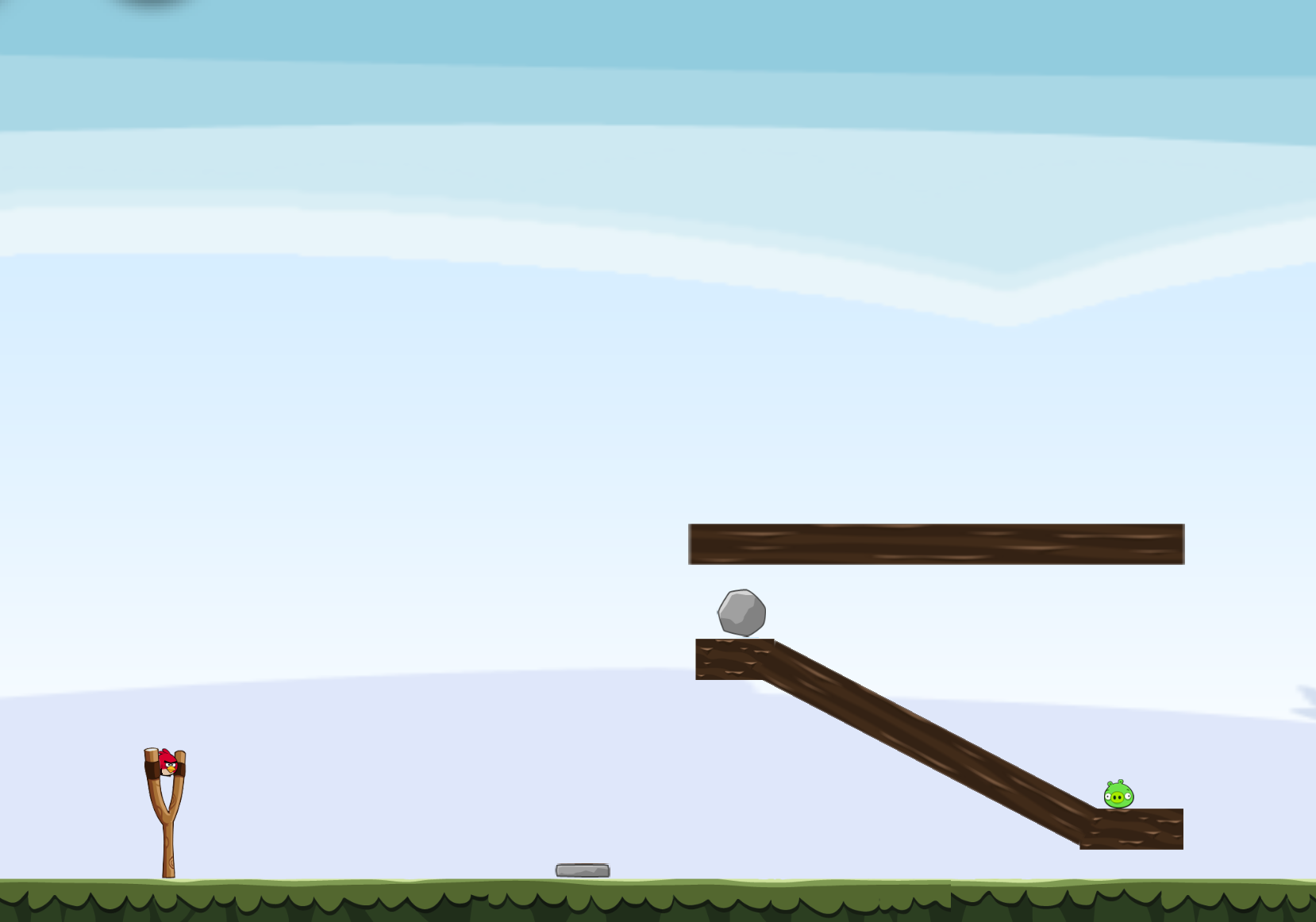}
      \end{subfigure}
      \begin{subfigure}[b]{0.49\columnwidth}
        \includegraphics[width=\linewidth]{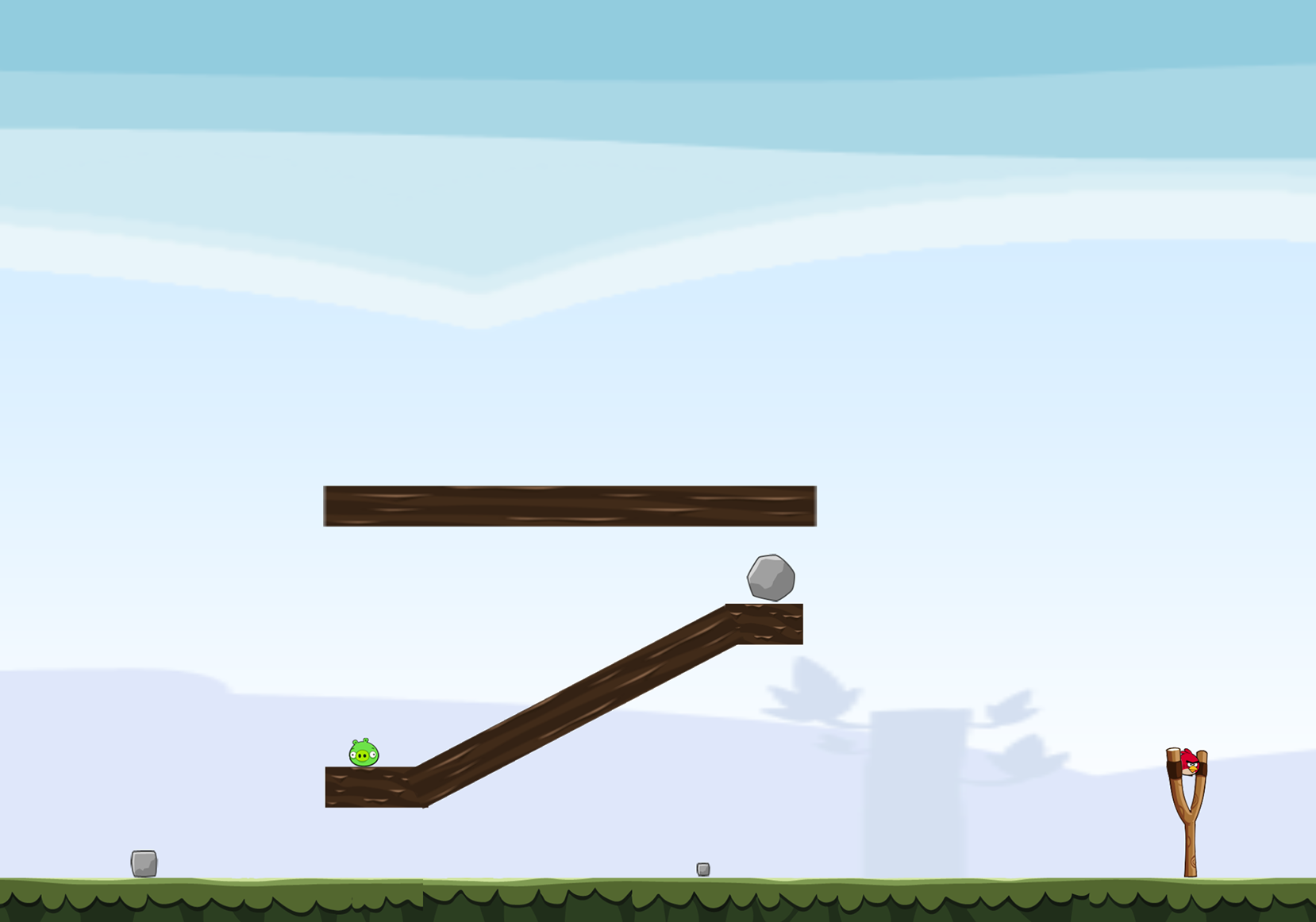}
      \end{subfigure}
  \caption{Relations}
  \end{subfigure}
  \begin{subfigure}[b]{0.49\columnwidth}
      \begin{subfigure}[b]{0.49\columnwidth}
        \includegraphics[width=\linewidth]{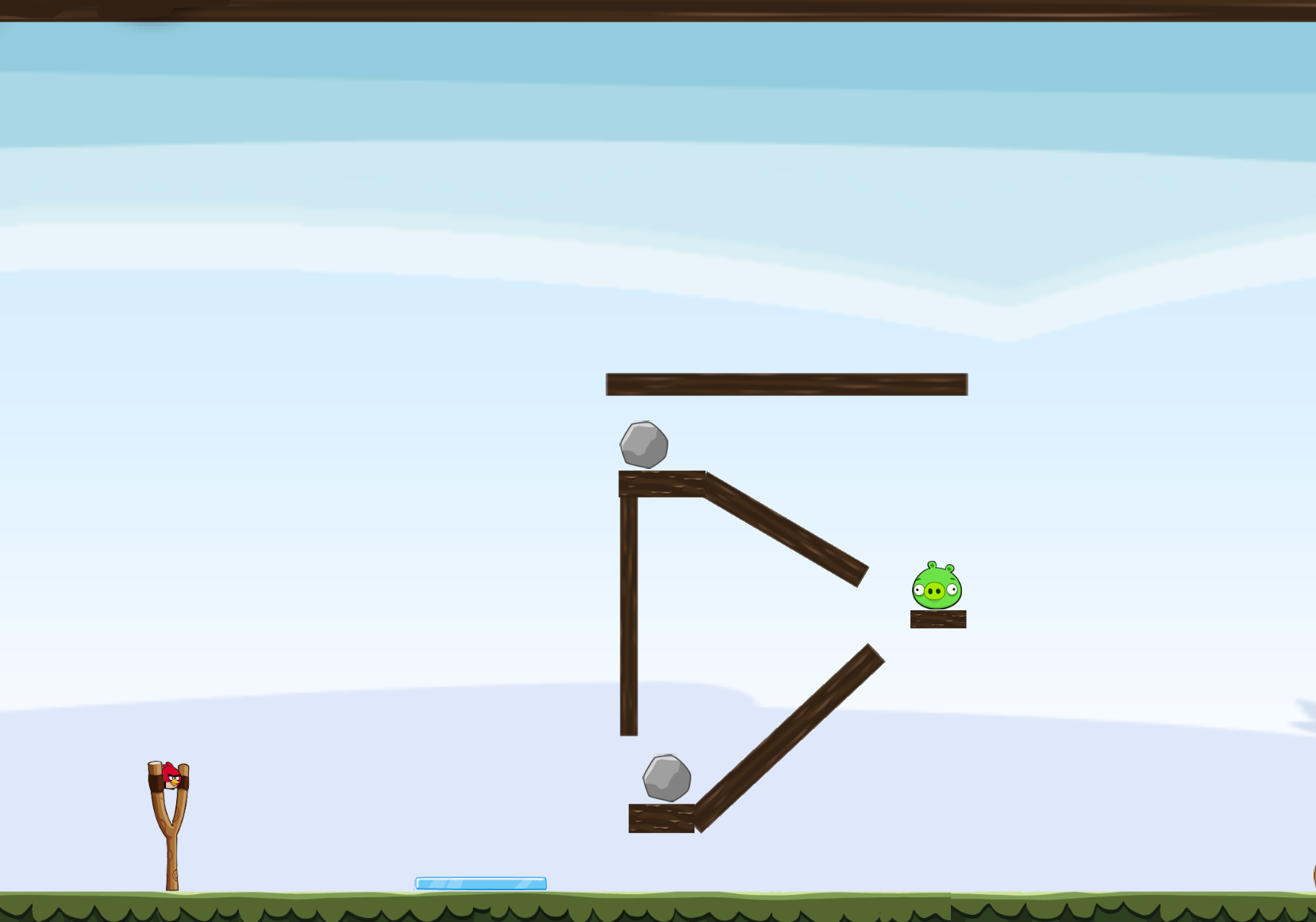}
      \end{subfigure}
      \begin{subfigure}[b]{0.49\columnwidth}
        \includegraphics[width=\linewidth]{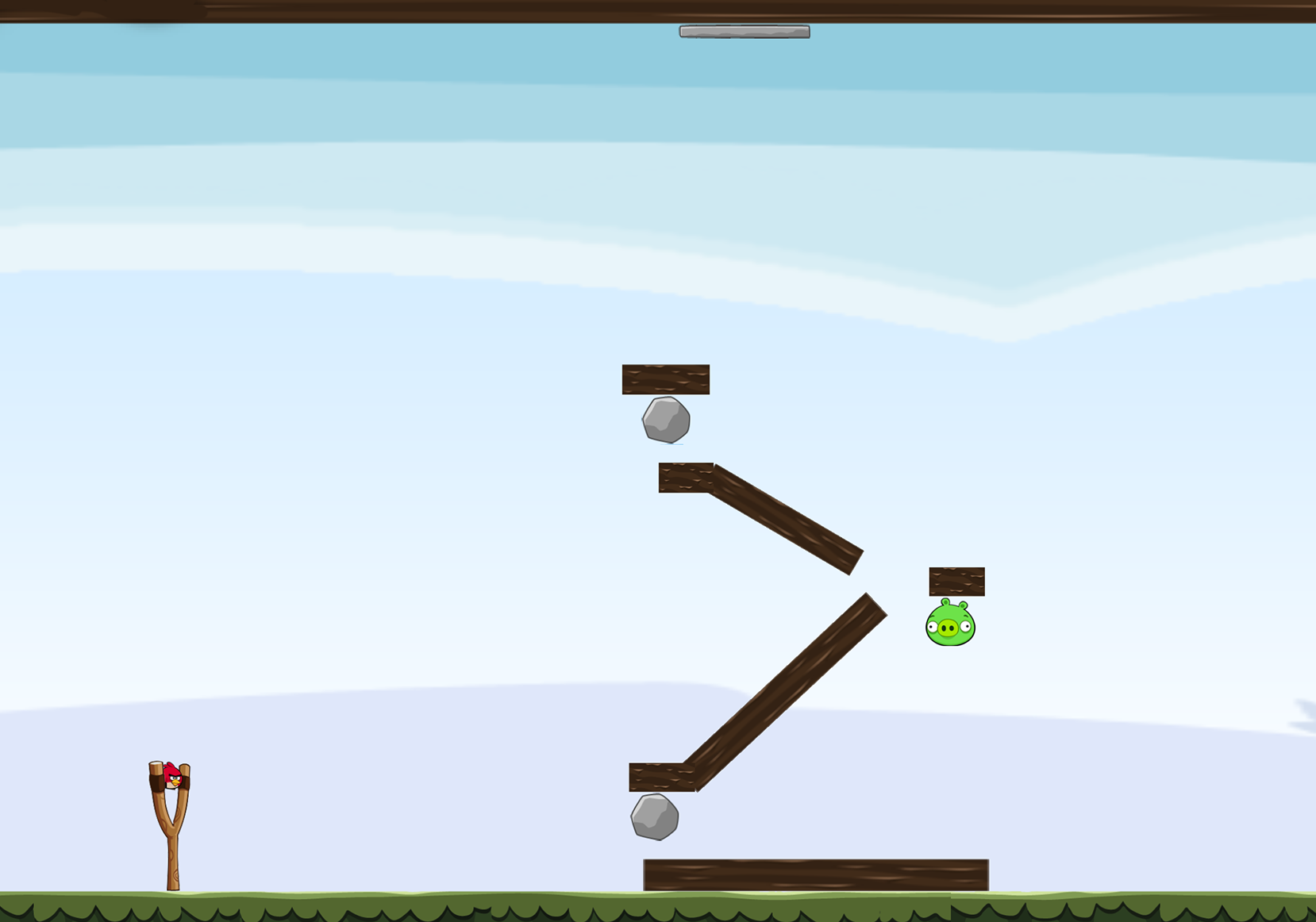}
      \end{subfigure}
  \caption{Environments}
  \end{subfigure}
  \begin{subfigure}[b]{0.49\columnwidth}
      \begin{subfigure}[b]{0.49\columnwidth}
        \includegraphics[width=\linewidth]{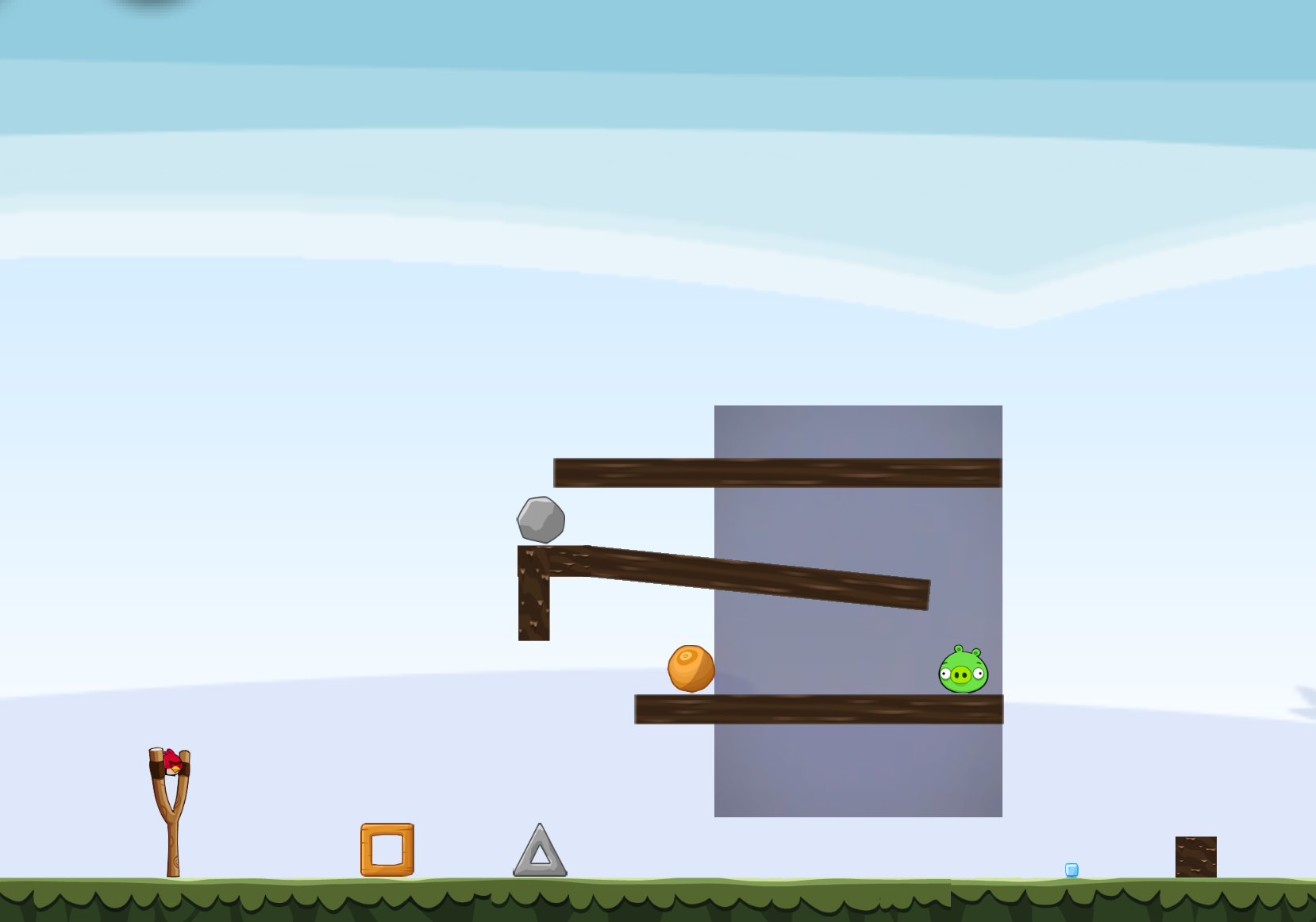}
      \end{subfigure}
      \begin{subfigure}[b]{0.49\columnwidth}
        \includegraphics[width=\linewidth]{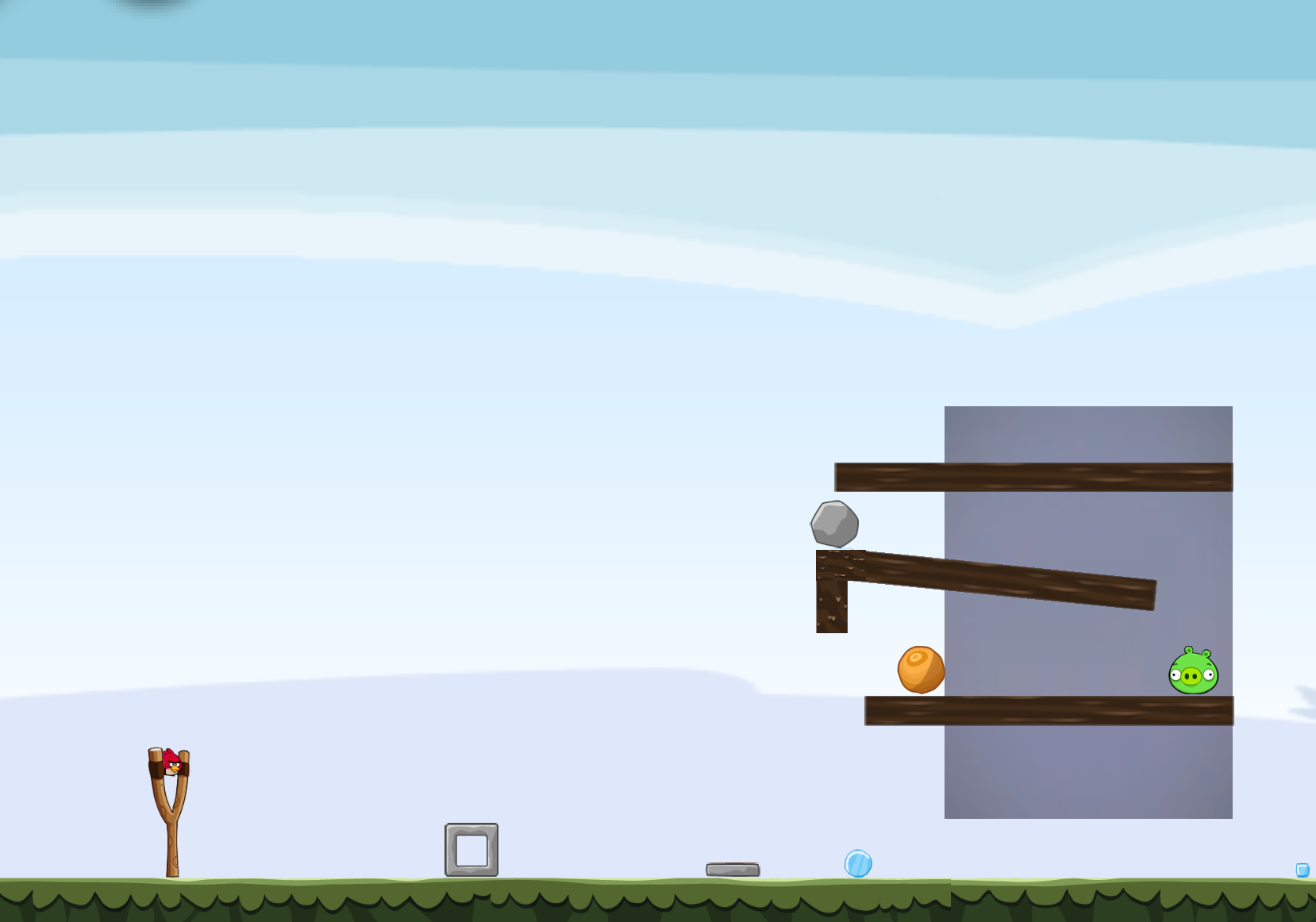}
      \end{subfigure}
  \caption{Goals}
  \end{subfigure}
  \begin{subfigure}[b]{0.49\columnwidth}
      \begin{subfigure}[b]{0.49\columnwidth}
        \includegraphics[width=\linewidth]{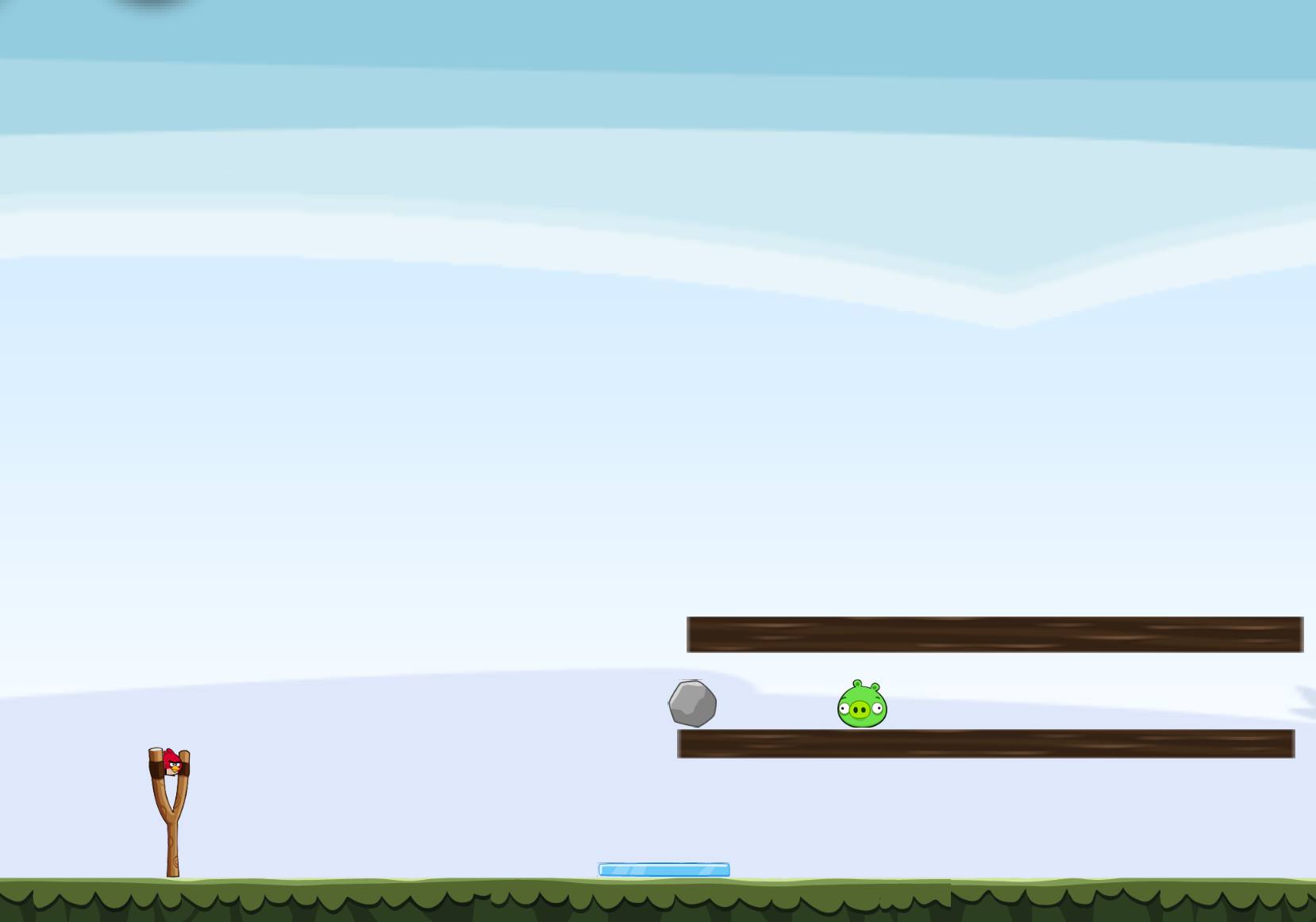}
      \end{subfigure}
      \begin{subfigure}[b]{0.49\columnwidth}
        \includegraphics[width=\linewidth]{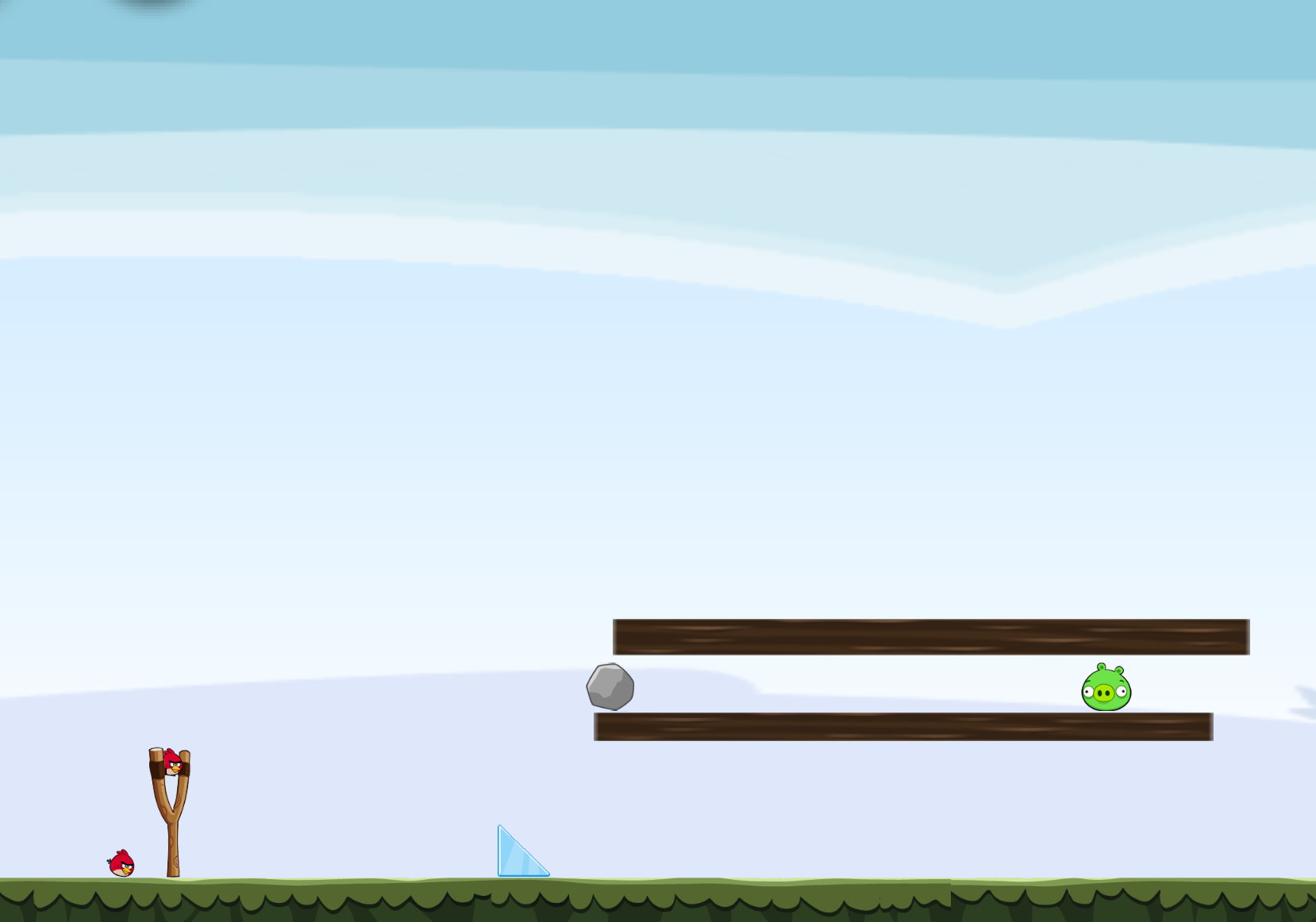}
      \end{subfigure}
  \caption{Events}
  \end{subfigure}
\caption{Task templates of the rolling scenario with eight novelties applied to them. In each subfigure, the left figure is the normal task and the right figure is the corresponding novel task with the novelty applied.}
\label{appendix_fig:rolling}
\end{figure}

\begin{figure}[h!]
  \centering
  \begin{subfigure}[b]{0.49\columnwidth}
      \begin{subfigure}[b]{0.49\columnwidth}
        \includegraphics[width=\linewidth]{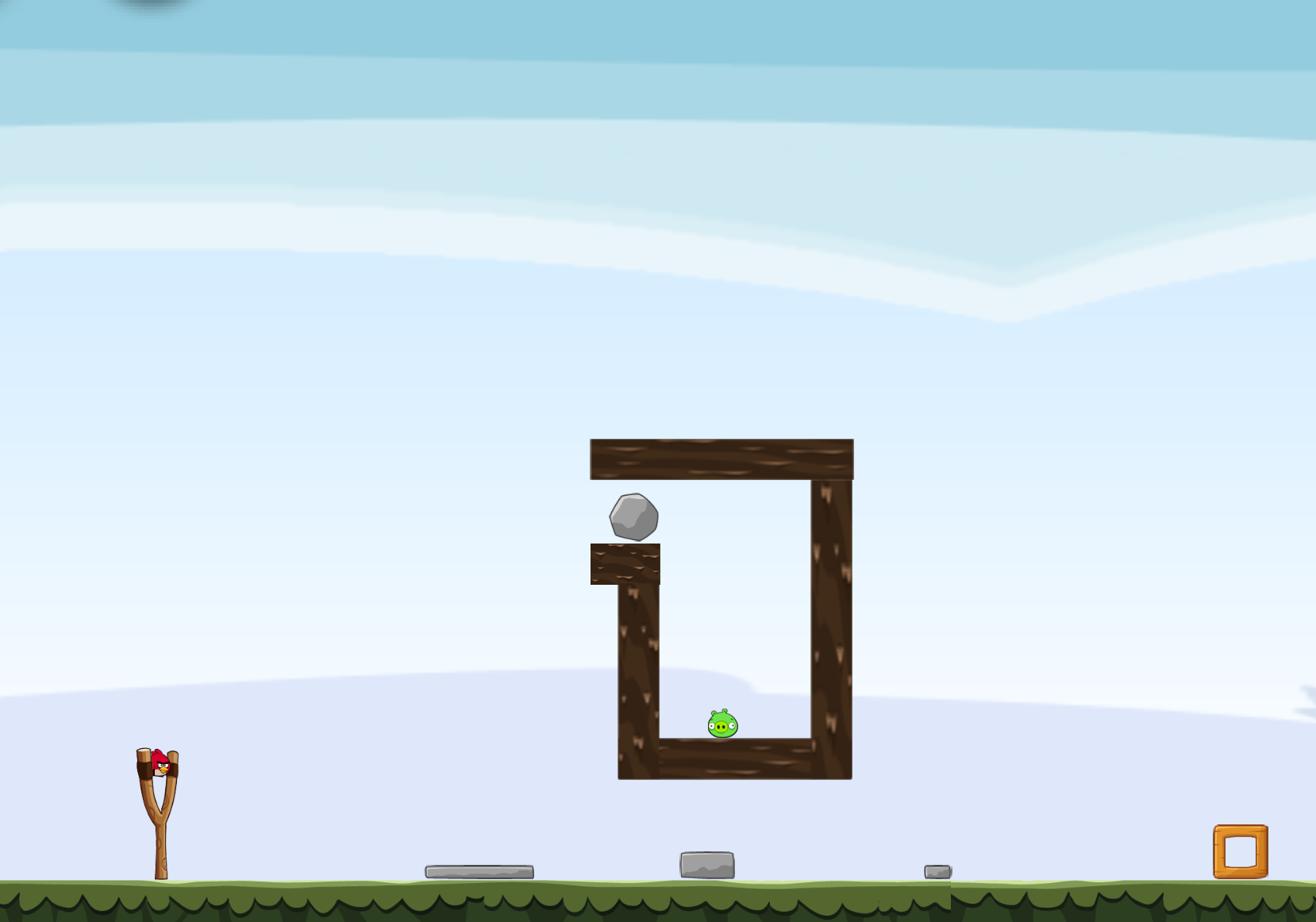}
      \end{subfigure}
      \begin{subfigure}[b]{0.49\columnwidth}
        \includegraphics[width=\linewidth]{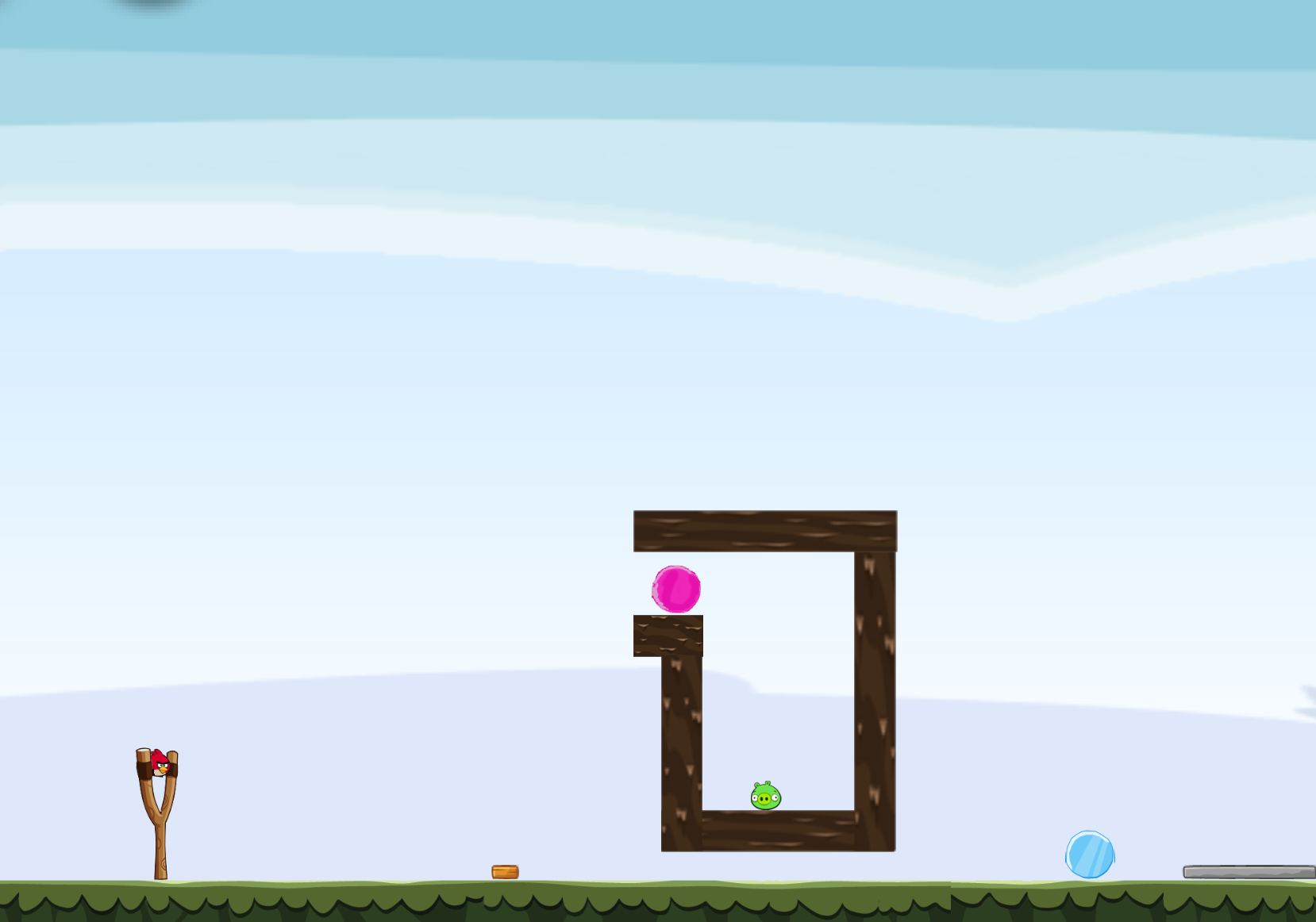}
      \end{subfigure}
  \caption{Objects}
  \end{subfigure}
  \begin{subfigure}[b]{0.49\columnwidth}
      \begin{subfigure}[b]{0.49\columnwidth}
        \includegraphics[width=\linewidth]{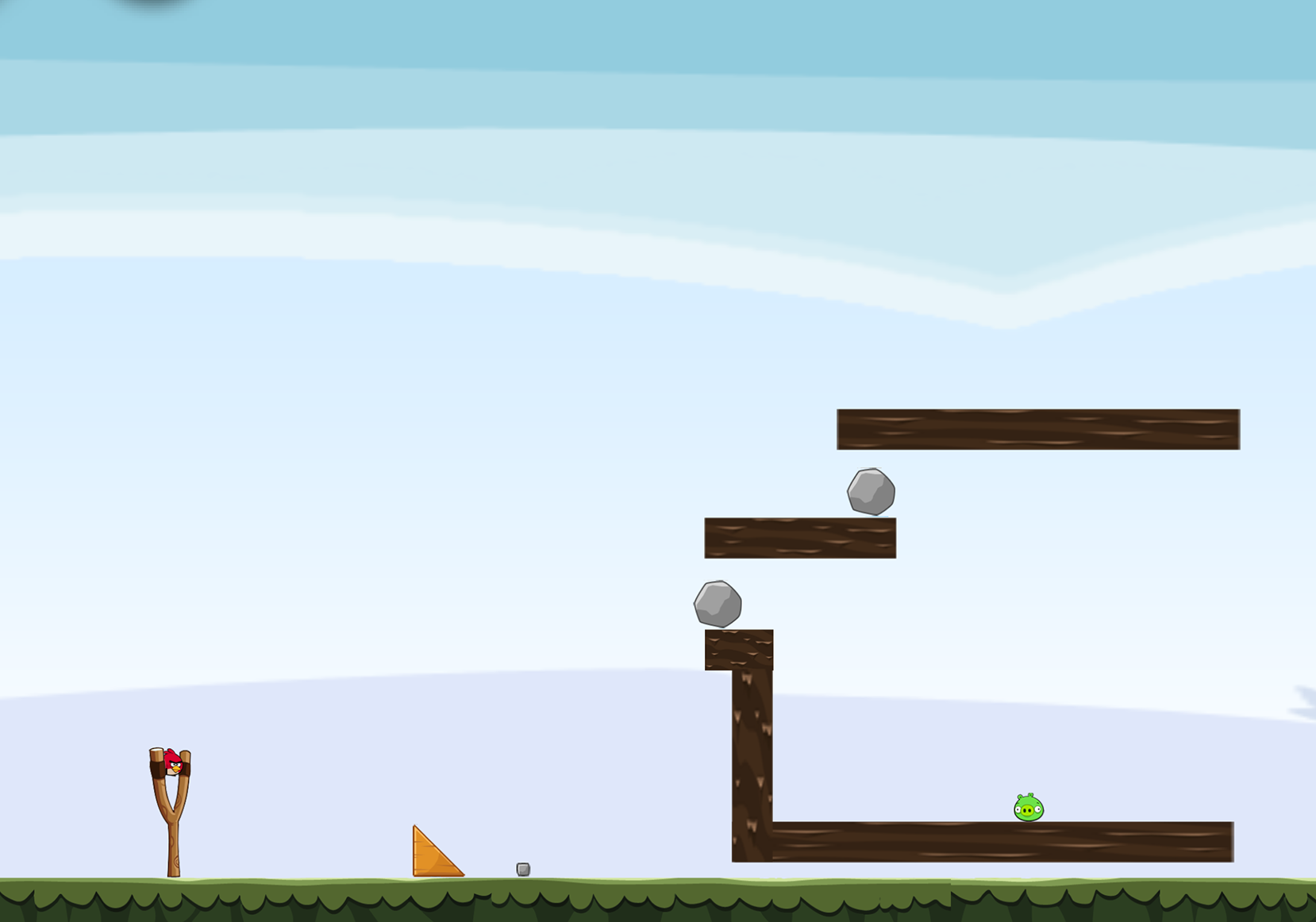}
      \end{subfigure}
      \begin{subfigure}[b]{0.49\columnwidth}
        \includegraphics[width=\linewidth]{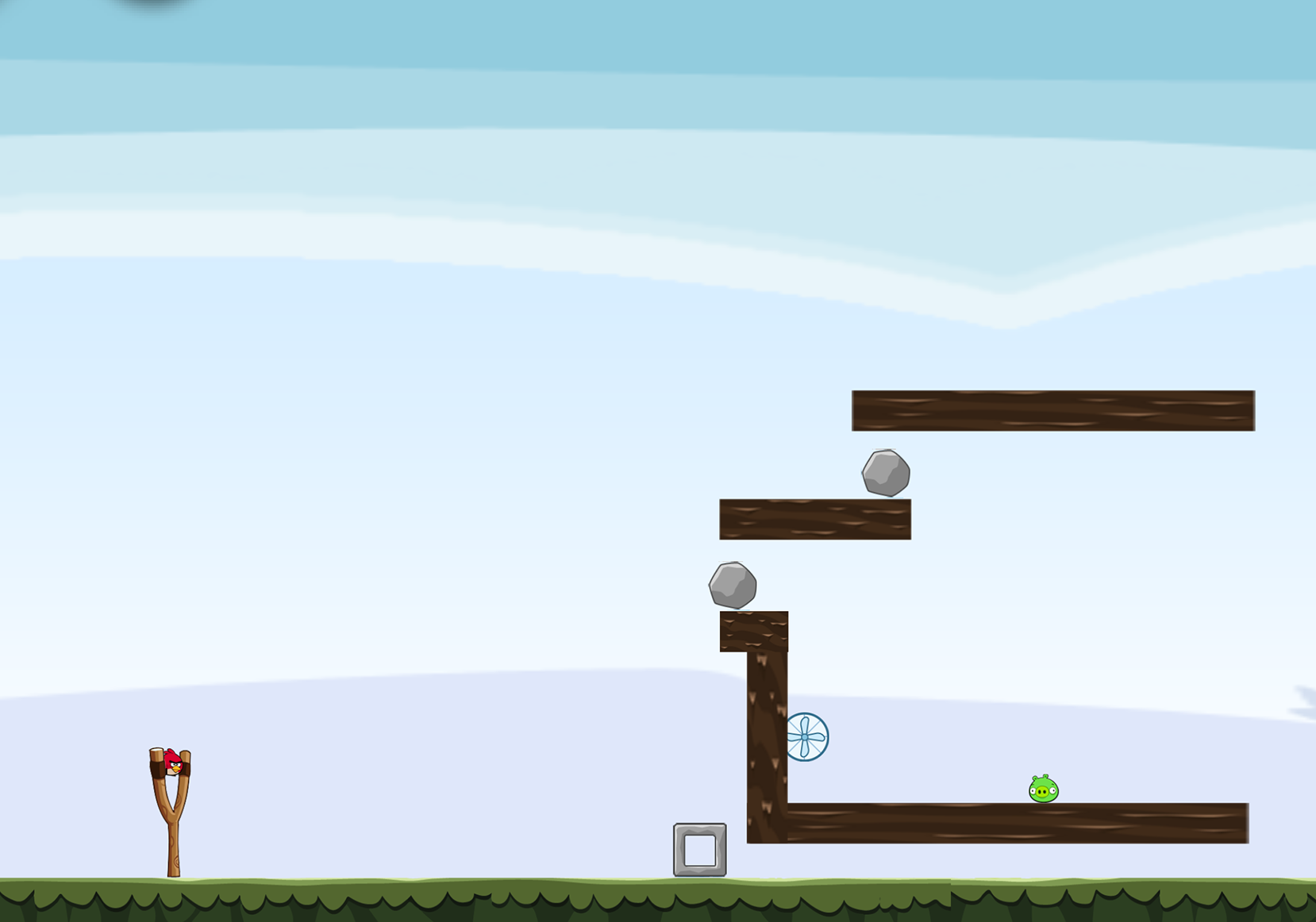}
      \end{subfigure}
  \caption{Agents}
  \end{subfigure}
  \begin{subfigure}[b]{0.49\columnwidth}
      \begin{subfigure}[b]{0.49\columnwidth}
        \includegraphics[width=\linewidth]{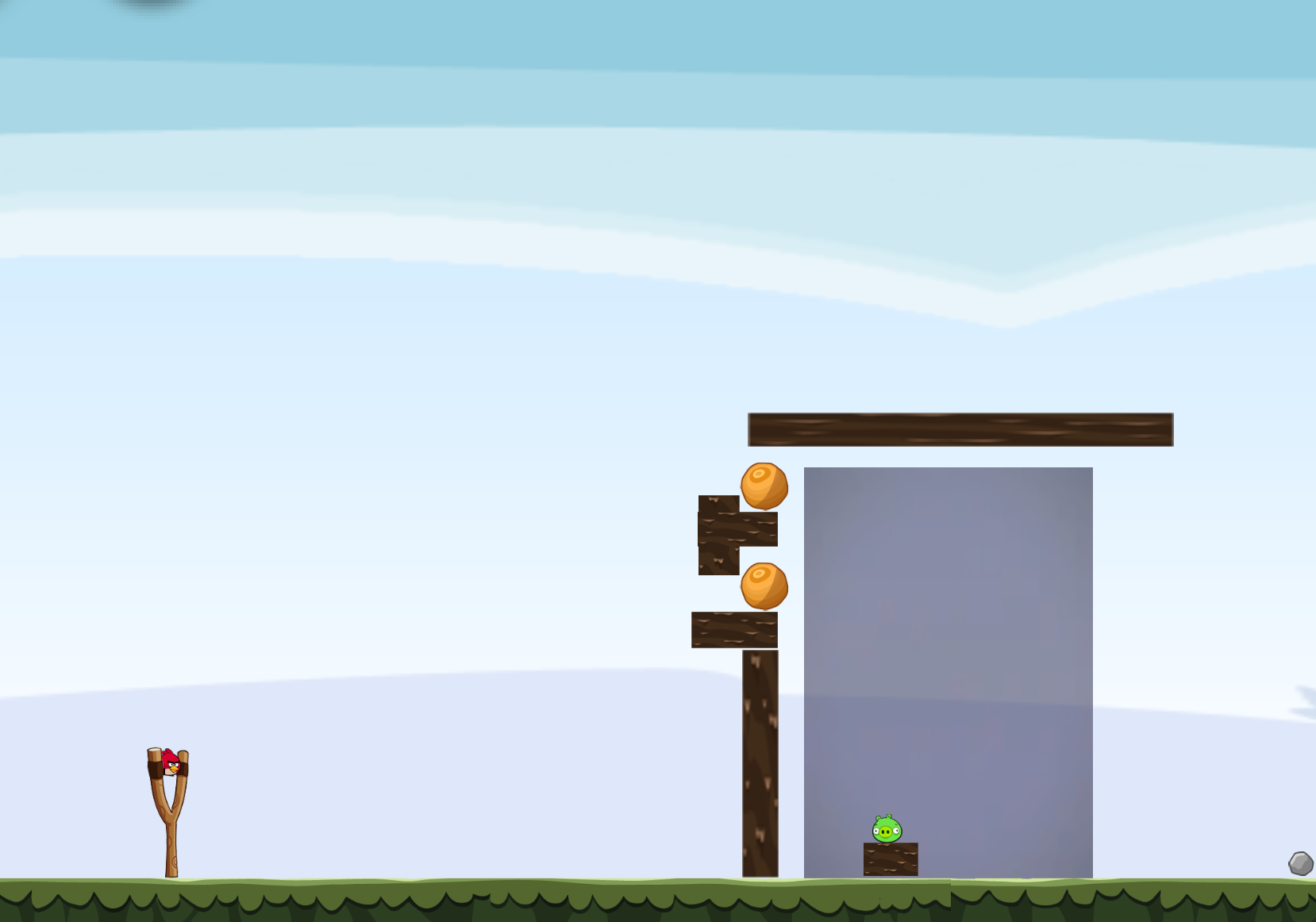}
      \end{subfigure}
      \begin{subfigure}[b]{0.49\columnwidth}
        \includegraphics[width=\linewidth]{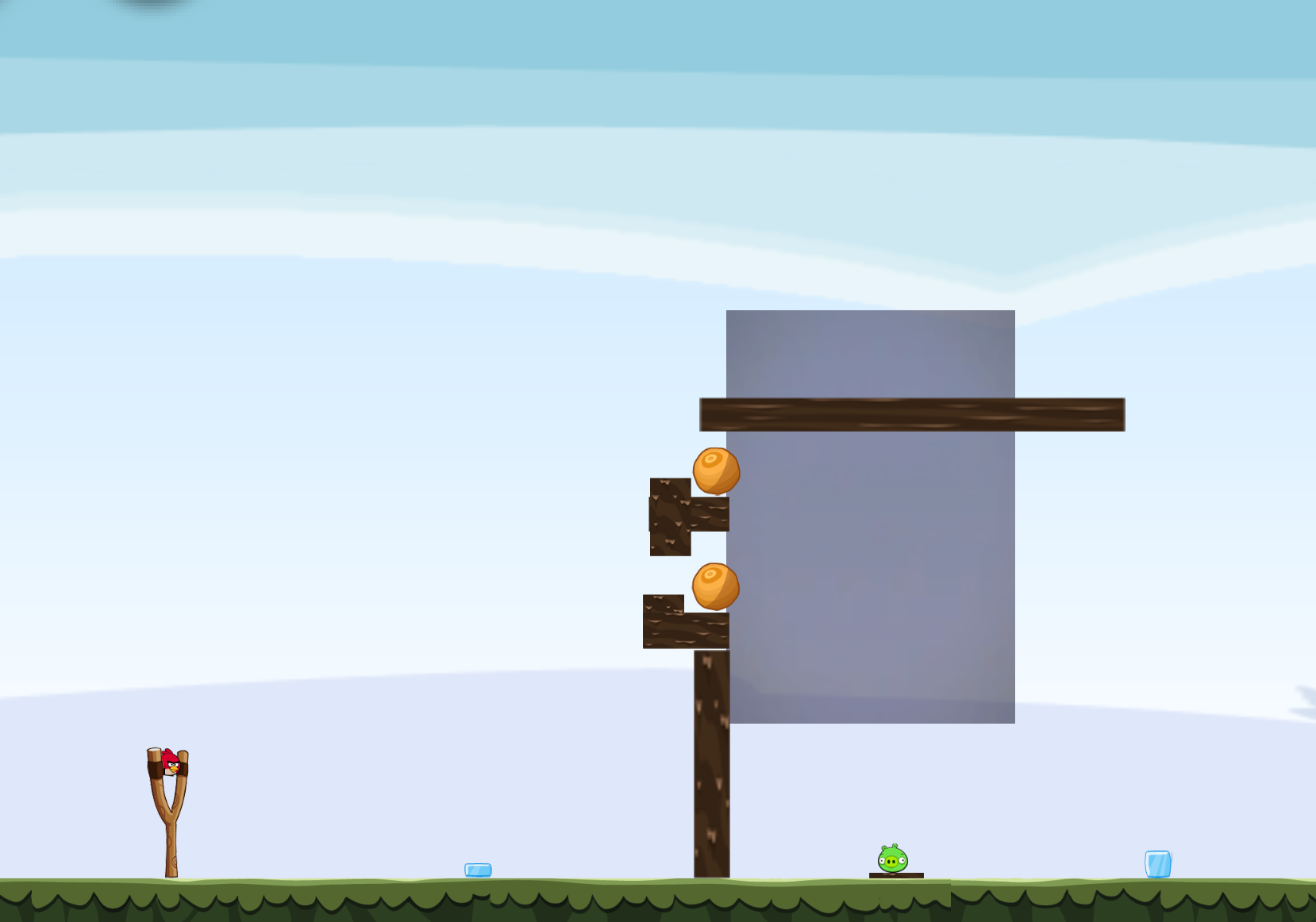}
      \end{subfigure}
  \caption{Actions}
  \end{subfigure}
  \begin{subfigure}[b]{0.49\columnwidth}
      \begin{subfigure}[b]{0.49\columnwidth}
        \includegraphics[width=\linewidth]{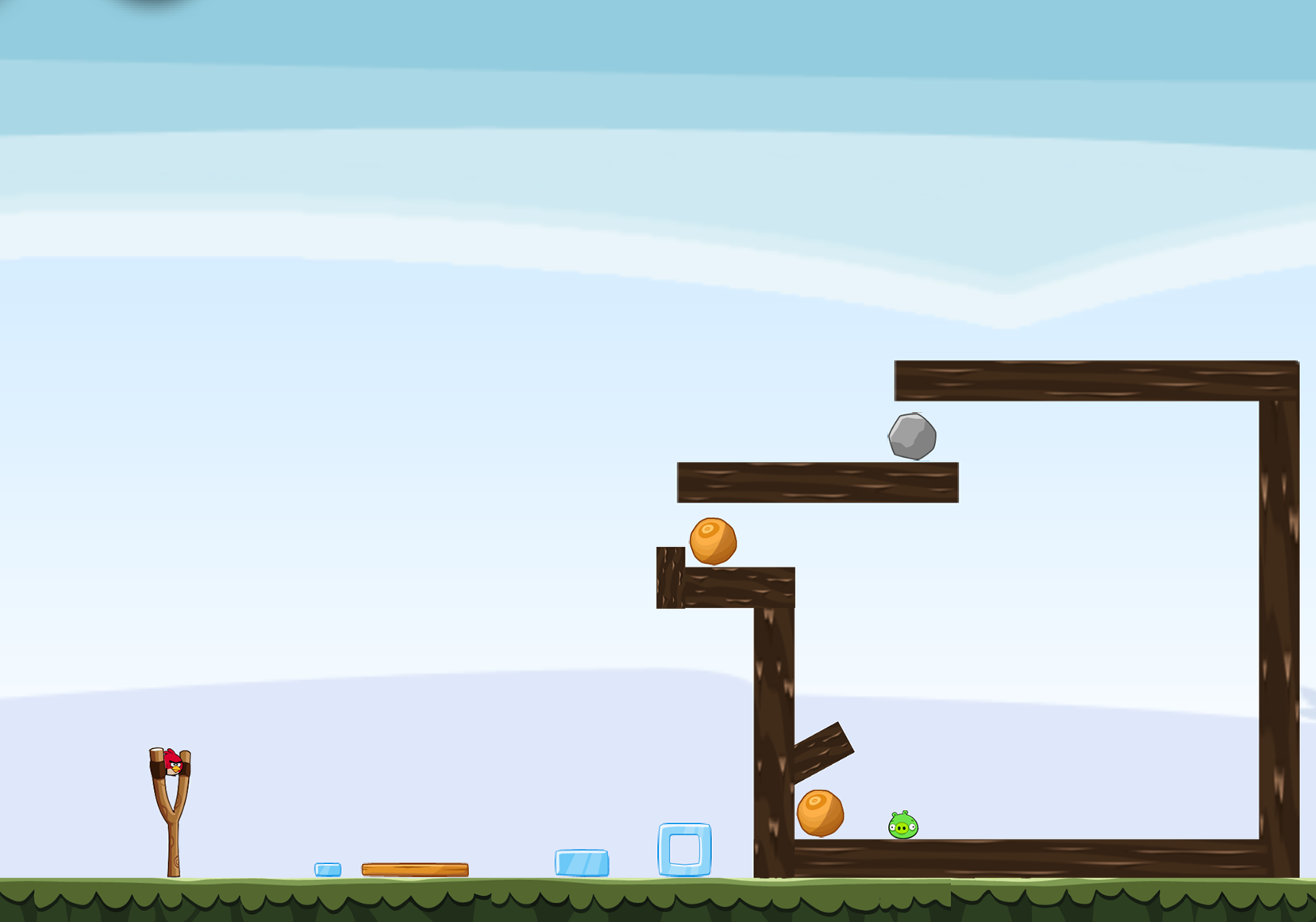}
      \end{subfigure}
      \begin{subfigure}[b]{0.49\columnwidth}
        \includegraphics[width=\linewidth]{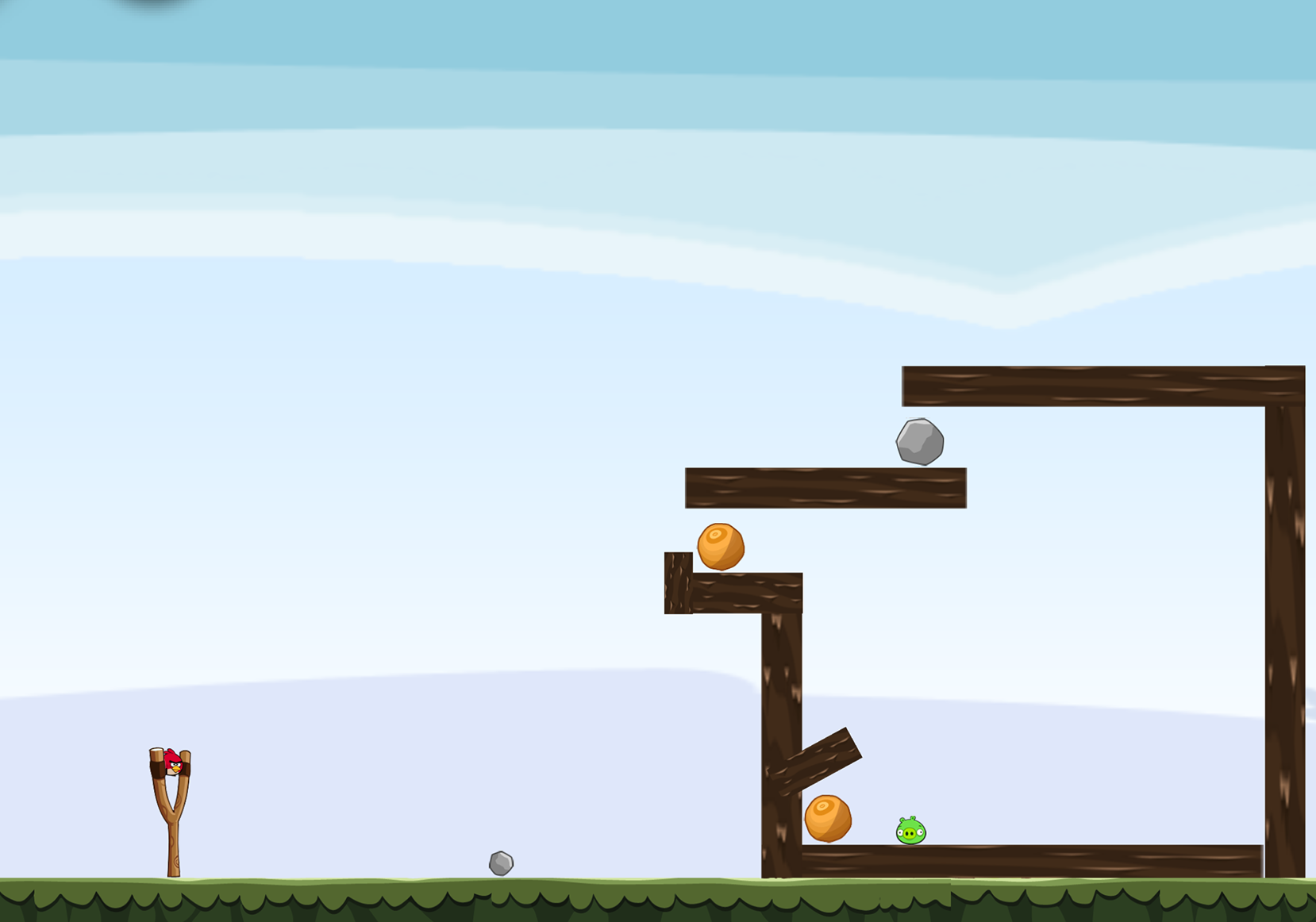}
      \end{subfigure}
  \caption{Interactions}
  \end{subfigure}
    \begin{subfigure}[b]{0.49\columnwidth}
      \begin{subfigure}[b]{0.49\columnwidth}
        \includegraphics[width=\linewidth]{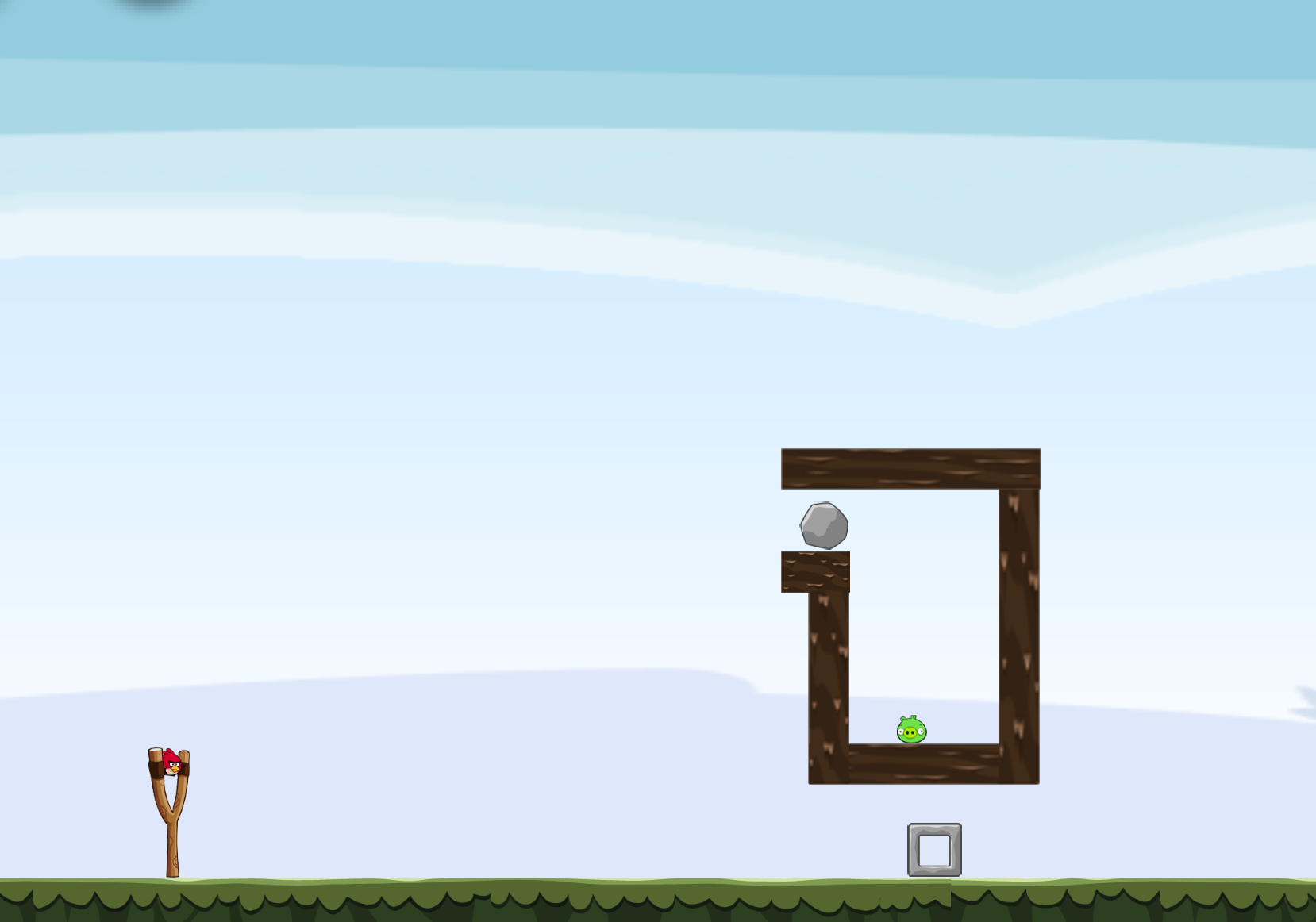}
      \end{subfigure}
      \begin{subfigure}[b]{0.49\columnwidth}
        \includegraphics[width=\linewidth]{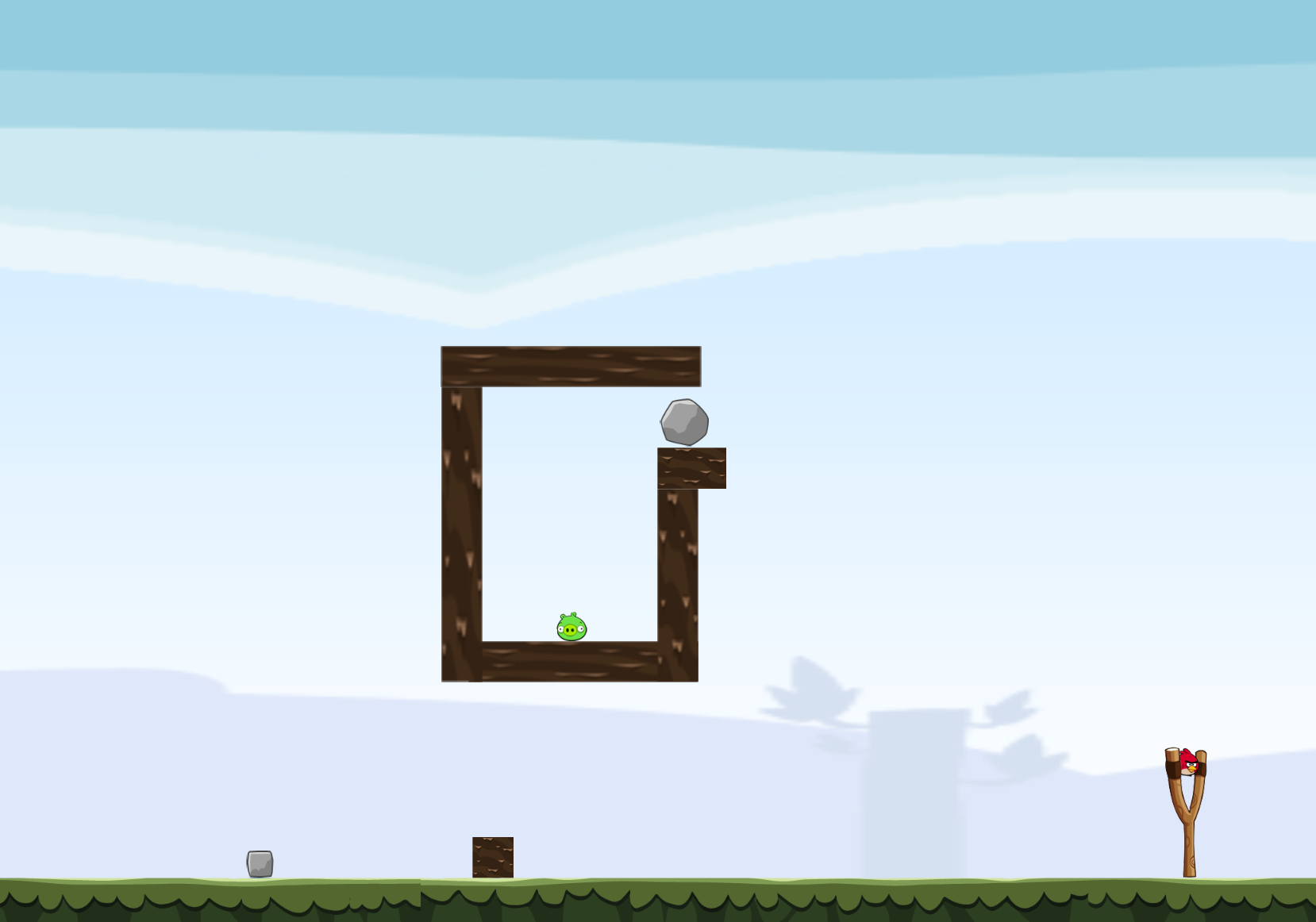}
      \end{subfigure}
  \caption{Relations}
  \end{subfigure}
  \begin{subfigure}[b]{0.49\columnwidth}
      \begin{subfigure}[b]{0.49\columnwidth}
        \includegraphics[width=\linewidth]{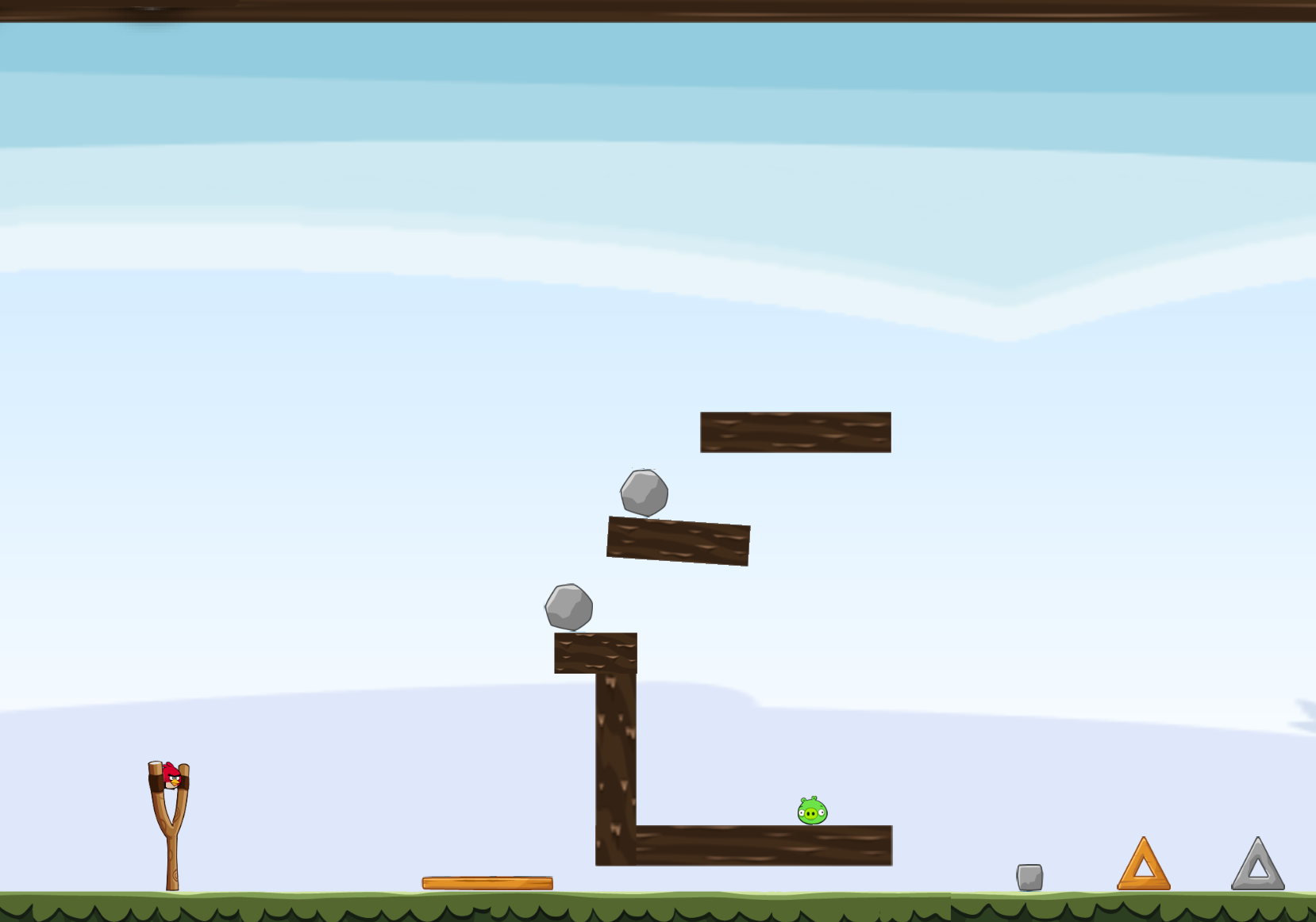}
      \end{subfigure}
      \begin{subfigure}[b]{0.49\columnwidth}
        \includegraphics[width=\linewidth]{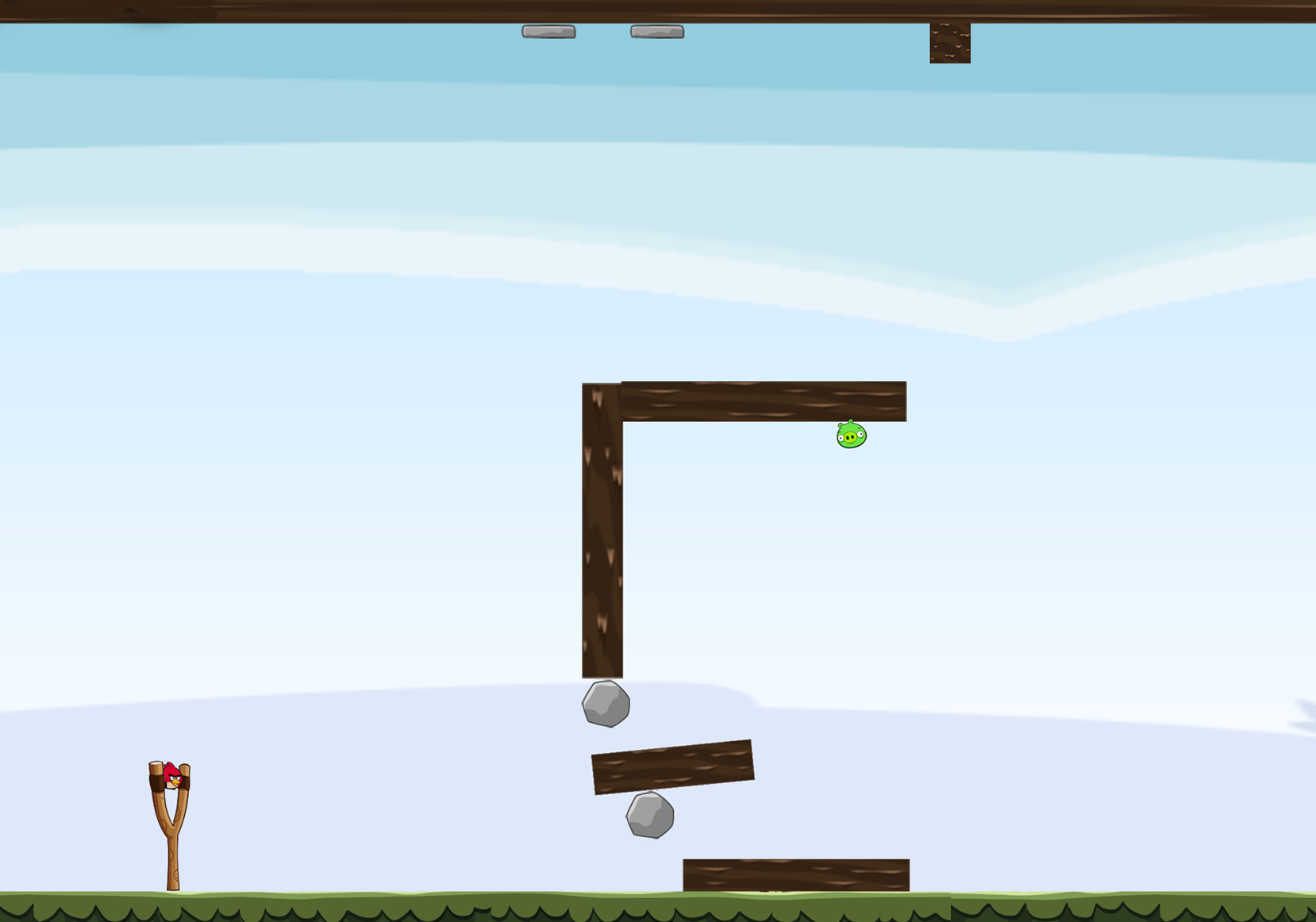}
      \end{subfigure}
  \caption{Environments}
  \end{subfigure}
  \begin{subfigure}[b]{0.49\columnwidth}
      \begin{subfigure}[b]{0.49\columnwidth}
        \includegraphics[width=\linewidth]{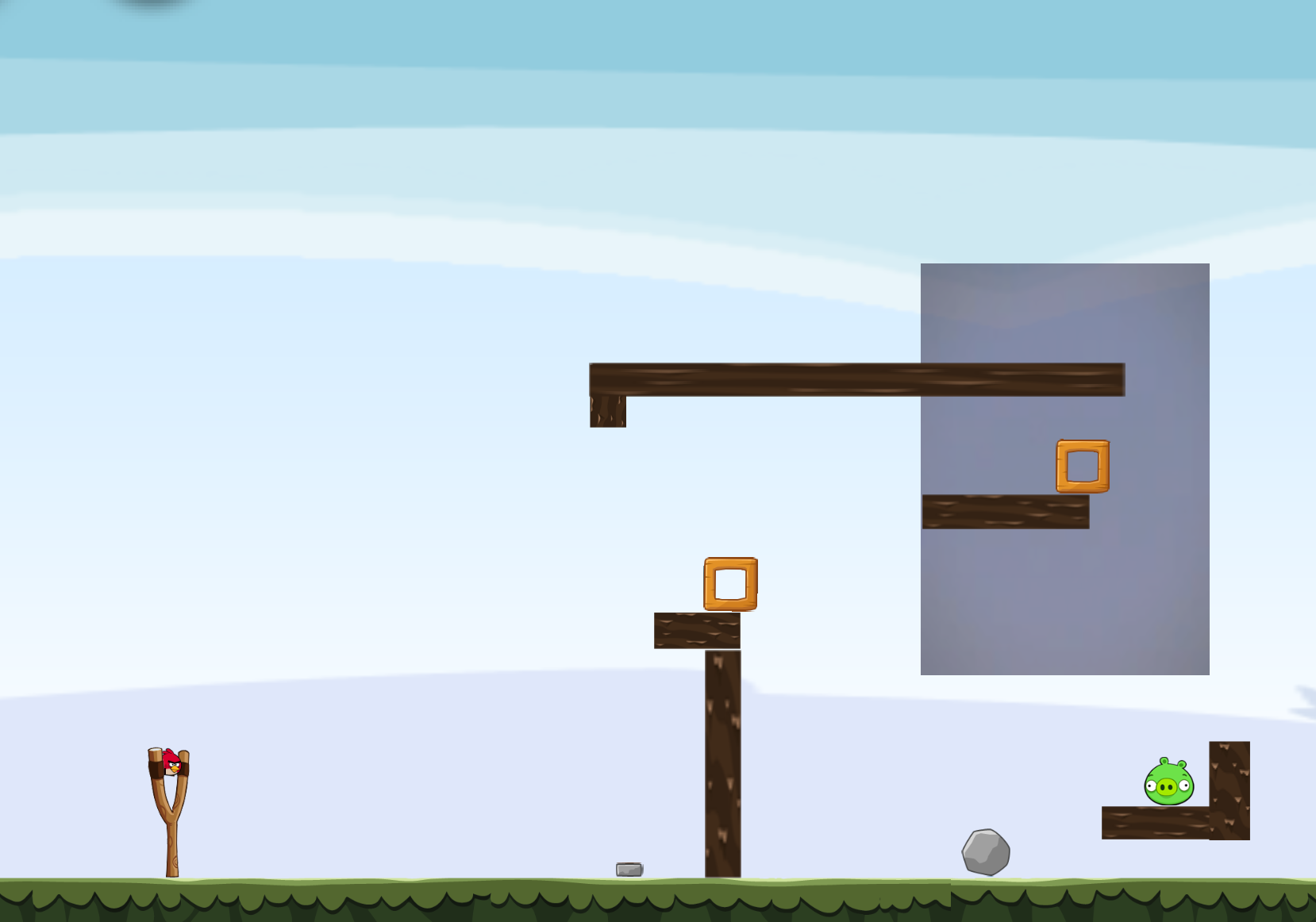}
      \end{subfigure}
      \begin{subfigure}[b]{0.49\columnwidth}
        \includegraphics[width=\linewidth]{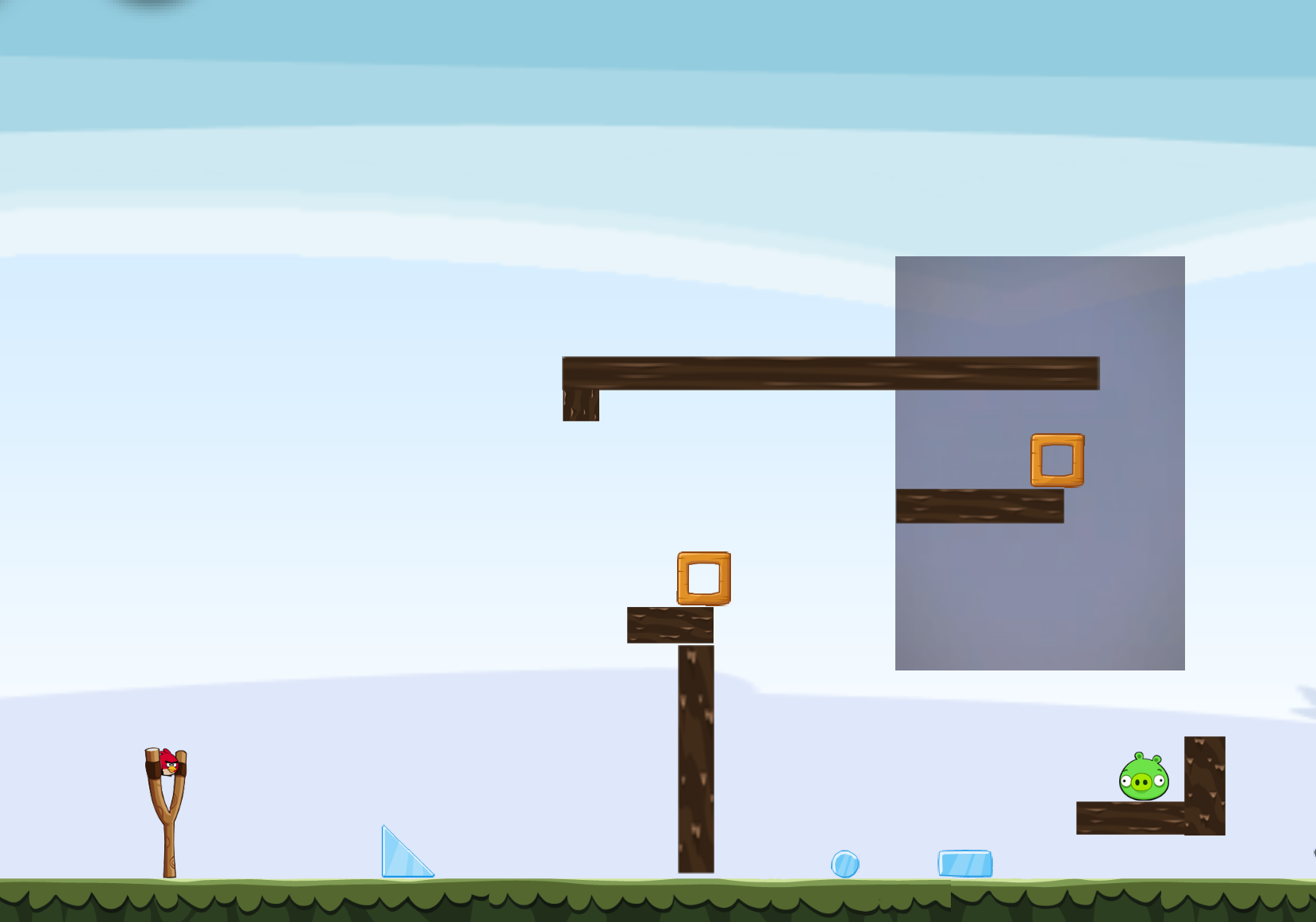}
      \end{subfigure}
  \caption{Goals}
  \end{subfigure}
  \begin{subfigure}[b]{0.49\columnwidth}
      \begin{subfigure}[b]{0.49\columnwidth}
        \includegraphics[width=\linewidth]{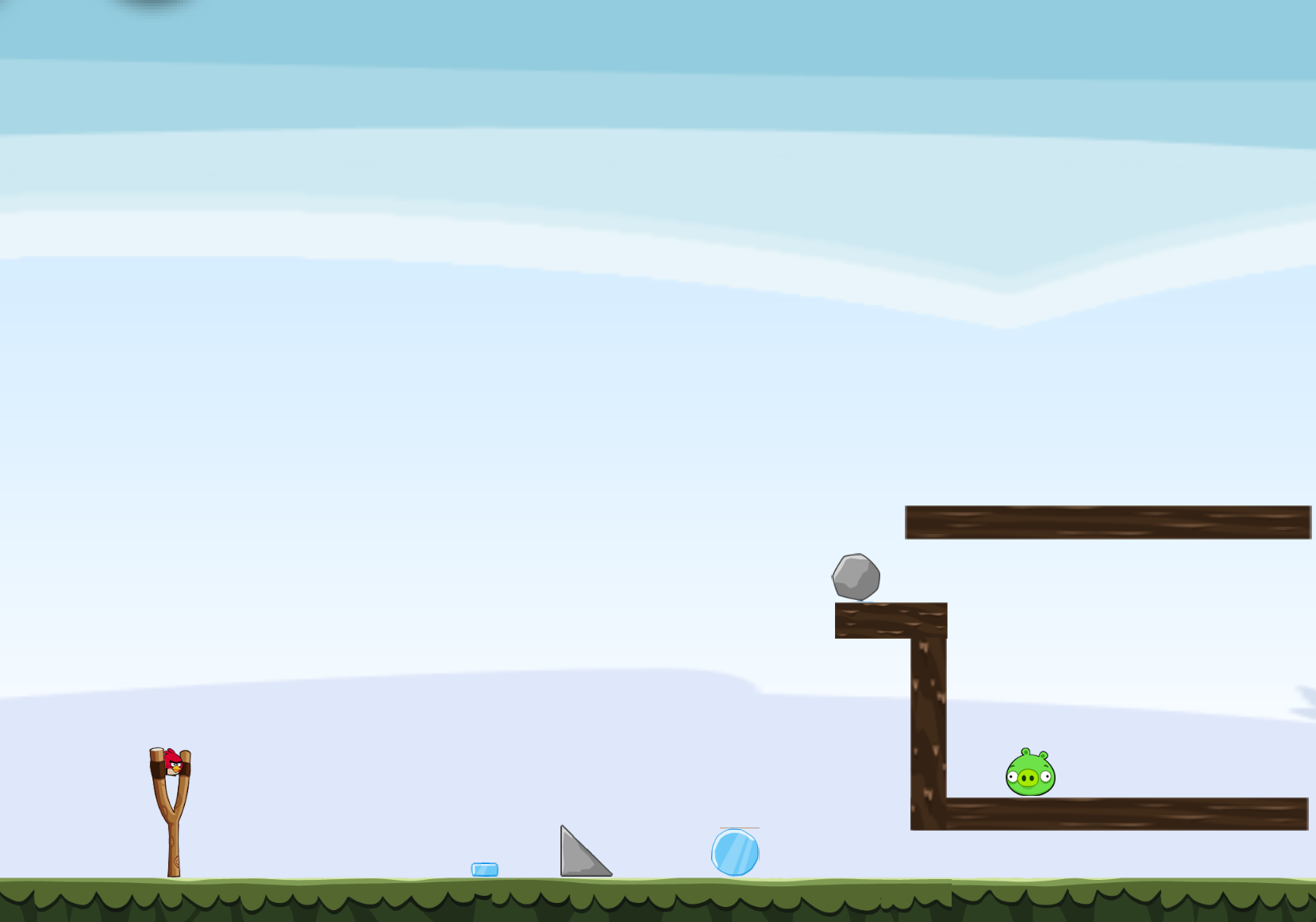}
      \end{subfigure}
      \begin{subfigure}[b]{0.49\columnwidth}
        \includegraphics[width=\linewidth]{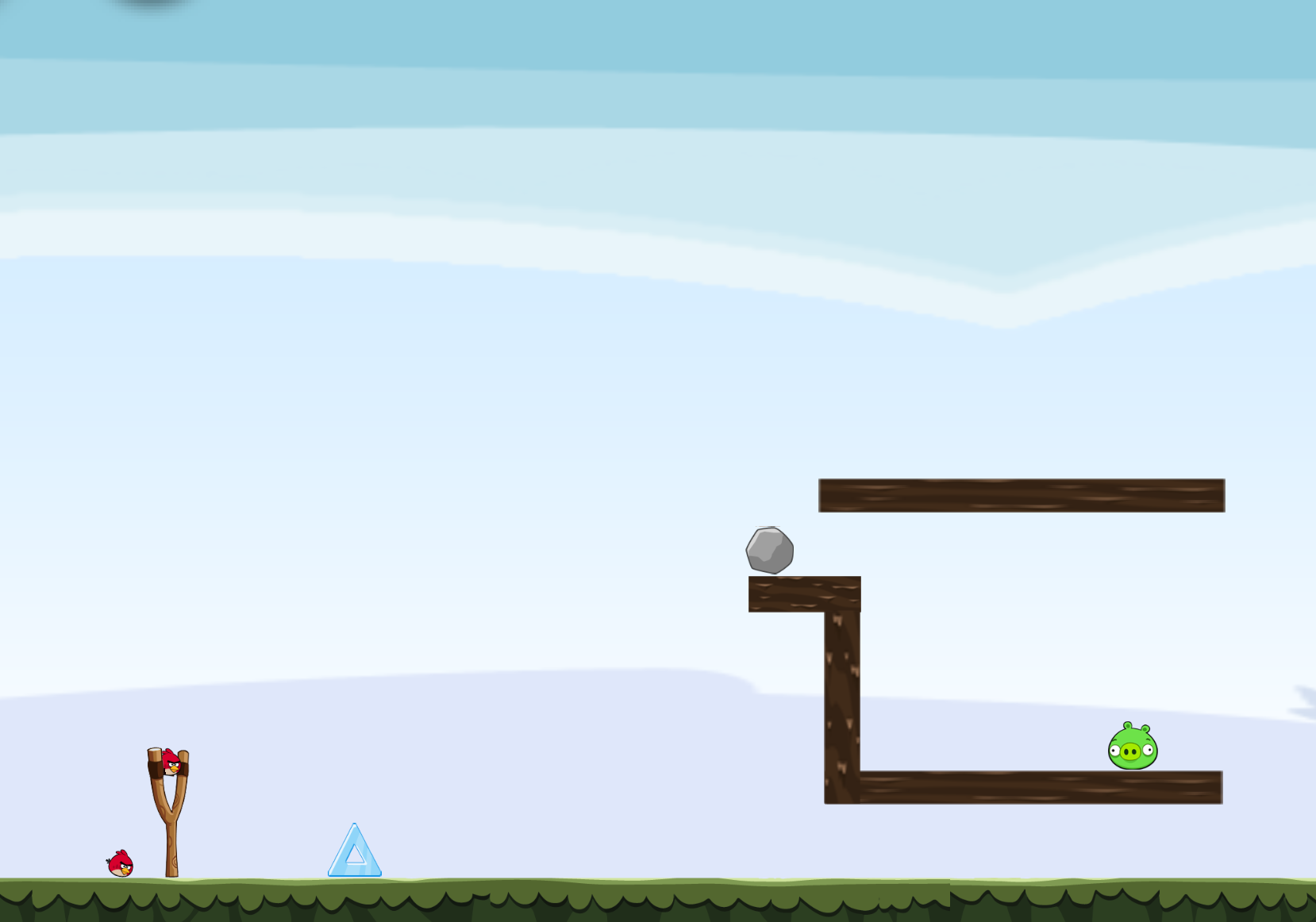}
      \end{subfigure}
  \caption{Events}
  \end{subfigure}
\caption{Task templates of the falling scenario with eight novelties applied to them. In each subfigure, the left figure is the normal task and the right figure is the corresponding novel task with the novelty applied.}
\label{appendix_fig:falling}
\end{figure}

\begin{figure}[h!]
  \centering
  \begin{subfigure}[b]{0.49\columnwidth}
      \begin{subfigure}[b]{0.49\columnwidth}
        \includegraphics[width=\linewidth]{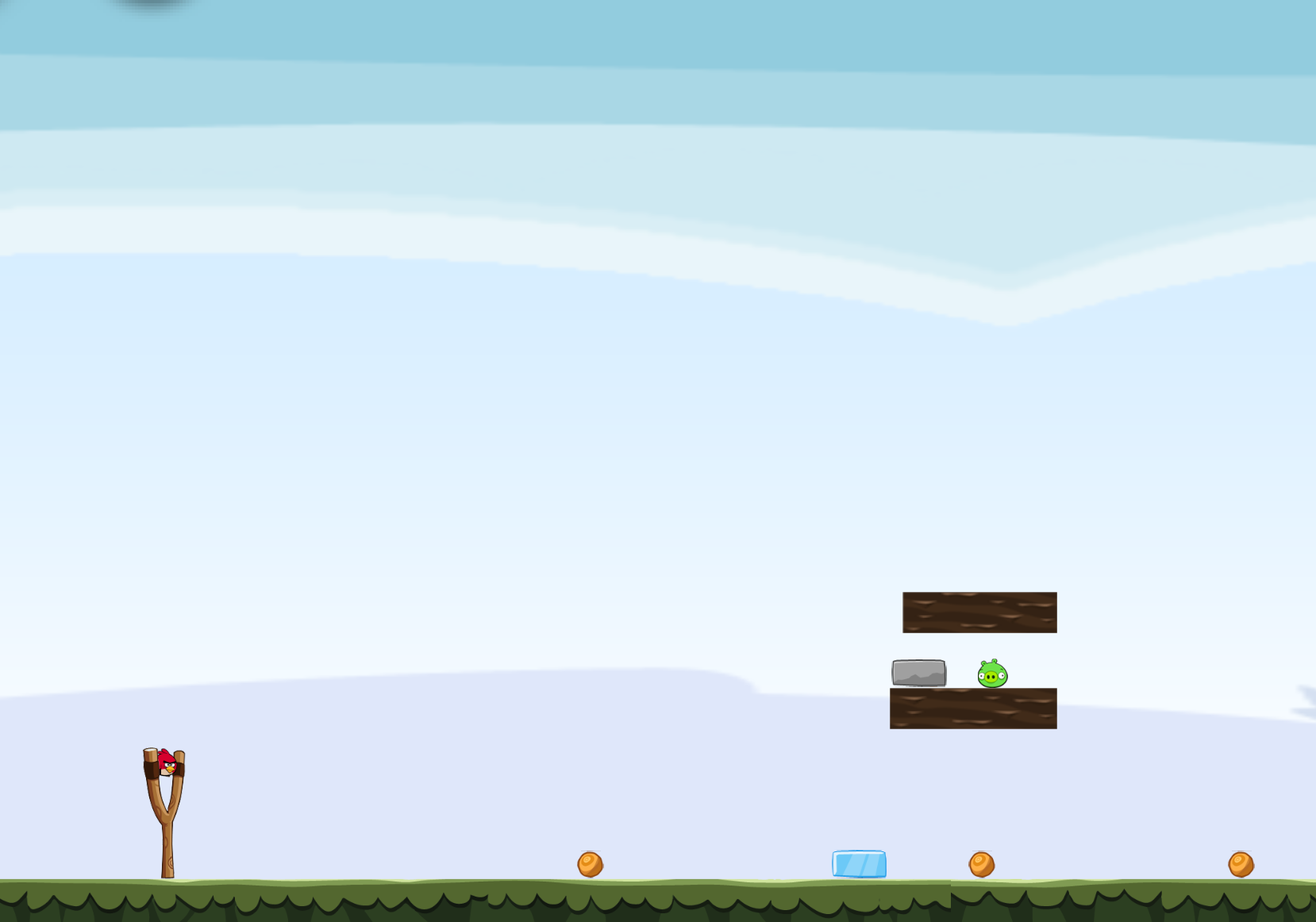}
      \end{subfigure}
      \begin{subfigure}[b]{0.49\columnwidth}
        \includegraphics[width=\linewidth]{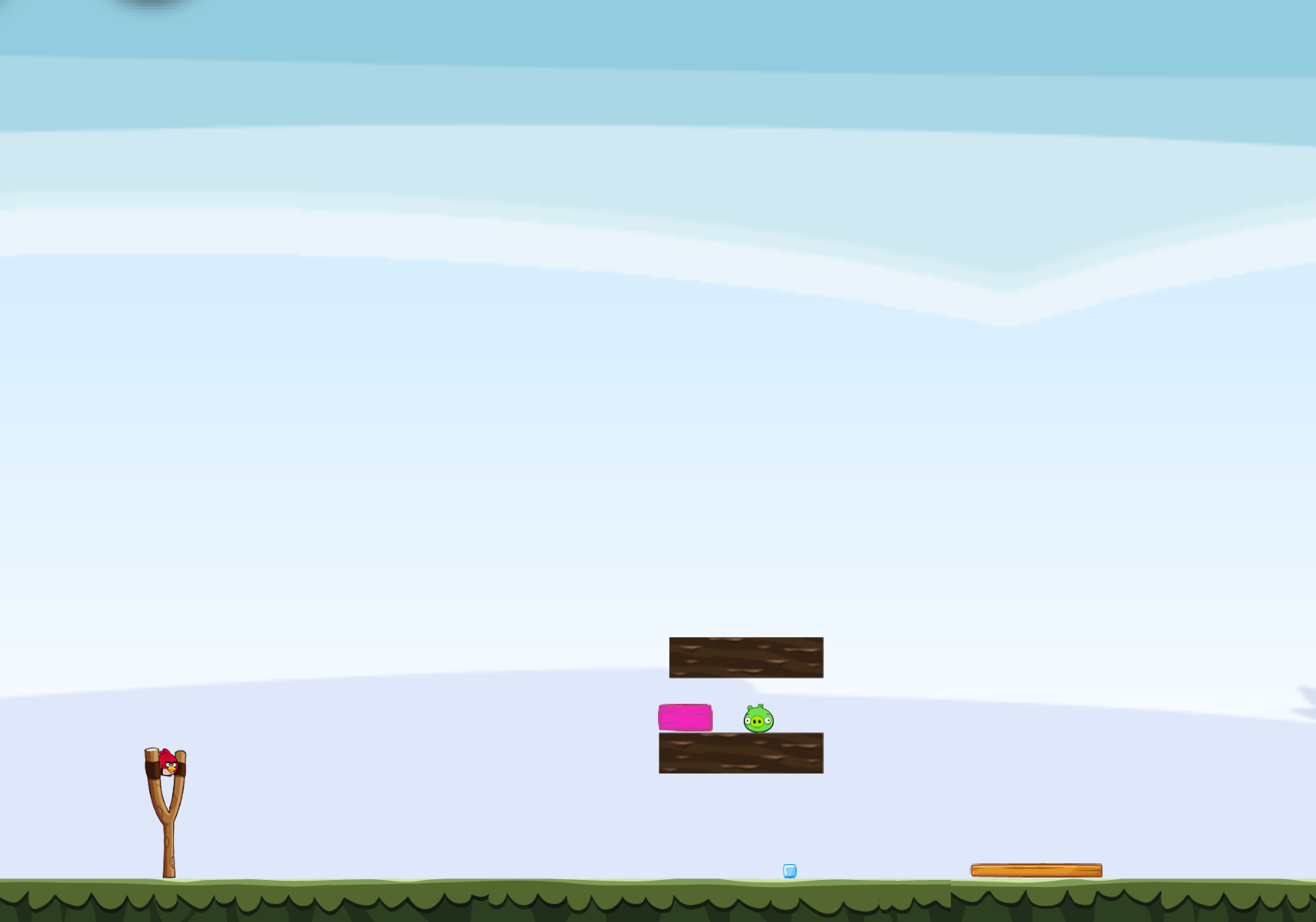}
      \end{subfigure}
  \caption{Objects}
  \end{subfigure}
  \begin{subfigure}[b]{0.49\columnwidth}
      \begin{subfigure}[b]{0.49\columnwidth}
        \includegraphics[width=\linewidth]{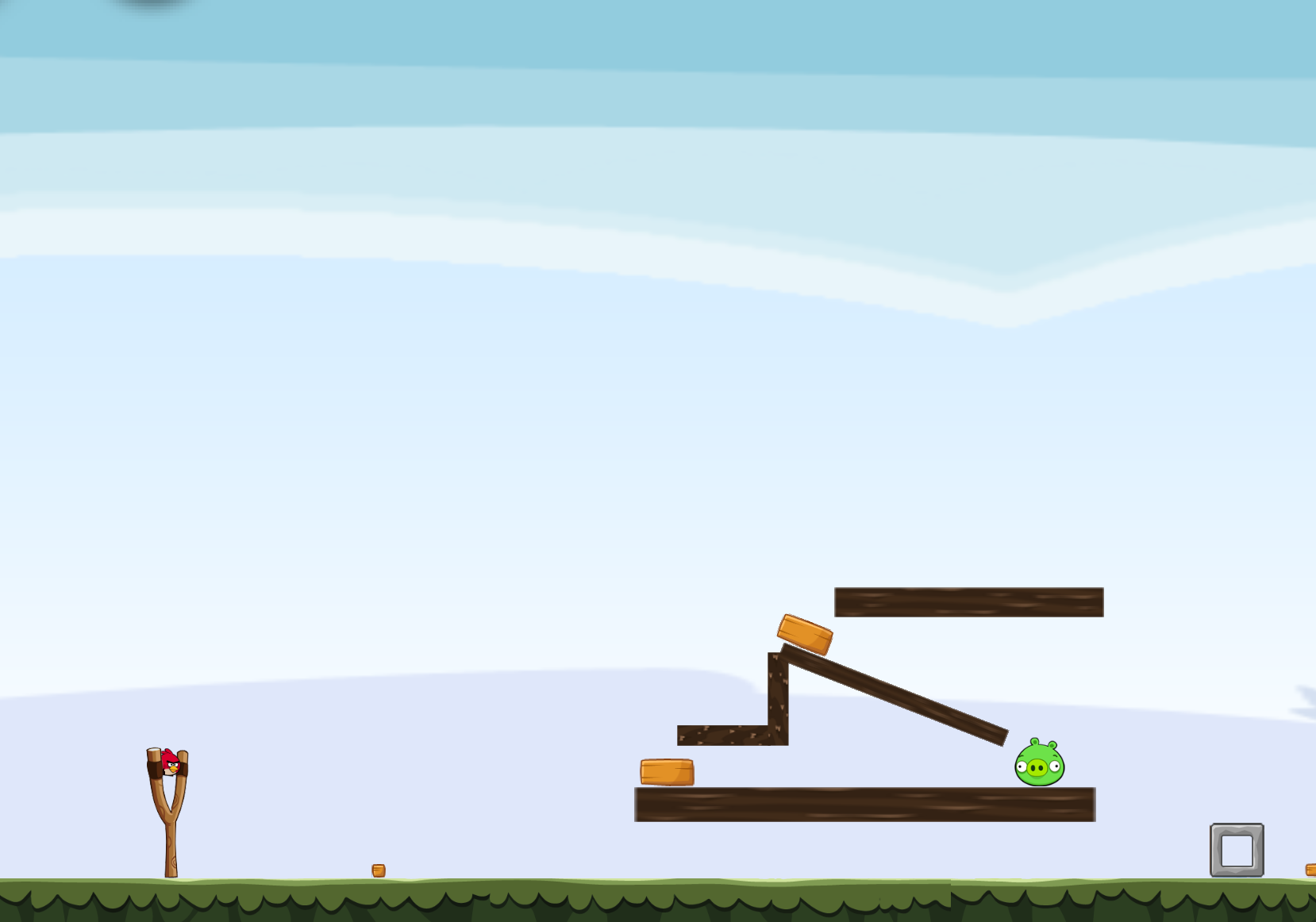}
      \end{subfigure}
      \begin{subfigure}[b]{0.49\columnwidth}
        \includegraphics[width=\linewidth]{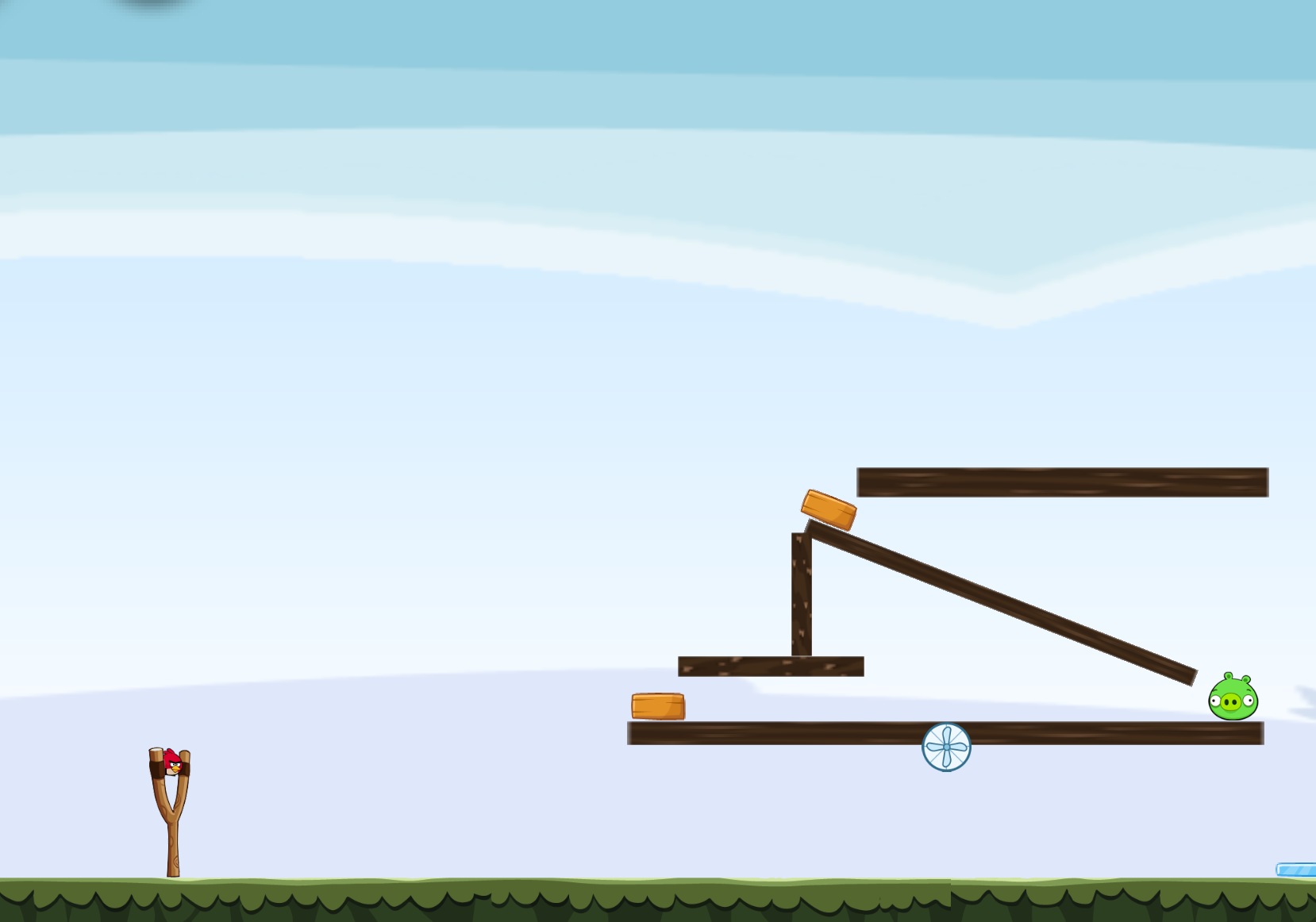}
      \end{subfigure}
  \caption{Agents}
  \end{subfigure}
  \begin{subfigure}[b]{0.49\columnwidth}
      \begin{subfigure}[b]{0.49\columnwidth}
        \includegraphics[width=\linewidth]{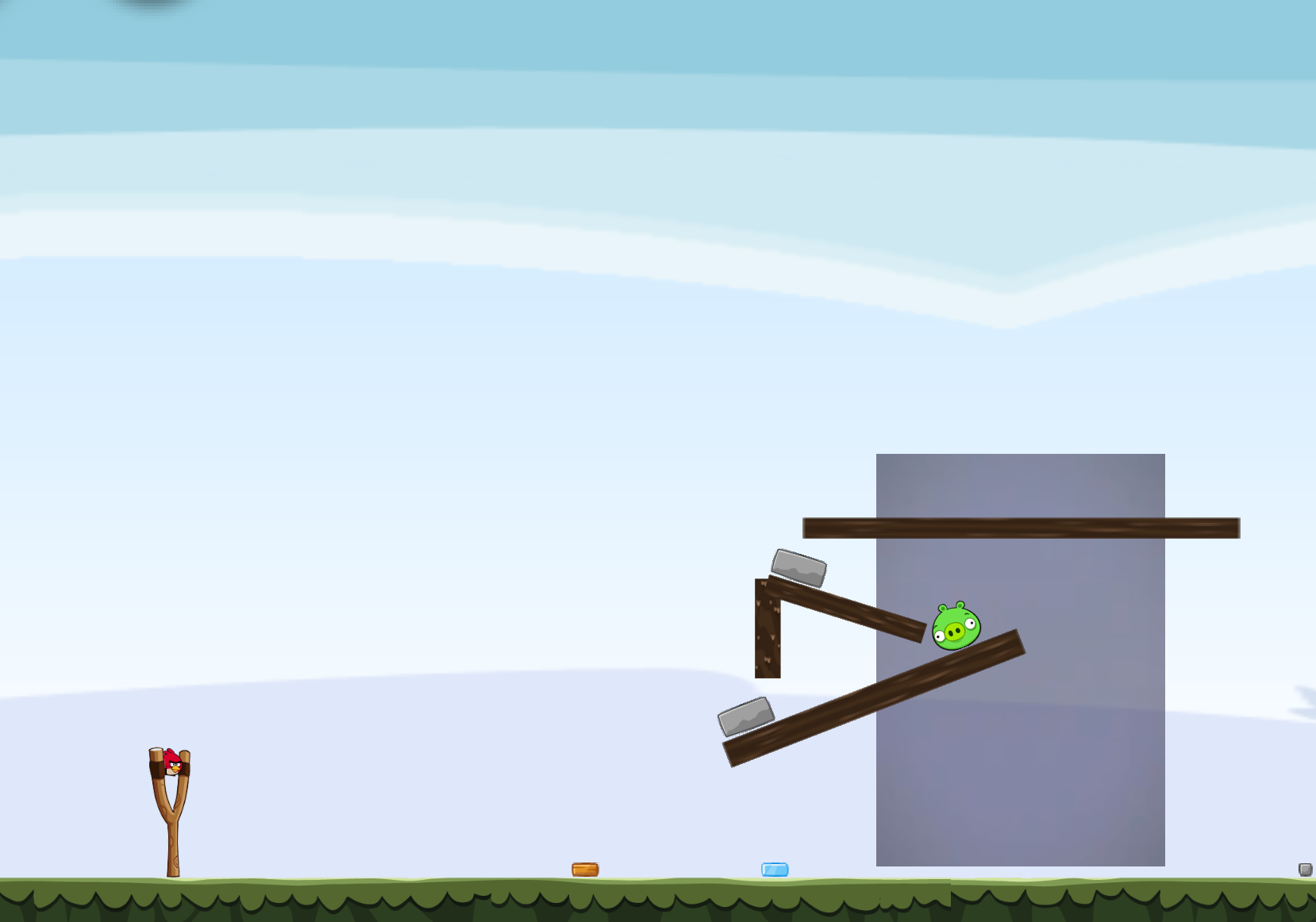}
      \end{subfigure}
      \begin{subfigure}[b]{0.49\columnwidth}
        \includegraphics[width=\linewidth]{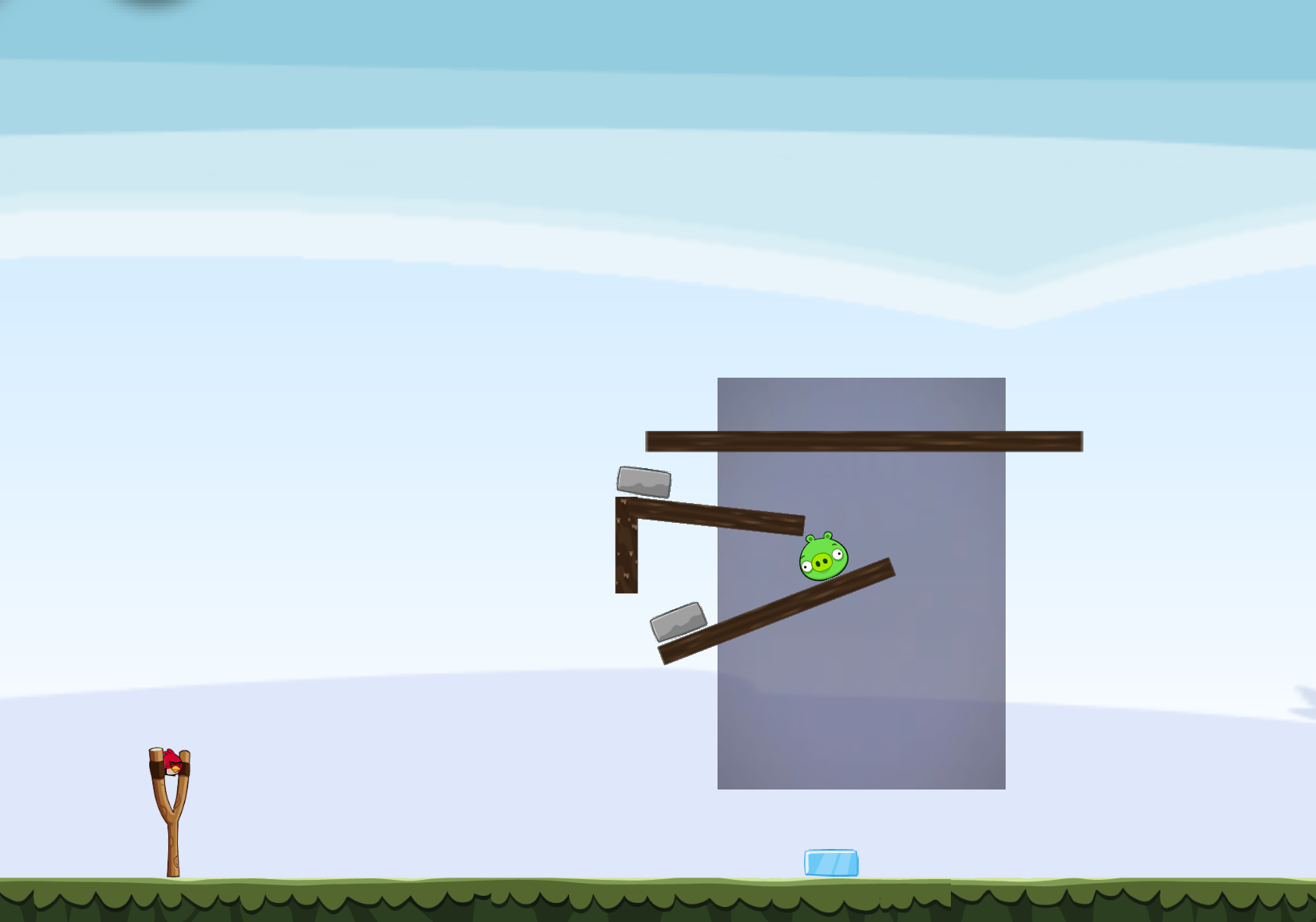}
      \end{subfigure}
  \caption{Actions}
  \end{subfigure}
  \begin{subfigure}[b]{0.49\columnwidth}
      \begin{subfigure}[b]{0.49\columnwidth}
        \includegraphics[width=\linewidth]{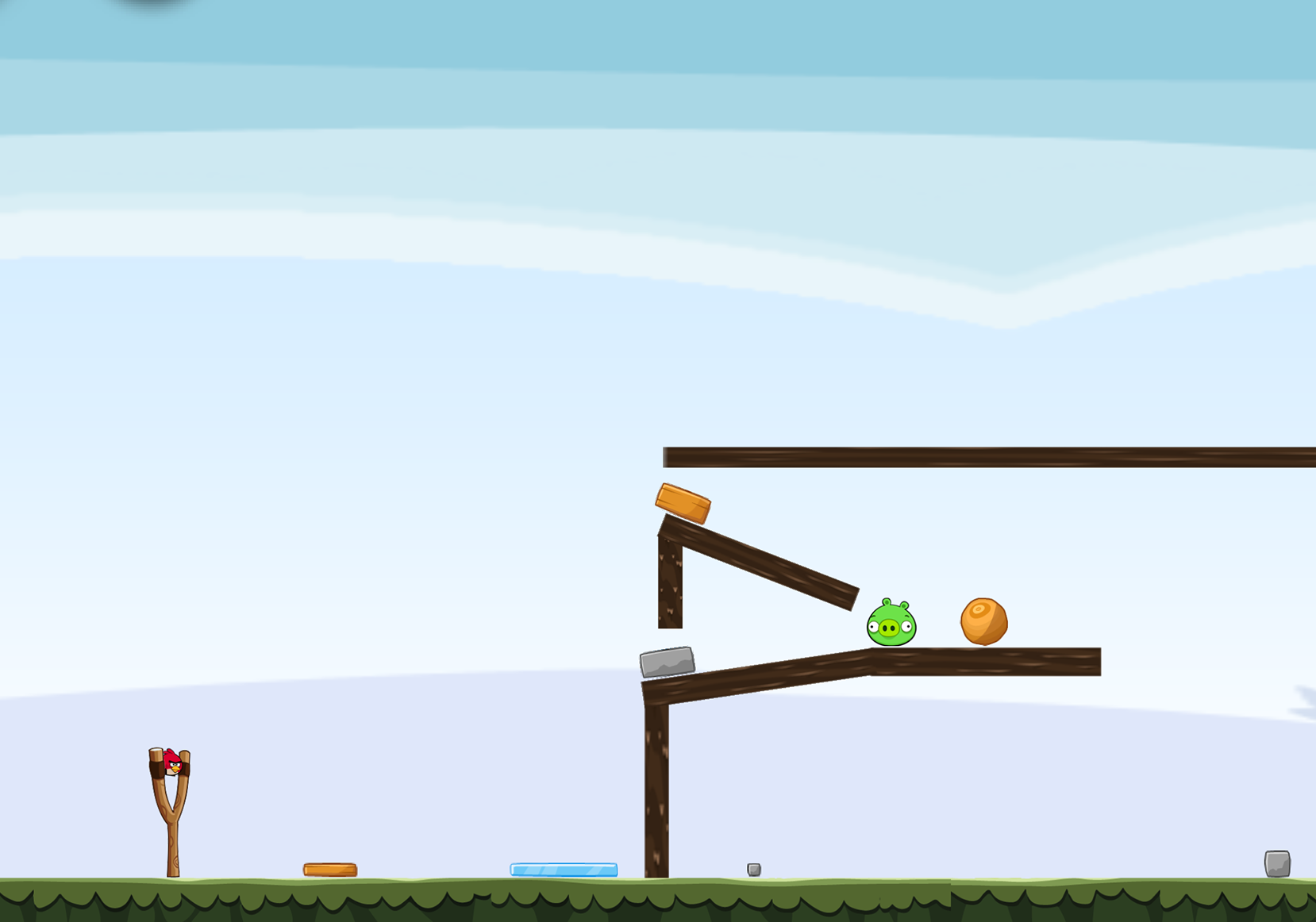}
      \end{subfigure}
      \begin{subfigure}[b]{0.49\columnwidth}
        \includegraphics[width=\linewidth]{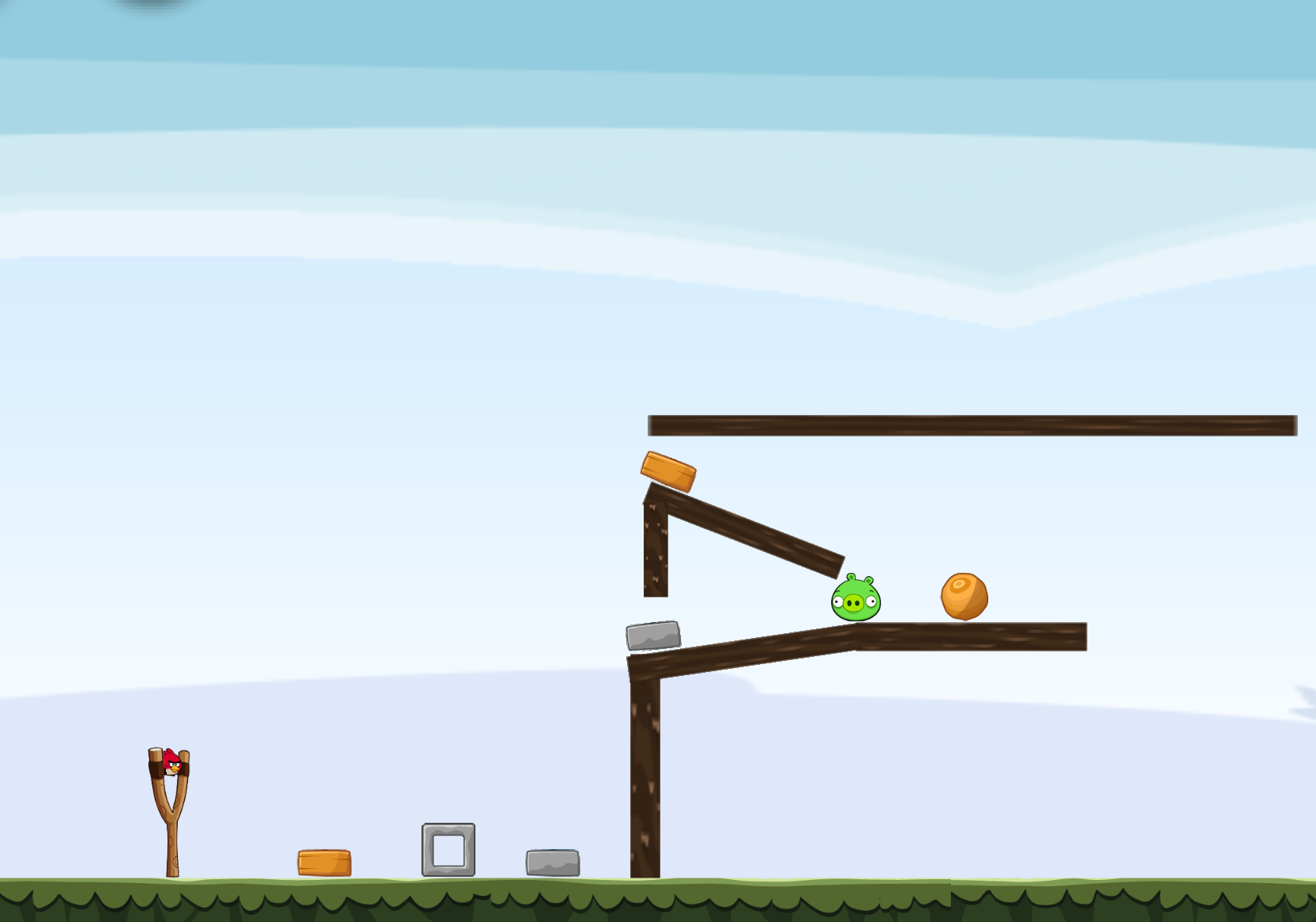}
      \end{subfigure}
  \caption{Interactions}
  \end{subfigure}
    \begin{subfigure}[b]{0.49\columnwidth}
      \begin{subfigure}[b]{0.49\columnwidth}
        \includegraphics[width=\linewidth]{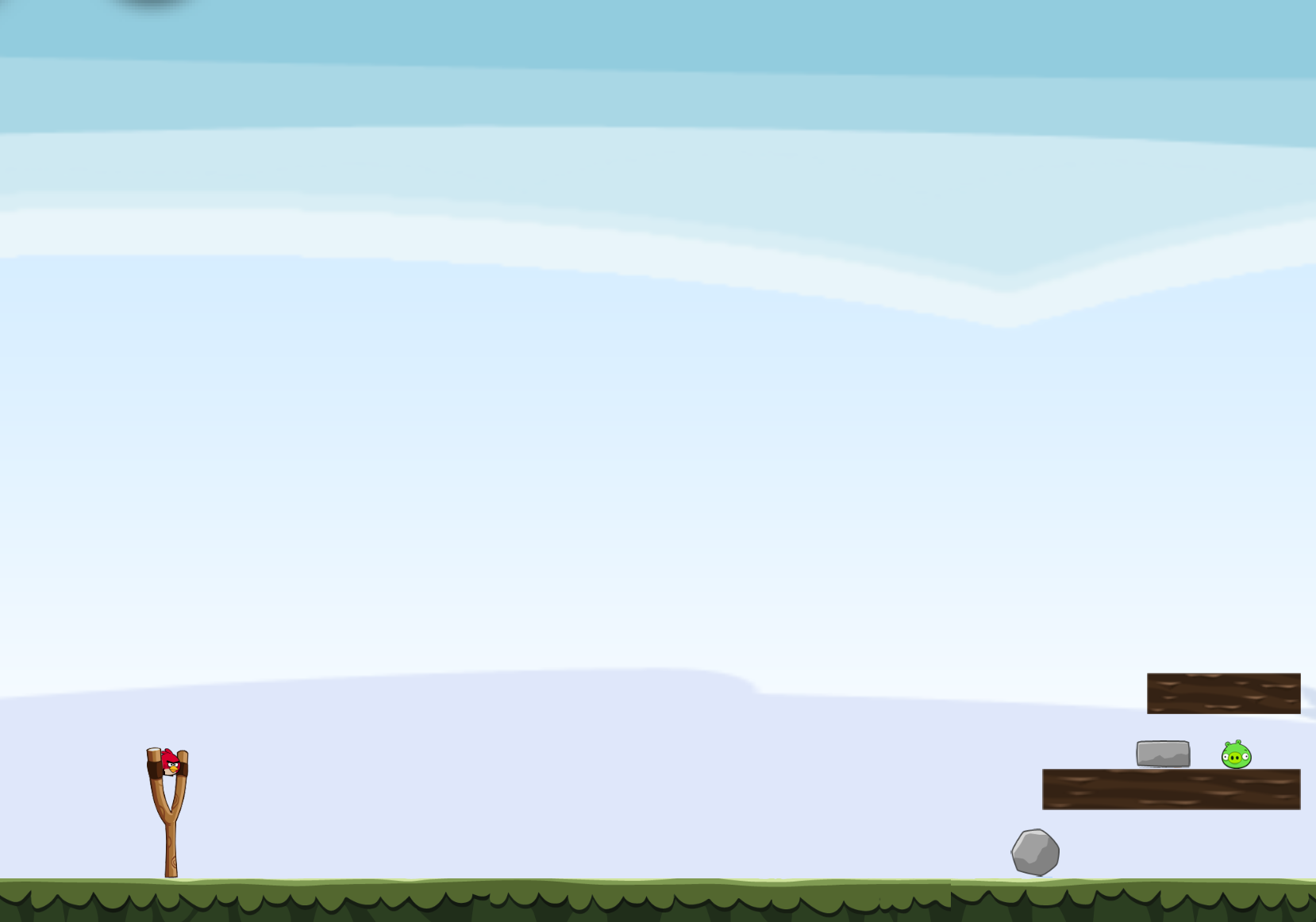}
      \end{subfigure}
      \begin{subfigure}[b]{0.49\columnwidth}
        \includegraphics[width=\linewidth]{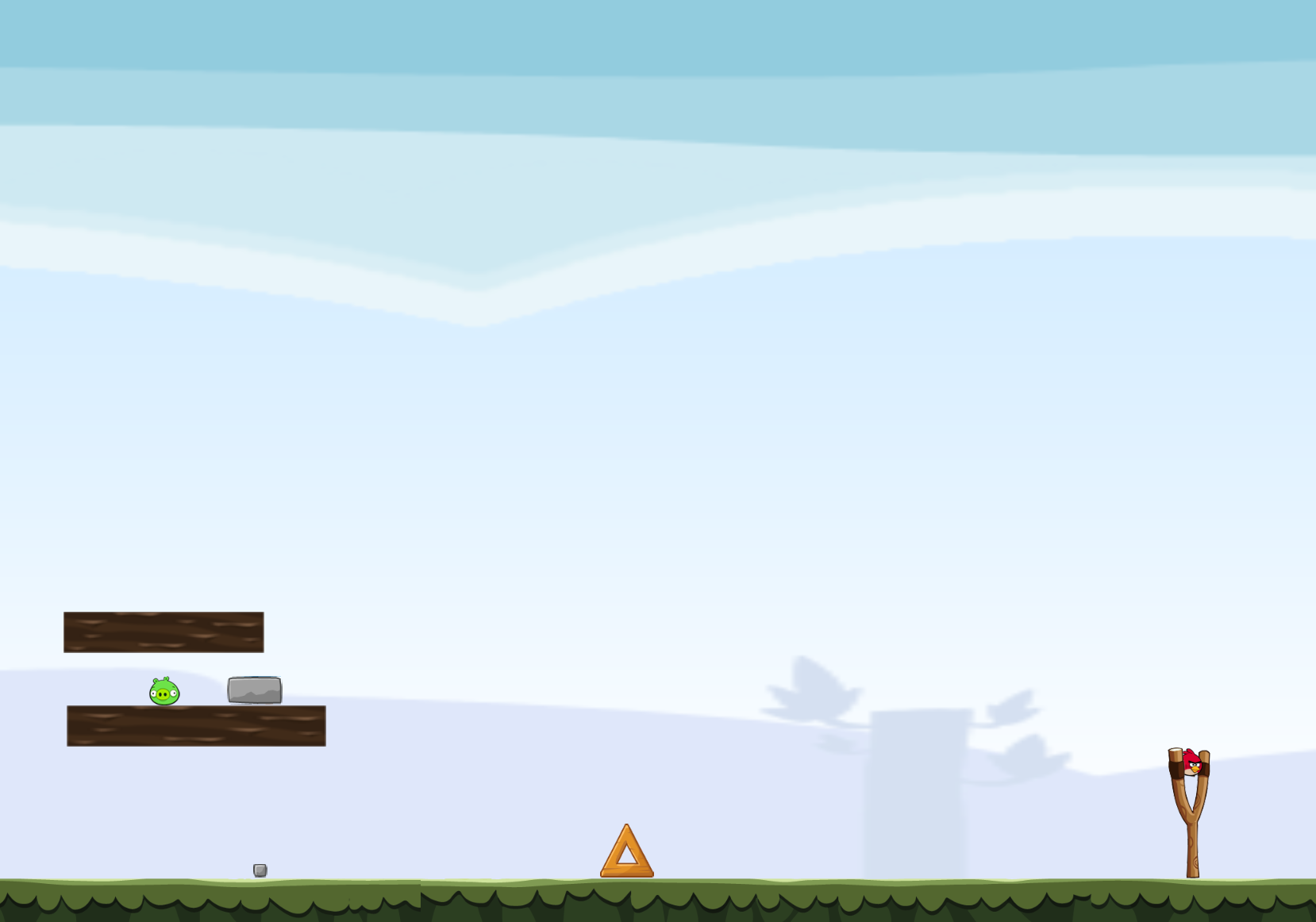}
      \end{subfigure}
  \caption{Relations}
  \end{subfigure}
  \begin{subfigure}[b]{0.49\columnwidth}
      \begin{subfigure}[b]{0.49\columnwidth}
        \includegraphics[width=\linewidth]{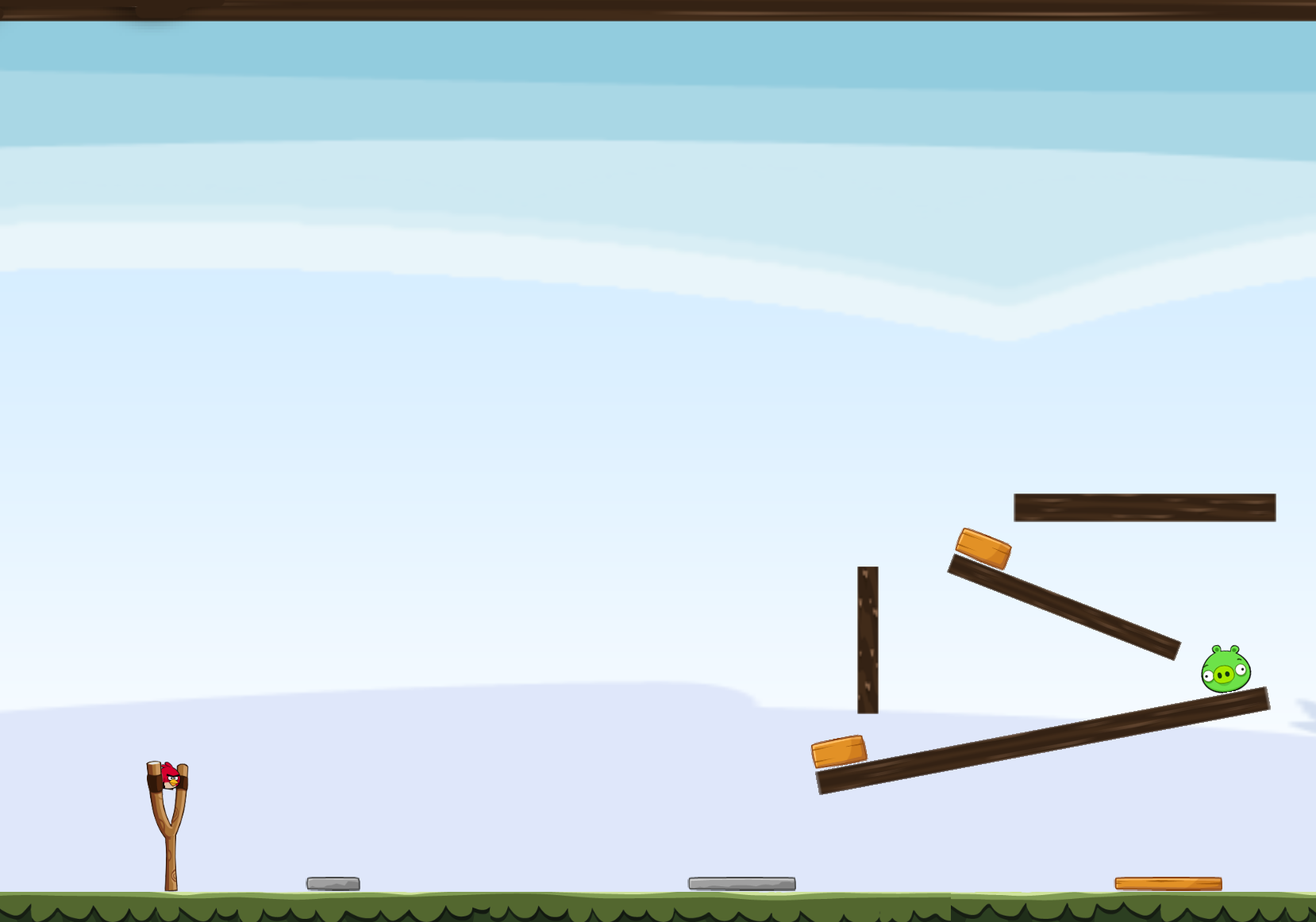}
      \end{subfigure}
      \begin{subfigure}[b]{0.49\columnwidth}
        \includegraphics[width=\linewidth]{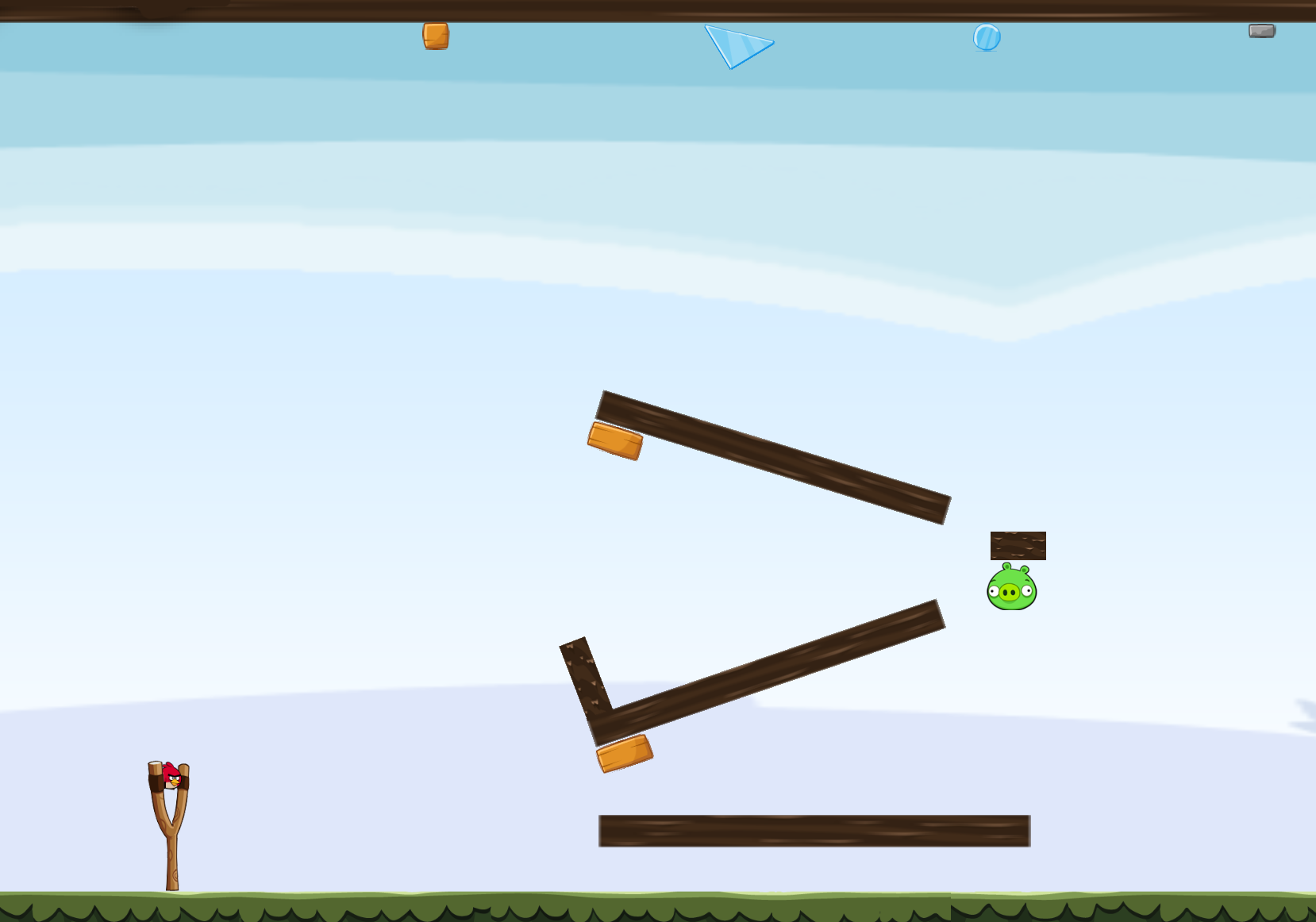}
      \end{subfigure}
  \caption{Environments}
  \end{subfigure}
  \begin{subfigure}[b]{0.49\columnwidth}
      \begin{subfigure}[b]{0.49\columnwidth}
        \includegraphics[width=\linewidth]{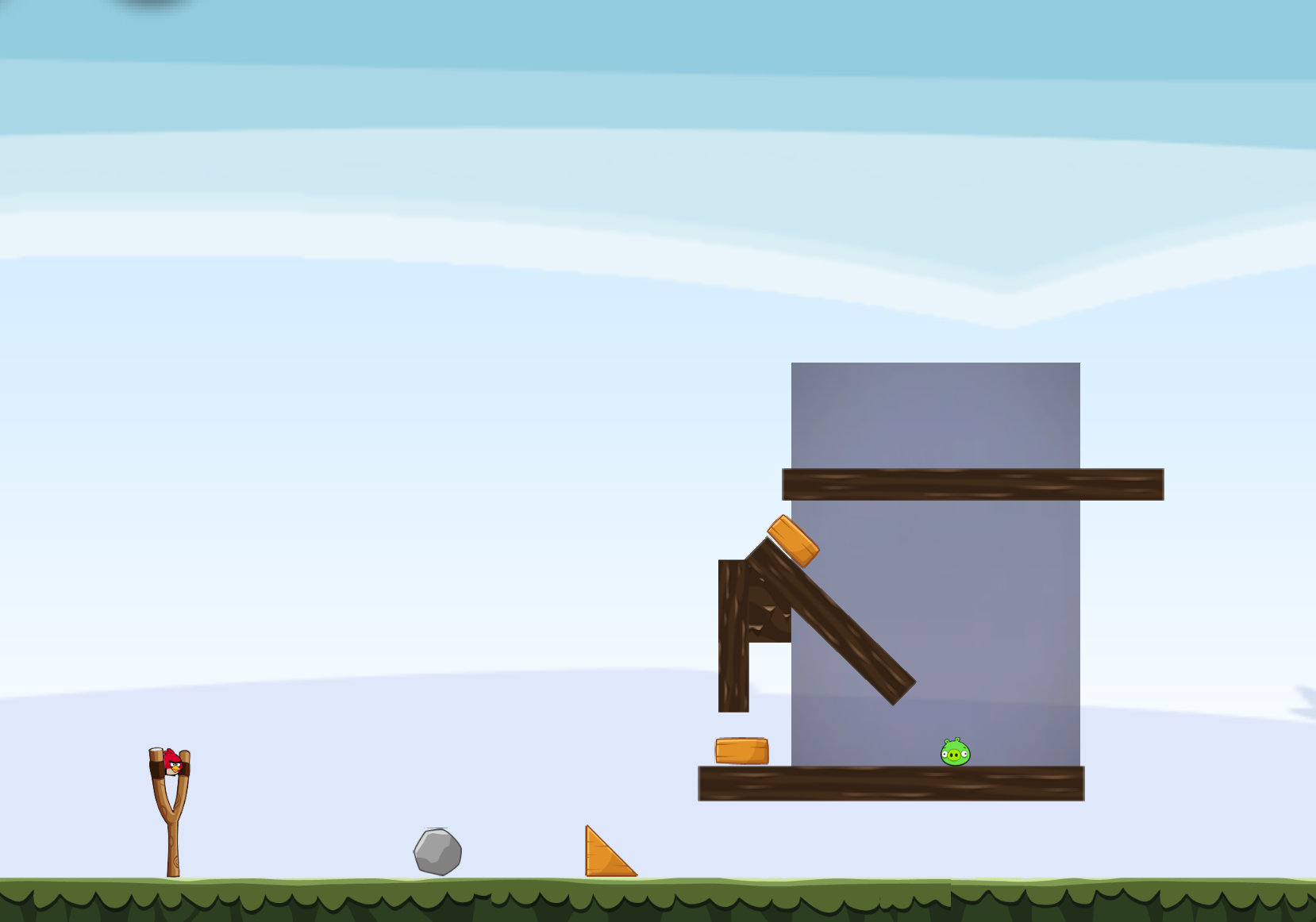}
      \end{subfigure}
      \begin{subfigure}[b]{0.49\columnwidth}
        \includegraphics[width=\linewidth]{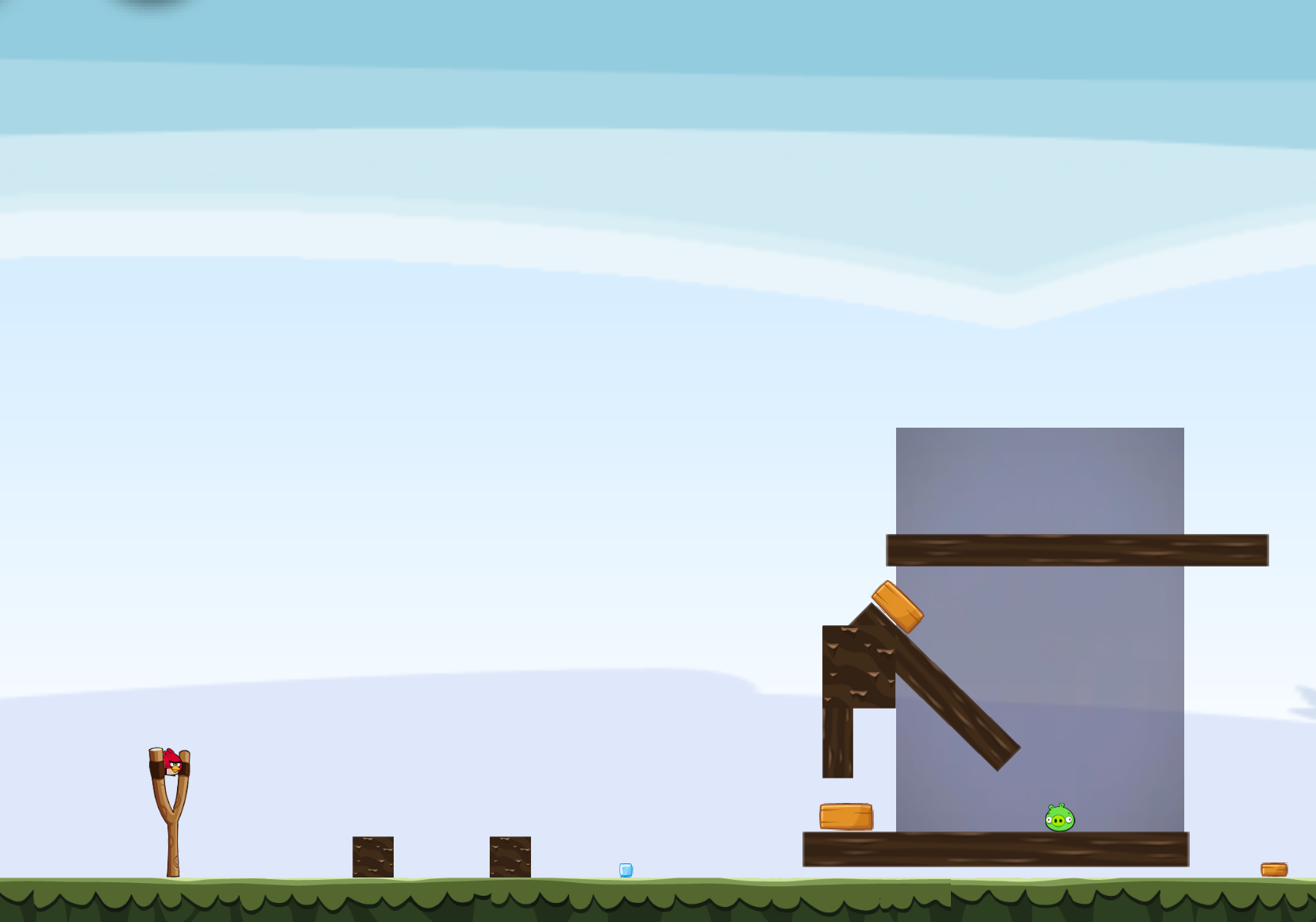}
      \end{subfigure}
  \caption{Goals}
  \end{subfigure}
  \begin{subfigure}[b]{0.49\columnwidth}
      \begin{subfigure}[b]{0.49\columnwidth}
        \includegraphics[width=\linewidth]{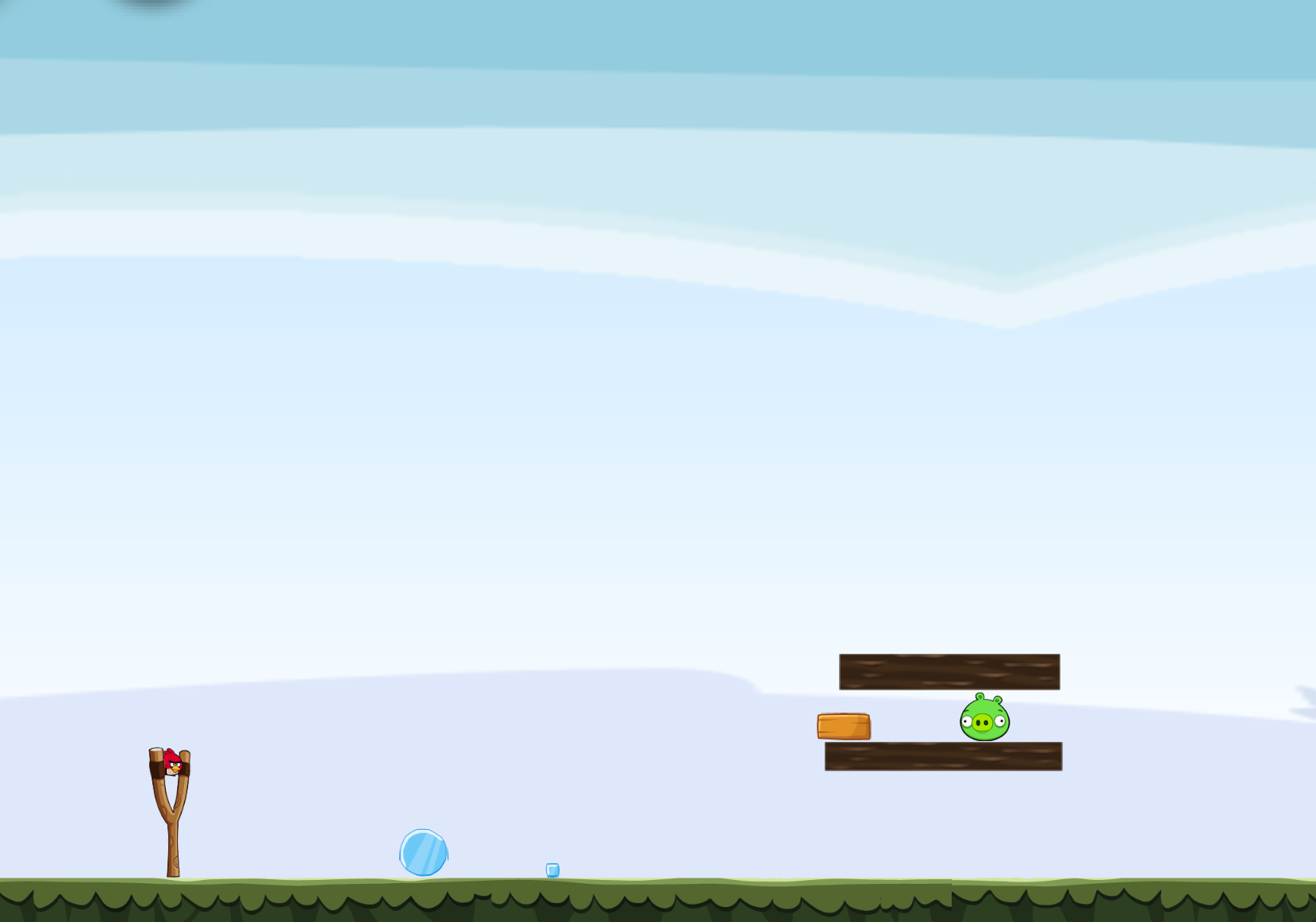}
      \end{subfigure}
      \begin{subfigure}[b]{0.49\columnwidth}
        \includegraphics[width=\linewidth]{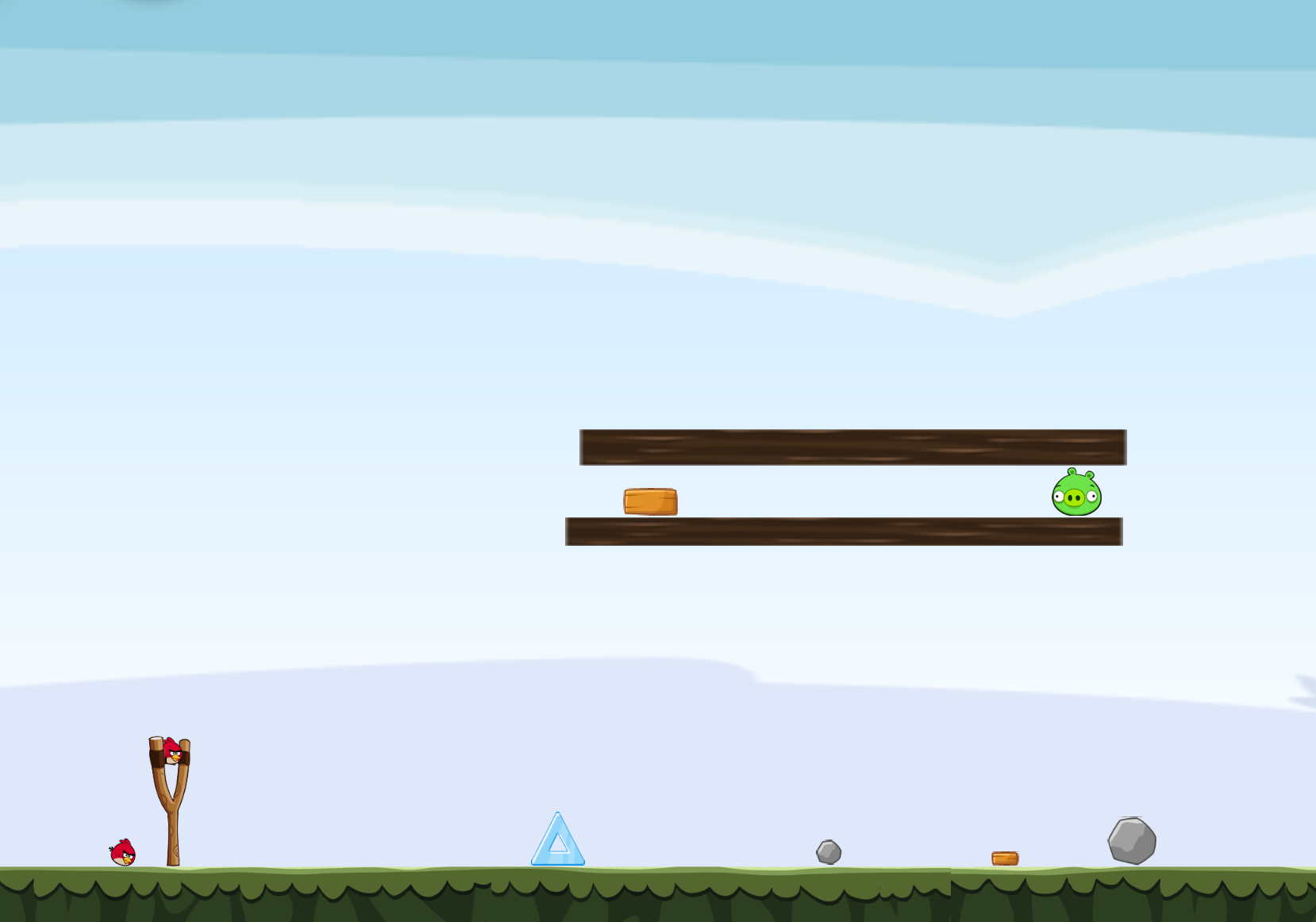}
      \end{subfigure}
  \caption{Events}
  \end{subfigure}
\caption{Task templates of the sliding scenario with eight novelties applied to them. In each subfigure, the left figure is the normal task and the right figure is the corresponding novel task with the novelty applied.}
\label{appendix_fig:sliding}
\end{figure}

\begin{figure}[h!]
  \captionsetup[subfigure]{labelformat=empty}
  \centering
  \begin{subfigure}[b]{0.24\columnwidth}
    \includegraphics[width=\linewidth]{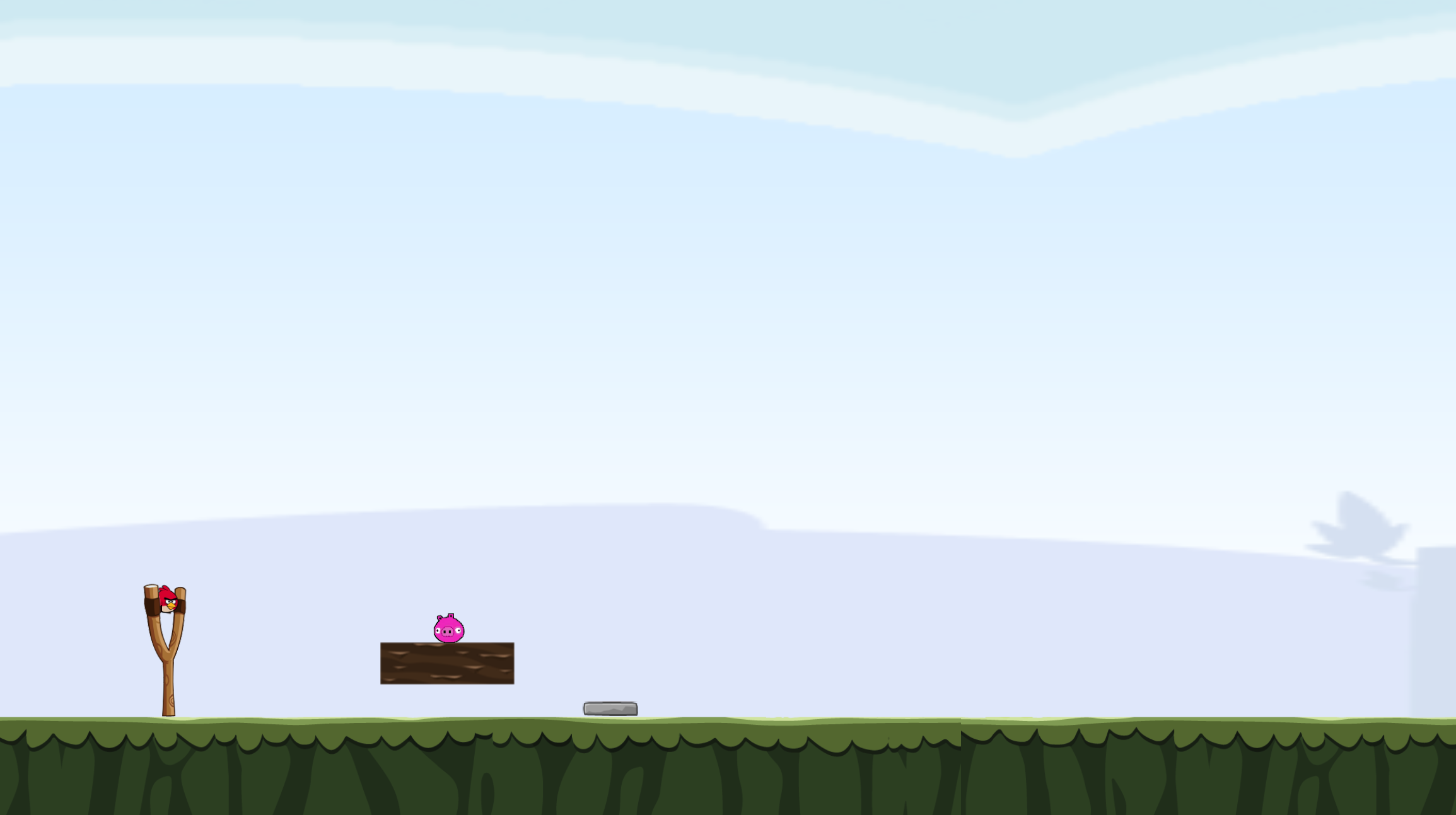}
    %\caption{1.1.1.1}
  \end{subfigure}
\begin{subfigure}[b]{0.24\columnwidth}
    \includegraphics[width=\linewidth]{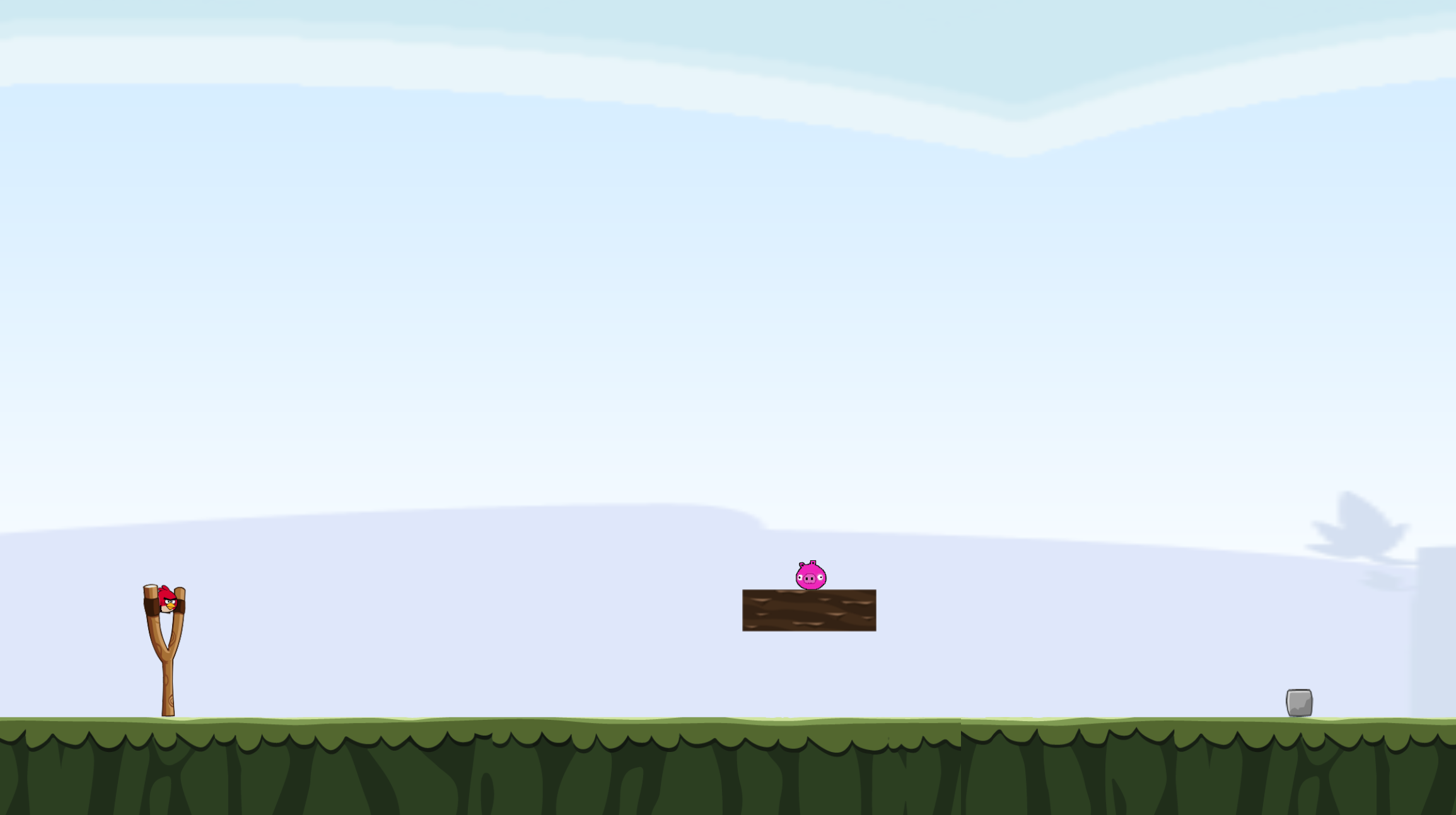}
    %\caption{2.1.1}
  \end{subfigure}
  \begin{subfigure}[b]{0.24\columnwidth}
    \includegraphics[width=\linewidth]{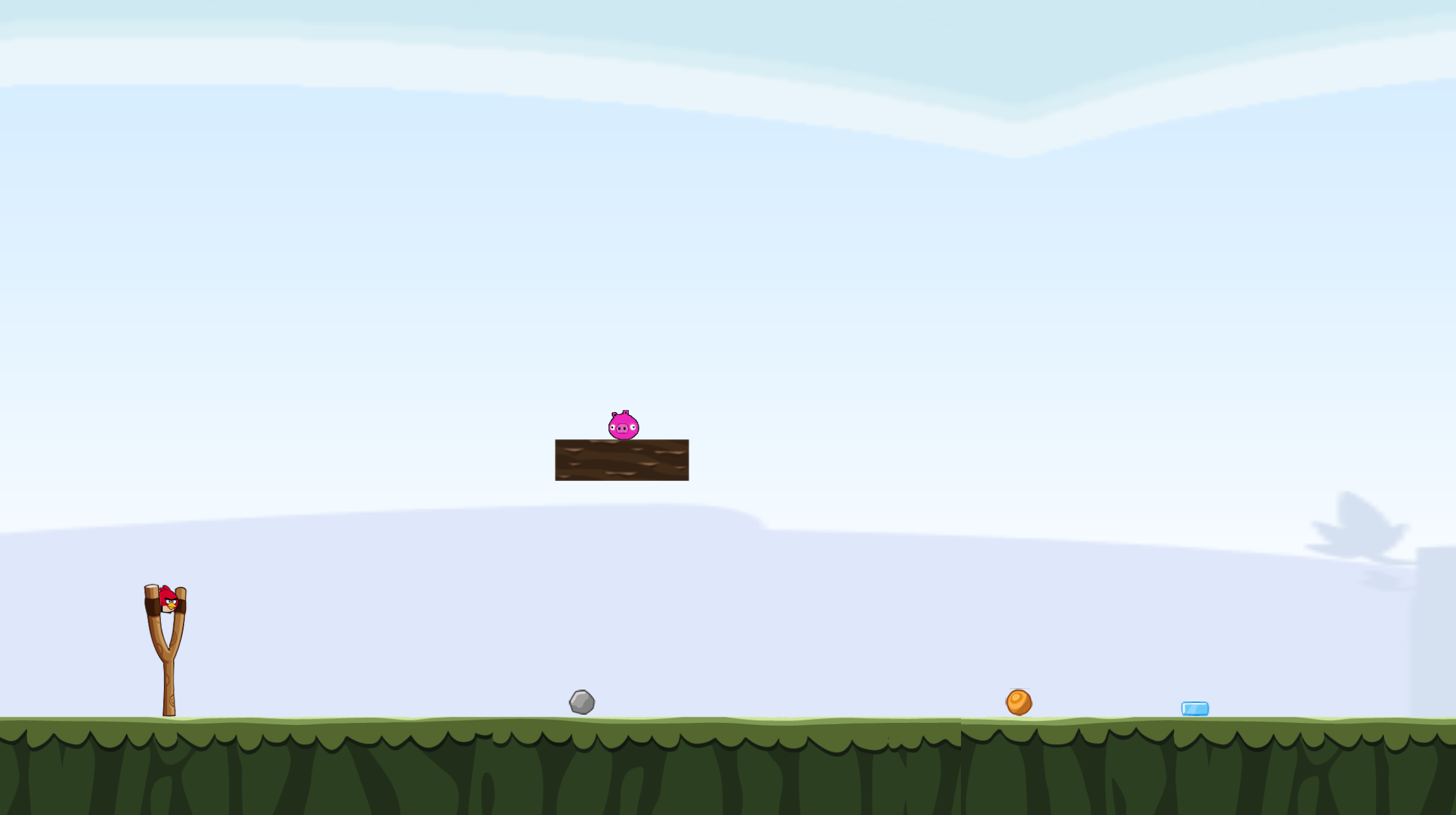}
    %\caption{2.1.4}
  \end{subfigure}
  \begin{subfigure}[b]{0.24\columnwidth}
    \includegraphics[width=\linewidth]{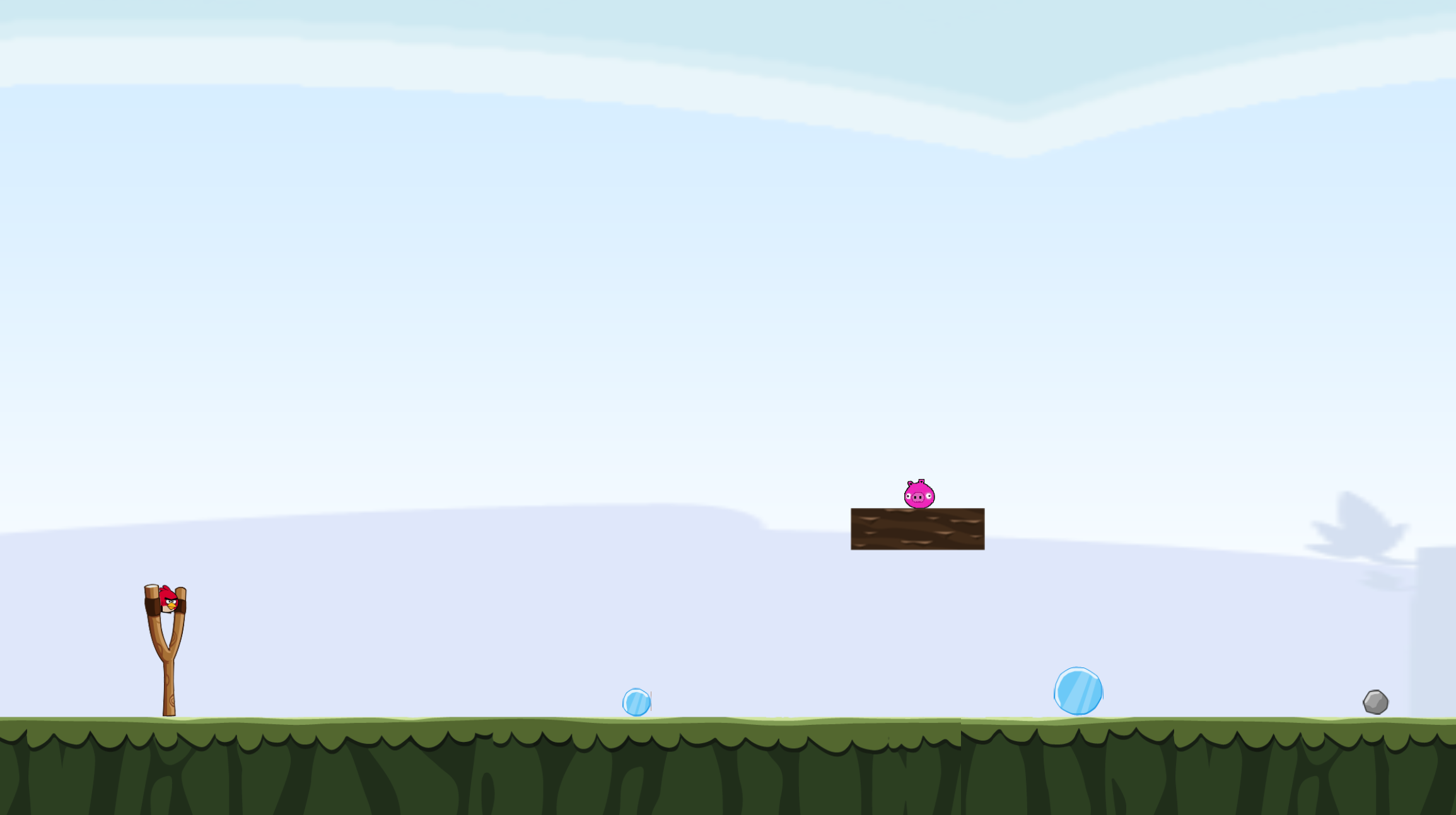}
    %\caption{2.1.5}
  \end{subfigure}
  \begin{subfigure}[b]{0.24\columnwidth}
    \includegraphics[width=\linewidth]{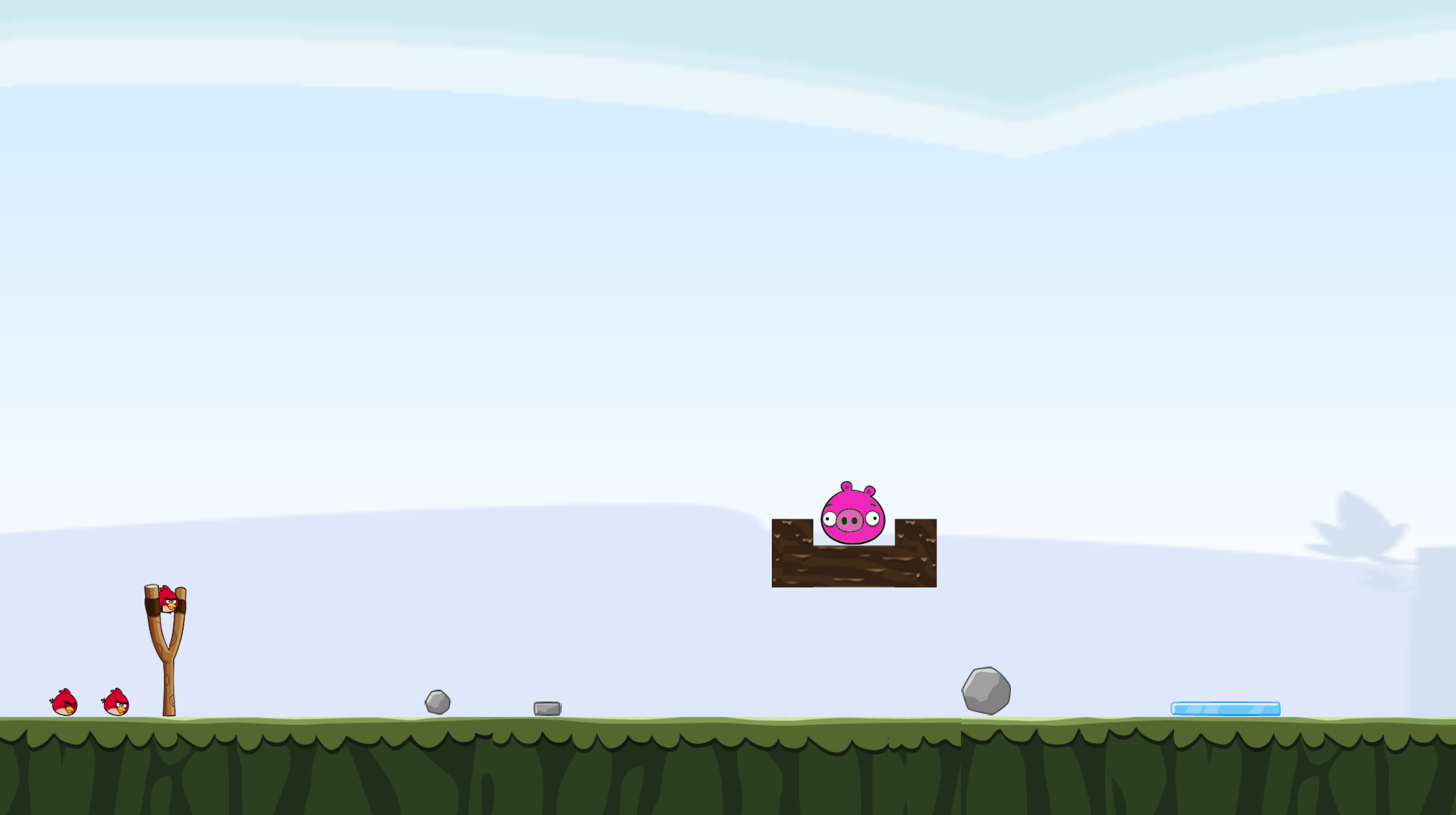}
    %\caption{2.2.1}
  \end{subfigure}
  \begin{subfigure}[b]{0.24\columnwidth}
    \includegraphics[width=\linewidth]{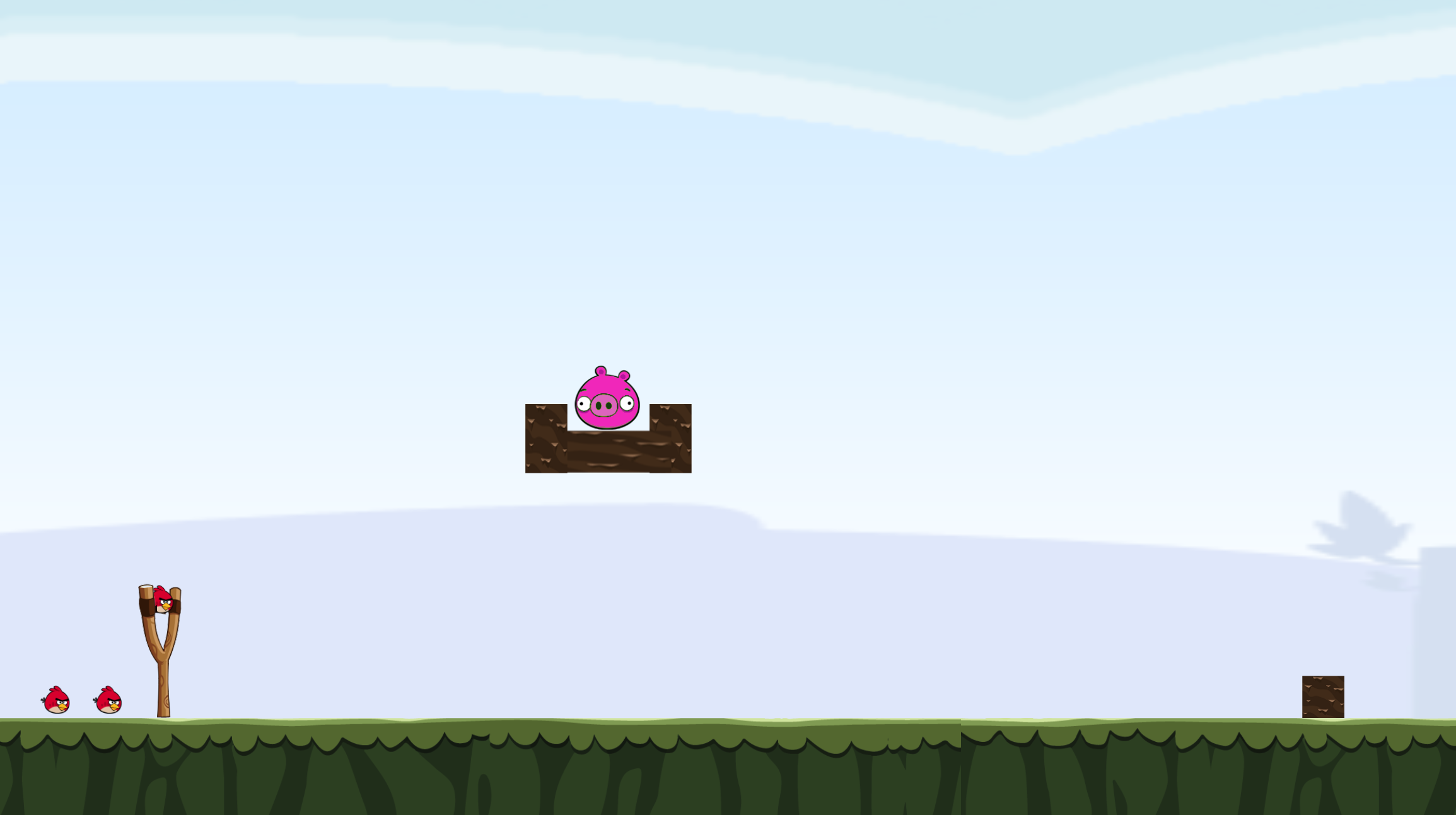}
    %\caption{2.2.2}
  \end{subfigure}
  \begin{subfigure}[b]{0.24\columnwidth}
    \includegraphics[width=\linewidth]{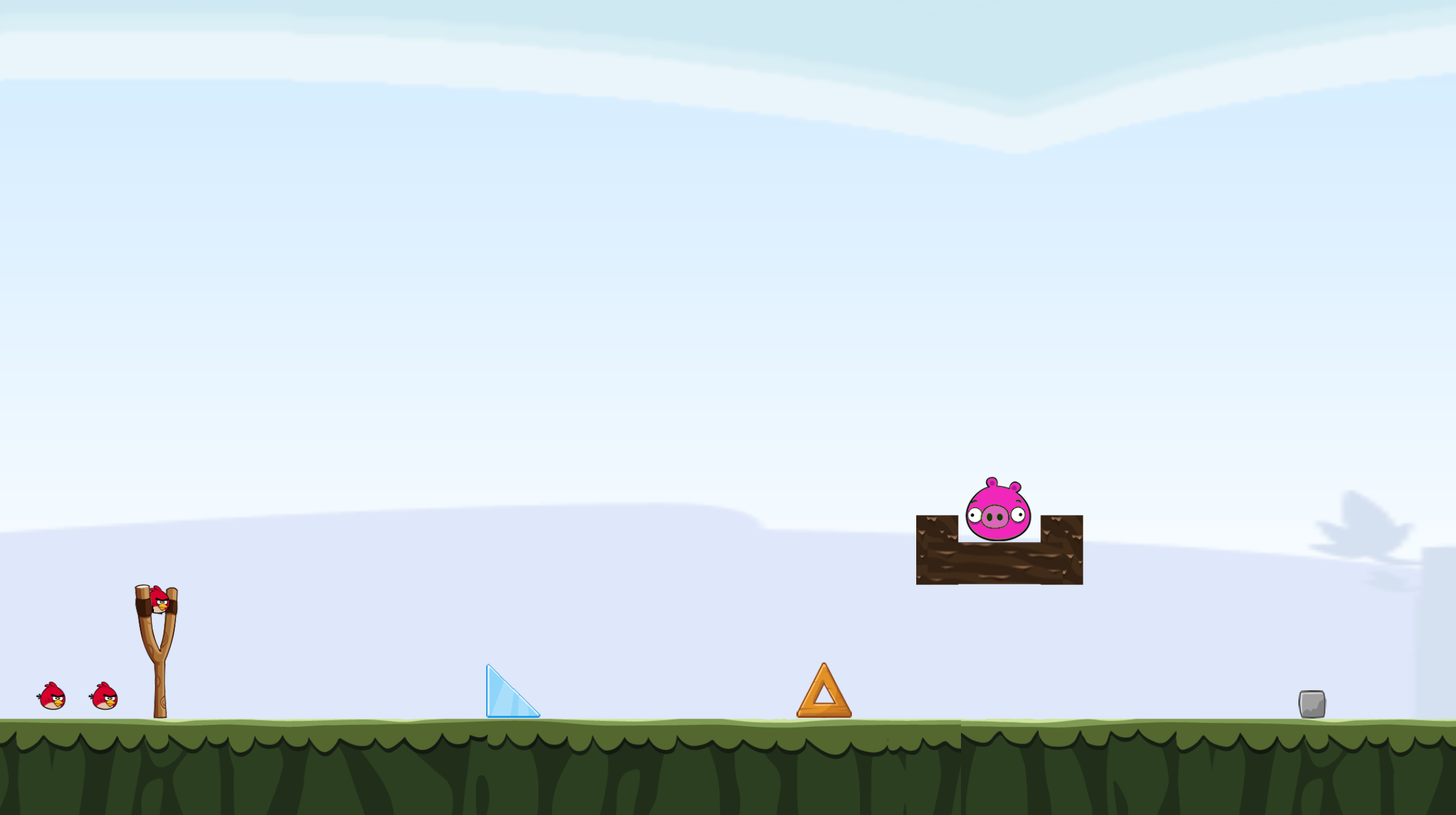}
    %\caption{2.2.3}
  \end{subfigure}
  \begin{subfigure}[b]{0.24\columnwidth}
    \includegraphics[width=\linewidth]{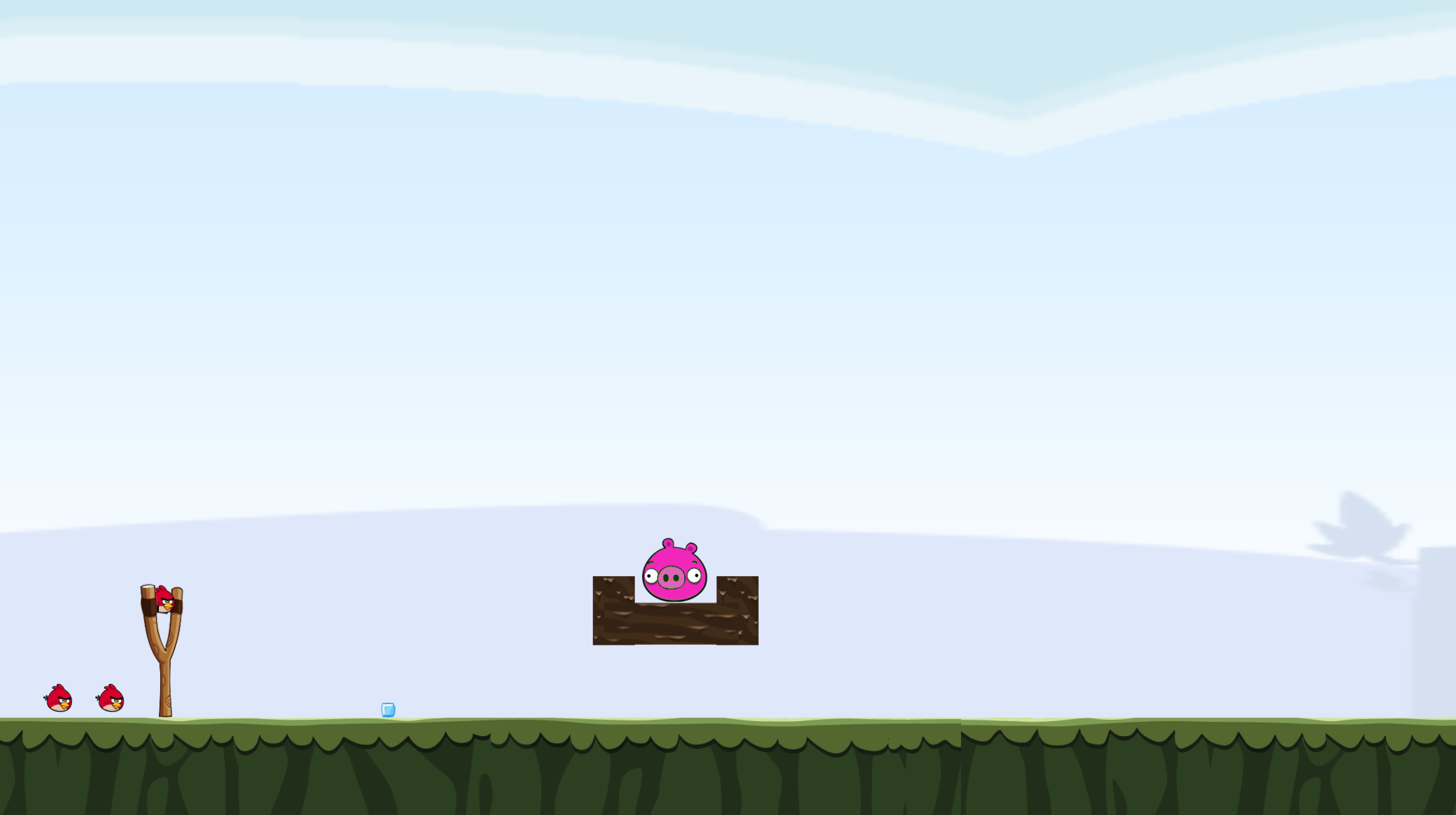}
    %\caption{2.2.4}
  \end{subfigure}
  \begin{subfigure}[b]{0.24\columnwidth}
    \includegraphics[width=\linewidth]{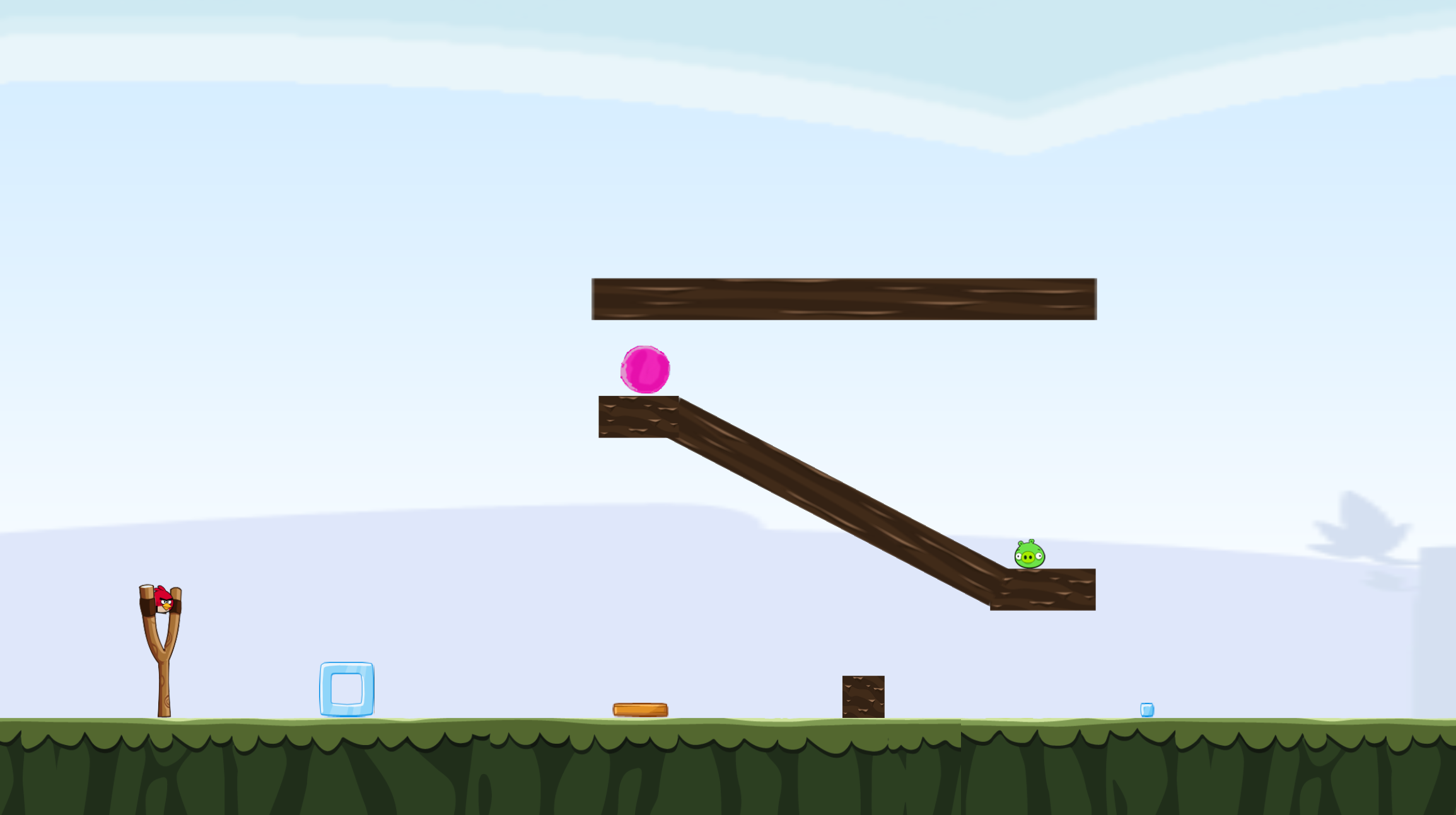}
    %\caption{2.2.5}
  \end{subfigure}
  \begin{subfigure}[b]{0.24\columnwidth}
    \includegraphics[width=\linewidth]{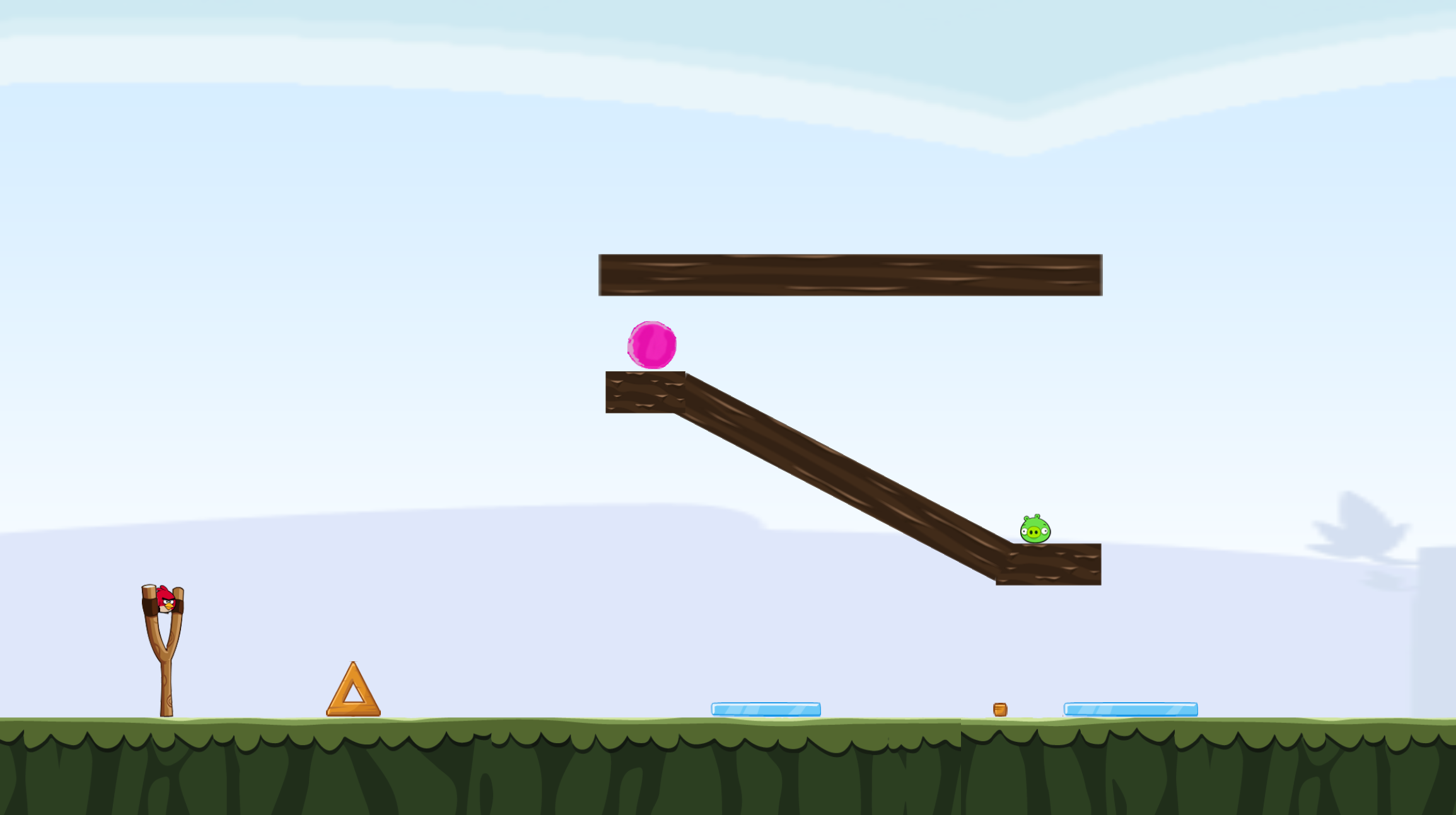}
    %\caption{2.3.2}
  \end{subfigure}
  \begin{subfigure}[b]{0.24\columnwidth}
    \includegraphics[width=\linewidth]{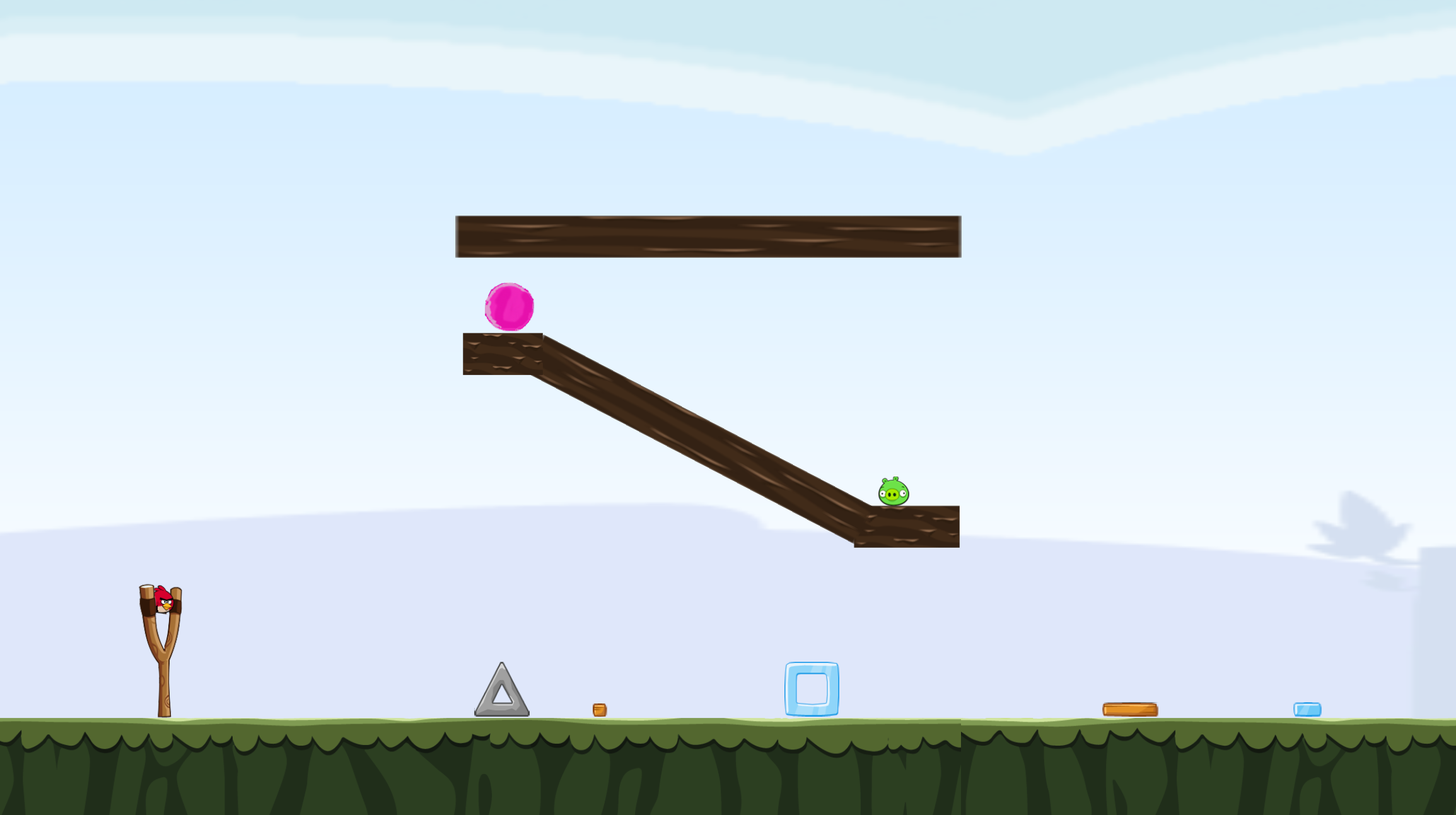}
    %\caption{2.3.4}
  \end{subfigure}
    \begin{subfigure}[b]{0.24\columnwidth}
    \includegraphics[width=\linewidth]{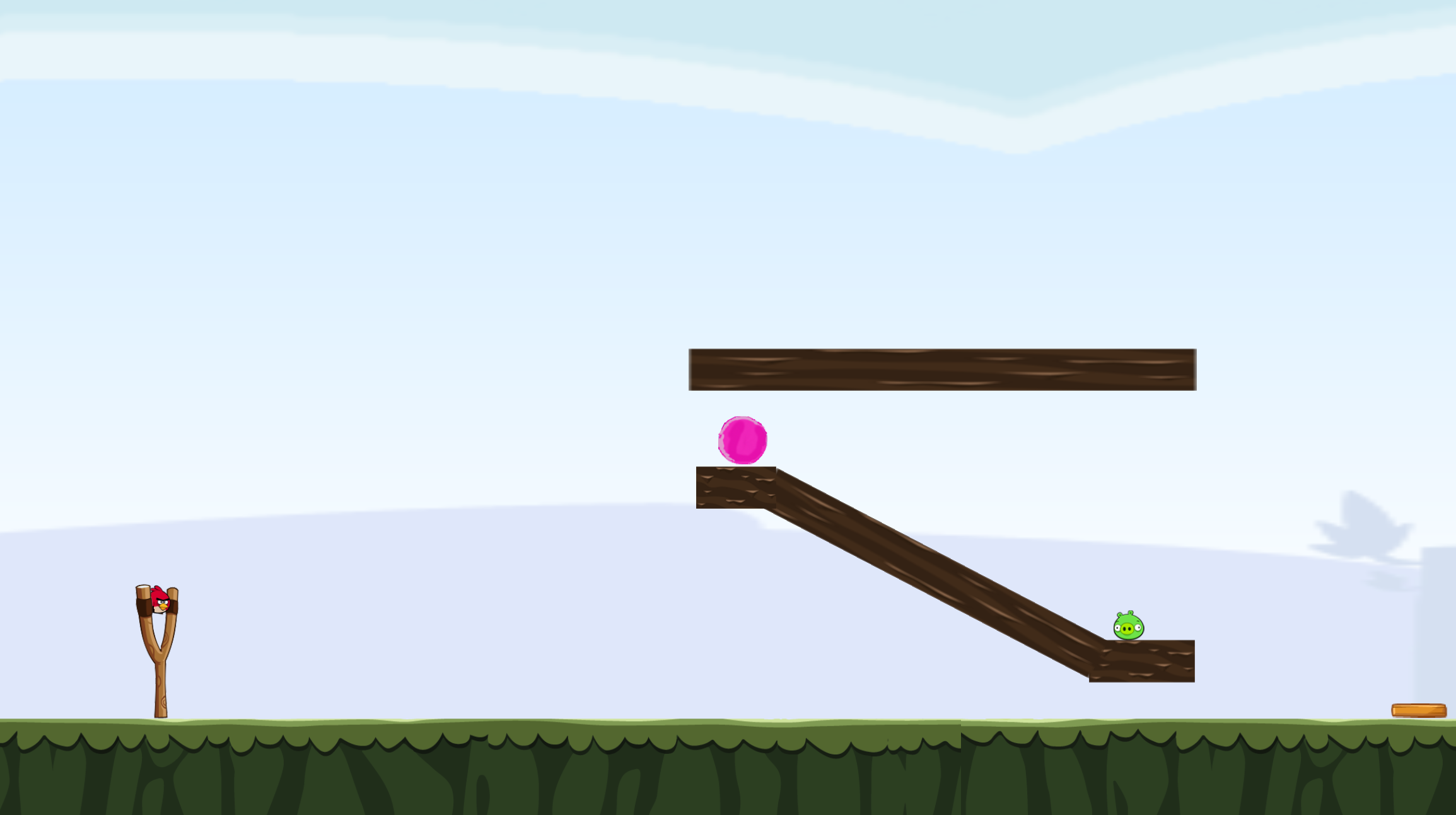}
    %\caption{2.3.1}
  \end{subfigure}
  \begin{subfigure}[b]{0.24\columnwidth}
    \includegraphics[width=\linewidth]{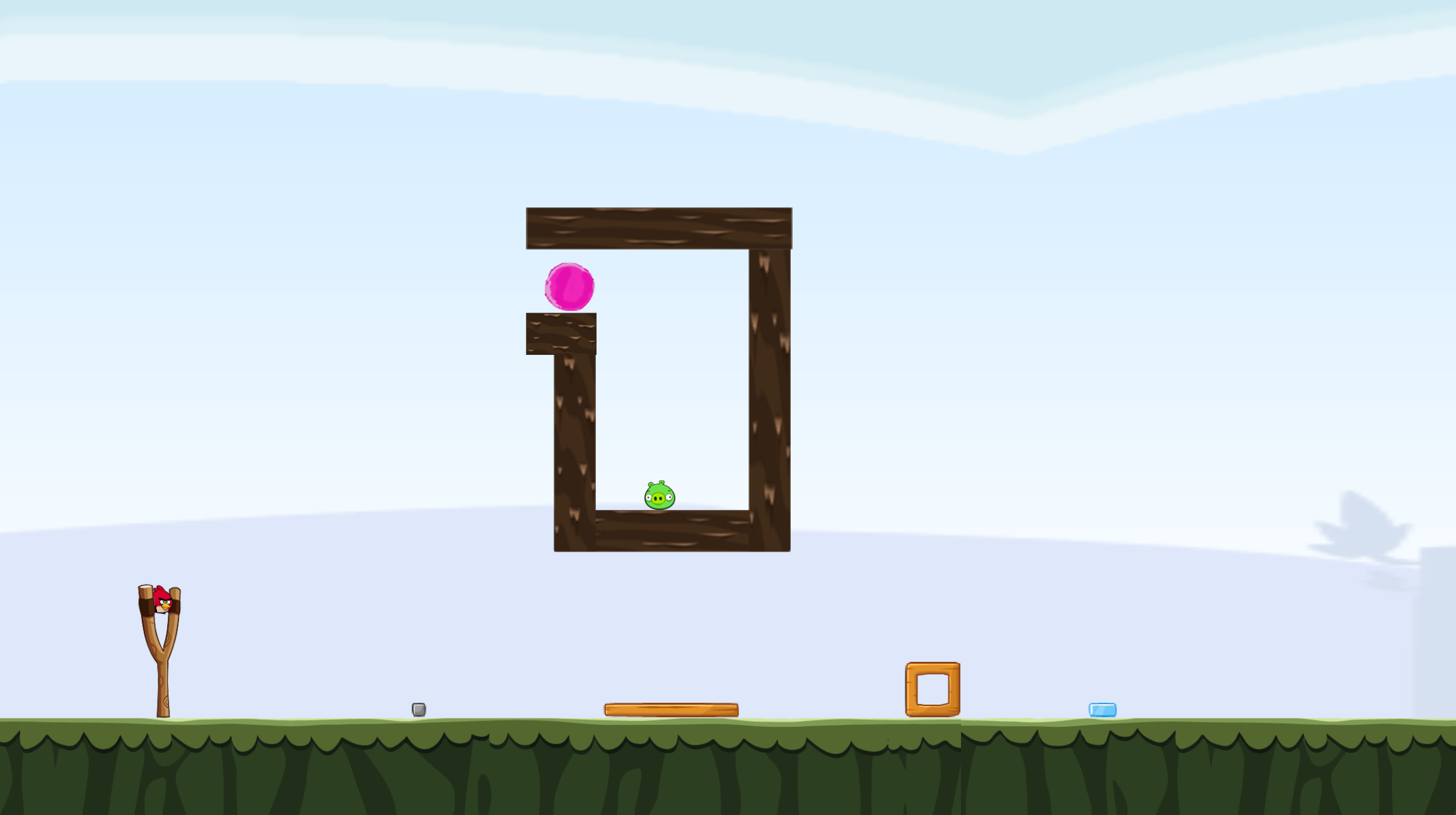}
    %\caption{2.4.1}
  \end{subfigure}
  \begin{subfigure}[b]{0.24\columnwidth}
    \includegraphics[width=\linewidth]{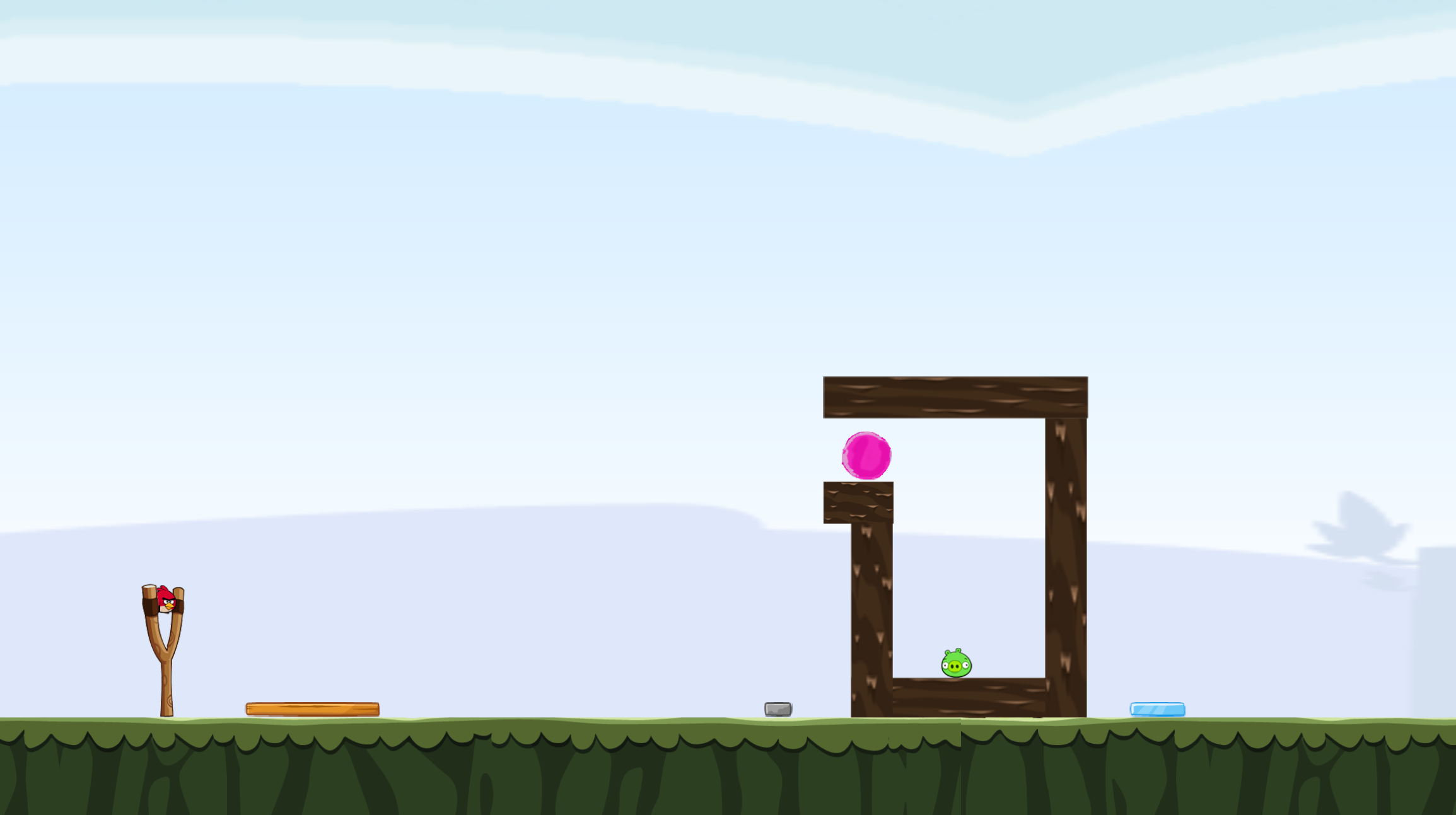}
    %\caption{2.4.2}
  \end{subfigure}
  \begin{subfigure}[b]{0.24\columnwidth}
    \includegraphics[width=\linewidth]{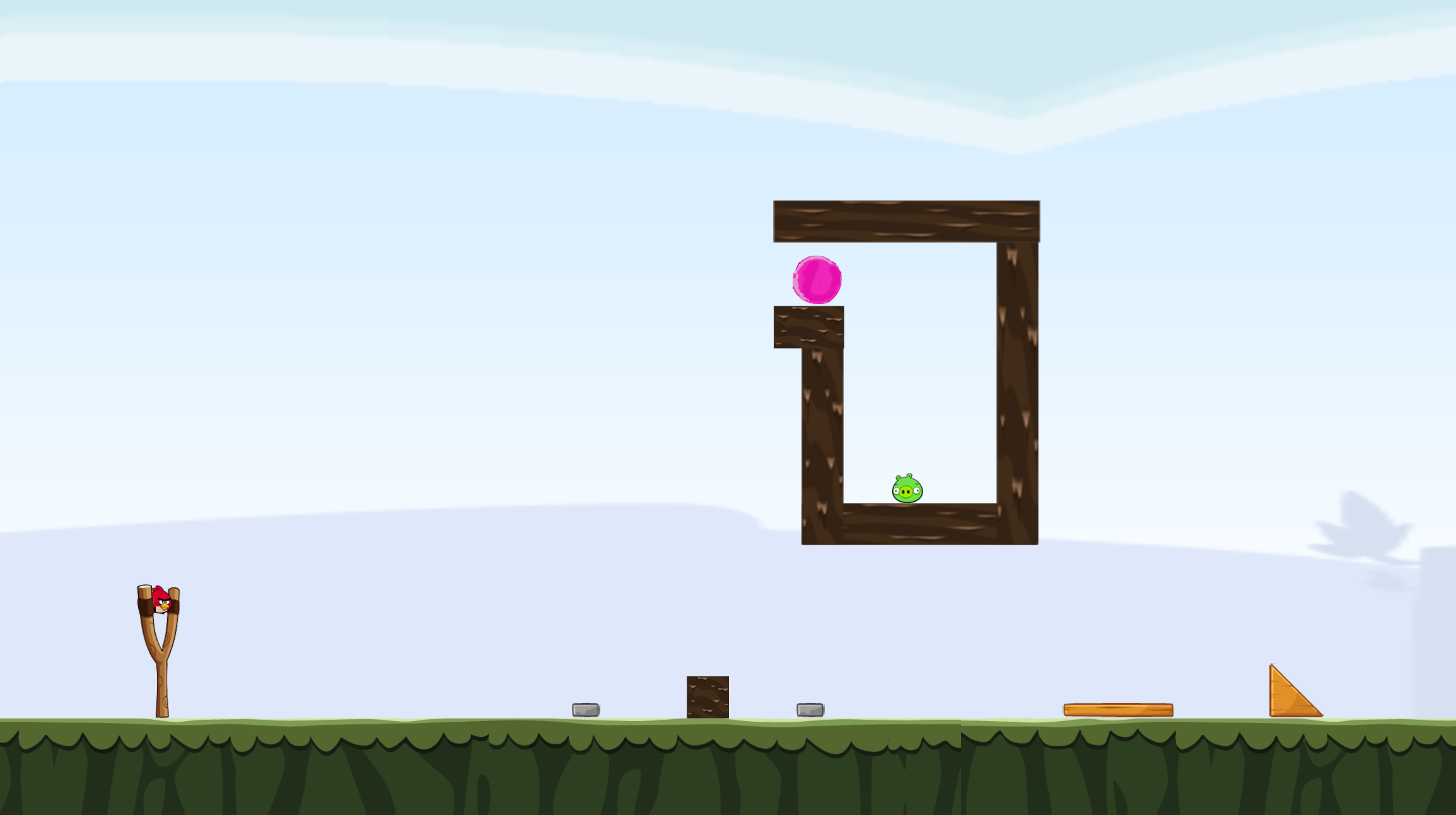}
    %\caption{2.4.3}
  \end{subfigure}
  \begin{subfigure}[b]{0.24\columnwidth}
    \includegraphics[width=\linewidth]{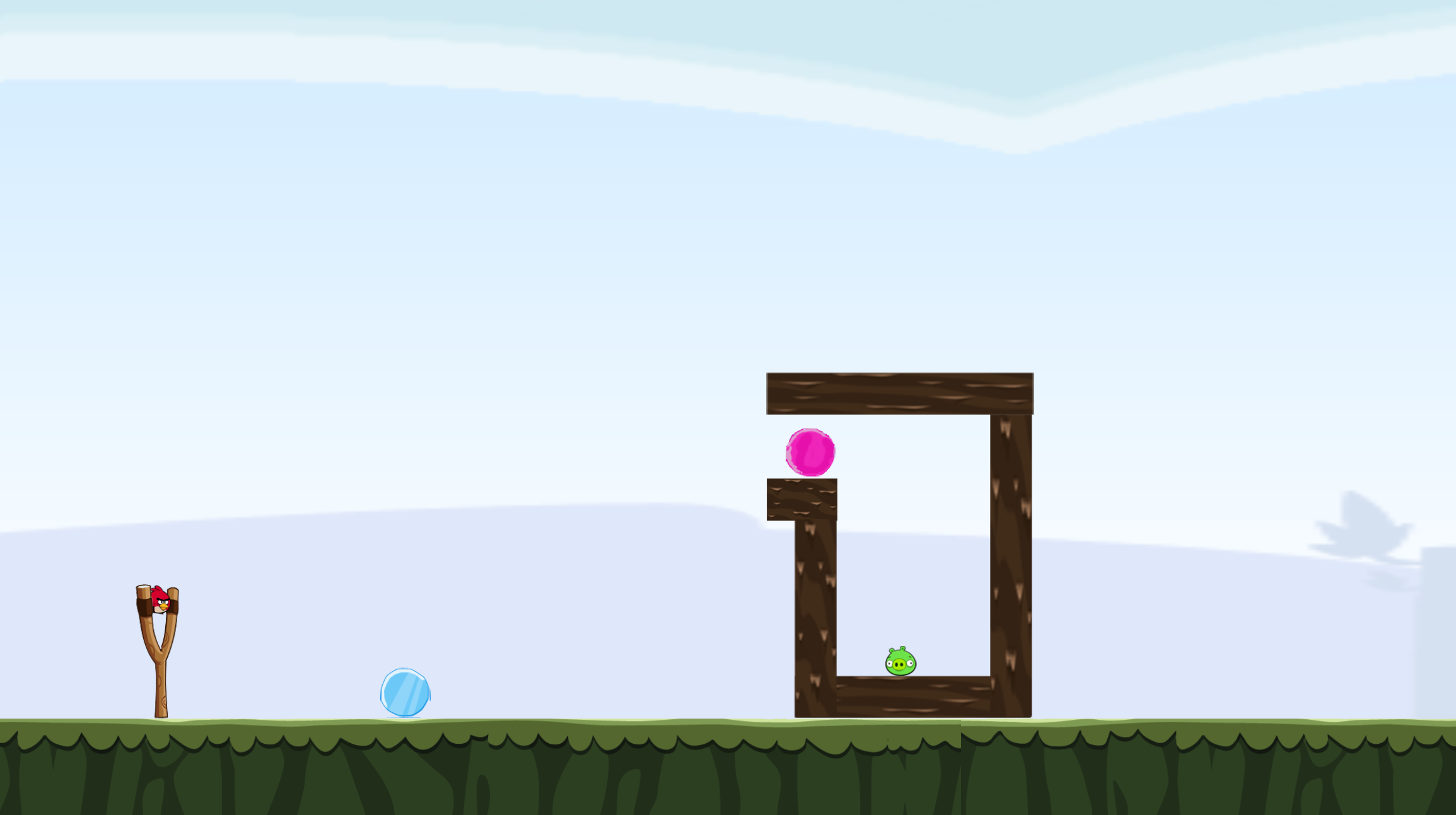}
    %\caption{2.3.4}
  \end{subfigure}
    \begin{subfigure}[b]{0.24\columnwidth}
    \includegraphics[width=\linewidth]{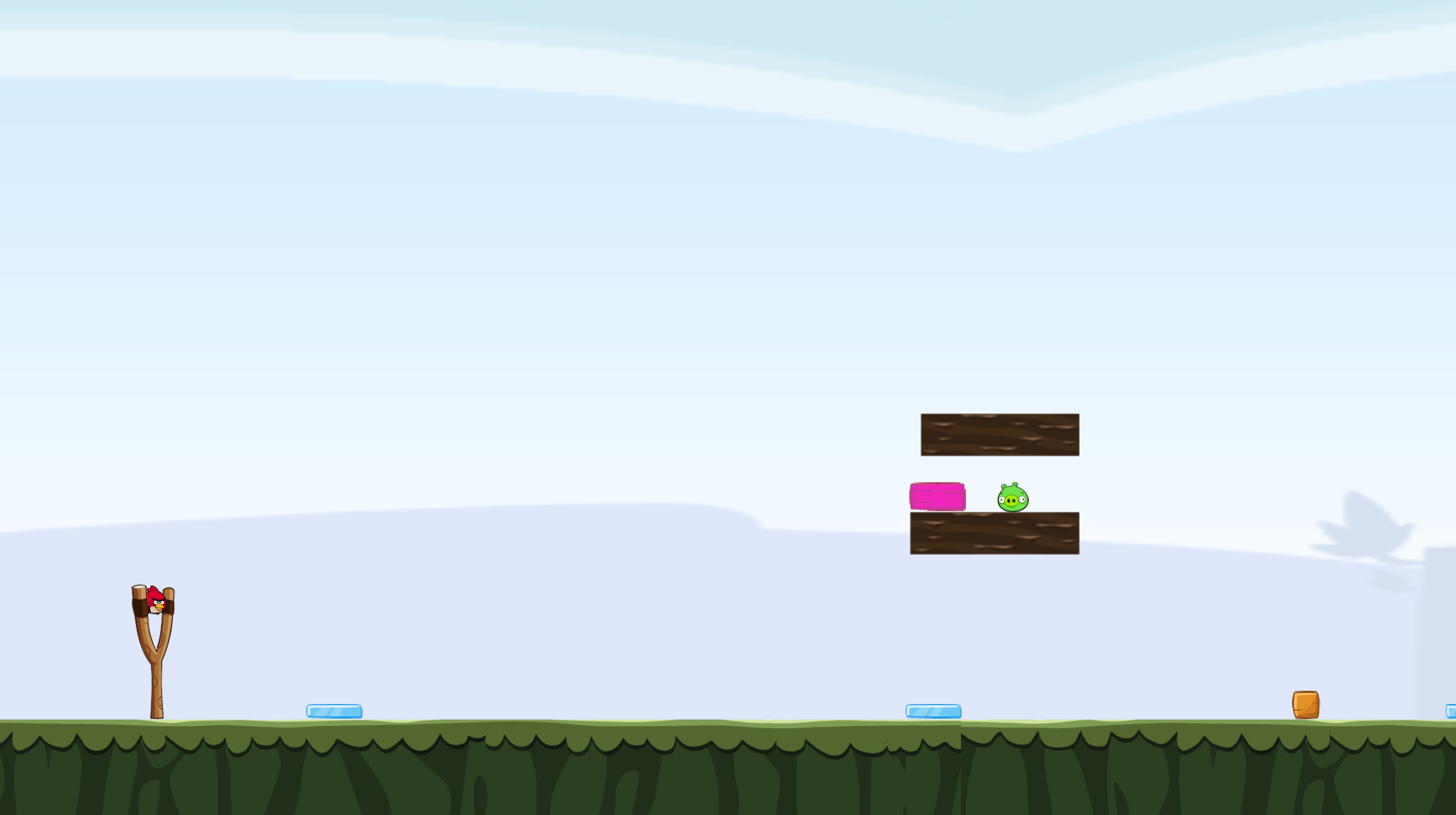}
    %\caption{2.3.4}
  \end{subfigure}
  \begin{subfigure}[b]{0.24\columnwidth}
    \includegraphics[width=\linewidth]{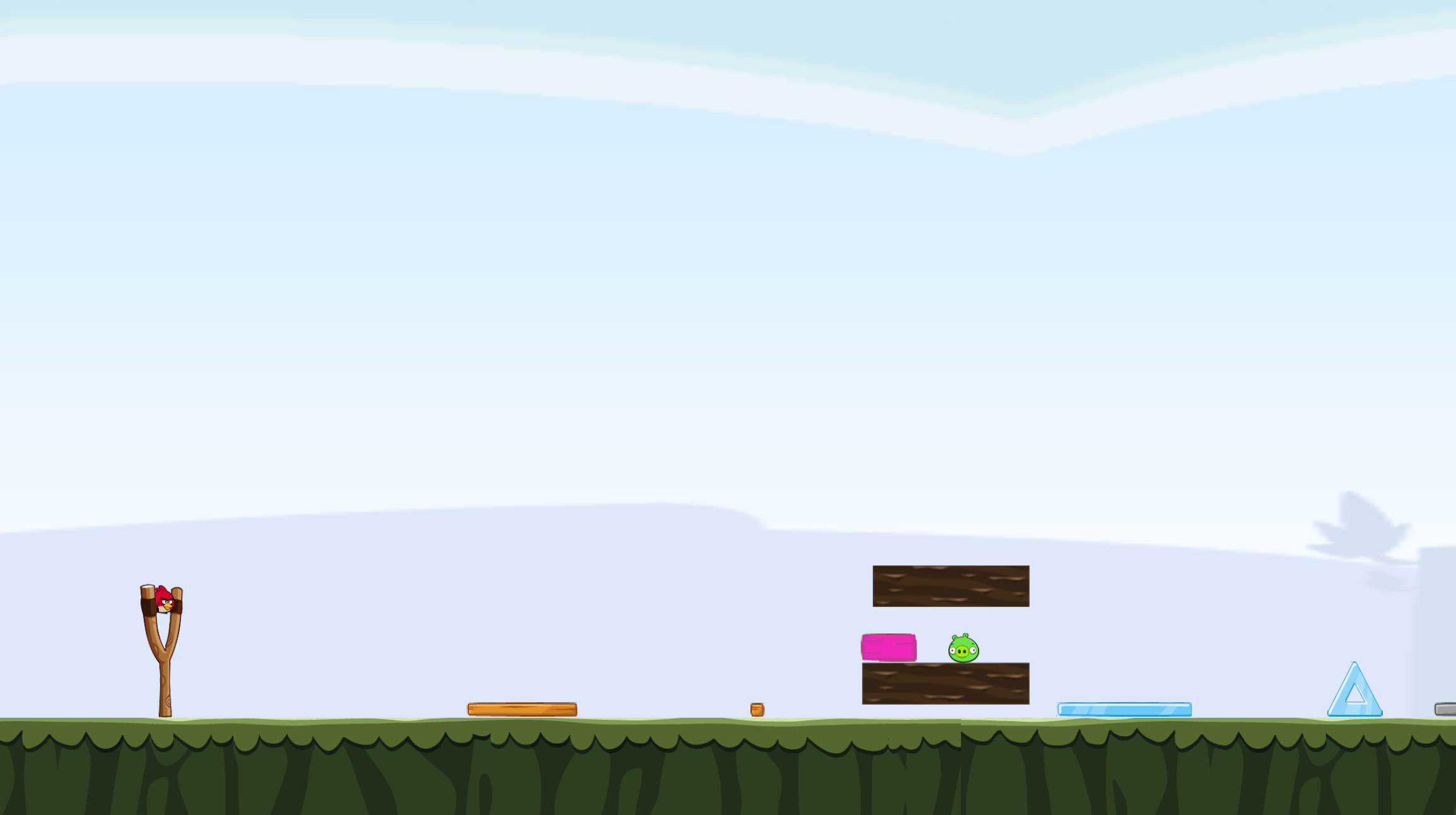}
    %\caption{2.4.1}
  \end{subfigure}
  \begin{subfigure}[b]{0.24\columnwidth}
    \includegraphics[width=\linewidth]{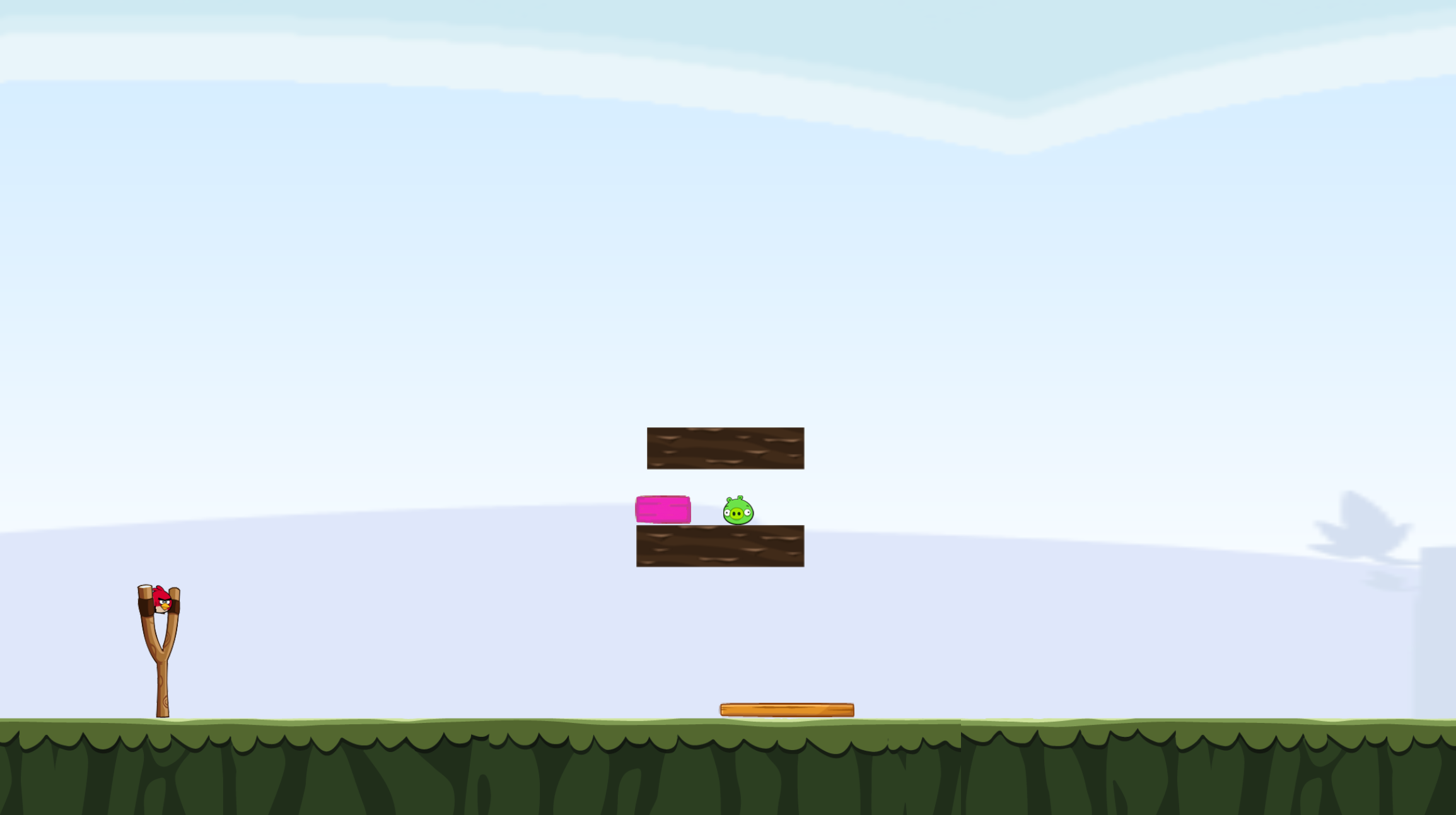}
    %\caption{2.4.2}
  \end{subfigure}
  \begin{subfigure}[b]{0.24\columnwidth}
    \includegraphics[width=\linewidth]{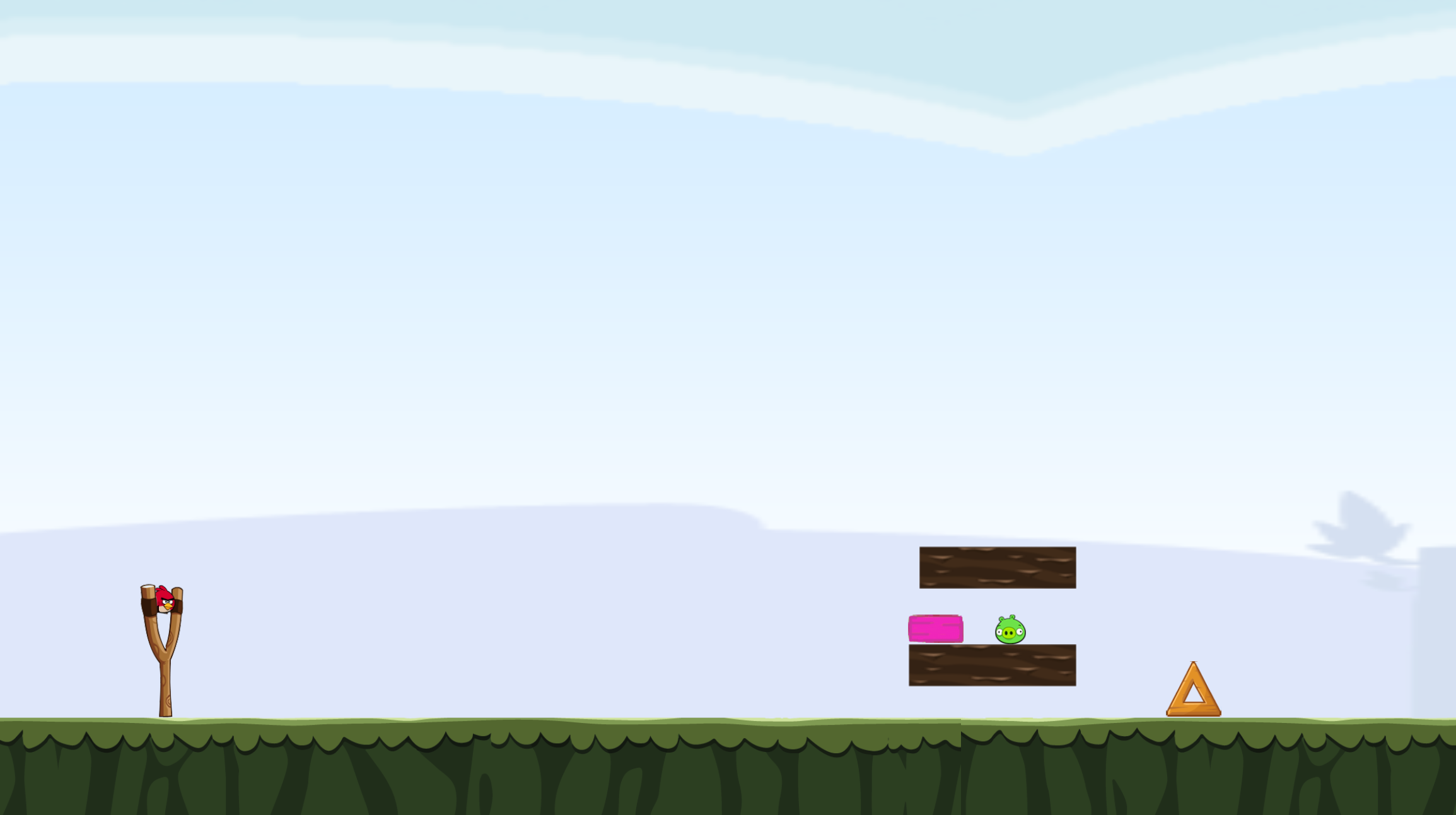}
    %\caption{2.4.3}
  \end{subfigure}
\caption{Each row shows four example tasks generated from the same task template. The positions of the objects and the distraction objects vary within the tasks of the same template, therefore each task has its own solution; however, all tasks of the same template can be solved by the same physical rule of the physical scenario that task template belongs to.}
\label{appendix_fig:task_variations_in_NovPhy}
\end{figure}

NovPhy contains 40 novel task templates and corresponding 40 normal task templates designed for five physical scenarios by applying eight novelties. The figures \ref{appendix_fig:single_force}, \ref{appendix_fig:multiple_forces}, \ref{appendix_fig:rolling}, \ref{appendix_fig:falling}, and \ref{appendix_fig:sliding} shows the normal and novel task templates of the five scenarios: single force, multiple forces, rolling, falling, and sliding, respectively. Figure \ref{appendix_fig:task_variations_in_NovPhy} shows tasks generated from five example task templates (of the objects novelty).

\newpage

\newpage

\newpage

\newpage

\newpage

\newpage

\section{Designing Novel Tasks in NovPhy} \label{Appendix: Section Task Designing Desiderata}
As discussed in Section 3.1 of the main paper, when designing novel tasks, we followed the desideratum - the agent has to work under the effects of the novelty to solve the task. To satisfy this desideratum, when designing novel tasks we ensure that the novelty is introduced to the task such that the novelty affects at least one of the physical interaction phases (initial, middle, or final) in the solution of the task. Table \ref{Tab:Interaction Affected by Novelty} shows the physical interaction phases that were affected when designing the 40 novelty-scenarios in NovPhy.

\begin{table}[h!]
\centering
%     \begin{adjustbox}{width=\textwidth,center}
    % \begin{adjustbox}{center}
    \resizebox{!}{170pt}{
        \begin{tabular}{lll}
            \hline
            Physical Scenario & Novelty & Physical Interaction Phases Affected \\ 
            \hline
            \multirow{8}{*}{Single force} & \multicolumn{1}{l}{1} & \multicolumn{1}{l}{final} \\\cline{2-3}
                 & \multicolumn{1}{l}{2} & \multicolumn{1}{l}{middle} \\\cline{2-3}
                 & \multicolumn{1}{l}{3} & \multicolumn{1}{l}{middle} \\\cline{2-3}
                 & \multicolumn{1}{l}{4} & \multicolumn{1}{l}{middle} \\\cline{2-3}
                 & \multicolumn{1}{l}{5} & \multicolumn{1}{l}{initial, middle} \\\cline{2-3}
                 & \multicolumn{1}{l}{6} & \multicolumn{1}{l}{initial, middle} \\\cline{2-3}
                 & \multicolumn{1}{l}{7} & \multicolumn{1}{l}{middle} \\\cline{2-3}
                 & \multicolumn{1}{l}{8} & \multicolumn{1}{l}{middle} \\\hline
            \multirow{8}{*}{Multiple forces} & \multicolumn{1}{l}{1} & \multicolumn{1}{l}{final} \\\cline{2-3}
                 & \multicolumn{1}{l}{2} & \multicolumn{1}{l}{middle} \\\cline{2-3}
                 & \multicolumn{1}{l}{3} & \multicolumn{1}{l}{middle, final} \\\cline{2-3}
                 & \multicolumn{1}{l}{4} & \multicolumn{1}{l}{middle, final} \\\cline{2-3}
                 & \multicolumn{1}{l}{5} & \multicolumn{1}{l}{initial, middle} \\\cline{2-3}
                 & \multicolumn{1}{l}{6} & \multicolumn{1}{l}{initial, middle} \\\cline{2-3}
                 & \multicolumn{1}{l}{7} & \multicolumn{1}{l}{middle, final} \\\cline{2-3}
                 & \multicolumn{1}{l}{8} & \multicolumn{1}{l}{middle} \\\hline
            \multirow{8}{*}{Rolling} & \multicolumn{1}{l}{1} & \multicolumn{1}{l}{initial} \\\cline{2-3}
                 & \multicolumn{1}{l}{2} & \multicolumn{1}{l}{middle} \\\cline{2-3}
                 & \multicolumn{1}{l}{3} & \multicolumn{1}{l}{middle} \\\cline{2-3}
                 & \multicolumn{1}{l}{4} & \multicolumn{1}{l}{middle} \\\cline{2-3}
                 & \multicolumn{1}{l}{5} & \multicolumn{1}{l}{initial, middle} \\\cline{2-3}
                 & \multicolumn{1}{l}{6} & \multicolumn{1}{l}{initial, middle} \\\cline{2-3}
                 & \multicolumn{1}{l}{7} & \multicolumn{1}{l}{middle} \\\cline{2-3}
                 & \multicolumn{1}{l}{8} & \multicolumn{1}{l}{middle} \\\hline
            \multirow{8}{*}{Falling} & \multicolumn{1}{l}{1} & \multicolumn{1}{l}{initial} \\\cline{2-3}
                 & \multicolumn{1}{l}{2} & \multicolumn{1}{l}{middle} \\\cline{2-3}
                 & \multicolumn{1}{l}{3} & \multicolumn{1}{l}{middle} \\\cline{2-3}
                 & \multicolumn{1}{l}{4} & \multicolumn{1}{l}{middle} \\\cline{2-3}
                 & \multicolumn{1}{l}{5} & \multicolumn{1}{l}{initial, middle} \\\cline{2-3}
                 & \multicolumn{1}{l}{6} & \multicolumn{1}{l}{initial, middle} \\\cline{2-3}
                 & \multicolumn{1}{l}{7} & \multicolumn{1}{l}{middle} \\\cline{2-3}
                 & \multicolumn{1}{l}{8} & \multicolumn{1}{l}{middle} \\\hline
            \multirow{8}{*}{Sliding} & \multicolumn{1}{l}{1} & \multicolumn{1}{l}{initial} \\\cline{2-3}
                 & \multicolumn{1}{l}{2} & \multicolumn{1}{l}{middle} \\\cline{2-3}
                 & \multicolumn{1}{l}{3} & \multicolumn{1}{l}{middle} \\\cline{2-3}
                 & \multicolumn{1}{l}{4} & \multicolumn{1}{l}{middle} \\\cline{2-3}
                 & \multicolumn{1}{l}{5} & \multicolumn{1}{l}{initial, middle} \\\cline{2-3}
                 & \multicolumn{1}{l}{6} & \multicolumn{1}{l}{initial, middle} \\\cline{2-3}
                 & \multicolumn{1}{l}{7} & \multicolumn{1}{l}{middle} \\\cline{2-3}
                 & \multicolumn{1}{l}{8} & \multicolumn{1}{l}{middle} \\
                
            \hline
        \end{tabular}
        }
%     \end{adjustbox}
%     \vspace{ - 05 mm}
    \caption{table}{The physical interaction phases of the solution that were affected by the novelty. Novelties 1 to 8 represents: 1. objects, 2. agents, 3. actions, 4. interactions, 5. relations, 6. environments, 7. goals, and 8. events.} 
    \label{Tab:Interaction Affected by Novelty}
\end{table}

\newpage

\section{Novelty Detection Performance} \label{Appendix: Novelty Detection Performance}
The detection performance is only measured in the agents that performed under the uninformed condition: Datalab, Eagle's Wing, Pig Shooter, and Random. The agent's detection module is based on the pass rate deviation. We have incorporated two algorithms, 1) simple moving average pass rate (sma) and 2) pre-assumed moving average pass rate (pma) that can be used as detection modules. 

The simple moving average method compares two consecutive moving average pass rates of a given window size and checks if the second pass rate deviates from the first pass rate more than the selected threshold. We have studied the variations of window sizes 5 and 10, and thresholds 0.2, 0.4, 0.6, and 0.8.  

In pre-assumed moving average method, we assume that a trained agent knows the pass rate of the normal tasks of a given task template as those tasks are provided to the agent for training beforehand. Therefore, in this method, we check if the pass rate within a given window size is deviated by a number of standard deviations (the detection thresholds). We have studied variations of window sizes 5 and 10, and thresholds 1.5 and 2.  

\begin{figure}[h!]
    \centering
    \includegraphics[width=\textwidth,height=\textheight,keepaspectratio]{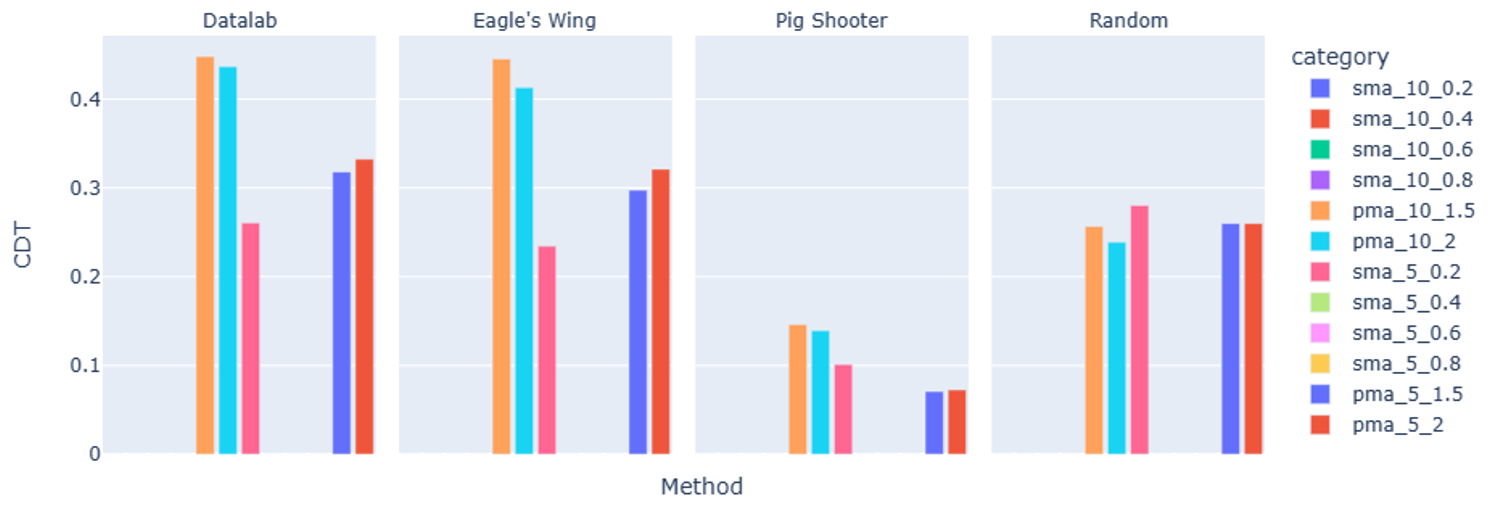}
    \caption{The CDT of the agents under the two algorithms sma and pma under variation of window size (the first value following the algorithm name) and detection thresholds (the last value).}
    \label{fig:best_detection_algorithm_determination}
\end{figure}

Figure \ref{fig:best_detection_algorithm_determination} shows the CDT results of the agents based on the variations of the window size and detection threshold. For Datalab, Eagle's Wing, and Pig Shooter, the pma method with window size 10 and 1.5 standard deviation detection threshold produced the highest CDT while for Random agent, the sma method with window size 5 and detection threshold 0.2 provided the highest CDT. The method that produces the highest CDT is used as the novelty detection method of the respective agent and the results achieved using that method are presented in the main paper and used in the rest of the appendix. 

\begin{figure} [h!]
    \centering
    \includegraphics[width=0.9\textwidth,height=\textheight,keepaspectratio]{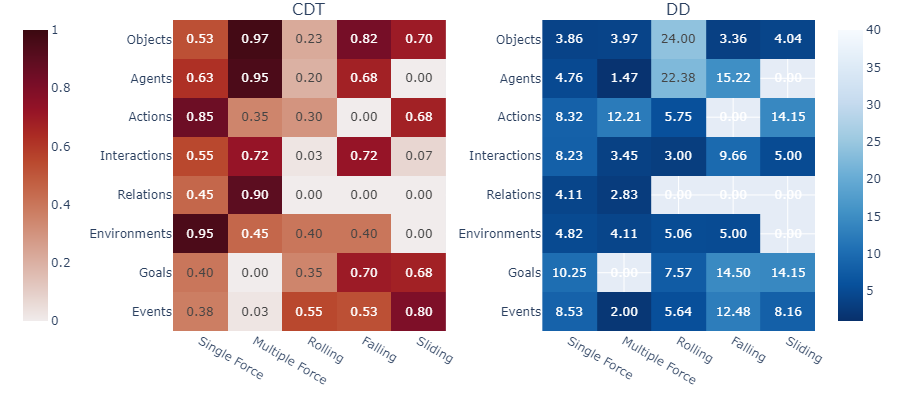}
    \caption{The CDT and DD results of the Datalab.}
    \label{fig:detection_datalab}
\end{figure}

\begin{figure}[h!]
    \centering
    \includegraphics[width=0.9\textwidth,height=\textheight,keepaspectratio]{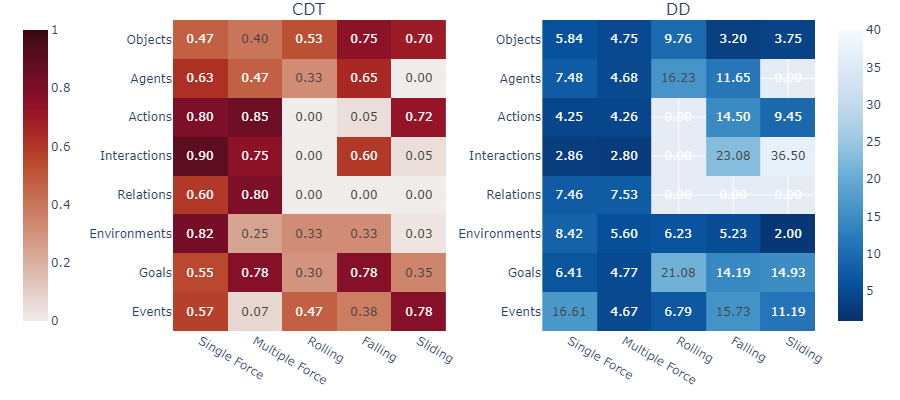}
    \caption{The CDT and DD results of the Eagle's Wing.}
    \label{fig:detection_eagleswing}
\end{figure}

\begin{figure}[h!]
    \centering
    \includegraphics[width=0.9\textwidth,height=\textheight,keepaspectratio]{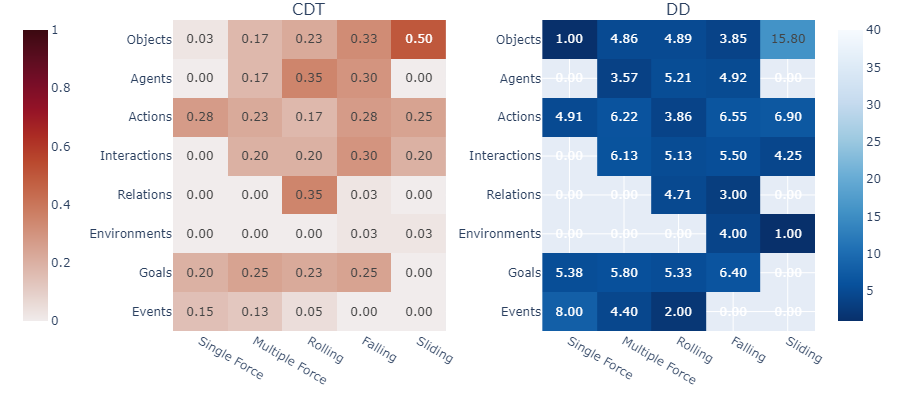}
    \caption{The CDT and DD results of the Pig Shooter.}
    \label{fig:detection_pigshooter}
\end{figure}

\begin{figure}[h!]
    \centering
    \includegraphics[width=0.9\textwidth,height=\textheight,keepaspectratio]{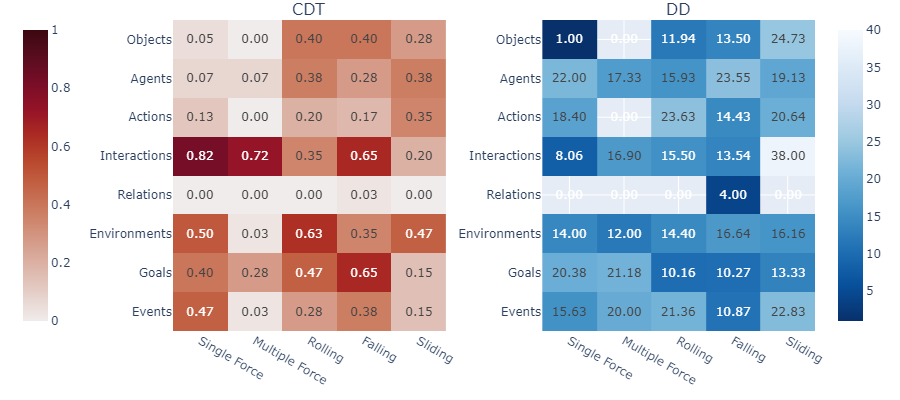}
    \caption{The CDT and DD results of the Random agent.}
    \label{fig:detection_random}
\end{figure}

\begin{figure}[h!]
    \centering    \includegraphics[width=0.9\textwidth,height=\textheight,keepaspectratio]{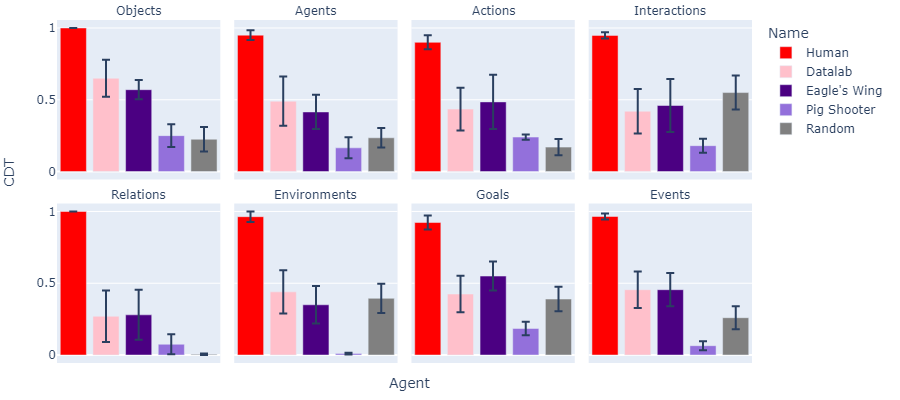}
    \caption{$NPS_{CDT}$ of the agents for the eight novelties.}
    \label{fig:cdt_per_novelty}
\end{figure}

\begin{figure}[h!]
    \centering    \includegraphics[width=\textwidth,height=\textheight,keepaspectratio]{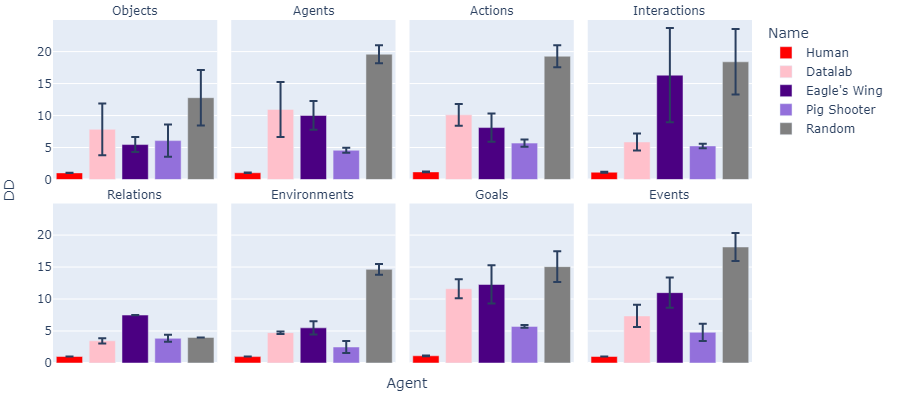}
    \caption{$NPS_{DD}$ of the agents for the eight novelties.}
    \label{fig:dd_per_novelty}
\end{figure}

\begin{figure}[h!]
    \centering    \includegraphics[width=\textwidth,height=\textheight,keepaspectratio]{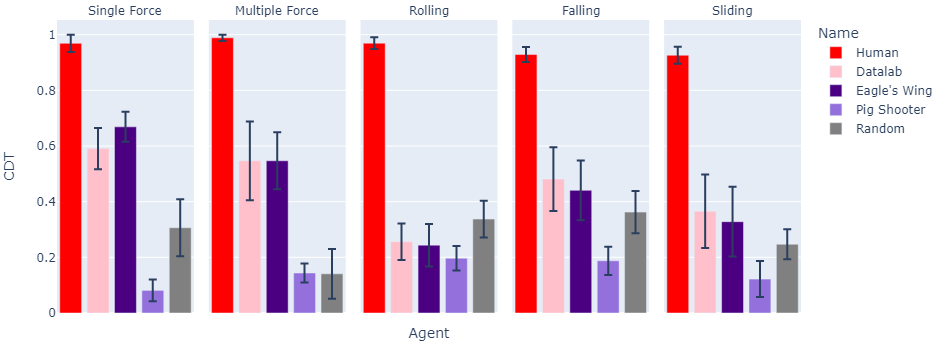}
    \caption{$SPN_{CDT}$ of the agents for the five scenarios.}
    \label{fig:cdt_per_scenario}
\end{figure}

\begin{figure}[h!]
    \centering    \includegraphics[width=\textwidth,height=\textheight,keepaspectratio]{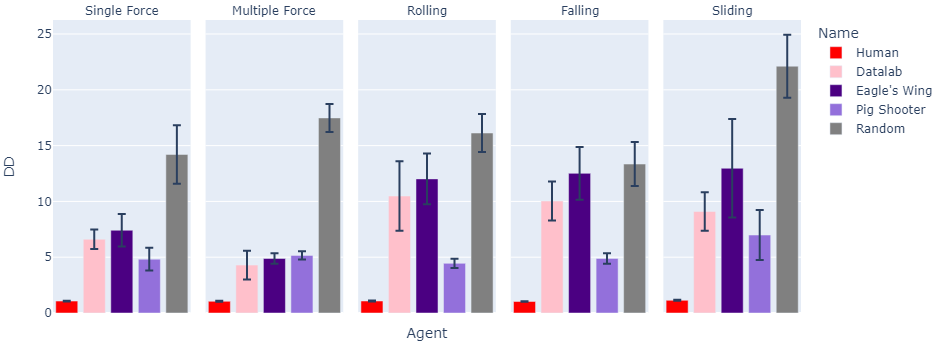}
    \caption{$SPN_{DD}$ of the agents for the five scenarios.}
    \label{fig:dd_per_scenario}
\end{figure}

Novelty detection performance is measured using the measures CDT and DD. Figures \ref{fig:detection_datalab}-\ref{fig:detection_random} present the heat maps of CDT (higher the better) and DD (lower the better) for each agent. Figures \ref{fig:cdt_per_novelty} and \ref{fig:dd_per_novelty} present the summary of detection performance per novelty ($NPS_{CDT}$) and Figures \ref{fig:cdt_per_scenario} and \ref{fig:dd_per_scenario} shows the summary per scenario ($SPN_{CDT}$). As it can be seen from the four summary figures and in comparison of agent heatmaps with humans, it can be seen that agents have a large room for improvement in novelty detection. 

\newpage

\section{Novelty Adaptation Performance}

\begin{figure}[h!]
    \centering
    \includegraphics[width=\textwidth,height=\textheight,keepaspectratio]{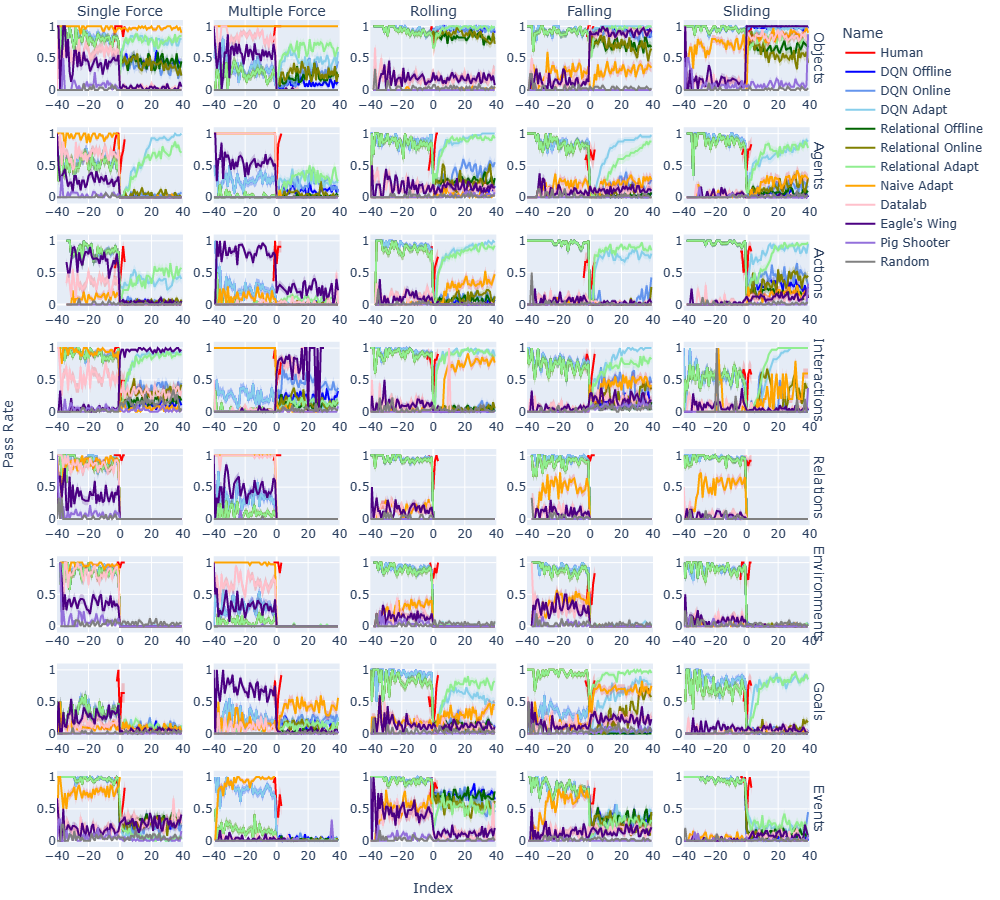}
    \caption{The pass rate of the agents in all novelty-scenarios. The x-axis represents the index of the task in trials. Indexes -40 to -1 are for normal tasks and 0 to 40 are for novel tasks. The y-axis represents the pass rate.}
    \label{fig:adaptation_plot_per_novelty-scenario}
\end{figure}

\begin{figure}
    \centering
    \includegraphics[width=0.9\textwidth,height=\textheight,keepaspectratio]{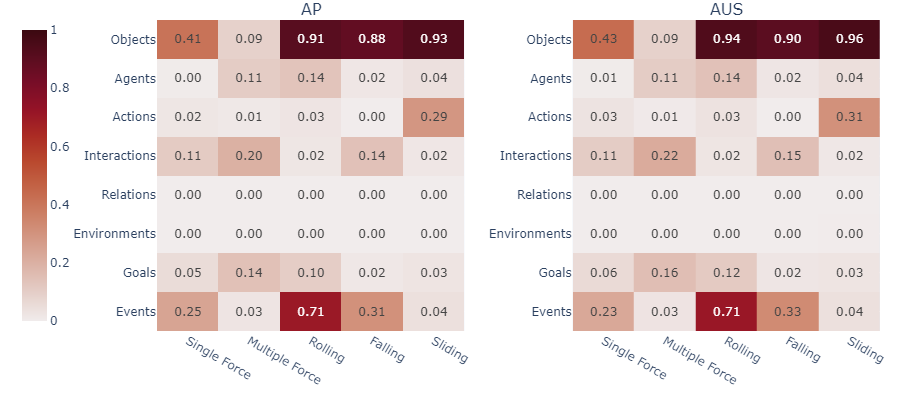}
    \caption{The AP and AUS results of the DQN Offline.}
    \label{fig:adaptation_dqn_offline}
\end{figure}

\begin{figure}
    \centering
    \includegraphics[width=0.9\textwidth,height=\textheight,keepaspectratio]{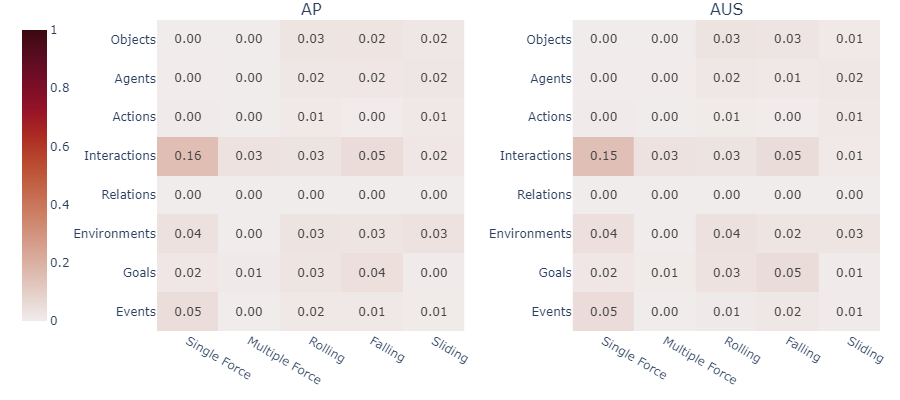}
    \caption{The AP and AUS results of the Random agent.}
    \label{fig:adaptation_random}
\end{figure}

\begin{figure}
    \centering
    \includegraphics[width=\textwidth,height=\textheight,keepaspectratio]{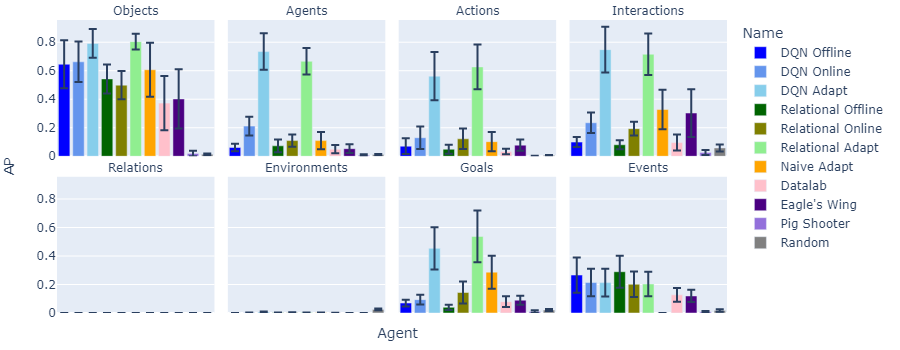}
    \caption{$NPS_{AP}$ of the agents for the eight novelties.}
    \label{fig:ap_per_novelty}
\end{figure}

\begin{figure}
    \centering
    \includegraphics[width=\textwidth,height=\textheight,keepaspectratio]{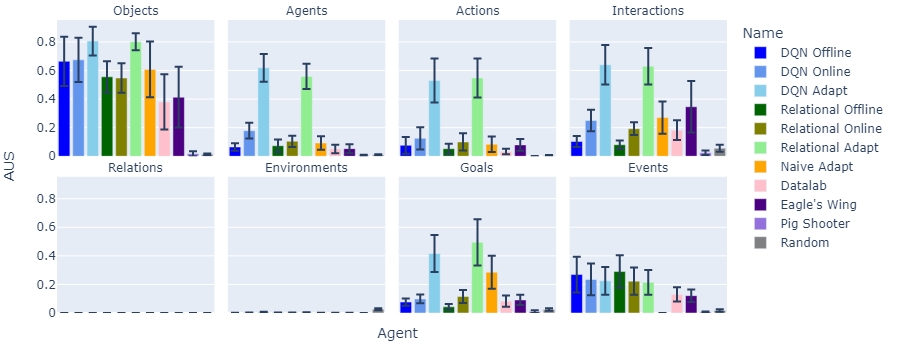}
    \caption{$NPS_{AUS}$ of the agents for the eight novelties.}
    \label{fig:aus_per_novelty}
\end{figure}

\begin{figure}
    \centering
    \includegraphics[width=\textwidth,height=0.8\textheight,keepaspectratio]{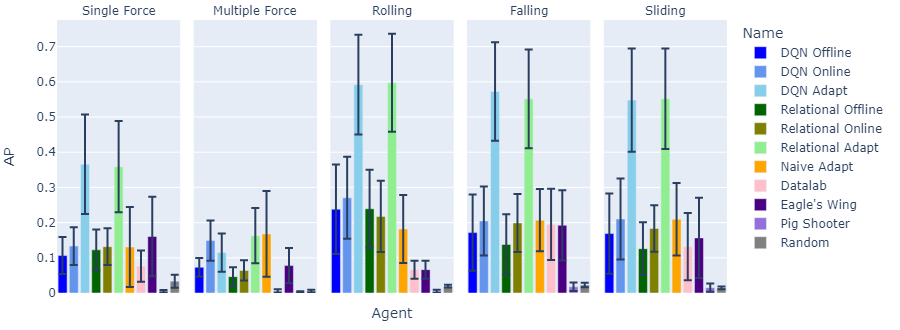}
    \caption{$SPN_{AP}$ of the agents for the five scenarios.}
    \label{fig:ap_per_scenario}
\end{figure}

\begin{figure}
    \centering
    \includegraphics[width=\textwidth,height=0.8\textheight,keepaspectratio]{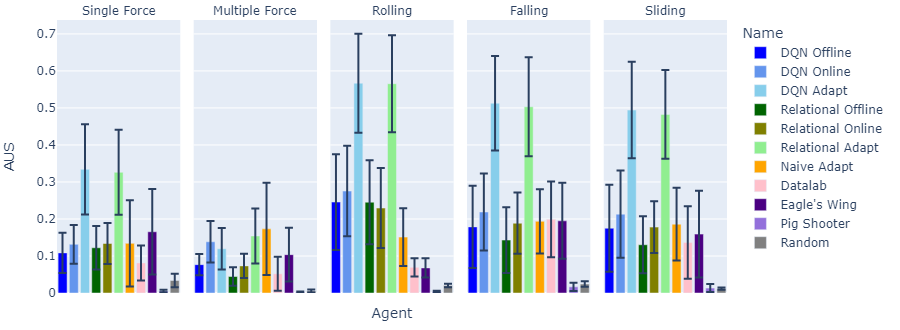}
    \caption{$SPN_{AUS}$ of the agents for the five scenarios.}
    \label{fig:aus_per_scenario}
\end{figure}

The adaptation curves of each novelty-scenario are presented in Figure \ref{fig:adaptation_plot_per_novelty-scenario}. The AP and AUS values derived from the adaptation curves for each agent are shown in the heat maps in Figures \ref{fig:adaptation_dqn_offline} - to \ref{fig:adaptation_random}. The AUS values are calculated from the performance of the agent in the last 50\% of the novel tasks (m=20). The adaptation results per novelty are shown in Figures \ref{fig:ap_per_novelty} and \ref{fig:aus_per_novelty} while Figures \ref{fig:ap_per_scenario} and \ref{fig:aus_per_scenario} show the adaptation results per scenario performance. It can be seen that agents DQN Adapt and DQN Relational could reach human AP and AUS results in some of the novelty-scenarios. However, looking at the adaptation plots, within the same number of tasks that were given to humans, those two agents have not reached the performance the humans could achieve, showing that there is room for improvement in terms of adaptation efficiency.

\begin{figure}
    \centering
    \includegraphics[width=0.9\textwidth,height=\textheight,keepaspectratio]{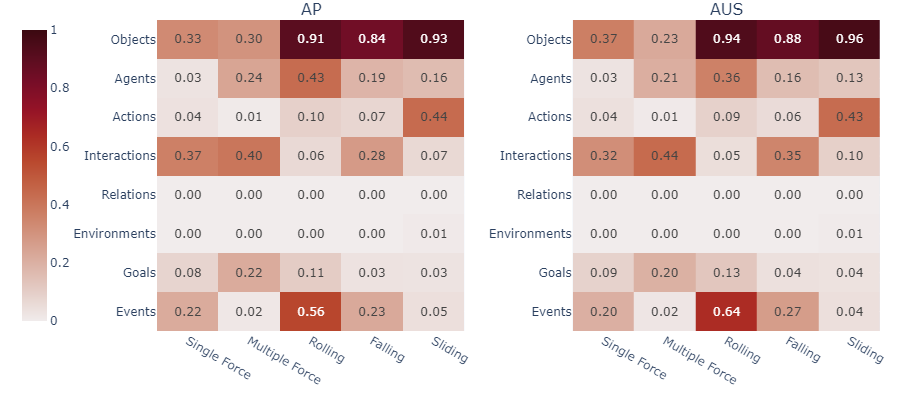}
    \caption{The AP and AUS results of the DQN Online.}
    \label{fig:adaptation_dqn_online}
\end{figure}

\begin{figure}
    \centering
    \includegraphics[width=0.9\textwidth,height=\textheight,keepaspectratio]{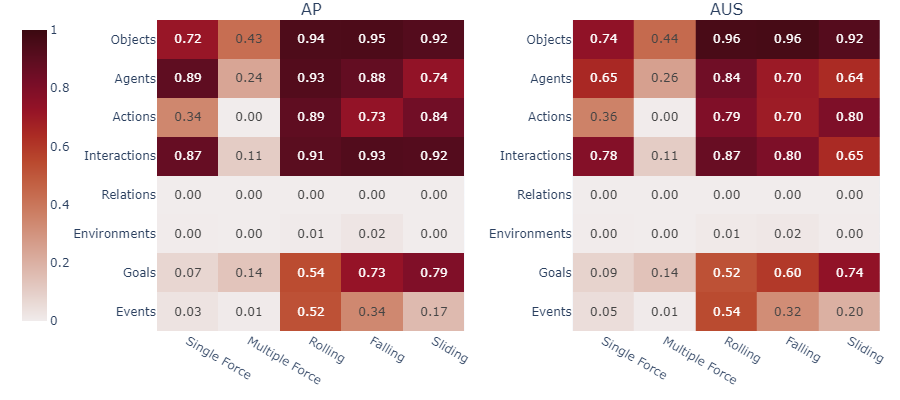}
    \caption{The AP and AUS results of the DQN Adapt.}
    \label{fig:adaptation_dqn_adapt}
\end{figure}

\begin{figure}
    \centering
    \includegraphics[width=0.9\textwidth,height=\textheight,keepaspectratio]{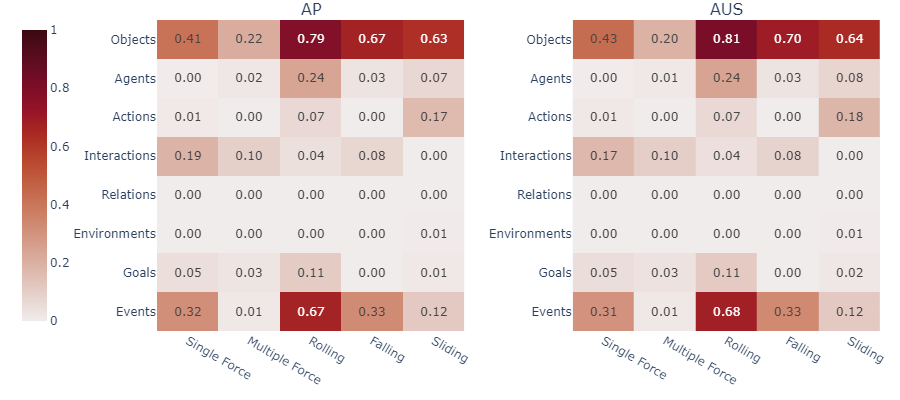}
    \caption{The AP and AUS results of the Relational Offline.}
    \label{fig:adaptation_rel_offline}
\end{figure}

\begin{figure}
    \centering
    \includegraphics[width=0.9\textwidth,height=\textheight,keepaspectratio]{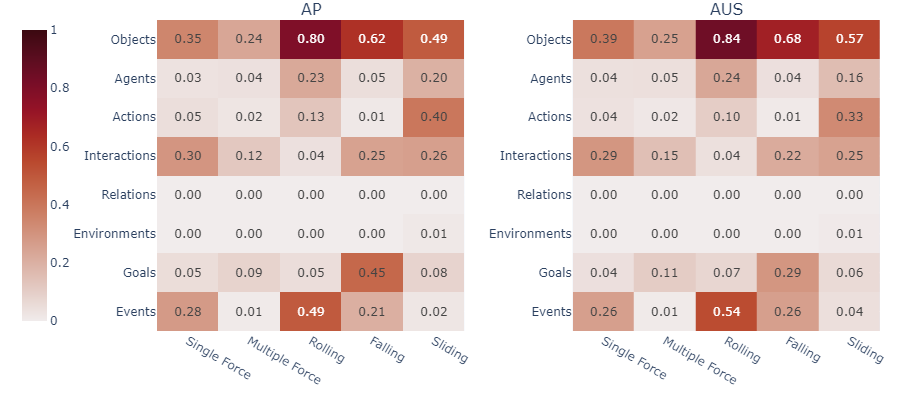}
    \caption{The AP and AUS results of the Relational Online.}
    \label{fig:adaptation_rel_online}
\end{figure}

\begin{figure}
    \centering
    \includegraphics[width=0.9\textwidth,height=\textheight,keepaspectratio]{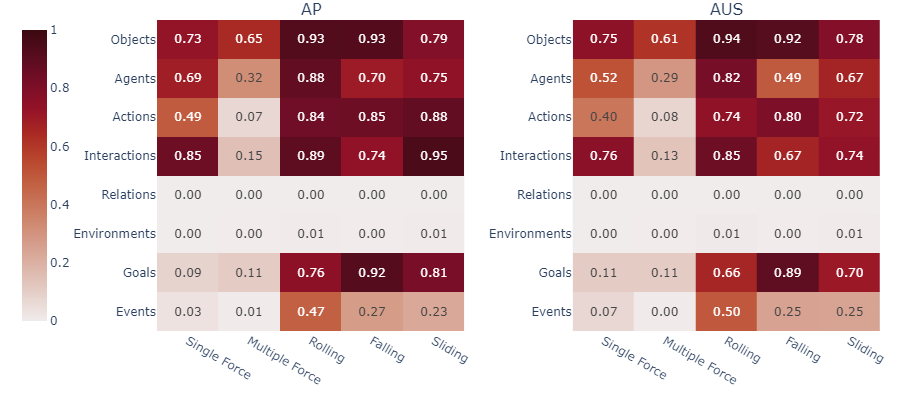}
    \caption{The AP and AUS results of the Relational Adapt.}
    \label{fig:adaptation_rel_adapt}
\end{figure}

\begin{figure}
    \centering
    \includegraphics[width=0.9\textwidth,height=\textheight,keepaspectratio]{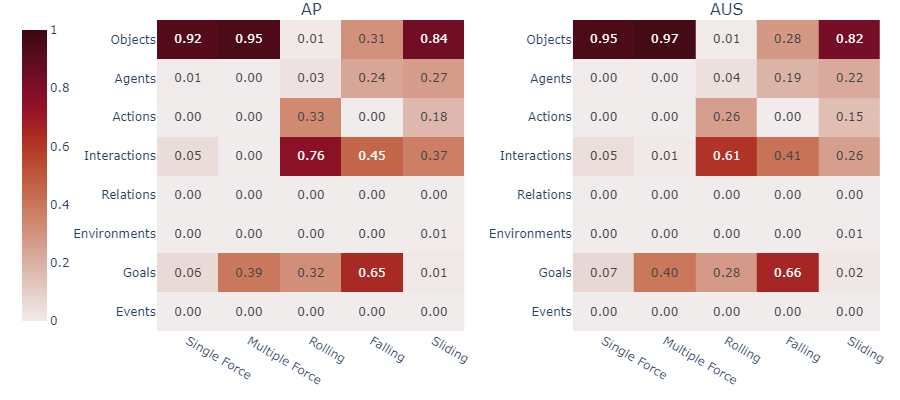}
    \caption{The AP and AUS results of the Naive Adapt.}
    \label{fig:adaptation_naive_adapt}
\end{figure}

\begin{figure}
    \centering
    \includegraphics[width=0.9\textwidth,height=\textheight,keepaspectratio]{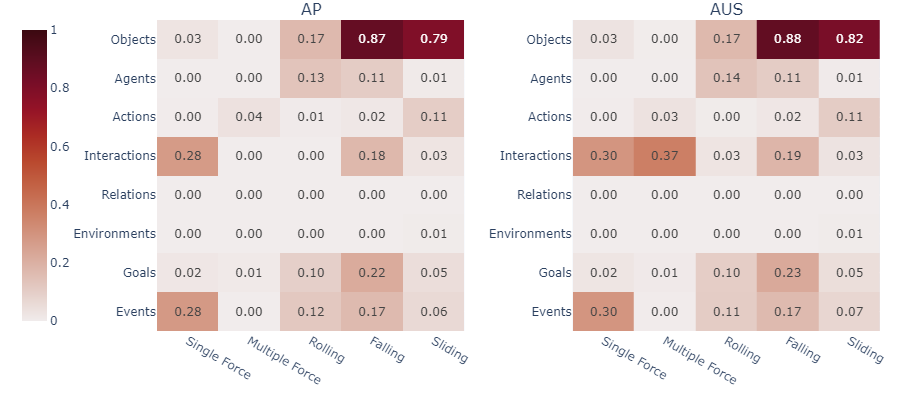}
    \caption{The AP and AUS results of the Datalab.}
    \label{fig:adaptation_datalab}
\end{figure}

\begin{figure}
    \centering
    \includegraphics[width=0.9\textwidth,height=\textheight,keepaspectratio]{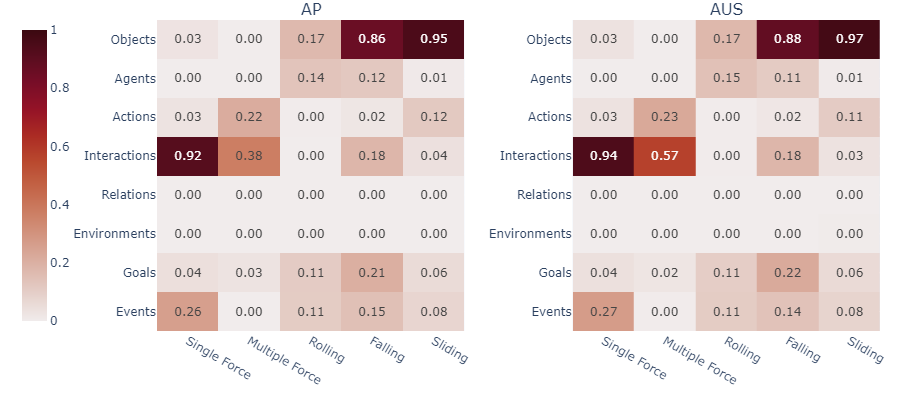}
    \caption{The AP and AUS results of the Eagle's Wing.}
    \label{fig:adaptation_eagleswing}
\end{figure}

\begin{figure}
    \centering
    \includegraphics[width=0.9\textwidth,height=\textheight,keepaspectratio]{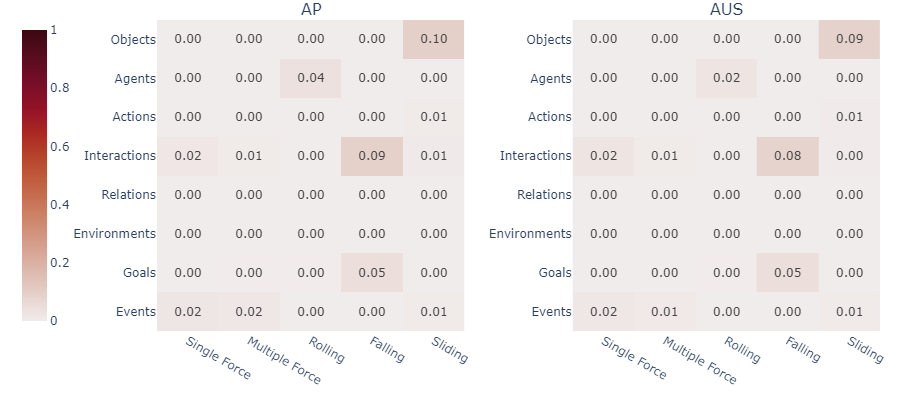}
    \caption{The AP and AUS results of the Pig shooter.}
    \label{fig:adaptation_pig_shooter}
\end{figure}

\newpage

\section{Human Detection vs Adaptation} \label{Appendix: Section Human Detection vs Adaptation}

In the previous sections, we have discussed humans' detection and adaptation performance and shown that humans' performance is far better than the agents' performance. This section analyses the relationship between novelty detection and adaptation performance of humans. 

\begin{figure}[h!]
    \centering    \includegraphics[width=\textwidth,height=0.9\textheight,keepaspectratio]{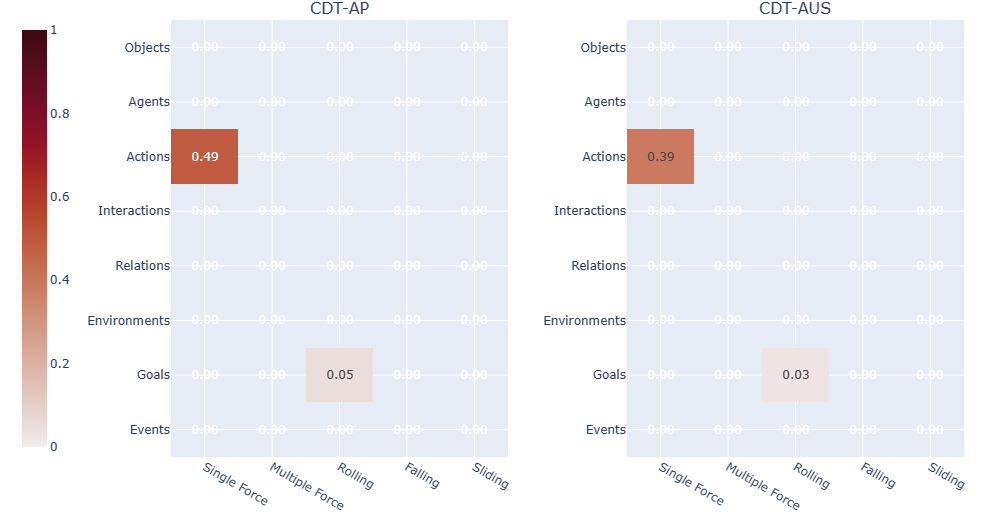}
    \caption{The p-value of the Mann-Whitney test for adaptation performance on the CDT=1 group and CDT=0 group}    \label{fig:CDT_vs_adaptation_significance}
\end{figure}

Figure \ref{fig:CDT_vs_adaptation_significance} shows the p-value of the Mann-Whitney test conducted to test if there is a difference in adaptation performance based on whether the person correctly detected the novelty or not (CDT = 1 or CDT = 0). For CDT = 0 group, we have only used the data from the players who did not detect novelty (excluded the data of the players who did wrong defections). The null hypothesis is, the adaptation performance of the two groups is equal. 
As Figure \ref{fig:CDT_vs_adaptation_significance} indicates, almost all the entries are not defined as all human players correctly detected novelty and hence there are no players who did not detect the novelty in those novelty-scenarios. For the actions-single force novelty-scenario, the p-values in both AP and AUS exceed the 5\% level of significance we consider. i.e., we do not have enough evidence to reject the null hypothesis, meaning that irrespective of the novelty detection, players performed similarly. The novelty here is the increased upward force of Air Turbulence, which is not visually detectable unless interacted. Some of the human players did not detect this novelty but they managed to solve the task as there is only a subtle change in the action from normal tasks to novel tasks. On the other hand, in the novelty-scenario goals-rolling, there is a significant difference in adaptation performance based on the CDT. In this novelty-scenario, the action has to be considerably adjusted when moving from normal to novel tasks, thus if the novelty is not detected, it would be difficult to adapt. 

\begin{figure}[h!]
    \centering    \includegraphics[width=\textwidth,height=0.9\textheight,keepaspectratio]{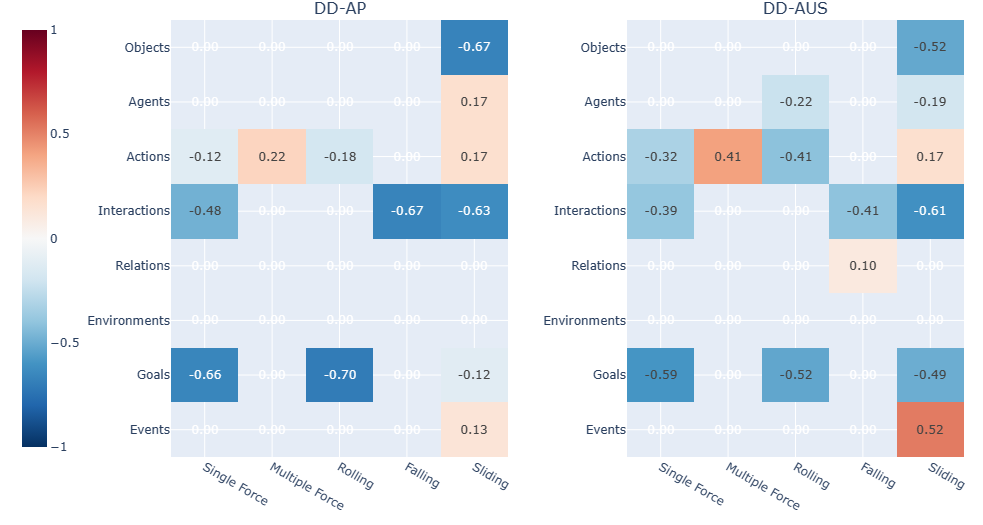}
    \caption{The Spearman's rank correlation for adaptation performance and detection delay}    \label{fig:DD_vs_adaptation_correlation}
\end{figure}

Figure \ref{fig:DD_vs_adaptation_correlation} shows the Spearman's rank correlation coefficient between detection delay and adaptation performance (AP and AUS). For this calculation, if the person did not detect novelty in the trial, we have assumed the detection delay to be five, which is the maximum number of novel instances in a trial + 1. As can be seen from the figures, the correlation coefficient is undefined for some novelty-scenarios. Those are the cases where all the human players detected the novelty in the first novel task (DD = 1). As shown in the DD-AP heatmap, there are five moderately strong negative correlations (between -0.6 and -0.7) and they are all statistically significant at 5\% level of significance. This indicates that, for those novelty-scenarios, the more tasks it takes to detect, the lesser the asymptotic adaptation performance. According to DD-AUS heatmap, two novelty-scenarios have moderate positive correlations (0.41 and 0.52). However, they are not statistically significant at 5\% level of significance. Only the moderately strong negative correlation value at novelty-scenario interactions-sliding is significant at 5\% level of significance.

%% else use the following coding to input the bibitems directly in the
%% TeX file.

% \begin{thebibliography}{00}

% %% \bibitem{label}
% %% Text of bibliographic item

% \bibitem{}

% \end{thebibliography}
%\fi
%\end{linenumbers}
\end{document}